%% file: arxiv.tex
\documentclass{article}[12pt]

\usepackage[margin=1.1in]{geometry}

\usepackage[utf8]{inputenc} % allow utf-8 input
\usepackage[T1]{fontenc}    % use 8-bit T1 fonts
\usepackage[hidelinks,colorlinks,linkcolor=black,citecolor=orange]{hyperref}   % hyperlinks
 \hypersetup{
     colorlinks=true,
     linkcolor=blue,
     filecolor=blue,    
     urlcolor=cyan,
     }
\usepackage{placeins}
\usepackage{url}            % simple URL typesetting
\usepackage{booktabs}       % professional-quality tables
\usepackage{amsfonts}       % blackboard math symbols
\usepackage{nicefrac}       % compact symbols for 1/2, etc.
\usepackage{microtype}      % microtypography
\usepackage{float}
\usepackage{xcolor}         % colors
\usepackage[citestyle=numeric,natbib=true,bibstyle=numeric,maxbibnames=99]{biblatex}
\usepackage{graphicx}
\usepackage{caption}
\usepackage{subcaption}
\usepackage{amsmath,amssymb,amsthm}
\usepackage{bbm}
\usepackage{enumitem}   
\usepackage{capt-of}
\usepackage{wrapfig}

\input{math_command.tex}

\title{Catapults in SGD: spikes in the training loss and their impact on  generalization  through feature learning}

\author{%\
  Libin Zhu\thanks{Department of Computer Science \& Halicioğlu Data Science Institute, UC San Diego. E-mail: \texttt{libinzhu@ucsd.edu}} ~~~
  Chaoyue Liu\thanks{Halicioğlu Data Science Institute, UC San Diego. E-mail: \texttt{chl212@ucsd.edu}} ~~~
  Adityanarayanan Radhakrishnan \thanks{Harvard University \& Broad Institute of MIT and Harvard. E-mail: \texttt{aradha@mit.edu}}~~~
  Mikhail Belkin\thanks{Halicioğlu Data Science Institute \& Department of Computer Science, UC San Diego. E-mail: \texttt{mbelkin@ucsd.edu}}
  \\
}
\date{}

\begin{document}
\maketitle
\begin{abstract}
In this paper, we first present an explanation regarding the common occurrence of spikes in the training loss when neural networks are trained with stochastic gradient descent (SGD). We provide evidence that the spikes in the training loss of  SGD are ``catapults'',  an optimization phenomenon originally observed in GD with large learning rates in~\cite{lewkowycz2020large}. We empirically show that these catapults  occur in a low-dimensional subspace spanned by the top eigenvectors of the tangent kernel, for both GD and SGD. Second, we posit an explanation for how catapults lead to better generalization by demonstrating that catapults promote feature learning by increasing alignment with the Average Gradient Outer Product (AGOP) of the true predictor.  Furthermore, we demonstrate that a smaller batch size in SGD induces a larger number of catapults, thereby improving \AGOP alignment and test performance. 
\end{abstract}

\section{Introduction}

\begin{wrapfigure}{r}{0.35\textwidth}
\vspace{-20pt}
  \begin{center}
    \includegraphics[width=0.30\textwidth]{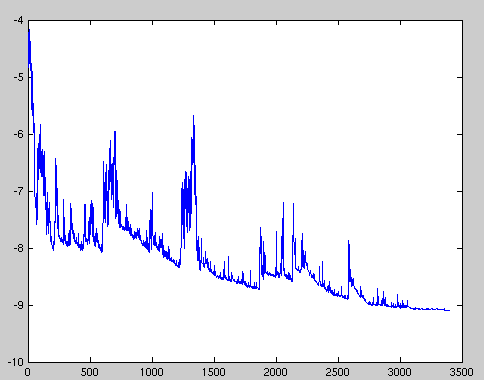}
  \end{center}\vspace{-10pt}
  \caption{Spikes in training loss when optimized using SGD (x-axis: iteration). (Source: \href{https://en.wikipedia.org/wiki/Stochastic_gradient_descent}{Wikipedia})}\label{fig:sgd_spikes_wiki}\vspace{-20pt}
\end{wrapfigure}
Training algorithms are a key ingredient to the success of deep learning.
%The success of deep learning has sparked significant interest in understanding its mechanism, with the training algorithm being a key ingredient. 
Stochastic gradient descent (SGD)~\citep{robbins1951stochastic}, a stochastic variant of gradient descent (GD), has been effective in finding parameters that yield good test performance despite the complicated nonlinear nature of neural networks.  Empirical evidence suggests that training networks using SGD with a larger learning rate results in better predictors~\citep{Frankle2020The, smith2019super, gilmer2021loss}.  In such settings, it is common to observe significant spikes in the training loss~\citep{lecun2015deep,ruder2016overview,keskar2017improving, xing2018walk} (see Fig.~\ref{fig:sgd_spikes_wiki} as an example).  One may not \textit{a priori} expect the training loss to decrease
back to its ``pre-spike'' level after a large spike.
Yet, this is what is commonly observed in training. Furthermore, the resulting ``post-spike'' model can yield improved generalization performance~\citep{he2016deep,zagoruyko2016wide,huang2017densely}.  
\begin{center}
    {\it Why do spikes occur during training, and how do the spikes relate to generalization? }
\end{center}

In this work, we answer these questions by connecting three common but seemingly unrelated phenomena in deep learning:
\begin{enumerate}[itemsep=0mm]
\item Spikes in the training loss of SGD,  
\item Catapult dynamics in GD~\citep{lewkowycz2020large},  \item Better generalization when training networks with small batch SGD as opposed to larger batch size or GD.
\end{enumerate}
In particular, we show that spikes in the training loss of SGD are caused by catapult dynamics, which were originally characterized in~\cite{lewkowycz2020large} as a single spike in the loss when training with GD and large learning rate.  We then show that smaller batch size in SGD results in a greater number of catapults.  We connect the optimization phenomena of catapults to generalization by showing that catapults improve generalization through increasing feature learning, which is quantified by the alignment between the Average Gradient Outer Product (AGOP) of the trained network and the true AGOP~\citep{hardle1989investigating,hristache2001structure,xia2002adaptive,trivedi2014consistent,radhakrishnan2024mechanism}.  Since decreasing batch size in SGD leads to more catapults, our result implies that SGD with small batch size yields improved generalization (see Table~\ref{tab:test_loss} for an example).  We outline our specific contributions in the context of optimization and generalization below.  

\begin{table}
\centering
\scalebox{1.0}{
\begin{tabular}{ccc}\\\toprule  
Batch size  & \shortstack{AGOP alignment} & Test loss  \\ \midrule\midrule
2000~(GD) & 0.81& 0.74 \\  \midrule
50 & 0.84&  0.71\\\midrule
10 & 0.89&  0.59\\\midrule
5  & 0.95& 0.42 \\  \bottomrule
\end{tabular}}
\caption{Smaller SGD batch size leads to a  higher (better) AGOP alignment and smaller (better) test loss. The results correspond to Fig.~\ref{fig:fl_sgd_agop}a (a synthetic dataset).\label{tab:test_loss}}

\end{table}

\paragraph{Optimization.}  We demonstrate that spikes in the training loss, specifically measured by Mean Squared Error, occur in the top eigenspace of the Neural Tangent Kernel, a kernel resulting from the linearization of a neural network~\citep{jacot2018neural}.  Namely, we project the residual (i.e., the difference between the predicted output and the target output) to the top eigenspace of the tangent kernel and show that spikes in the total loss function correspond to the spikes in the components of the loss in this low-dimensional subspace (see Section~\ref{sec:cata_gd}).  In contrast, the components of the loss in the space spanned by the remaining eigendirections decrease monotonically.  
Thus, the catapult phenomenon occurs in the span of the top eigenvectors while the remaining eigendirections are not affected. This explains why the loss drops quickly to pre-spike levels, namely the loss value right before the spike, from the peak of the spike.   We further show that multiple catapults can be generated in GD by increasing the learning rate during training (see Section~\ref{subsec:multi_cata}).   While prior work~\cite{lewkowycz2020large} observed that the spectral norm of the tangent kernel decreased for one catapult, we extend that observation by  showing that the norm decreases after each catapult.
%thereby preventing divergence of the loss. 

% \adit{add back contribution on decreasing spectral norm for multiple GD and SGD.}

We further provide evidence for catapults in SGD with large learning rates (see Section~\ref{sec:cata_sgd}).  Namely, we demonstrate that spikes in the loss when training with SGD correspond to catapults by showing that similarly to GD:
\begin{enumerate}[itemsep=0mm]
\item The spikes occur in the top eigenspace of the tangent kernel,
\item Each spike results in a decrease in the spectral norm of the tangent kernel.
\end{enumerate}
We corroborate our findings across several network architectures including Wide ResNet~\citep{zagoruyko2016wide} and ViT~\citep{dosovitskiy2021an}  and datasets including CIFAR-10~\citep{krizhevsky2009learning} and SVHN~\citep{netzer2011reading}. 

Moreover, as  small batch size leads to higher variance in the eigenvalues of the tangent kernel for any given batch,  small batch size results in an increased number of catapults.

\paragraph{Generalization.} We posit that catapults improve the generalization performance by alignment between the AGOP of the trained network with that of the true model\footnote{{When the underlying model is not available, we use a SOTA model as a substitute.} }.  The AGOP identifies the features that lead to greatest change in predictor output when perturbed and has been recently posited as the mechanism through which neural networks learn features~\citep{radhakrishnan2024mechanism,beaglehole2023mechanism}.    We  use AGOP alignment to provide an explanation for prior empirical results from~\cite{lewkowycz2020large,zhu2024quadratic} showing that a single catapult can lead to better test performance in GD.  Moreover, we extend these prior results to show that test performance continues to improve as the number of catapults increases in GD.  Thus, we show that decreasing batch size with SGD can lead to better test performance due to an increase in the number of catapults. We further demonstrate that AGOP alignment is an effective measure of generalization by showing that test error is highly correlated with the AGOP alignment when training on the same task across different optimization algorithms including Adagrad~\citep{duchi2011adaptive}, Adadelta~\citep{zeiler2012adadelta} and Adam~\citep{KingBa15} etc.  We corroborate our findings on CelebA~\citep{liu2015faceattributes} and SVHN~\citep{netzer2011reading} datasets and architectures including fully-connected and convolutional neural networks.  See Section~\ref{sec:feature_learning}.

\subsection{Related works}

\paragraph{Linear dynamics and catapult phase phenomenon.} Recent studies have shown that (stochastic) GD for wide neural networks provably converges to global minima with an appropriately small learning rate~\citep{du2019gradient,zou2019improved,liu2020loss}. These works leveraged the fact that neural networks with sufficiently large widths, under specific initialization conditions, can be accurately approximated by their linearization obtained by the first-order Taylor expansion~\citep{jacot2018neural, liu2020linearity,liu2022transition,zhu2024quadratic}. Therefore, their training dynamics are close to the dynamics of the corresponding linear models, under which the training loss decreases monotonically. Such a training regime is commonly referred to as the kernel regime. However, under the same setup of the kernel regime except using a large learning rate, GD will experience a catapult phase~\cite{lewkowycz2020large}: the training loss increases drastically in the beginning stage of training then decreases, while GD still converges. {Recent studies focusing on understanding catapults in GD include~\cite{zhu2024quadratic}, which considers quadratic approximations of neural networks, and~\cite{meltzer2023catapult}, examining two-layer homogeneous neural networks. Our work investigates the impact of catapults in SGD on both optimization and generalization through experimental approaches. }
% \vspace{-10pt}
% \paragraph{Catapult phase.} When training networks with GD and large learning rate, recent work~\cite{lewkowycz2020large} identified a striking phenomenon that cannot be manifested in the kernel regime. This phenomenon, referred to as the ``catapult phase'',  is characterized by an increase in loss during the early stages of training, followed by a decrease that forms a single spike in the training loss. After the catapult, the spectral norm of the tangent kernel, i.e., its top eigenvalue, decreases and thus, remarkably prevents divergence. 

\paragraph{Edge of stability.} A phenomenon related to catapults is the ``Edge of Stability'' (EoS), which describes the dynamics of the training loss and the sharpness, i.e., eigenvalues of the Hessian of the loss, at the later stage of training networks with GD~\citep{cohen2021gradient} and SGD~\citep{jastrzębski2018on,Jastrzebski2020The}.  {There is a growing body of work analyzing the mechanism of EoS in training dynamics with GD~\citep{arora2022understanding, ahn2022understanding,damian2023selfstabilization,wang2022analyzing,agarwala2023second,agarwala2023sam,wang2021large}, and SGD~\cite{kalra2023phase}. It was conjectured in~\cite{cohen2021gradient} that at EoS for GD the spikes in the training loss are micro-catapults. Our work provides evidence that the spikes in the training loss using SGD are catapults and demonstrates the connection between the loss spikes and feature learning. }
% We believe it is important to understand how EoS relates to catapults to better understand the optimization and generalization of neural networks.

\paragraph{Generalization and sharpness.} It has been observed that networks trained with SGD generalize better than GD, and smaller batch sizes often lead to better generalization performance ~\citep{lecun2002efficient,keskar2017on,goyal2017accurate,jastrzkebski2017three,masters2018revisiting,kandel2020effect,smith2020origin}. Empirically, it has been observed that training with SGD results in flat minima~\citep{hochreiter1994simplifying,hochreiter1997flat}. However, we noticed that it is not always the case, e.g.,~\cite{geiping2022stochastic}. A number of works  been argued that flatness of the minima is connected to the generalization performance~\citep{neyshabur2017exploring,wu2017towards,kleinberg2018alternative,xie2020diffusion,Jiang*2020Fantastic,dinh2017sharp}, however we know only one theoretical result in that direction~\citep{ding2024flat}. Training algorithms aiming to find a flat minimum were developed and shown to perform well on a variety of tasks~\citep{izmailov2018averaging,foret2021sharpnessaware}.  As an explanation for empirically observed improved generalization, prior work~\cite{lewkowycz2020large} argued that a single catapult with GD resulted in flatter minima.  In this work we propose a different line of investigation to understand generalization properties of GD-based algorithms based on feature learning as measured by the alignment with AGOP.

\section{Preliminaries}
\paragraph{Notation.} We use bold letters (e.g., $\rvw$) to denote vectors and capital letters (e.g., $K$) to denote matrices.  For a matrix, we use $\|\cdot\|_F$ to denote its Frobenius norm and use $\|\cdot\|_2$ to denote its spectral norm. For trainable parameters, we use superscript $t$, as in $\rvw^t$, to denote the time stamp during training. We use the big-$O$ notation $O(\cdot)$ to hide constants, and  use $\tilde{O}(\cdot)$ to further hide logarithmic factors.
For a map $f(\rvw):\mathbb{R}^p \rightarrow \mathbb{R}^c$, we use $\nabla_\rvw f(\rvv)$ and $\nabla_\rvw^2f(\rvv)$ to denote the first and second order derivative of $f$ w.r.t. $\rvw$ evaluated at $\rvv$ respectively.

\paragraph{Optimization task.}  Consider a parameterized model $f(\rvw;\cdot):\mathbb{R}^p \rightarrow \mathbb{R}$ (e.g., a neural network) with parameters $\rvw$ and a
training dataset $\mathcal{D}=\{(\vx_i,y_i)\}_{i=1}^n$ with data $\vx_i \in \mathbb{R}^d$ and labels $y_i \in\mathbb{R}$ for $i\in[n]$. Denote $X\in \mathbb{R}^{n\times d}$ as the collection of training input data, with each row of $X$ representing an input $\vx_i$, and $\vy:=(y_1,\cdots,y_n)^T$. We further write $\rvf \in \mathbb{R}^{n}$ as the predictions of $f$ on $X$. The goal of the optimization task is to  minimize the Mean Square Error (MSE) 
{\small
\begin{align}\label{eq:mse}
    \L(\rvw;(X,\vy)) = \frac{1}{n}\sum_{i=1}^n (f(\rvw;\vx_i)-y_i)^2=\frac{1}{n}\|\rvf-\vy\|^2.
\end{align} 
}
Let $\rvw_0$ be the weight parameters at initialization. Mini-batch SGD is conducted as follows: at each step $t$, randomly sample a batch $\mathcal{B}\subset \mathcal{D}$ (of batch size $b$), and perform the update following
\begin{align*}
    \rvw^{t+1} = \rvw^t - \frac{\eta}{b} \frac{\partial }{\partial \rvw}\sum_{(\vx_j,y_j)\in \mathcal{B}}(f(\rvw^t;\vx_j) -y_j)^2 ,
\end{align*}
where $\eta$ is the learning rate. When $b=n$, mini-batch SGD reduces to the full-batch gradient descent (GD).

% \vspace{-10pt}\paragraph{Paramterization.} Unless specified, we use NTK parameterization~\citep{jacot2018neural} for neural networks, that is, we initialize each trainable parameter i.i.d. from standard normal distribution, i.e., $\rvw_0 \sim \mathcal{N}(\mathbf{0},I_p)$, and multiply each pre-activated neuron with an extra scaling factor $1/\sqrt{m}$ where $m$ is the fan-in of that neuron. For example, for neurons in fully connected networks, the fan-in is the width of the previous hidden layer.  In convolutional nets, fan-in is the number of channels multiplied by the size of the window.  We also use Pytorch~\citep{paszke2019pytorch} default parameterization in some settings.

\paragraph{Neural Tangent Kernel (NTK).} Proposed in \cite{jacot2018neural}, NTK is a useful tool in understanding and analyzing over-parameterized neural networks. 

\begin{definition}[(Neural) Tangent Kernel]
The (neural) tangent kernel $K(\rvw;\cdot,\cdot)$ for a parameterized machine learning model $f(\rvw;\cdot):\mathbb{R}^p\times \mathbb{R}^d \rightarrow \mathbb{R}$ is defined  as: 
\begin{align*}
    \forall \vx,\vz\in\mathbb{R}^d, \quad K(\rvw;\vx,\vz) =\inner{  {\frac{\partial f(\rvw;\vx)}{\partial \rvw},\frac{\partial f(\rvw;\vz)}{\partial\rvw}}}.
\end{align*}
\end{definition}

Given the training data inputs $X$, the NTK can be evaluated on any pair of inputs $\vx_i$ and $\vx_j$, which results in a $n\times n$ matrix $K$, called the NTK matrix.
% It was shown in~\cite{jacot2018neural} that for infinitely wide neural networks, the evolution of the output $\rvf^t$ at time $t$ under GD with learning rate $\eta $ takes the form $\rvf^{t+1} - \vy = (I_n - \eta K_0/n) (\rvf^t - \vy)$, where $K_0 = K(w_0)$. In this paper, we analyze the training dynamics for finite-width neural networks using their linear approximation whose training dynamics can be described by a time-dependent NTK $K_t$ using a second-order Taylor expansion of $\rvf^{t+1}$:
By definition, the NTK matrix $K$ is symmetric and positive semi-definite. Therefore, it can be decomposed as  $K = \sum_{j=1}^n \lambda_{j}\rvu_{j}{\rvu_{j}}^T$,
with $\lambda_{j}\in \mathbb{R}$ and $\rvu_{j}\in\mathbb{R}^n$, $j \in \{ 1,\cdots, n\}$, being the eigenvalues and unit-length eigenvectors, respectively. Without loss of generality, we assume $\lambda_1 \geq \lambda_2 \geq \cdots \geq \lambda_n\ge 0$. 

\paragraph{Top-eigenspace and decomposition of the loss.}
Given an integer $s$, $1\le s < n$, we call the {\it top eigenspace} (or {\it top-$s$ eigenspace}) of NTK as the subspace spanned by the top eigenvectors $\rvu_j$ with $1\le j \le s$. We also define projection operators $\P_{\leq s}: \mathbb{R}^n \rightarrow \mathbb{R}^n$ and $\P_{> s}: \mathbb{R}^n \rightarrow \mathbb{R}^n$, such that for any vector $\rvv \in \mathbb{R}^n$ the followings hold:
\begin{align*}
    \P_{\leq s} \rvv = \sum_{i=1}^s \inner{\rvv,\rvu_i}\rvu_i,~~~ \P_{>s}\rvv = \sum_{i=s+1}^n \inner{\rvv,\rvu_i}\rvu_i.
\end{align*}
% \end{align*}
The MSE Eq.~(\ref{eq:mse}) can be decomposed as 
\begin{align}\label{eq:loss_decomp}
    \L = \frac{1}{n}\norm{\rvf - \vy}_2^2& = \frac{1}{n}\norm{\P_{\leq s}(\rvf-\vy)}_2^2 + \frac{1}{n}\norm{\P_{>s} (\rvf-\vy)}_2^2 =: \L_{{\leq s}} + \L_{{> s}}.
\end{align} 

\paragraph{Critical learning rate.} When a constant learning rate of the algorithm is used throughout the training, it is important to select the learning rate $\eta$, as a large $\eta$ easily leads to a divergence of loss and a small $\eta$  slows down the training procedure. A conventional wisdom is to set $\eta$ no larger than the {\it critical learning rate} $\etc(\rvw):= \frac{2}{\lambda_{\max}(H_\L(\rvw))}$, where $H_\L:= \nabla_\rvw^2\L(\rvw)$ denotes the Hessian of the loss. This intuition follows from the well-known lemma in optimization: 
\begin{lemma}[Descent Lemma~\cite{nesterov1983method}]
For a smooth loss $\L(\rvw):\mathbb{R}^p\rightarrow \mathbb{R}$, suppose $\lambda_{\max}(H_\L(\rvw)) \leq \beta$ for all $\rvw\in \mathbb{R}^p$, then GD satisfies:
\begin{align*}
    \L(\rvw^{t+1}) \leq \L(\rvw^t) - {\eta \round{1-\frac{\eta \beta}{2}}}\norm{\nabla_\rvw \L(\rvw^t)}^2.
\end{align*}
\end{lemma}

% This is because learning rates larger than $\etc$ are believed to trigger the divergence of the algorithm, while smaller learning rates are considered safe for convergence. This can be seen from the simple case of quadratic loss. ????add explanation????? 
 For $\eta < 2/\beta$, the descent lemma guarantees the decrease of the loss.
 Note that this inequality is tight for quadratic loss, e.g., loss for linear models.  
 % The inequality still holds if we relax the condition on $\rvw$ such that $\sup_{\tau\in[0,1]}\lambda_{\max}\round{H_\L(\tau\rvw^t + (1-\tau)\rvw^{t+1})} \leq \beta$. Therefore we define $\etc(\rvw^t)$ as $2/\lambda_{\max}(H_\L(\rvw^t))$. Note that without specifying the dependence on the time step $t$, the critical learning rate is evaluated at initialization $\rvw_0$.
 For neural networks with sufficient width trained with a constant learning rate smaller than $\etc$, due to {\it{transition to linearity}}~\cite{liu2020linearity}, the critical learning rate $\eta_{\mathrm{crit}}$ almost does not change during training~\cite{lee2019wide}.  Furthermore, by decomposing the Hessian of the loss, it can be seen that $\etc$ can be well-approximated by NTK (exact, for linear models): $\eta_{\mathrm{crit}} \approx n/{\|K\|_2} = 
{n}/{\lambda_1}$, as detailed in Appendix~\ref{subsec:h_K_init}. For neural networks that are not wide, \cite{papyan2019measurements,agarwala2023second,wang2022analyzing} showed the approximation still holds and we provide additional evidence for SGD trained with a large learning rate in Appendix~\ref{subsec:h_K_whole}. 

Note that unless specified, the critical learning rate is evaluated at initialization $\rvw_0$.
% Unless specified otherwise, the critical learning rate is evaluated at initialization $\rvw_0$.

% \paragraph{Critical learning rate.} When a constant learning rate of algorithm is used throughout the training, it is important to select the learning rate $\eta$, as a large $\eta$ may easily lead to a divergence of loss and a small $\eta$ may slow down the training procedure. A common wisdom is to set $\eta$ no larger than the {\it critical learning rate} $\etc:= \frac{2}{\lambda_{\max}(H_\L(\rvw))}$, where $H_\L:= \frac{d^2\L(\rvw)}{d\rvw^2}$ denotes the Hessian of the loss. This is because learning rates larger than $\etc$ are believed to trigger the divergence of algorithm, while smaller learning rates are considered safe for convergence. This can be seen from the simple case of quadratic loss. ????add explanation????? 

% For neural networks with sufficient width (also for over-parameterized linear models), the critical learning rate $\etc$ can be well-approximated (exact, for linear models) by $\etc \approx 2/\|K\|_2 = 2/\lambda_1$ (see ????).

\begin{wrapfigure}{r}{0.4\textwidth}
\vspace{-25pt}
  \begin{center}
    \includegraphics[width=0.35\textwidth]{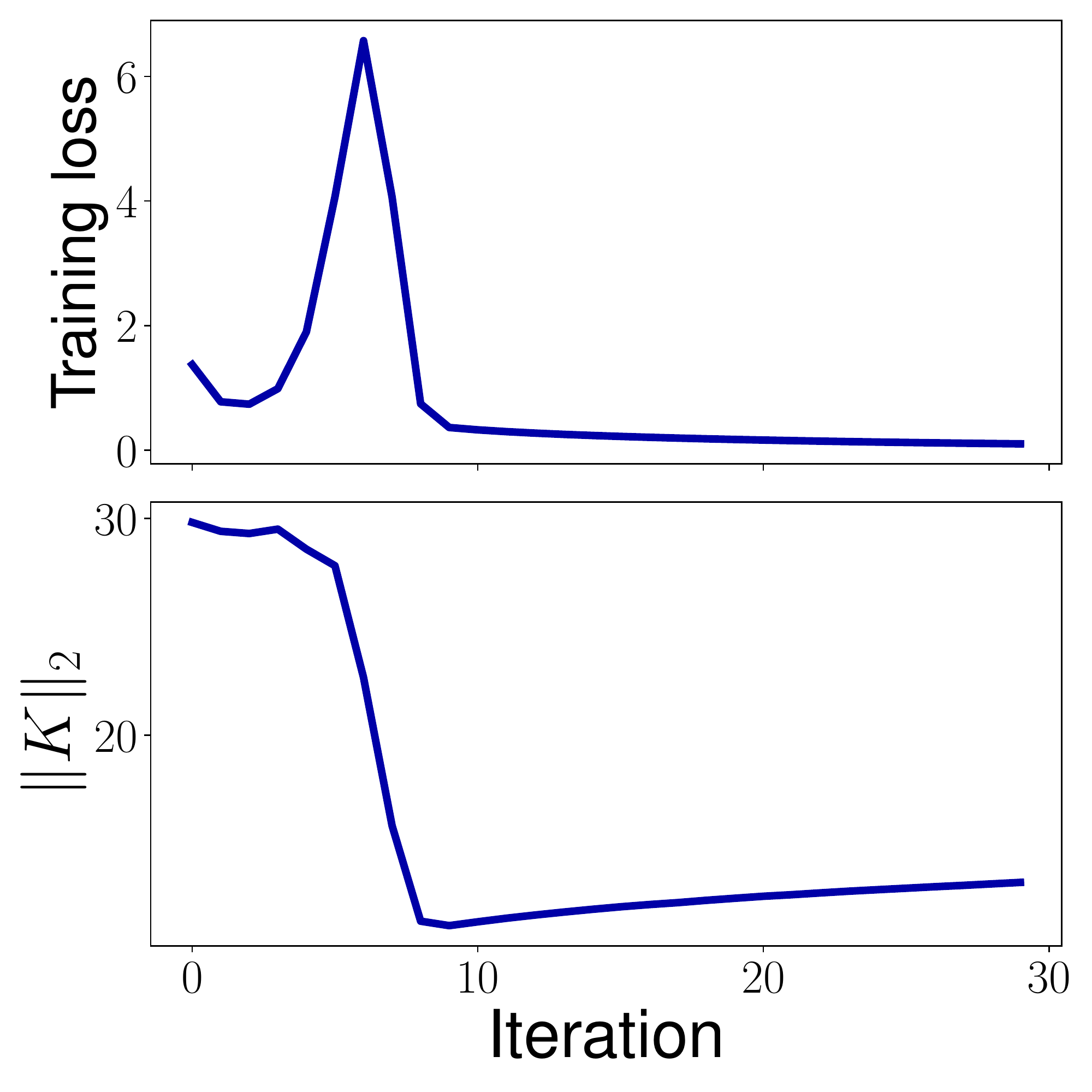}
  \end{center}\vspace{-15pt}
  \caption{An illustration of the catapult. This experiment corresponds to Fig.~\ref{fig:cata_fcn}a.}\label{fig:catapult_illu}\vspace{-35pt}
\end{wrapfigure}
\vspace{0pt}
\paragraph{Catapult dynamics.}
It was recently observed in~\cite{lewkowycz2020large} that, for wide neural network, full batch GD with a learning rate that is larger than $\etc$ (e.g., $\eta \in (\etc, 2\etc)$ as shown in \cite{lewkowycz2020large}) surprisingly ends up with a convergence. Instead of the expected divergence, the loss decreases after a drastic increase at the beginning stage of training, forming a loss spike (see Fig.~\ref{fig:catapult_illu}). Moreover, $\|K\|_2$ is observed to be smaller at the end of the spike.  Interestingly, the solution found by this large-learning-rate GD turns out to perform better in terms of test loss. Intuitively,  the decrease in $\|K\|_2$ raises the divergence threshold ${n}/{\|K\|_2}$ which allows a final convergence. 

In this paper, we refer {\it catapult dynamics} as the phenomenon of a drastic increase followed by a fast decrease in the training loss which is triggered by a learning rate larger than $\etc$ and accompanied by a decreasing $\|K\|_2$.

\section{Catapults in optimization}\label{sec:cata_op}

\subsection{Catapults occur in  the top eigenspace of the tangent kernel for GD}\label{sec:cata_gd}
% In this section, we show that catapults occur in the subspace spanned by the top eigen-directions of the tangent kernel.

The training dynamics of the machine learning model, e.g., a neural network, are closely related to its NTK $K^t:=K(\rvw^t;X,X)\in\mathbb{R}^{n\times n}$. Specifically, when the loss is optimized by gradient flow (continuous-time GD) with learning rate $\eta$, the output follows the dynamic equation~\cite{lee2019wide}:
\begin{align*}
    \frac{d\rvf^{t}}{dt} = -2\eta \frac{K^t}{n} (\rvf^t -\vy).
\end{align*}
By discrete time GD, this becomes 
% Similarly, for GD with learning rate $\eta$, we can use a first-order Taylor expansion with Lagrange remainder term to approximate neural networks at each time step: 
% \begin{align*}
%     \rvf^{t+1} = \rvf^t + \nabla_\rvw \rvf(\rvw^t)(\rvw^{t+1}-\rvw^t)+ \Delta_{H_\rvf^t},
% \end{align*}
% then $K^t$ appears in the update equation of GD by taking $\rvw^{t+1} - \rvw^t = -\frac{2\eta}{n}\nabla_\rvw\rvf(\rvw^t)^T  (\rvf^t - \vy)$:
\begin{align}\label{eq:gd_linear}
\rvf^{t+1} - \vy = \round{I_n - 2\eta\frac{K^t}{n}} (\rvf^t - \vy) + \Delta_{H_\rvf^t},
\end{align}
with $\Delta_{H_\rvf^t}:=\inner{\rvw^{t+1}-\rvw^t, \nabla_\rvw^2 \rvf(\xi)(\rvw^{t+1}-\rvw^t)}\in\mathbb{R}^n$ and $\xi = \tau\rvw^t + (1-\tau)\rvw^{t+1}, \tau \in(0,1)$.

Note that for finitely wide neural networks, $\norm{\Delta_{H_\rvf^t}}_2$ is small compared to the first term~\cite{papyan2019measurements,wang2022analyzing}  and is exactly zero for infinitely wide neural networks~\cite{lee2019wide}. Therefore,  the training dynamics of neural networks are mainly determined by the first term in R.H.S. of the above equation, which relies on the spectral information of the NTK $K^t$.
% {\color{blue}I am not sure whether it is the good place to mention the following related work. If yes, we should point out what did they do. Do they have the same results?}
This data-dependent NTK is also useful for understanding the generalization performance of neural networks~\cite{fort2020deep,atanasov2022neural,ortiz2021can,loo2022evolution}.

\begin{figure}[t!]
     \centering
     % \begin{subfigure}[b]{0.27\textwidth}
     %     \centering
     %     \includegraphics[width=\textwidth]{figure/loss_fcn_gd.pdf}
     %     \caption{Training loss(FCN)}
     % \end{subfigure}\hspace*{-0.9em}
          \begin{subfigure}[b]{0.4\textwidth}
         \centering
         \includegraphics[width=\textwidth]{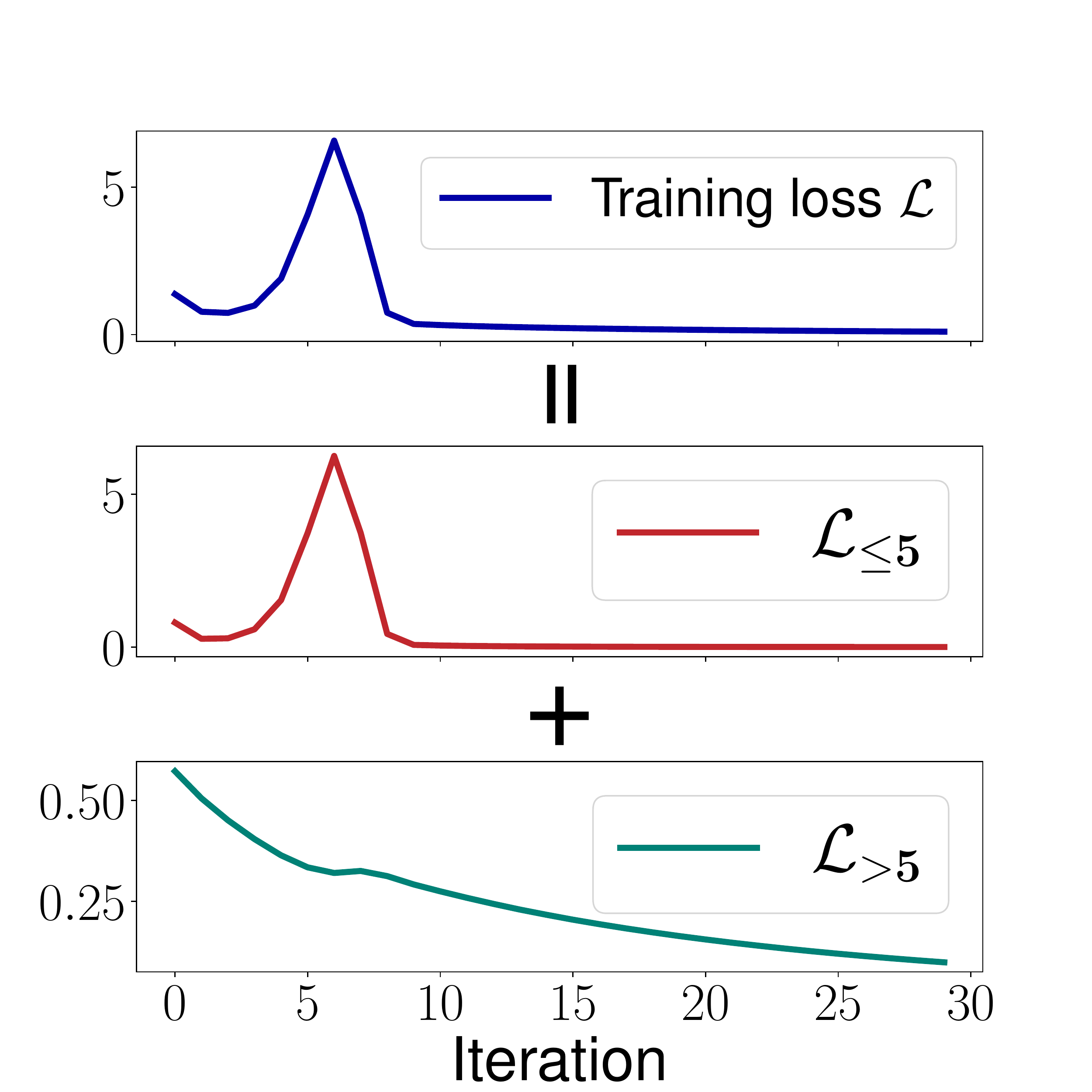}
         \caption{Loss decomposition(FCN)}
         \end{subfigure}
    % \begin{subfigure}[b]{0.27\textwidth}
    %      \centering
    %      \includegraphics[width=\textwidth]{figure/loss_cnn_gd.pdf}
    %     \caption{Training loss(CNN)}
    %  \end{subfigure}\hspace*{-0.9em}
          \begin{subfigure}[b]{0.4\textwidth}
         \centering
         \includegraphics[width=\textwidth]{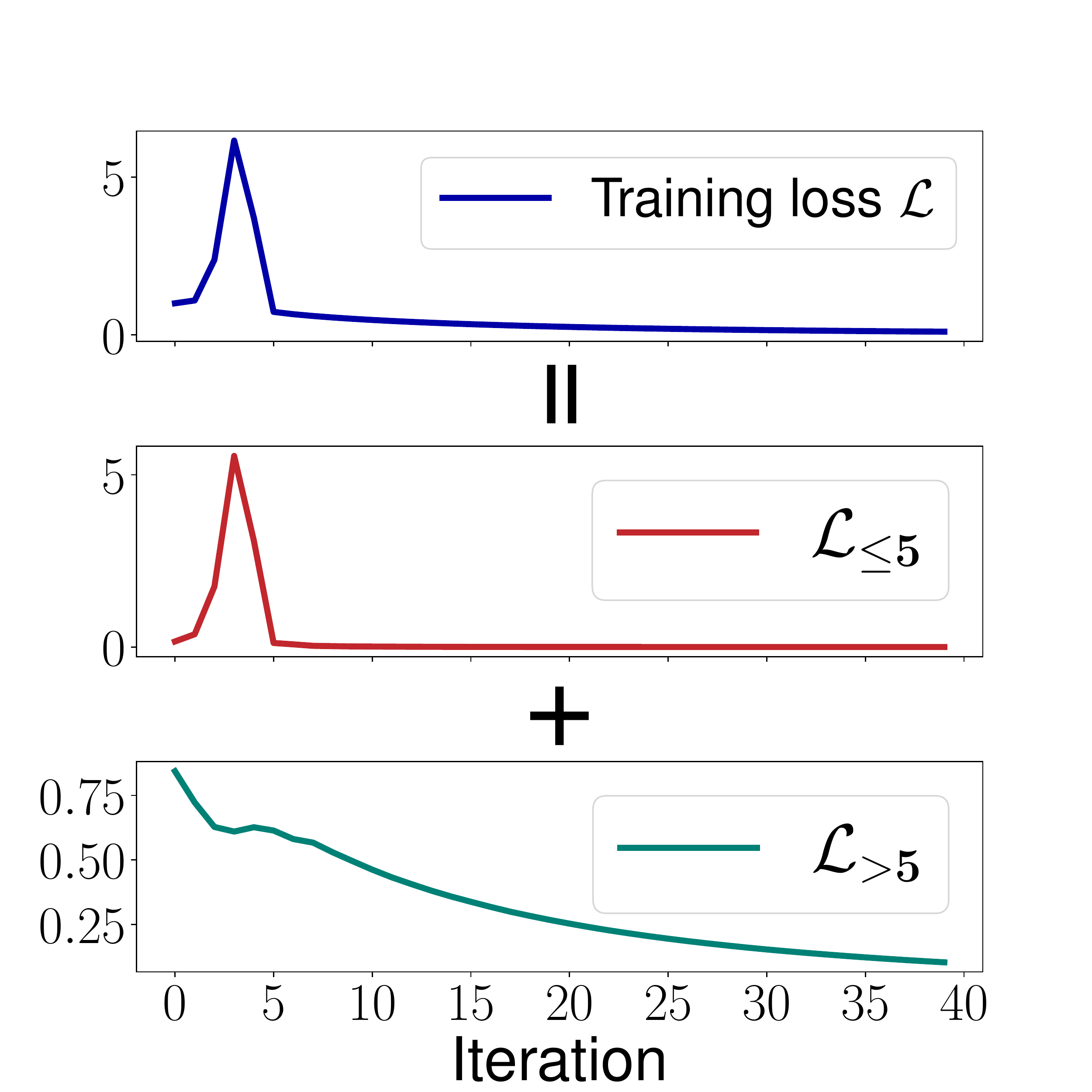}
          \caption{Loss decomposition(CNN)}
     \end{subfigure}
     \caption{{\bf  Catapult occurring in the top eigenspace of NTK in GD for 5-layer FCN (a) and CNN (b).} The training loss is decomposed into the eigenspace of NTK, i.e., $\L = \L_{\leq 5} + \L_{>5}$. In the experiment, both networks are trained by GD on  $128$ data points from CIFAR-10 with learning rate $6$ and $8$ respectively (the critical learning rates are $3.6$ for FCN and $4.5$ for CNN). }\label{fig:cata_fcn}\vspace{-10pt}
\end{figure}

% In works \cite{fort2020deep,loo2022evolution,atanasov2021neural,ortiz2021can}, they analyzed the evolution of the data-dependent NTK $K^t$ for understanding the generalization performance of neural networks. 
% Additionally, it is not hard to see that for SGD, a similar equation can be derived for each mini-batch.

% Here we omit the dependence of $\P\L$ and $\P\L^\bot$ on $K^t$ for the simplicity of notation. 
Consider decomposing Eq.~(\ref{eq:gd_linear}) into eigendirections of the NTK $K^t$, i.e., $\inner{\rvf^t-\vy,\rvu^t_i}$.  Supposing the dynamics among eigendirections are not interacting and $\rvu_i$ is constant, we expect that the increase of training loss during catapult occurs in the top few eigendirections where $\eta>n/{\lambda_i}$, while the loss corresponding to the remaining eigendirections remain decreasing. Indeed, this has been theoretically shown to be true on quadratic models that approximate wide neural networks~\cite{zhu2024quadratic}.
% We formulate our intuition as follows: 

\begin{claim}%\vspace{-10pt}
\label{claim: Claim 1}
 The catapult occurs in the top eigenspace of the tangent kernel: the loss component corresponding to the top-s eigenspace $\L_{\leq s}$ has a spike during the catapult, while the loss component in the complementary eigenspace $\L_{>s}$ decreases monotonically.
 % for the spike in the training loss when catapult occurs, $\L_{\leq s}$ corresponds to the loss spike while $\L_{>s}$ decreases monotonically, with a small $s$. 
 % the loss spike disappears in $\L_{{>s}}$ with a small $s$.
\end{claim}
\begin{remark} We note that the catapult does not occur in all eigendirections, as the learning rate $\eta$ cannot be arbitrarily large. Instead, there is a maximum learning rate $\etm$ such that if $\eta>\etm$ the algorithm will diverge. 
  % As analyzed above, the loss corresponding to the eigendirection that satisfies $\eta > n/\lambda_i$ will increase when trained with learning rate $\eta$. However, catapults cannot occur in all eigendirections.  For non-linear models, there is a maximum learning rate $\etm$ with which the optimization can stably converge, where $\etm= c\etc = cn/\lambda_1$ with $c>1$ dependent on the model architecture~\cite{iyer2023maximal}. 
  For instance, $\etm= 2\etc$ for quadratic models~\cite{zhu2024quadratic} and $\etm\approx 6\etc$ for ReLU networks~\cite{lewkowycz2020large}. Therefore, for any learning rate $\eta \in (\etc,\etm$) such that catapult occurs, only the top few eigendirections satisfy $ n/\lambda_i<\etm$. We consistently observe that $s$ is a small constant no larger than $10$ in all our experiments.   
\end{remark}

% , as indicated by the linear dynamics before the spike. Therefore, in general, we need a larger $s$ to capture the spikes in the training loss for larger learning rates. However, for $\eta\in(\etc,\etm)$ such that the GD can stably converge with catapults, we consistently observe that $s$ is a small constant no larger than $10$ in all our experiments.
% %\libin{namely a small constant compared to $n$ e.g., $s=5$ in Figure~\ref{fig:cata_fcn}}, 
% The remaining loss $\L_{\pi_s}^\bot$  decreases nearly monotonically with the number of iterations (See Figure~\ref{fig:cata_fcn}b and d). 

We empirically justify Claim~\ref{claim: Claim 1} for neural networks. In particular, we consider three neural network architectures: a 5-layer Fully Connected Neural Network (FCN), a 5-layer Convolutional Neural Network (CNN), and  Wide ResNets 10-10; and three datasets CIFAR-10, SVHN, and a synthetic dataset.  The details of experimental setup can be found in Appendix~\ref{sec:exp_details}. We present a selection of the results in Fig.~\ref{fig:cata_fcn} with the remaining results in Fig.~\ref{fig:cata_fcn_svhn} and \ref{fig:cata_wrn} in Appendix~\ref{sec:gd_add}. We can see that $\L_{{\leq 5}}$ corresponds to the spike in the training loss while $\L_{{>5}}$ decreases almost monotonically. 
 Concurrently with this study, \cite{zhang2023loss} showed that the loss spike in GD is primarily due to the low-frequency component, corroborating our findings through a frequency perspective.

We note that the same phenomenon holds for multidimensional outputs. 
% In particular, for $k$-class classification tasks, we project the flattened vector of predictions of size $kn$ to the top eigenspaces of the empirical NTK, which is of size $kn \times kn$.
% Correspondingly, we empirically observe that catapults occurs in the top $ks$ eigenspace with a small $s$. 
See more details in  Fig.~\ref{fig:cata_fcn_full} in Appendix~\ref{sec:gd_add}.
%  \begin{wrapfigure}{r}{0.4\textwidth}
% \vspace{-20pt}
%   \begin{center}
%     \includegraphics[width=0.3\textwidth]{figure/multi_toy_loss}
%   \end{center}\vspace{-10pt}
%   \caption{{\bf The mechanism of multiple catapults in GD.} Extra spike in the training loss occurs when the learning rate is increased to satisfy $\eta>\etc$. 
%   % We train a shallow network on a subset of CIFAR-10 with increased learning rates.
%   } \label{fig:shallow_multi_cata}\vspace{-40pt} 
% \end{wrapfigure}
% \adit{Libin can you double check this last sentence?}.

% (add experiments to estimate $s$)

\begin{figure}[h]
     % \centering\hspace*{-2em}
     % \begin{subfigure}[b]{0.25\textwidth}
     %     \centering
     %     \includegraphics[width=\textwidth]{figure/multi_toy_loss.pdf}
     %     \caption{Shallow network}
     % \end{subfigure}\hspace*{1em}
     % \begin{subfigure}[b]{0.27\textwidth}
     %     \centering
     %     \includegraphics[width=\textwidth]{figure/multi_toy_proj.pdf}
     %     \caption{Shallow network}
     % \end{subfigure}\hspace*{-0.9em}
     \vspace{-10pt}
     \centering
     \begin{subfigure}[b]{0.4\textwidth}
         \centering
         \includegraphics[width=\textwidth]{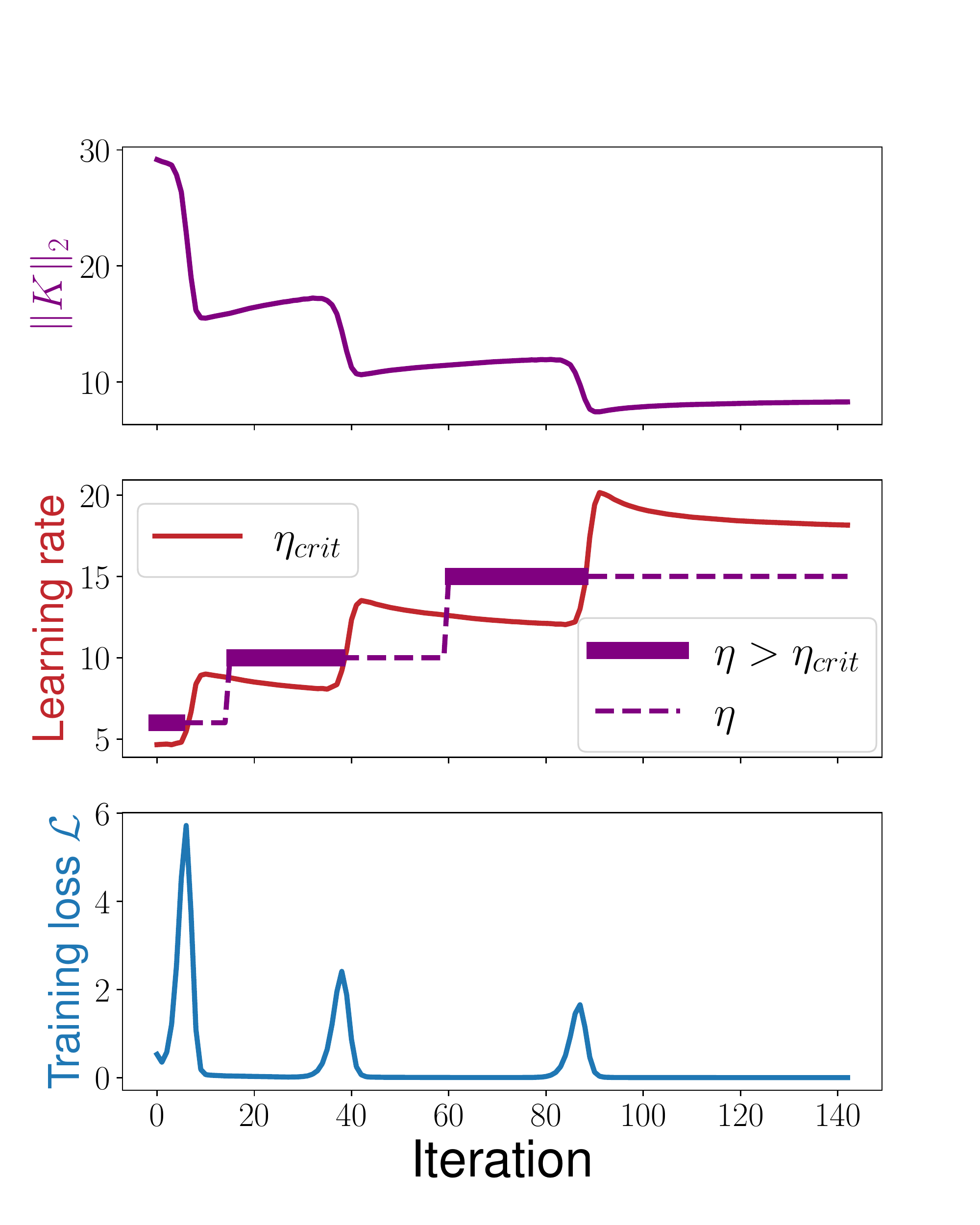}
         \caption{5-layer FCN}
     \end{subfigure}
     \begin{subfigure}[b]{0.4\textwidth}
         \centering
         \includegraphics[width=\textwidth]{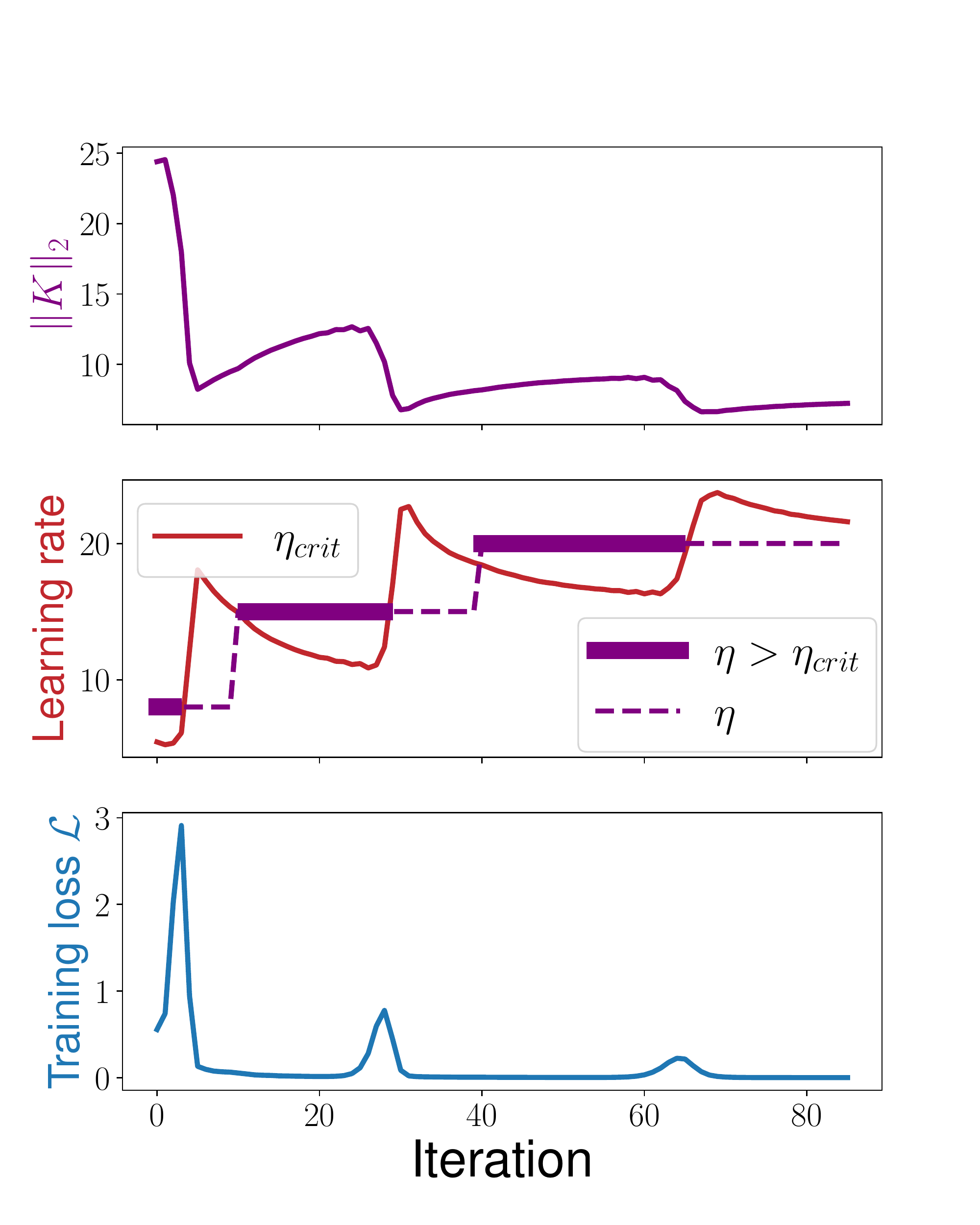}
         \caption{5-layer CNN}
     \end{subfigure}\hspace*{1em}
     \caption{{\bf Multiple catapults during GD with increased learning rates}. We train a 5-layer FCN and CNN on a subset of CIFAR-10 using GD. The learning rate is increased two times for each experiment. The experimental details can be found in Appendix~\ref{exp:muli_cata}.}
     \label{fig:multi_cata}
\end{figure}

\subsection{Inducing multiple catapults in GD}\label{subsec:multi_cata}

While prior work showed a single catapult during training with gradient descent~\citep{lewkowycz2020large,zhu2024quadratic,kalra2023phase}, we present that catapults can be induced multiple times by repeatedly increasing the learning rate during training. 
 % As a result, $\|K\|_2$ will accordingly decrease multiple times.
% We also observe that the test performance will be improved with multiple catapult phases.

% ( allows a larger leearning rate...?, intuition?? why???)
Specifically, during a catapult, the norm of NTK $\norm{K}_2$ decreases, which leads to an increase in the critical learning rate $\etc\approx n/\norm{K}_2$, see Fig.~\ref{fig:multi_cata}. When the loss starts to decrease during a catapult, $\etc$ surpasses the current learning rate $\eta$ of the algorithm. Hence, after each catapult, one can reset the algorithmic learning rate $\eta$ to be greater than the current $\etc$ to trigger another catapult. 
% Note that during the later catapults the norm of NTK $\norm{K}_2$ further decreases, which allows the possibility of inducing more catapults afterward.
In practice, we observe that a sequence of catapults can be triggered by repeating the above procedure. See Fig.~\ref{fig:multi_cata} for a demonstration of various neural network architectures. 

Interestingly, with multiple catapults, the gradient descent can ultimately  
converge with a much larger learning rate, which leads to a divergence, instead of a catapult, if set as the initial learning rate of gradient descent (see Fig.~\ref{fig:multi_cata_diverge} in Appendix~\ref{app:multi_cata}). Furthermore, thanks to the relation $\etc\approx n/\norm{K}_2$, this indicates that the multiple catapults achieve a much smaller  $\norm{K}_2$ which can not be obtained in the scenario of a single catapult. See Fig.~\ref{fig:multi_cata} for an experimental demonstration.
 Moreover, the multiple catapults lead to better generalization performance than a single catapult. We defer this discussion of generalization performance to Section~\ref{sec:feature_learning}.

\subsection{Catapults in SGD}\label{sec:cata_sgd}

% In this section, we leverage the mechanism of catapults in GD to demonstrate the occurrence of catapults in batches of SGD, which manifest as spikes in the training loss.
In this section, we consider the stochastic setting, and argue that the spikes often observed in the training loss of SGD (e.g., Fig.~\ref{fig:sgd_spikes_wiki}) are in fact {\it catapults}.

% \chaoyue{I think it is better to have a separate (sub)subsection for each evidence.}
% In this section, we provide evidence that catapults also occur in SGD and manifest as spikes in the training loss. Specifically, we leverage the mechanism of catapults in GD to demonstrate the occurrence of catapults in batches of SGD, and  we further show that the empirical phenomenon observed in the catapults of GD occurs in SGD.
% Specifically, we leverage the mechanism of catapults in GD to demonstrate the occurrence of catapults in batches of SGD, since an SGD step can be viewed as a GD step on a batch. We then empirically show that when training using SGD with a large learning rate,  we can consistently observe spikes in the training loss, and the spikes mainly exist in a low-dimensional subspace, i.e., the top eigenspace of the tangent kernel, which indicates that loss spikes are catapults.
% \vspace{-5pt}

% Corresponding to Figure~\ref{fig:sign_cata} with learning rate $0.8$ where we observe catapults in SGD, we show the loss decomposition for the training loss in Figure~\ref{fig:sign_cata_app}.

% \begin{figure}[H]
     
%      \caption{{\bf The spectral norm of NTK and the loss decomposition during catapults for shallow network.} This figure corresponds to Figure~\ref{fig:sign_cata} with learning rate $0.8$. The training loss is decomposed based on the eigendirections of the NTK: $\L_{\leq 5}$ and $\L_{>5}$.\label{fig:sign_cata_app}}
% \end{figure}

\paragraph{Mechanism of catapults in SGD.}
Recall that the catapults are triggered when $\eta>\etc$. Unlike in deterministic gradient descent, the mini-batch stochastic training dynamics is determined by the NTK matrix evaluated on the given batch $X_{\batch}$. Specifically, the update equation of mini-batch SGD becomes (c.f. Eq.(\ref{eq:gd_linear}) of GD):
\begin{align}\label{eq:gd_linear_batch}
\rvf^{t+1}_{\batch} - \vy_{\batch} = &\round{I_b -  2{\eta}\frac{K_t(X_{\batch})}{b}}(\rvf^t_{\batch} - \vy_{\batch})+ \Delta_{H_{\rvf_\batch^t}},
\end{align}
where $b$ is the mini batch size, $\rvf_{\batch}:=\rvf(X_\batch)$ and $\vy_\batch$ is the label corresponds to $X_\batch$.
It is important to note that in mini-batch SGD the critical learning rate $\etc(X_{\batch})$ becomes batch dependent: for batches that have relatively large (small, respectively) $\norm{K_{\batch}}_2$, the corresponding critical learning rate $\etc(X_{\batch})$ is relatively small (large, respectively). Then,  if  $\etc(X_\batch)$ of a given batch is smaller than the algorithmic learning rate $\eta$ of SGD, we expect a   catapult will happen: an increase in the running training loss.

% Hence, we expect that catapults will occur when the model is trained with mini-batch SGD as long as the algorithmic learning rate is not smaller than $\min_\batch \etc(X_{\batch})$. 

\begin{figure}[htb!]
    \centering
    \includegraphics[width=0.5\textwidth]{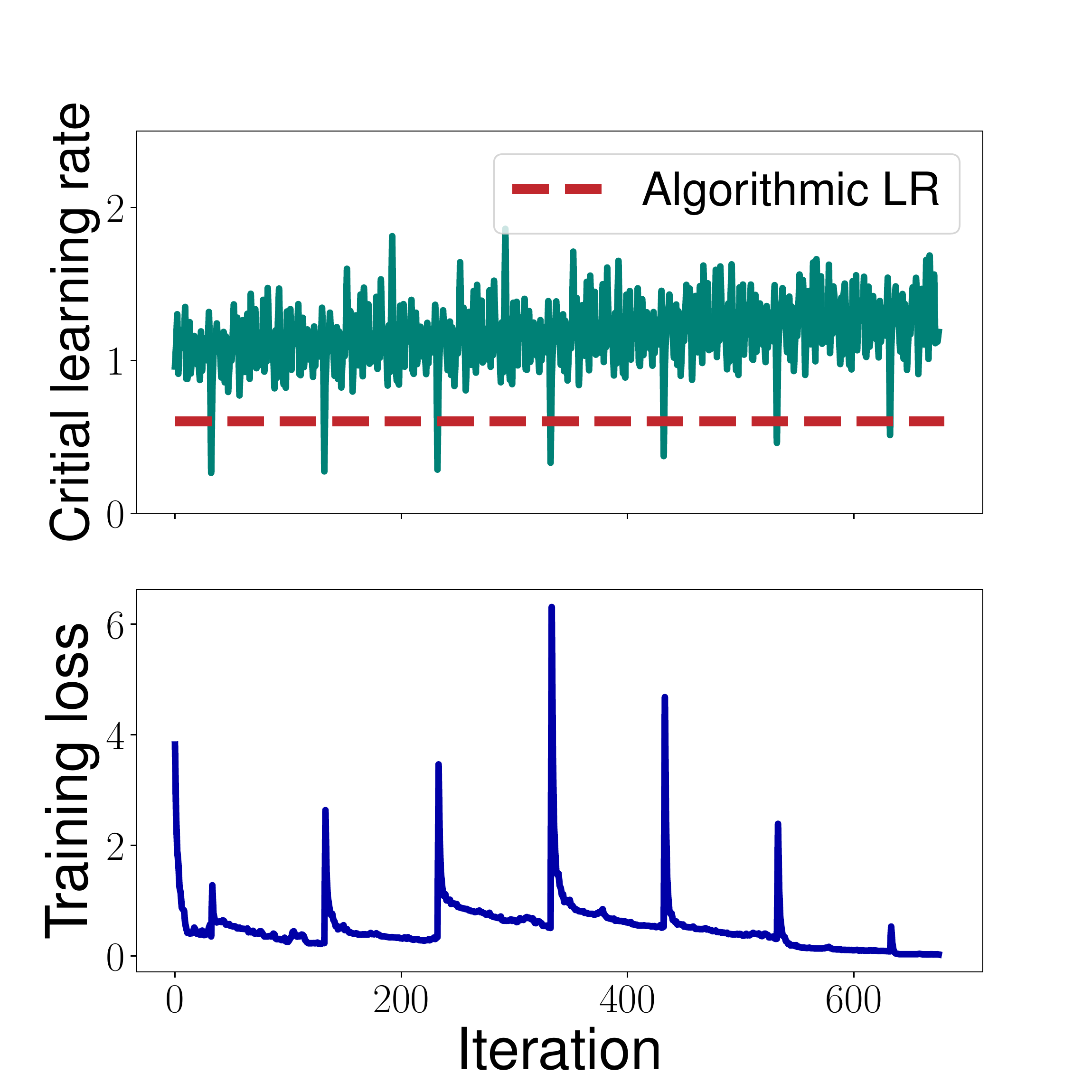}
    \caption{\textbf{Exact match between the occasion when $\eta>\etc(X_\batch)$ and loss spike for SGD.} We train a two-layer neural network on a synthetic dataset using SGD with batch size one.}
    \label{fig:loss_lr_match_sgd}
\end{figure}

Indeed, this expectation is confirmed in our experiments. Specifically, we train the network on a synthetic dataset with SGD and consider batch size one.  We set the algorithmic learning rate higher than the critical learning rate for only one training example. As expected, we observe that the loss spikes only occur when the gradient is computed based on that particular training example. See the result in Fig.~\ref{fig:loss_lr_match_sgd} and the detailed experimental setup in Appendix~\ref{exp:cata_sgd}.

\begin{figure}[b!]
     \centering
     \captionsetup[subfigure]{justification=centering}
          \begin{subfigure}[b]{0.5\textwidth}
         \centering
         \includegraphics[width=\textwidth]{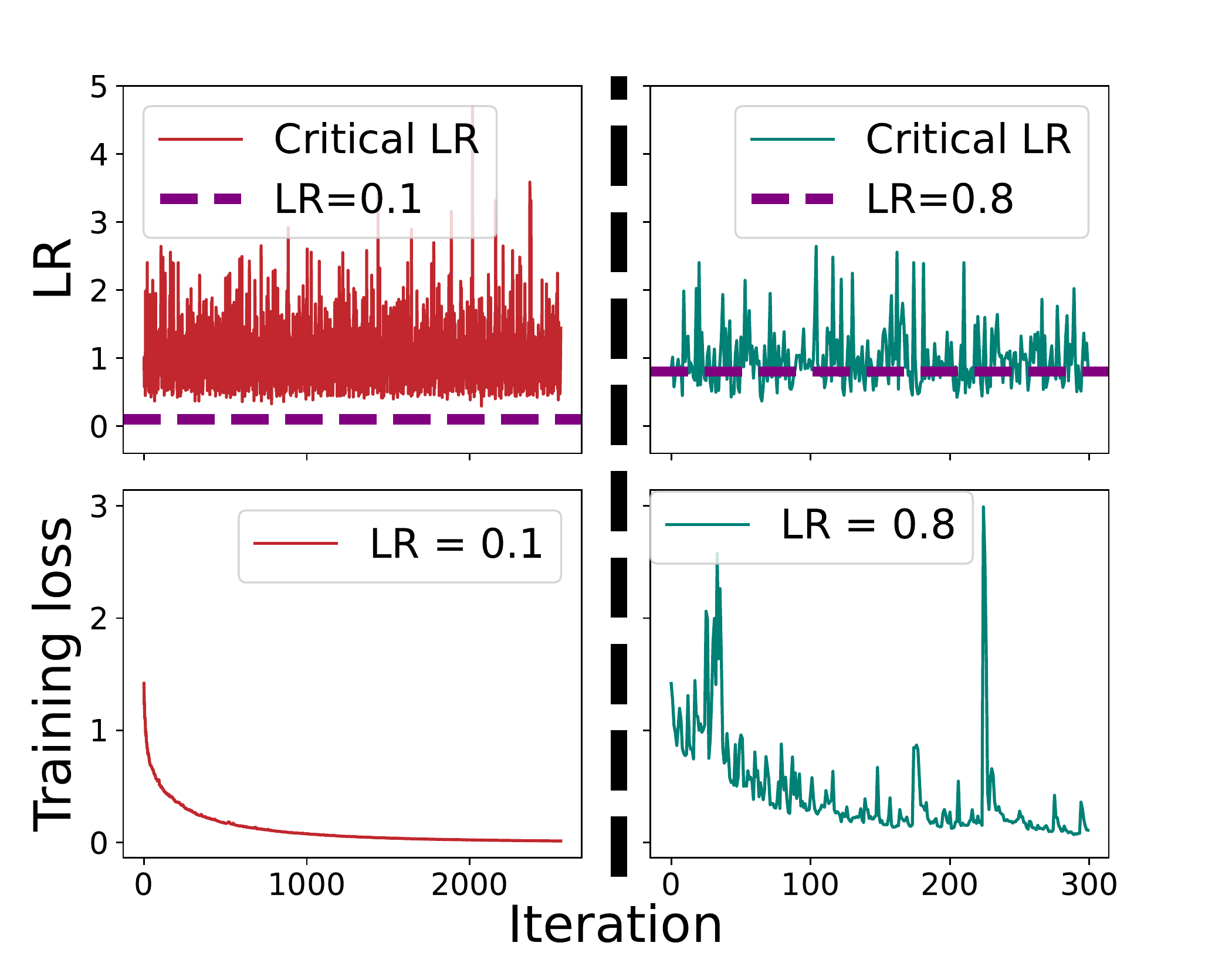}
         \caption{Critial learning rates and training loss}
     \end{subfigure}
               \begin{subfigure}[b]{0.4\textwidth}
         \centering
         \includegraphics[width=\textwidth]{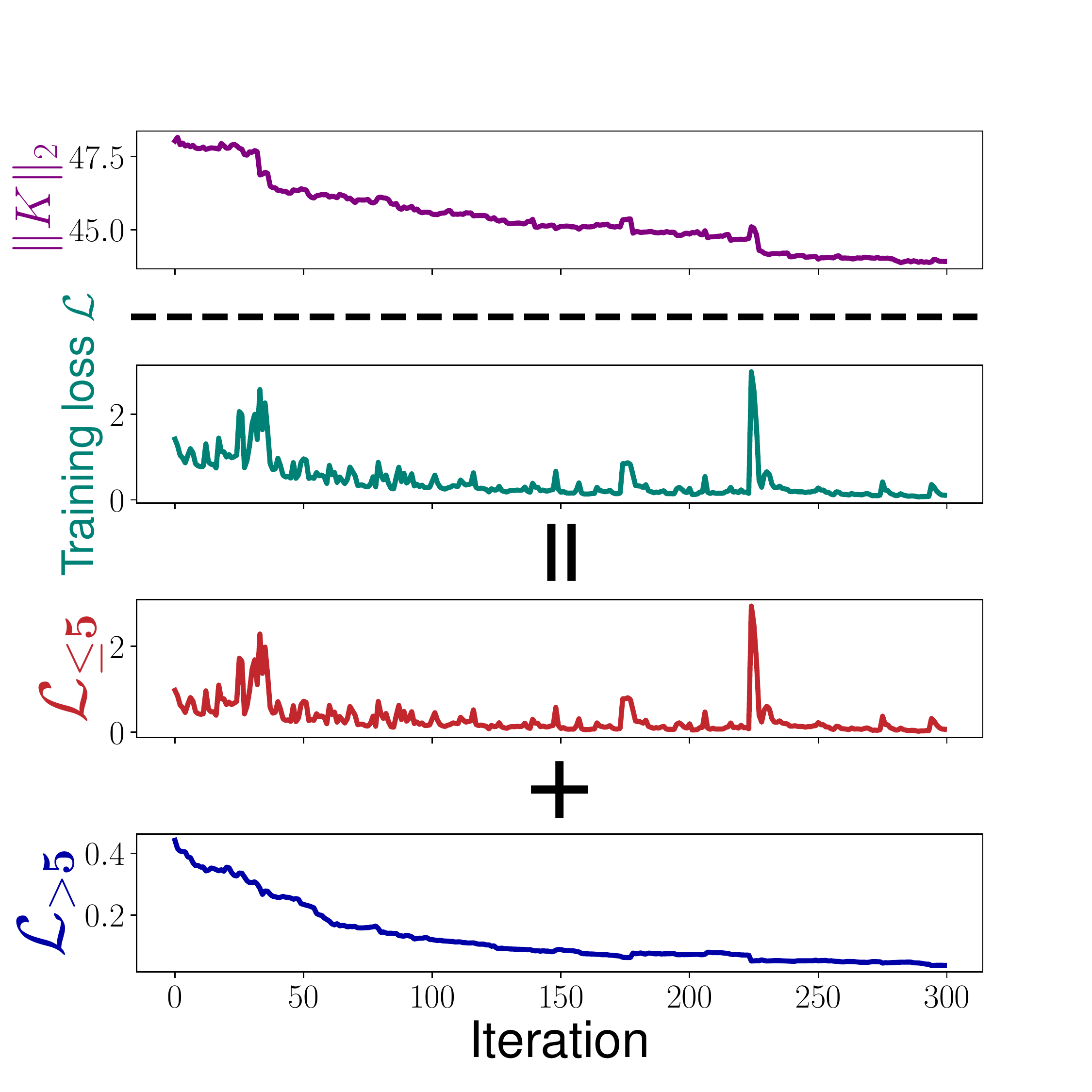}
         \caption{Loss decomposition with $\eta = 0.8$}
         % \caption{Loss decomposed into $\L_{\leq 5}$ and $\L_{> 5}$ with $\eta = 0.8$}
     \end{subfigure}
     \caption{{\bf Panel(a): Critical learning rates of batches and the training loss of SGD with learning rate $0.1$ (left two subfigures) and $0.8$ (right two subfigures). Panel(b): Loss decomposition with $\eta = 0.8$. } LR is an abbreviation for learning rate.  We train a two-layer neural network on $128$ data points of CIFAR-10 using SGD with batch size $32$. We further decompose the loss into the top-5 eigendirections of the NTK $\L_{\leq 5}$ and the remaining eigendirections $\L_{> 5}$ corresponding to $\eta = 0.8$. }\label{fig:sign_cata}
\end{figure}

In more practical scenarios, we train a shallow network by SGD with mini-batch size $32$, on a subset of CIFAR-10 with training size $128$. First, when the algorithmic learning rate $\eta$ is smaller than $\etc(X_{\batch})$ of all the batches (as shown in the case of $\eta=0.1$ in (Fig.~\ref{fig:sign_cata} upper left)), we observe that the training loss of mini-batch SGD monotonically decreases until convergence without any spike; when $\eta$ becomes greater than $\etc(X_{\batch})$ for some of the batches (as shown in the case of $\eta=0.8$ in Fig.~\ref{fig:sign_cata} upper right), many spikes appear in the training loss. Moreover, we show that these spikes in the (total) training loss are caused by large learning rates for batches. Specifically,  for the case of $\eta = 0.8$, we verify that whenever the (total) training loss increases, the algorithmic learning rate $\eta$ is larger than the critical learning rate $\etc(X_{\batch})$ for the current batch $X_{\batch}$. This phenomenon is further verified for 5-layer FCN and CNN. See Table \ref{tab:match_rate}.

\begin{table}[t!]
\centering
{
\begin{tabular}{cc}\\\toprule  
Network Architecture& \parbox[t]{5cm}{Match rate between $\Delta\L >0$\\ and  $\eta>\eta_{\mathrm{crit}}(X_\batch)(\%)$   }   \\ \midrule\midrule
Shallow network & $97.32\pm 0.45$\\  \midrule
5-layer FCN & $96.17\pm 1.46$\\\midrule
5-layer CNN  & $94.67\pm 3.27$ \\  \bottomrule
\end{tabular}
}
\caption{ \textbf{The match rate between $\Delta\L^t := \L^{t+1}-\L^t >0$ and $\eta>\eta^t_{\mathrm{crit}}(X_\batch)$}. For each network architecture, we calculate the match rate as the ratio of occurrences where $\eta > \eta^t_{\mathrm{crit}}(X_{\mathrm{batch}})$ for all $t$ such that $\L^{t+1} > \L^t$ until convergence of SGD (see the training loss in Fig.~\ref{fig:sign_cata}(c) for shallow net and Fig.~\ref{fig:sgd_deep} (a,b) for deep nets). Each result is the average of 3 independent runs. \label{tab:match_rate}}

\end{table}

\paragraph{Decreases in the spectral norm of the tangent kernel correspond to spikes.} As shown in prior work \cite{lewkowycz2020large} and in the multiple catapults in Section \ref{subsec:multi_cata}
an important characterization of the catapult dynamics is the decreasing NTK norm $\norm{K}_2$.
Here, we experimentally show that the spectral norm of the NTK decreases whenever there is a spike in the SGD training loss.

\begin{figure}[t!]
\centering
\hspace*{-4em}
\includegraphics[width=1.2\textwidth]{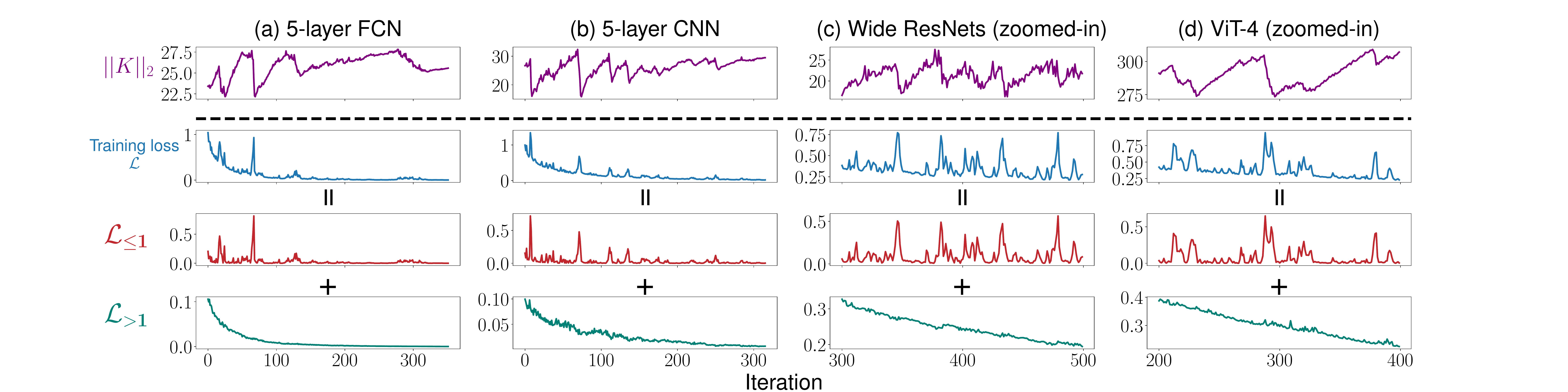}
\caption{{\bf Catapult dynamics in SGD for modern deep architectures.} The training loss is decomposed based on the eigendirections of the NTK: $\L_{\leq 1}$ and $\L_{>1}$.  We train the networks on a subset of CIFAR-10 using SGD. The complete versions of Panel (c) and (d) can be found in Fig.~\ref{fig:sgd_deep_complete} in Appendix~\ref{sec:sgd_add}.}\label{fig:sgd_deep}
\hspace*{-3em}
\end{figure}

Specifically, we consider four network architectures: (1) 5-layer FCN, (2) 5-layer CNN (the same as the ones in Fig.~\ref{fig:cata_fcn}), (3) Wide ResNets 10-10 and (4) ViT-4. We train neural networks on a subset of CIFAR-10 using SGD.   Fig.~\ref{fig:sgd_deep} shows some of the results (more results on various datasets and parameterizations are available in Appendix~\ref{sec:sgd_add}). One can easily see that at each spike of the training loss, there is a significant drop in the spectral norm of NTK $\norm{K}_2$, while $\norm{K}_2$ are mostly increasing or staying unchanged at other steps. This empirical evidence corroborates that these spikes are indeed (mini-)catapults, instead of some random fluctuations in the training loss. All experimental details can be found in Appendix~\ref{sec:exp_details}.

% In~\cite{lewkowycz2020large}, the catapult led to a decrease in the spectral norm of the tangent kernel in GD, and our experimental results in Section~\ref{sec:cata_gd} extended this finding to settings with multiple catapults in GD. We observe a consistent phenomenon in SGD.  
% Specifically, in our experiments with shallow networks, we observe a significant decrease in the spectral norm of the tangent kernel whenever there is a spike in the training loss during SGD (Figure~\ref{fig:sign_cata}c). This observation remains consistent for deep networks (Figure~\ref{fig:sgd_deep}) as well.  Therefore, this finding further justifies that spikes in the training loss of SGD are catapults. 
% We further validate our empirical observations on additional datasets in Appendix~\ref{sec:sgd_add}. All experimental details can be found in Appendix~\ref{sec:exp_details}.

\paragraph{Catapults occur in the top eigenspace of the tangent kernel for SGD.}
As discussed in Section~\ref{sec:cata_gd}, another characteristic of the catapults is that they occur in the top eigenspace of the tangent kernel. We show that these loss spikes in SGD also occur in the top eigenspace, as another evidence that these spikes are catapults.

% Extending the findings for catapults in GD, we now show that (1) spikes during SGD arise as a result of the learning rate being larger than the critical learning rate and (2) the spikes in the training loss of SGD also exist in the top eigenspace. 

In the experiments, we decompose the training loss of SGD into $\L_{\leq 1}$ and $\L_{> 1}$ based on the eigendirections of the tangent kernel. We observe that $\L_{\leq 1}$ corresponds to the spikes in the training loss, while the decrease of $\L_{> 1}$ is nearly monotonic, with only small oscillations present. See  Fig.~\ref{fig:sign_cata}b for the shallow network with $\eta = 0.8$  and Fig.~\ref{fig:sgd_deep} for deep networks. Note that for deep neural networks,  compared to the catapults in GD where they occur in the top-$5$ eigendirections of the NTK (Fig.~\ref{fig:cata_fcn}), we consistently observe that for SGD, catapults occur {only} in the top-1 eigendirection.   Additional empirical validation can be found in Appendix~\ref{sec:sgd_add}.

% In the experiment for shallow networks, we similarly decompose the loss with $\eta=0.8$ into  $\L_{\leq 5}$ and $\L_{> 5}$ based on the eigenspace of the tangent kernel.  We observe that $\L_{\leq 5}$ corresponds to the spikes in the training loss, while the decrease of $\L_{\leq 5}$ is nearly monotonic, with only small oscillations present (Figure~\ref{fig:sign_cata}c).
% We also observe the same phenomenon for deep networks as shown in Figure~\ref{fig:sgd_deep}. Specifically, we observe that  $\L_{\leq 1}$ corresponds to the spikes in the training loss while the remaining loss $\L_{\leq 1}$ decreases nearly monotonically. 

% This observation, along with the results in Figure~\ref{fig:sgd_deep} for deep networks, is consistent with our findings for GD and provides evidence that the spikes in training loss for neural networks are caused by catapults. 

This observation, along with the results that the spectral norm of the NTK decreases corresponding to the loss spike, is consistent with our findings for GD and provides evidence that the spikes in training loss for neural networks are caused by catapults.

\begin{remark}[Top eigenspace accounts for the sharp loss spikes in SGD]
In SGD training loss, the sharp spikes we observe last only a few iterations before rapidly returning to their pre-spike levels. These spikes can be attributed to catapults occurring in the top-1 eigendirection of the tangent kernel. Consider the loss change in each eigendirection of the tangent kernel.  We expect that the rate of loss change in each eigendirection depends on the corresponding eigenvalue's size. Therefore, with a constant learning rate, changes happen faster in the top eigendirections, which accounts for the sharp loss spikes in SGD as they occur in the top-1 eigendirection.
\end{remark}

\begin{remark}[Catapults in SGD with cyclical learning rate schedule]
    Training neural networks with the learning rate cyclically varying between selected boundary values was widely shown to improve the generalization performance of neural networks with less tuning~\cite{izmailov2018averaging,smith2017cyclical}. We empirically show that the increasing phase of the cyclical learning rate schedule induces catapults in SGD.
    Specifically, we observe that there is a spike in the training loss when the learning rate is increased. 
    We demonstrate that the loss spikes are caused by catapults, by providing similar evidence to the case of SGD with a constant learning rate.  See the results in Fig.~\ref{fig:cyc_lr} in Appendix~\ref{subsec:cyclic}. 
\end{remark}

\section{ Catapults lead to better generalization through feature learning}\label{sec:feature_learning}

Previous empirical results from~\cite{lewkowycz2020large,zhu2024quadratic} show that a single catapult can lead to better test performance in GD for wide neural networks. In this section, we observe a similar trend in our experiments for both GD and SGD with multiple catapults.  We posit an explanation for this phenomenon by demonstrating that catapults improve feature learning by increasing alignment between the Average Gradient Outer Products (AGOP) of the trained network and the true model, therefore improving generalization.  We formalize this claim as follows.  Let $\{(x_i, f^*(x_i))\}_{i=1}^{n}$ denote training data with $f^*(x)$ denoting the true model.  Then, for any predictor $f$, the AGOP, $G(f, \{x_1, \ldots, x_n\})$ is given as follows:     
\begin{align}
\label{eq:agop}
    G(f, \{x_1, \ldots, x_n\}) = \frac{1}{n} \sum_{i=1}^{n} \nabla_x f(x_i) \nabla_x f(x_i)^T~;
\end{align}
where $\nabla_x f$ denotes the gradient of $f$ with respect to the input $x$.\footnote{For predictors with multivariate outputs, we consider the Jacobian instead of the gradient.}  We will suppress the dependence on the data $\{x_i\}_{i=1}^{n}$ to simplify notation.  Assuming the data $x_i$ are i.i.d. samples from an underlying data distribution, in the limit as $n \to \infty$, Eq.~\eqref{eq:agop} converges to a quantity referred to as the Expected Gradient Outer Product (EGOP).  Letting $G^*$ denote the EGOP of $f^*$ and $G$ denote the AGOP of $f$, we define \textit{AGOP alignment} using the cosine similarity between $G, G^*$ as follows:
\begin{align}\label{eq:agop_align}
 \mathbf{AGOP~alignment:}\cos(G,G^*) :=  \frac{\Tr(G^T G^*)}{\norm{G}_F\norm{G^*}_F}.
\end{align}

\begin{remark}
$G^*$ captures the directions along which $f^*$ varies the most and those along which it varies least.  When training a predictor on data generated using low rank $G^*$, it is possible to improve sample efficiency by first estimating $G^*$.  Indeed, this has been theoretically shown in the case of multi-index models, i.e., functions of the form $f^*(\vx) = g(U\vx)$ where the index space $U$ is a low-rank matrix~\cite{hardle1989investigating, trivedi2014consistent, yuan2023efficient}.  Additionally, a recent line of work connected AGOP with feature learning in neural networks and further demonstrated that training predictors on data transformed by AGOP can lead to substantial increases in test performance~\cite{radhakrishnan2024mechanism,beaglehole2023mechanism, radhakrishnan2024linear}.  Thus, we believe that AGOP alignment is a key measure for generalization, and we next corroborate our claim empirically across a broad class of network architectures and prediction tasks. 
\end{remark}

\paragraph{Experimental settings.} We work with a total of seven datasets: three synthetic datasets and four real-world datasets. For synthetic datasets, we consider  true functions $f^*(\vx)=$  $(1) x_1x_2$ (rank-2),  $(2) x_1x_2(\sum_{i=1}^{10}x_i)$ (rank-3) and $(3) \sum_{j=1}^4 \prod_{i=1}^j x_i$ (rank-4)~\citep{abbe2021staircase}.  For the four real-world datasets, we use (1) CelebA~\citep{liu2015faceattributes}, (2) SVHN dataset~\citep{netzer2011reading}, (3) Fashion-MNIST~\cite{xiao2017fashion} and (4) USPS dataset~\cite{uspsdataset}.  When the underlying model is not available, we use a state-of-the-art model as a substitute. We present the results for a selection of the datasets in this section and put the results for the remaining datasets in Appendix~\ref{sec:add_data}.
 The experimental details can be found in Appendix~\ref{sec:exp_details}.

\begin{figure}[H]
     \centering
     \begin{subfigure}[b]{0.35\textwidth}
         \centering
         \includegraphics[width=\textwidth]{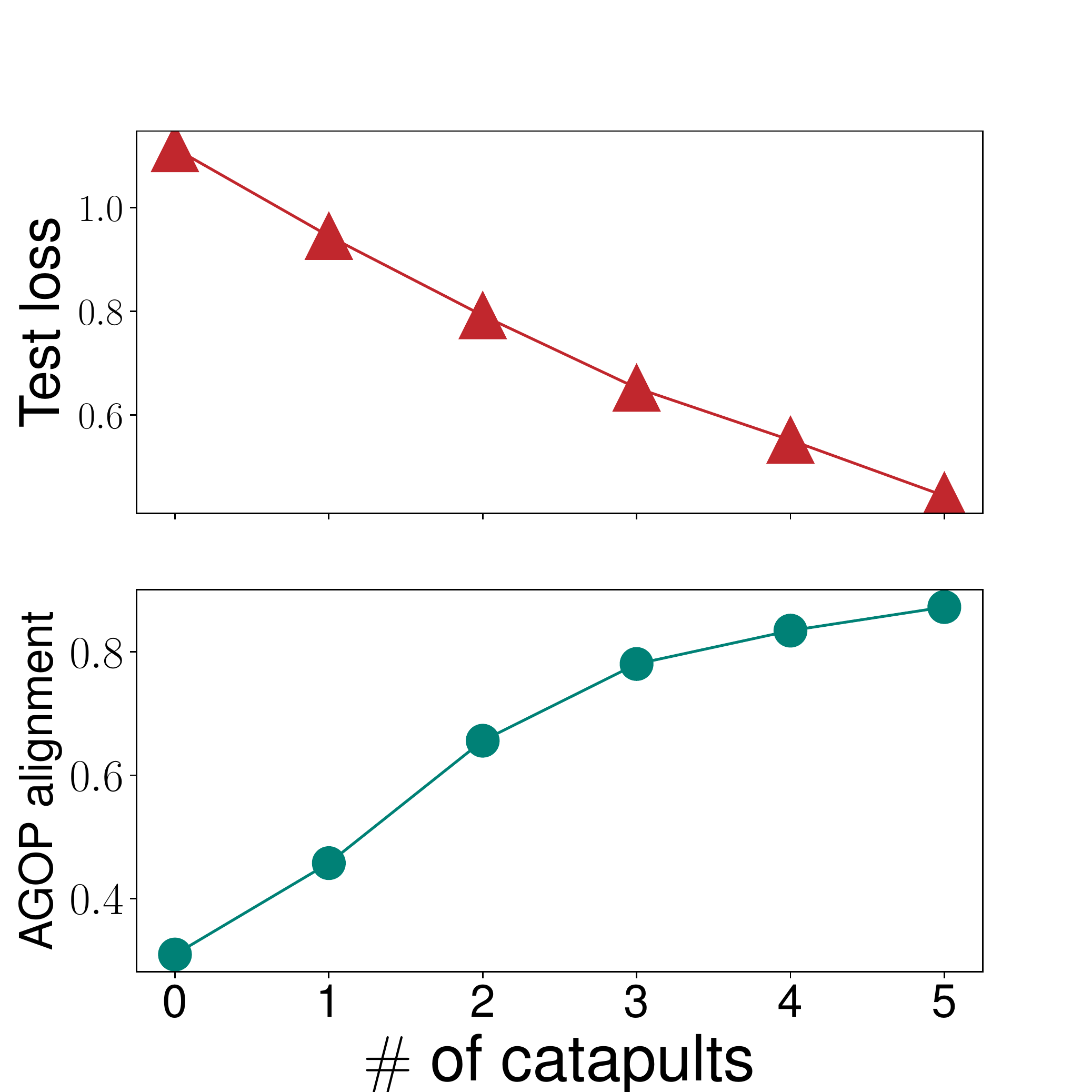}
         \caption{Rank-2 regression}
     \end{subfigure}
          \begin{subfigure}[b]{0.35\textwidth}
         \centering
         \includegraphics[width=\textwidth]{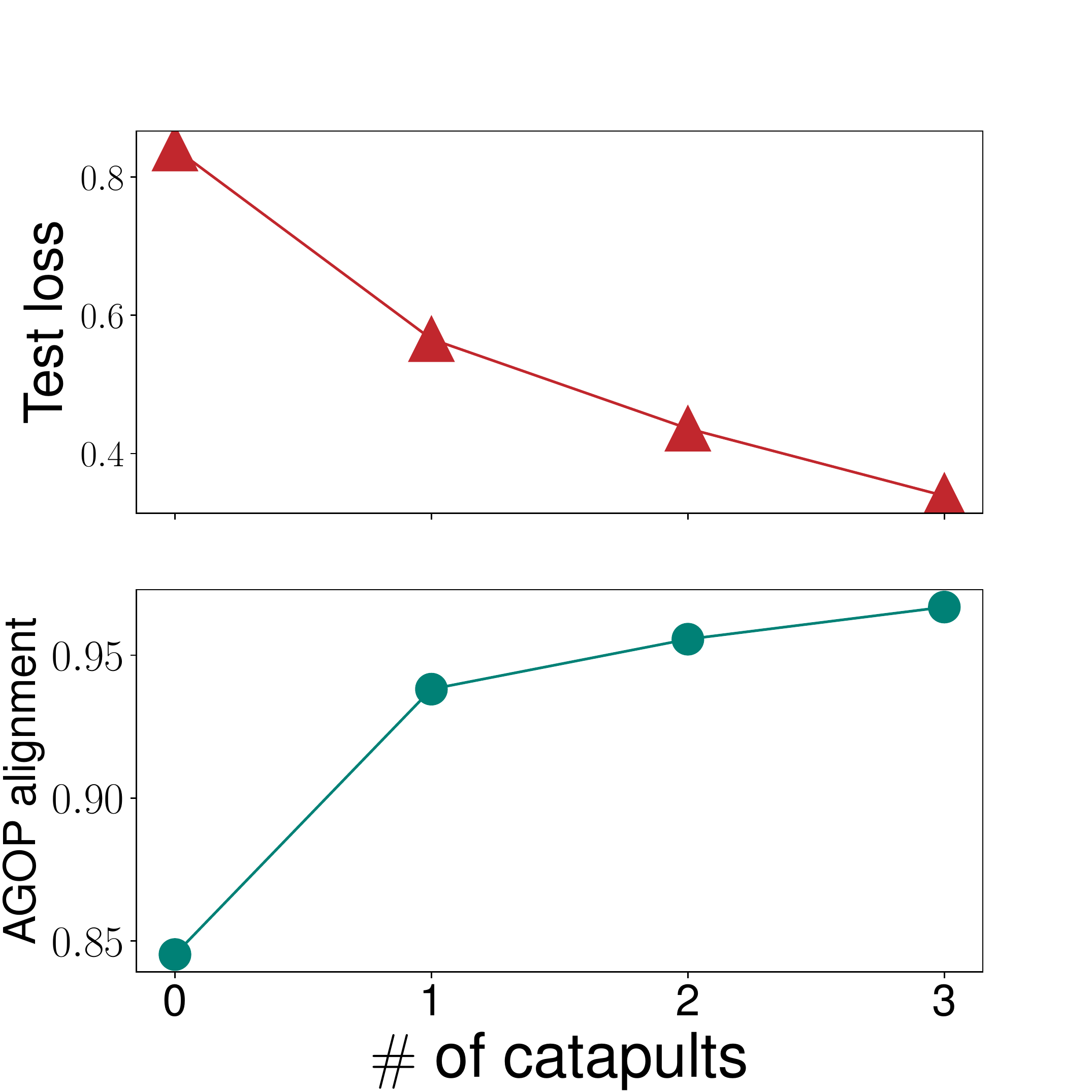}
         \caption{Rank-3 regression}
     \end{subfigure}
     \begin{subfigure}[b]{0.35\textwidth}
         \centering
         \includegraphics[width=\textwidth]{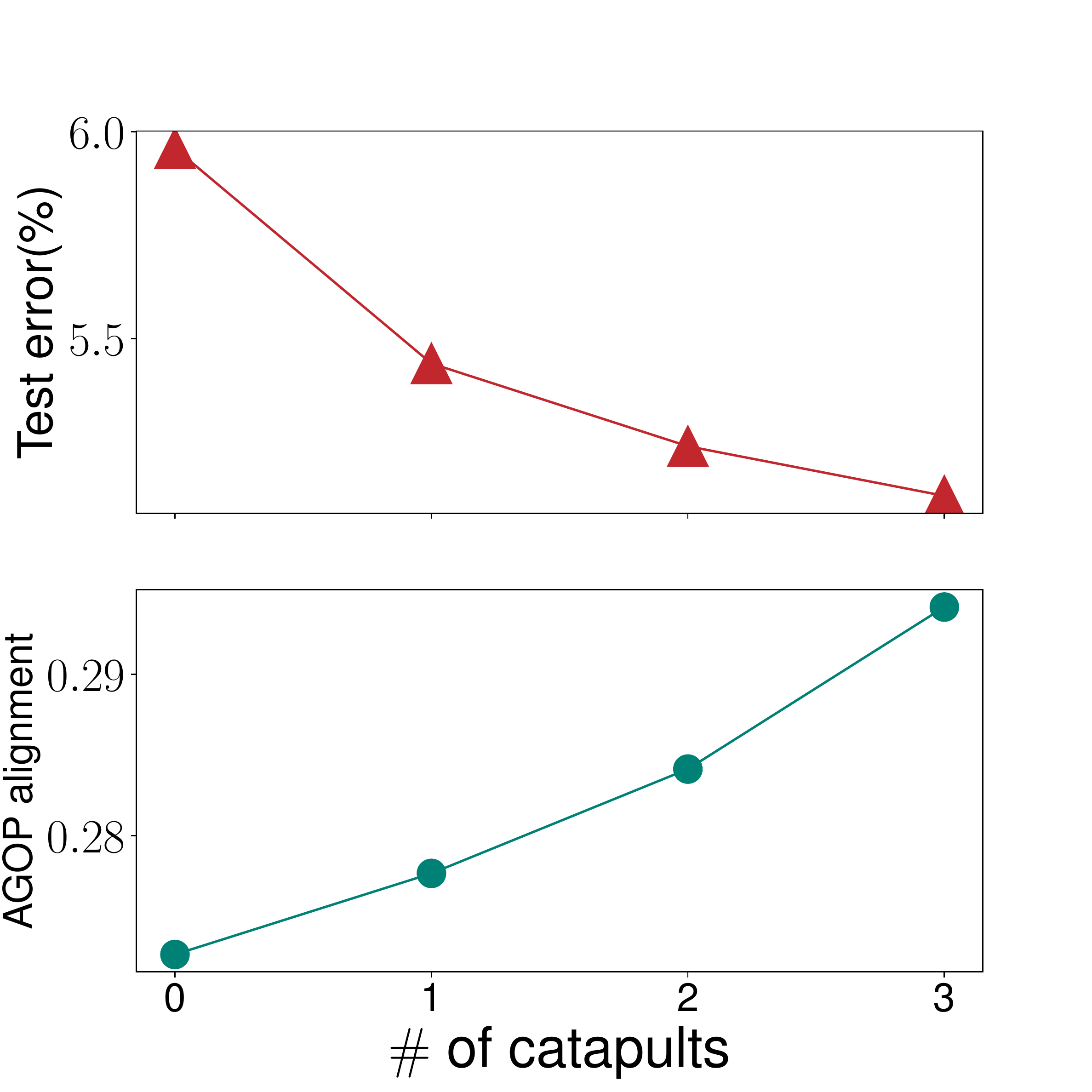}
         \caption{SVHN}
     \end{subfigure}
     \begin{subfigure}[b]{0.35\textwidth}
         \centering
         \includegraphics[width=\textwidth]{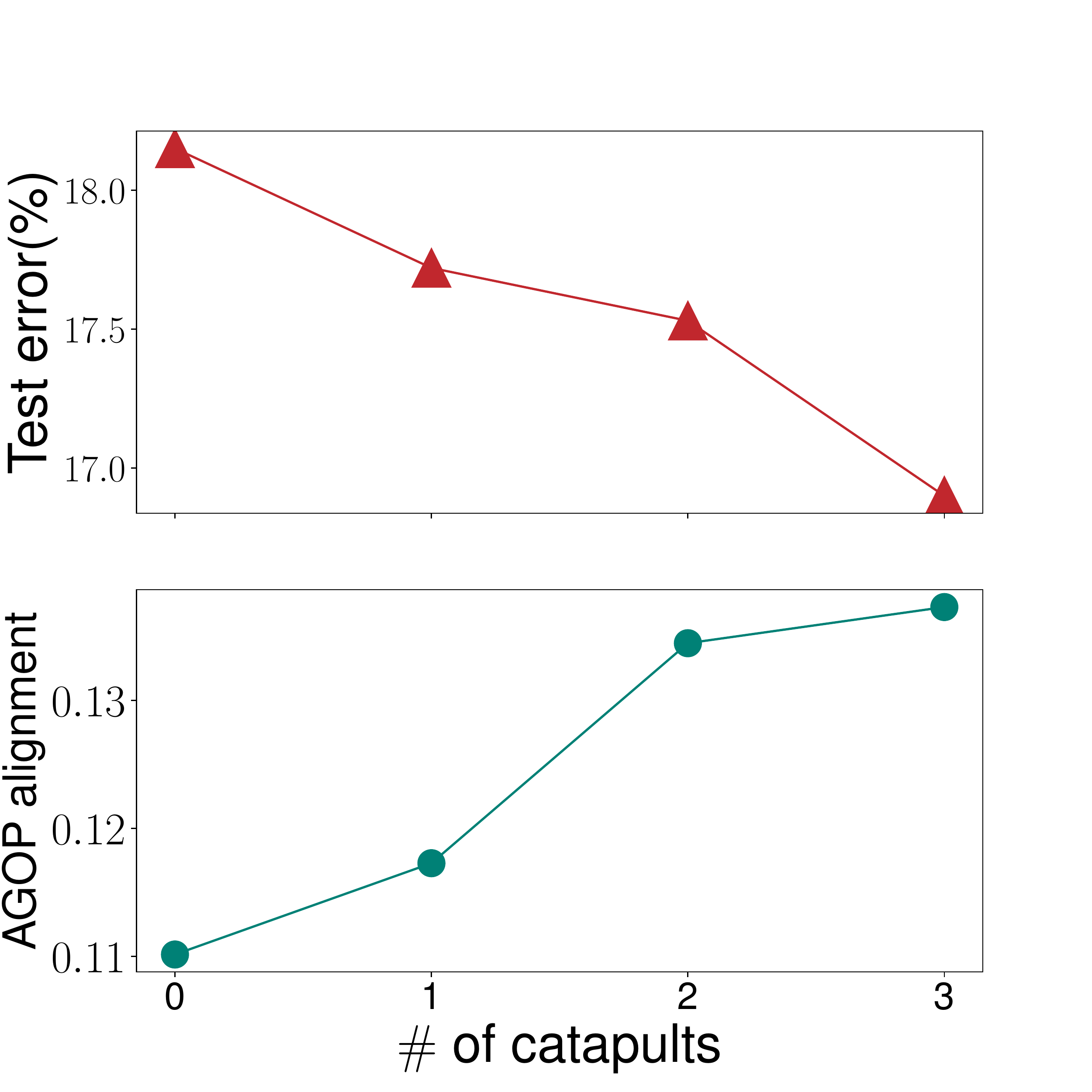}
         \caption{CelebA}
     \end{subfigure}
\caption{{\bf Correlation between AGOP alignment and test performance in GD with multiple catapults. } The learning rate is increased multiple times during training to generate multiple catapults. 
We train 2-layer FCN in Panel(a), 4-layer FCN in Panel(b,d) and 5-layer CNN in Panel(c).  Experimental details can be found in Appendix~\ref{exp:feature_learning}.\label{fig:multi_cata_gd_low_rank}}
\end{figure}

\paragraph{Improved test performance by catapults in GD.}In Section~\ref{subsec:multi_cata}, we showed that catapults can be generated multiple times.  We now show that generating multiple catapults leads to improved test performance of neural networks trained with GD by leading to increased \AGOP alignment. 
% For the rank-2 regression task (panel(a) and panel(b)), we increase the learning rate to $[8,16,30,50,75,80]$ at iteration $[50,150,220,280,350,420]$ correspondingly. For the rank-4 regression task (panel(c) and panel(d)), we increase the learning rate to $[50,75,110]$ at iteration $[15,40,60]$ correspondingly. add SVHN??? celeb faces??? 
In Fig.~\ref{fig:multi_cata_gd_low_rank}, we can see for all tasks,  the test loss/error decreases as the number of catapults increases while \AGOP alignment increases. This indicates that learning the EGOP strongly correlates with test performance. 

\begin{remark}
As discussed earlier, AGOP alignment is a means of improving sample efficiency when training on data from multi-index models with low-rank index space.  Our results on synthetic datasets show that catapults increase AGOP alignment, thereby leading to improved test performance.  Additionally, we show that when the index space is full-rank, which can be effectively learned by neural networks in the NTK regime, catapults do not improve the test performance as well as the AGOP alignment. See Fig.~\ref{fig:multi_cata_gd_full_rank} in Appendix~\ref{sec:add_agop_gd}.
\end{remark}

\begin{figure}[H]
     \centering
     \begin{subfigure}[b]{0.4\textwidth}
         \centering
         \includegraphics[width=\textwidth]{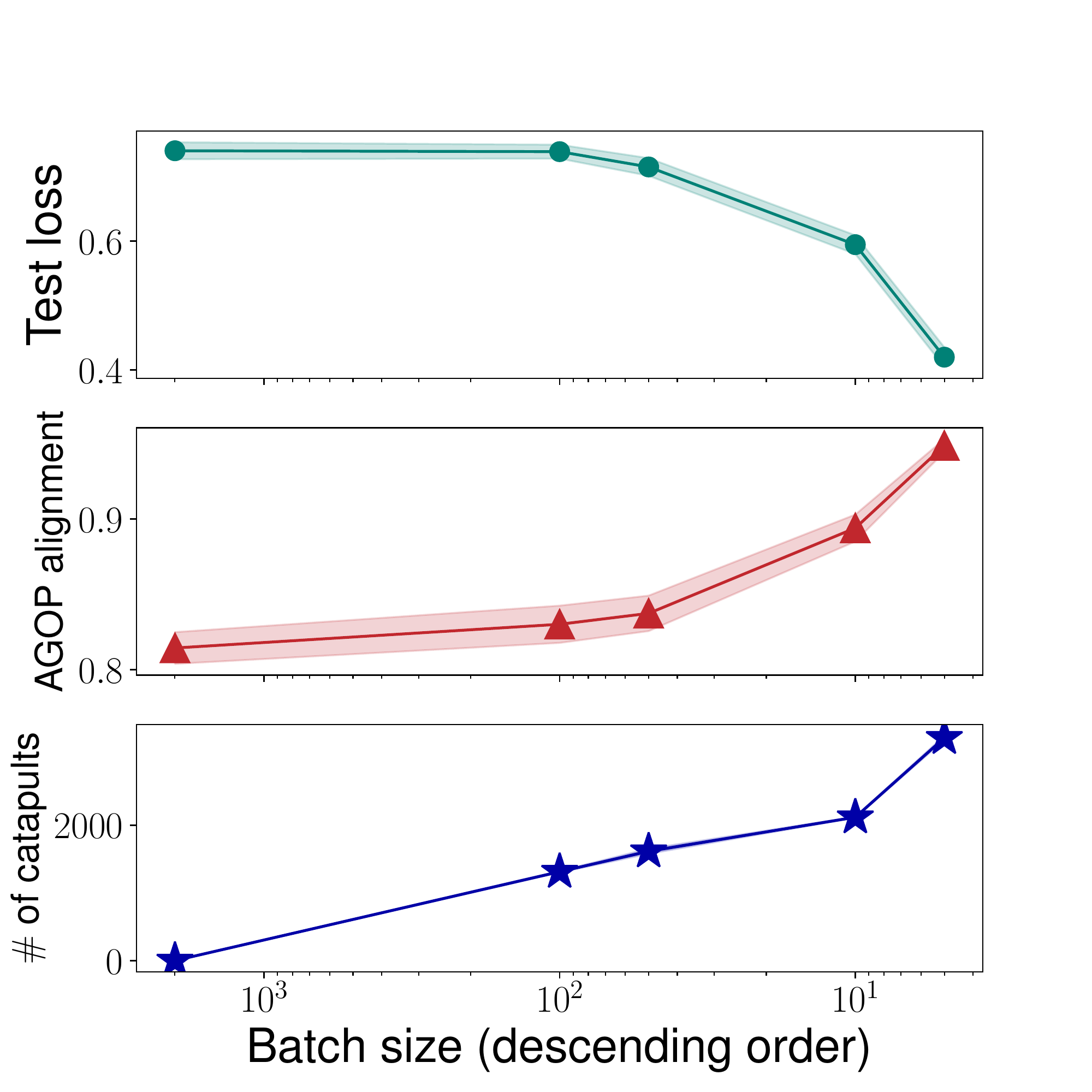}
             \caption{Rank-2 regression}
     \end{subfigure}
      \begin{subfigure}[b]{0.4\textwidth}
         \centering
         \includegraphics[width=\textwidth]{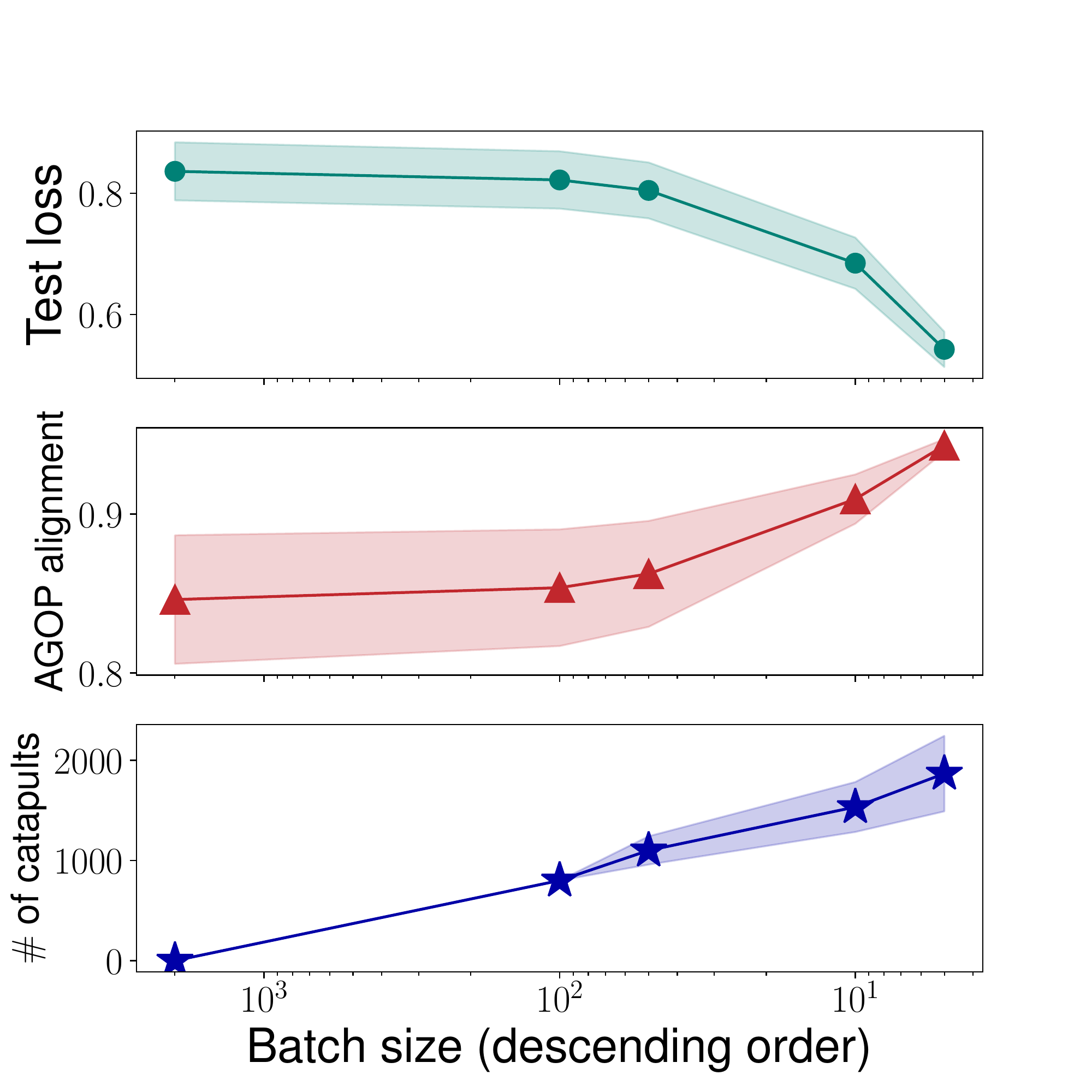}
             \caption{Rank-3 regression}
     \end{subfigure}

     % \begin{subfigure}[b]{0.22\textwidth}
     %     \centering
     %     \includegraphics[width=\textwidth]{figure/fl_sgd_test_loss_cata_agop_4_digit.pdf}
     %     \caption{Rank-4 regression}
     % \end{subfigure}\hspace*{-0.9em}
     \begin{subfigure}[b]{0.4\textwidth}
         \centering
         \includegraphics[width=\textwidth]{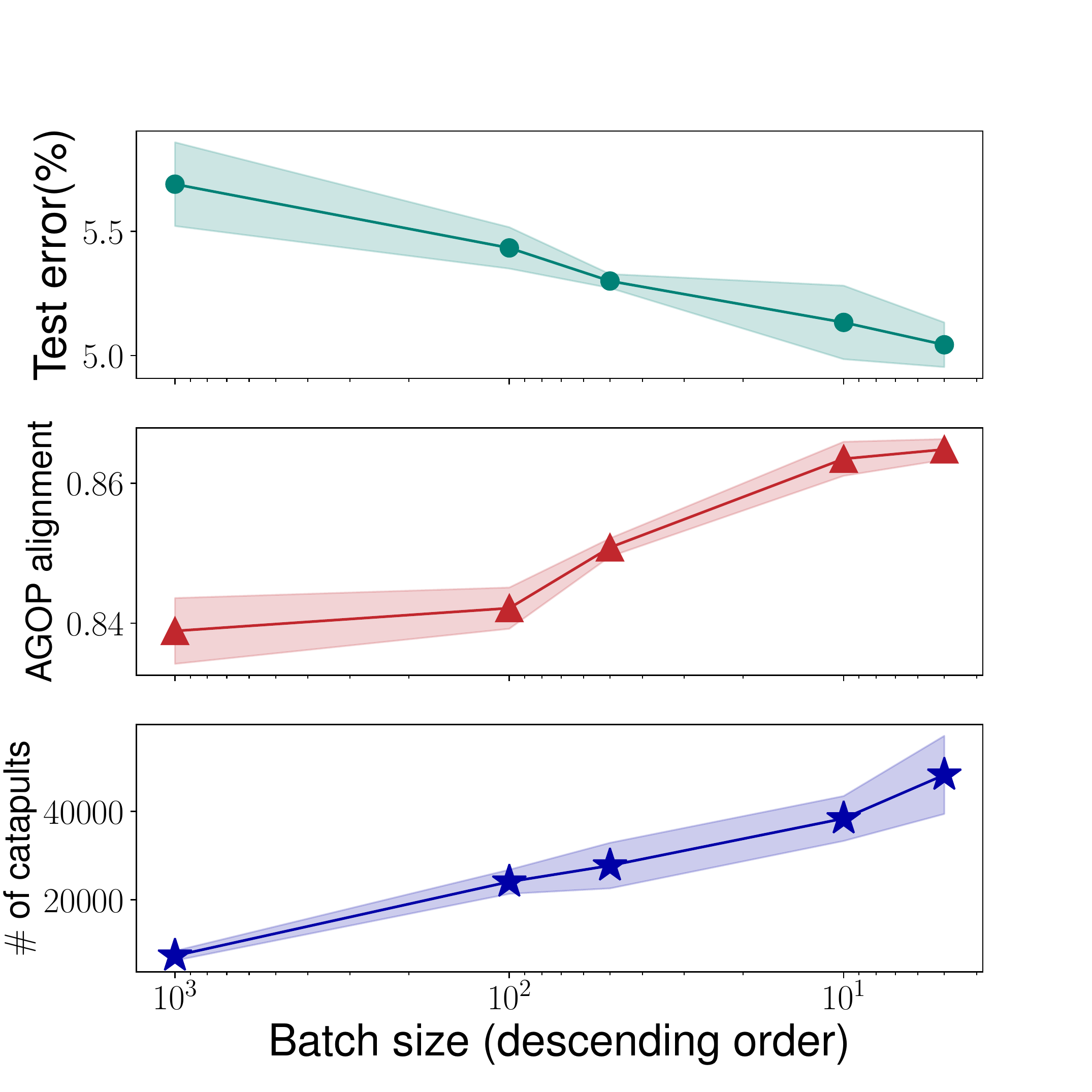}
         \caption{SVHN}
     \end{subfigure}
        \begin{subfigure}[b]{0.4\textwidth}
         \centering
         \includegraphics[width=\textwidth]{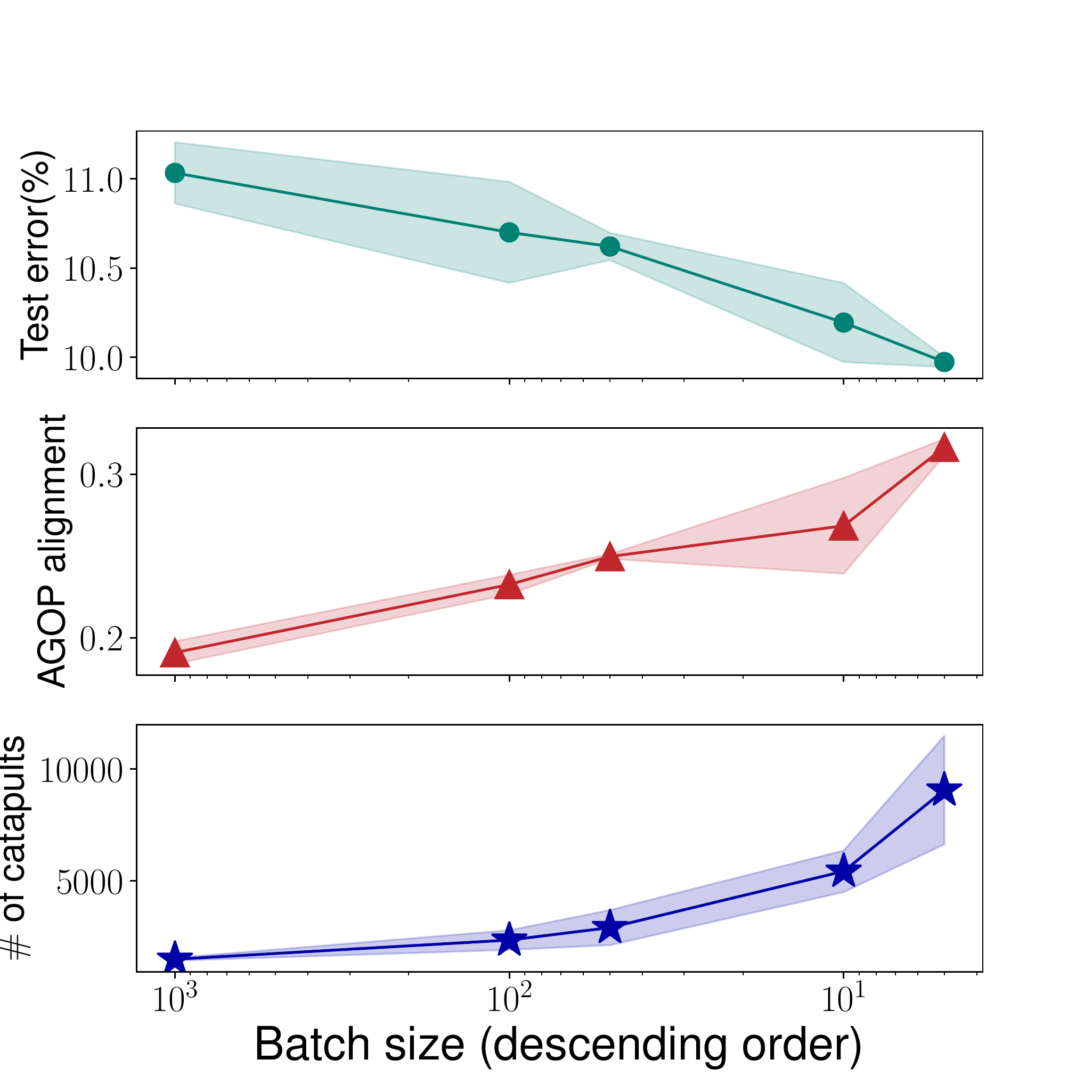}
         \caption{CelebA}
     \end{subfigure}
\caption{{\bf Correlation between AGOP alignment and test performance in SGD.}  We train a 2-layer FCN in Panel(a), 4-layer FCN in Panel(b,d) and 5-layer CNN in Panel(c) by SGD with a constant learning rate. We report the results as the average of 3 independent runs.  Experimental details can be found in Appendix~\ref{exp:feature_learning}.\label{fig:fl_sgd_agop}}
\end{figure}

\paragraph{Improved test performance by catapults in SGD.}
% \label{subsec:fl_sgd}

 In Section~\ref{sec:cata_sgd}, we have demonstrated the occurrence of catapults in SGD. We now show that decreasing batch size in SGD leads to better test performance as a result of an increase in the number of catapults and thus, increased \AGOP alignment. 
% We  consider the same training tasks as in~\ref{subsec:fl_gd} except that we use larger synthetic datasets and full SVHN and CelebA datasets. 
% We train the network with a large constant learning rate and stop training when the training loss is less than $10^{-3}$. 
We estimate the number of catapults during training by counting the number of the occurrence of the event $\eta - \etc(X_\batch)>\epsilon$ with $\epsilon=10^{-8}$ until the best validation loss/error. 
% Recall that when catapults occur, the component of the loss in the top eigenspace of the tangent kernel will increase, as discussed in section~\ref{sec:cata_sgd}. 
% More details of the experiments can be found in Appendix~\ref{sec:exp_details}.

In Fig.~\ref{fig:fl_sgd_agop}, we can see that across all tasks, as the batch size decreases, (1) the number of catapults increases, (2) the test loss/error decreases and (3) the \AGOP alignment improves. These findings indicate that in SGD, a smaller batch size leads to more catapults which in turn improves the test performance through alignment with the AGOP. These observations are consistent with our findings in GD.

\begin{figure}[H]
     \centering
     \begin{subfigure}[b]{0.4\textwidth}
         \centering
         \includegraphics[width=\textwidth]{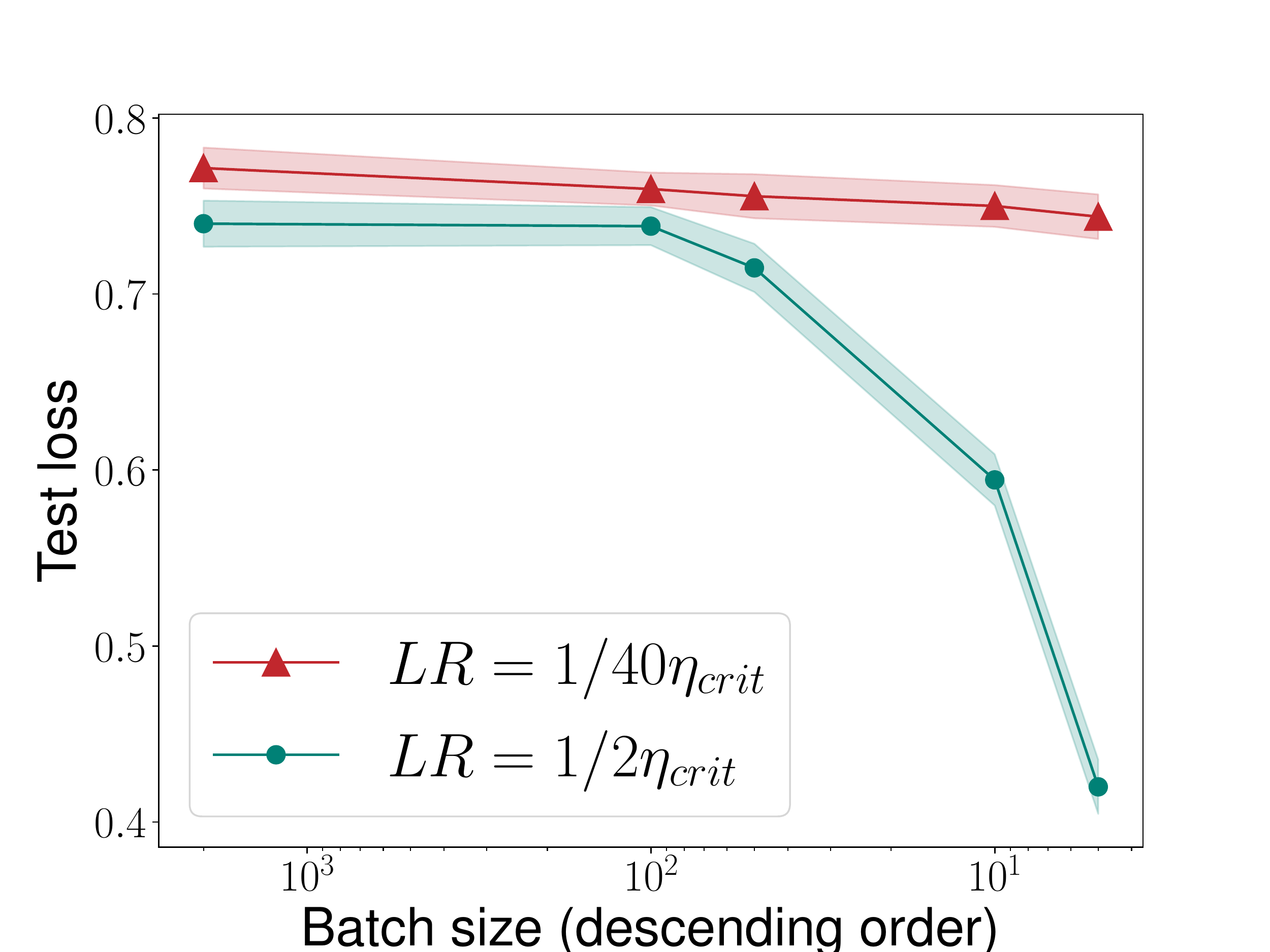}
             \caption{Rank-2 regression}
     \end{subfigure}
     \begin{subfigure}[b]{0.4\textwidth}
         \centering
         \includegraphics[width=\textwidth]{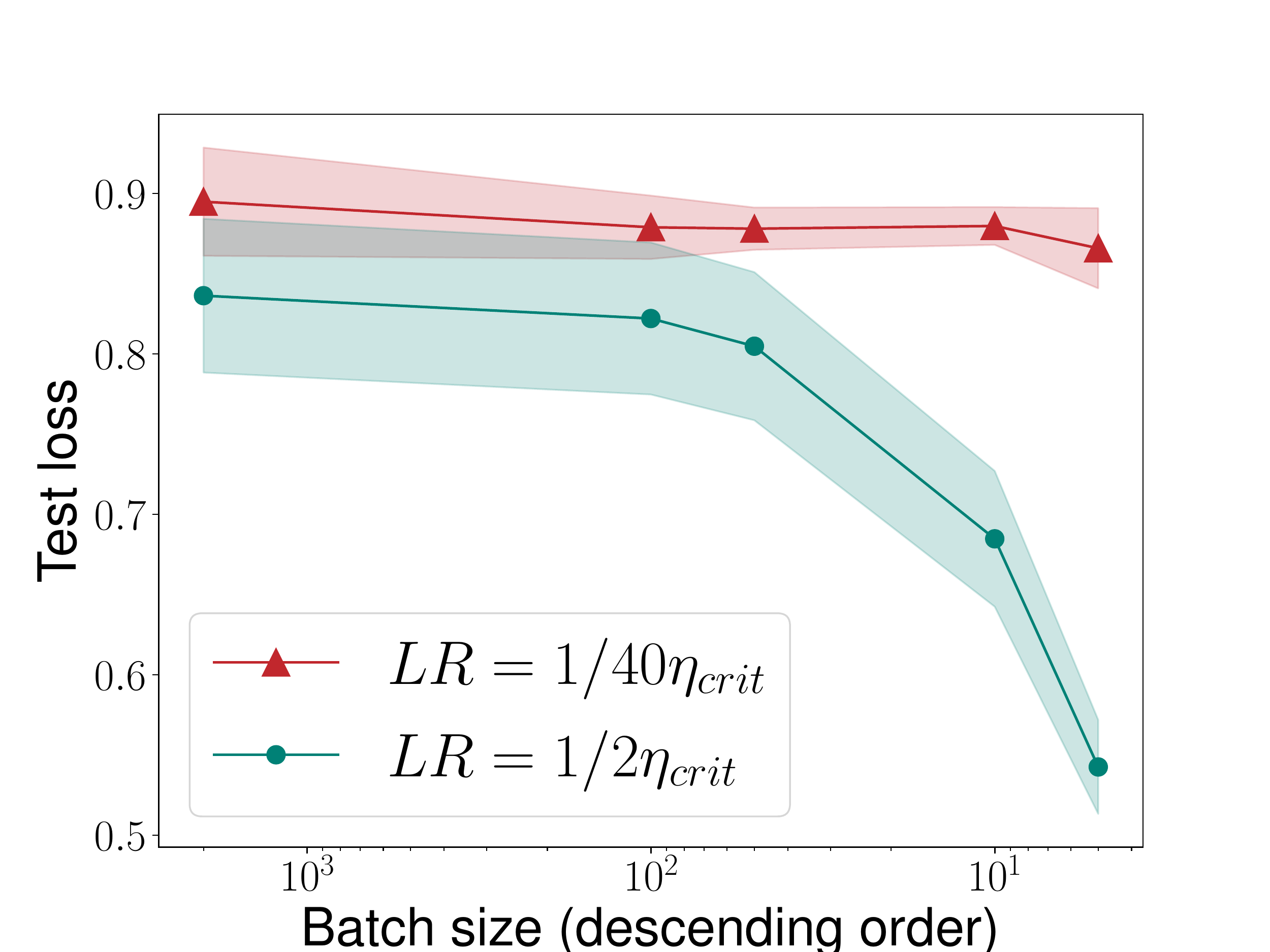}
         \caption{Rank-3 regression}
     \end{subfigure}
     % \begin{subfigure}[b]{0.24\textwidth}
     %     \centering
     %     \includegraphics[width=\textwidth]{figure/fl_sgd_test_loss_cata_agop_4_digit_std.pdf}
     %     \caption{Rank-4 regression}
     % \end{subfigure}\hspace*{-0.9em}
     \begin{subfigure}[b]{0.4\textwidth}
         \centering
         \includegraphics[width=\textwidth]{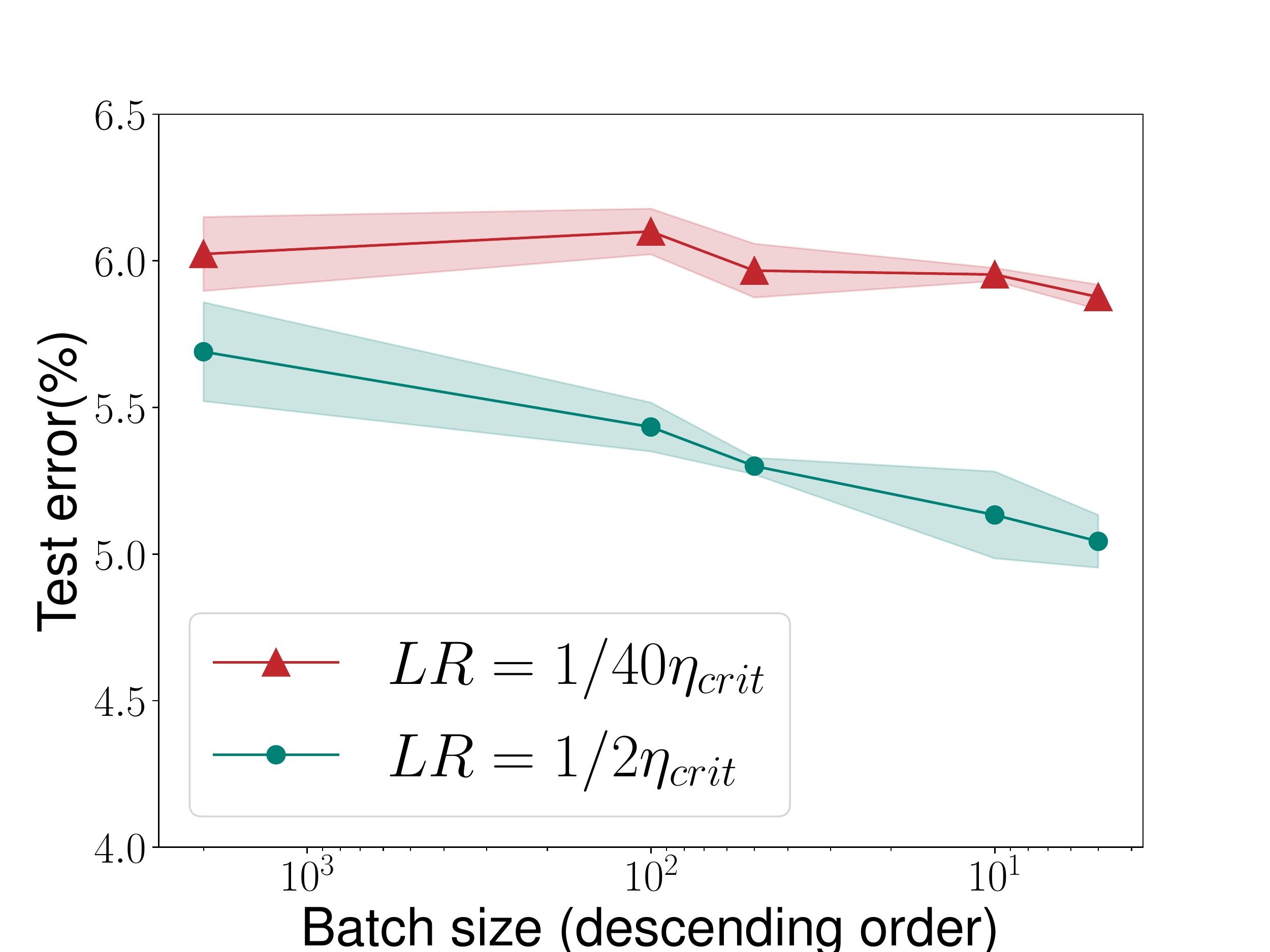}
         \caption{SVHN}
     \end{subfigure}
        \begin{subfigure}[b]{0.4\textwidth}
         \centering
         \includegraphics[width=\textwidth]{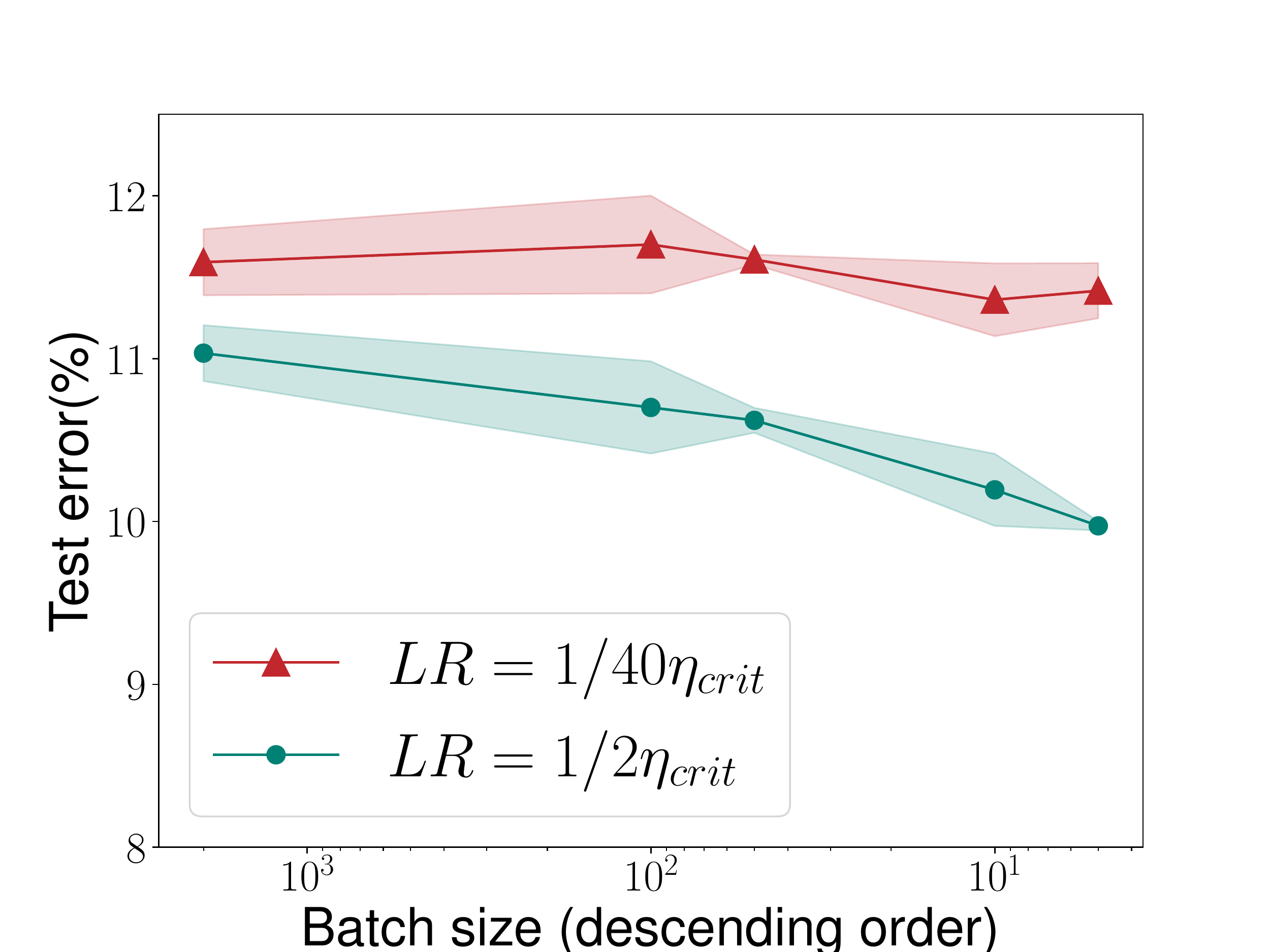}
         \caption{CelebA}
     \end{subfigure}
\caption{{\bf ``Large'' vs. ``small'' learning rate on test performance with different batch sizes.} We consider the same setting as in Fig.~\ref{fig:fl_sgd_agop} except for selecting a smaller learning rate $\etc/40$ compared to  $ \etc/2$ in Fig.~\ref{fig:fl_sgd_agop}.  Here $\etc$ is the critical learning rate for the whole dataset.\label{fig:fl_sgd_test_loss_small_lr}}
\end{figure}

\paragraph{Batch size does not affect generalization when the learning rate is small.} Given the discussion above, sufficiently small learning rates will result in no catapults for any batch size. Thus we expect that all batch sizes will provide similar generalization performance for sufficiently small learning rates. This, indeed, is what we observe in the experiments presented in Fig.~\ref{fig:fl_sgd_test_loss_small_lr} where we keep the same experimental setting as for Fig.~\ref{fig:fl_sgd_agop} except for a smaller learning rate. Specifically, we observe that while decreasing batch size consistently improves generalization for large learning rates, it has little effect on generalization for small learning rates.

% In the experiments, we train a 2-layer FCN on Rank-2 dataset and a 4-layer FCN on Rank-4 and CelebA dataset, and train a 4-layer CNN on SVHN-2 dataset. 

\begin{figure}[t!]
     \centering
     \begin{subfigure}[b]{0.4\textwidth}
         \centering
         \includegraphics[width=\textwidth]{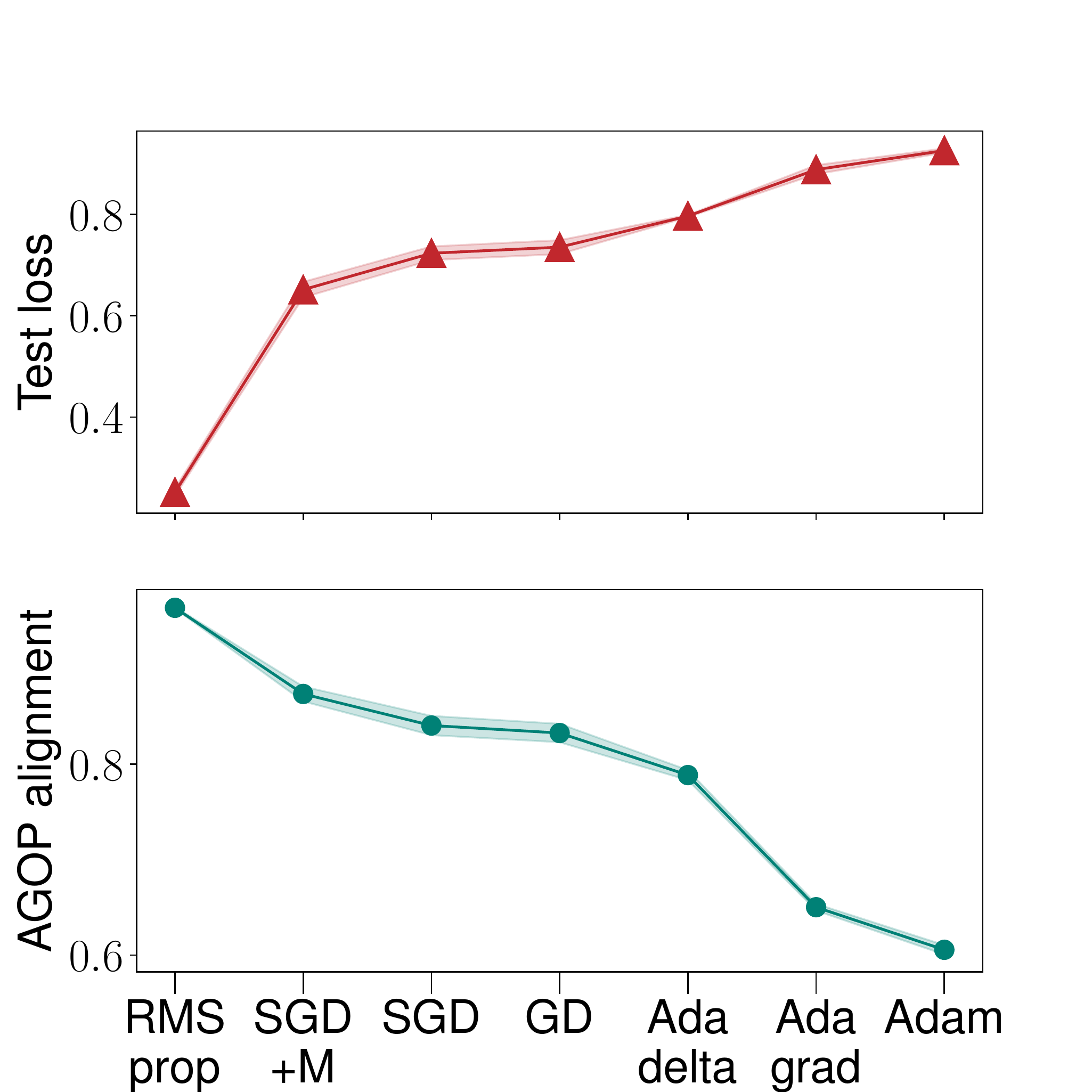}
         \caption{Rank-2 regression}
     \end{subfigure}
      \begin{subfigure}[b]{0.4\textwidth}
         \centering
         \includegraphics[width=\textwidth]{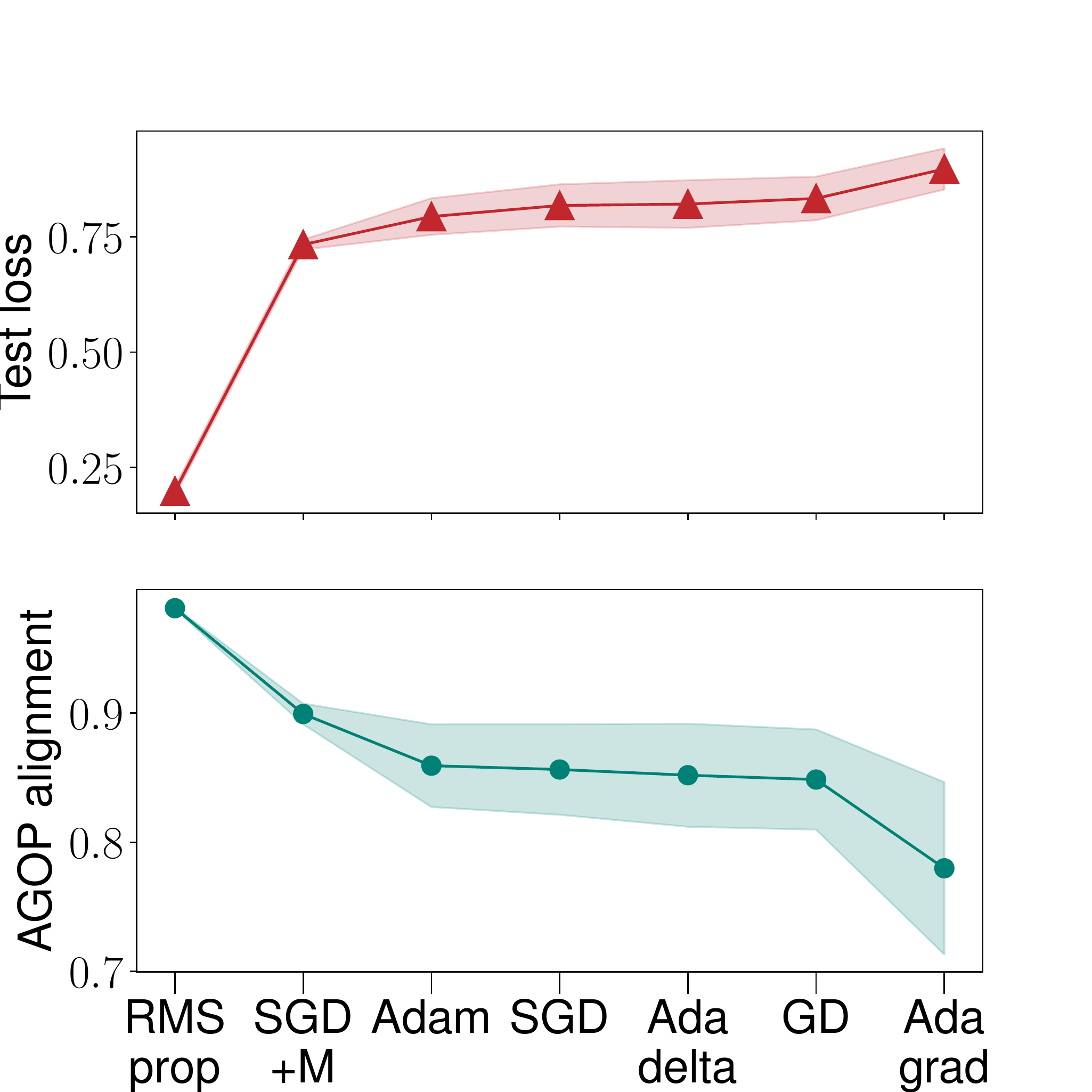}
         \caption{Rank-3 regression}
     \end{subfigure}
     %      \begin{subfigure}[b]{0.22\textwidth}
     %     \centering
     %     \includegraphics[width=\textwidth]{figure/optim_test_agop_4_digit.pdf}
     %     \caption{Rank-4 regression}
     % \end{subfigure}\hspace*{-0.9em}
     \begin{subfigure}[b]{0.4\textwidth}
         \centering
         \includegraphics[width=\textwidth]{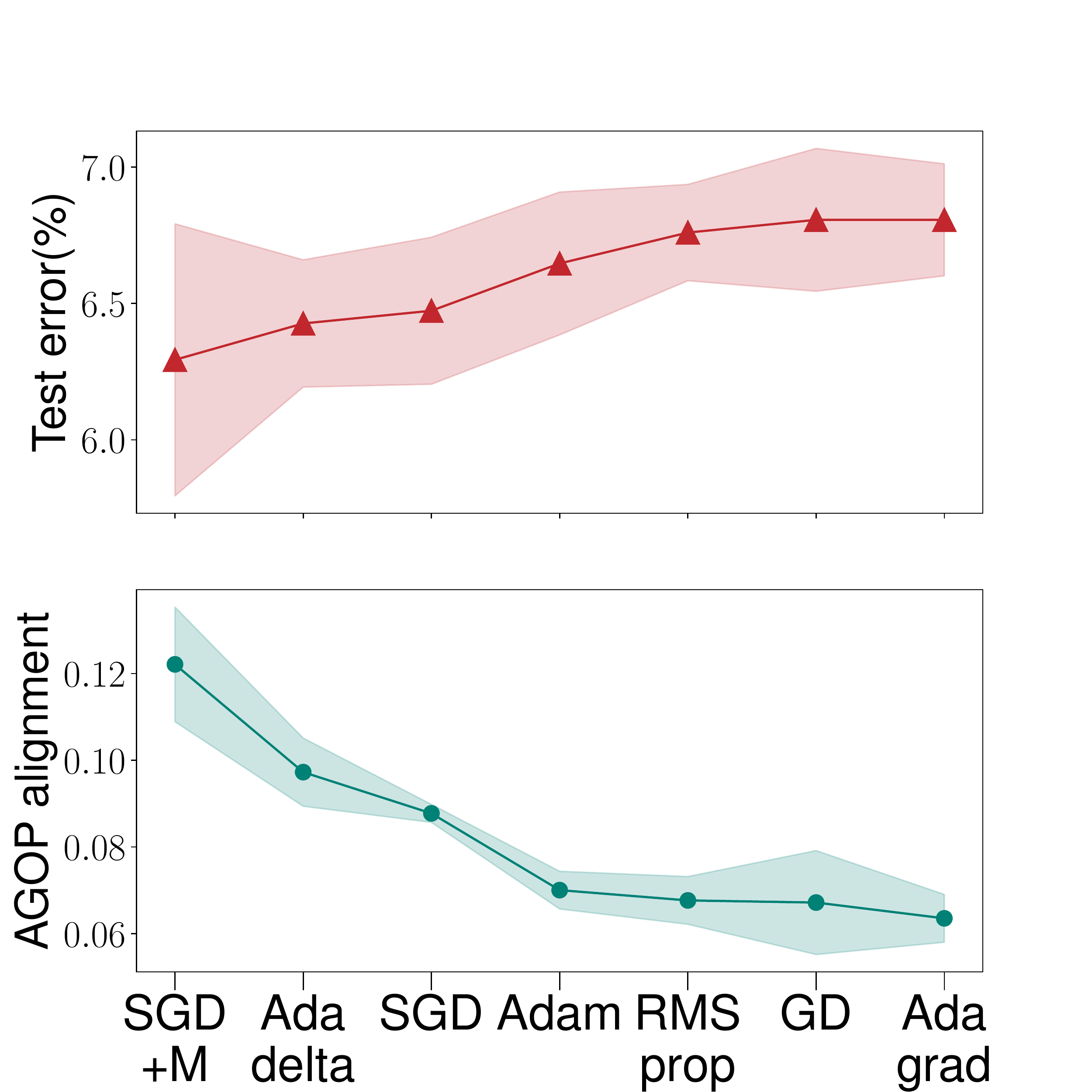}
         \caption{SVHN}
     \end{subfigure}
     \begin{subfigure}[b]{0.4\textwidth}
         \centering
         \includegraphics[width=\textwidth]{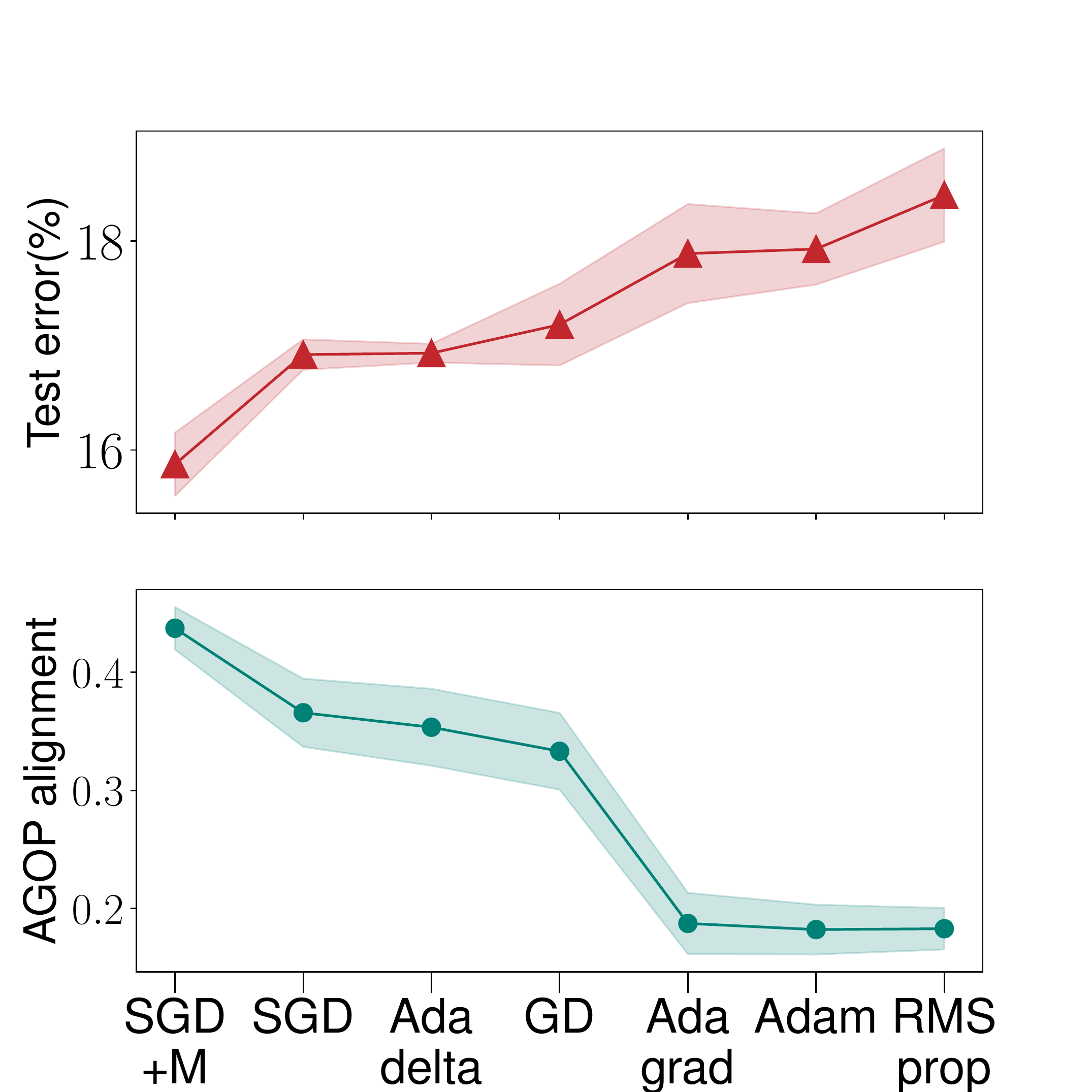}
         \caption{CelebA}
     \end{subfigure}
\caption{{\bf Correlation between test performance and \AGOP{} alignment for different optimization algorithms.} We train a 2-layer FCN in Panel(a), a 4-layer FCN in Panel(b,d) and a 5-layer CNN in Panel(c).  We use GD, SGD,   SGD with Momentum~\citep{qian1999momentum}(SGD+M), RMSprop~\citep{rmsprop},  Adagrad~\citep{duchi2011adaptive}, Adadelta~\citep{zeiler2012adadelta} and Adam~\citep{KingBa15} for training. Experimental details can be found in Appendix~\ref{exp:feature_learning}.\label{fig:optim_agop}} 
\end{figure}
\paragraph{Generalization  with different optimizers correlates with  \AGOP alignment.} We further demonstrate the strong correlation between the test performance and \AGOP alignment by comparing the predictors trained on the same task with a number of  different optimization algorithms.  
From the results shown in Fig.~\ref{fig:optim_agop}, we can see that the \AGOP alignment strongly correlates with the test performance, which suggests that models  learning the \AGOP is useful for learning the problem.

\section{Conclusions}

% In this work, we show that the fluctuations in the training loss of SGD are catapults, and we utilize the catapult mechanism to explain the improved test performance when training with smaller batch sizes in SGD. 
% Finally, we show that catapults improve the generalization by feature learning as \AGOP{} progressively aligns with the true \AGOP{} when catapults occur.
 In this work, we addressed the following questions: (1) why do spikes in training loss occur during training with SGD and (2) how do the spikes relate to generalization? For the first question,  we demonstrate that the spikes in the training loss are caused by the catapult dynamics in the top eigenspace of the tangent kernel. 
 %correspond to catapults by showing the spikes occur in the top eigenspace of the tangent kernel and each loss spike corresponds to a decrease in the spectral norm of the tangent kernel. 
 For the second question, we show that catapults lead to increased alignment between the \AGOP of the model being trained and the  AGOP of the underlying model or its state-of-the-art approximation.  A consequence of our results is the explanation for the observation that SGD with small batch size often leads to improved generalization. 
 This is due to an increase in the number of catapults for small batch sizes, due to increased batch variability, which, in turn, leads to better \AGOP alignment.

\section*{Acknowledgements}

A.R. is supported by the Eric and Wendy Schmidt Center at the Broad Institute. We are grateful for the support from the National Science Foundation (NSF) and the Simons Foundation for the Collaboration on the Theoretical Foundations of Deep Learning (\url{https://deepfoundations.ai/}) through awards DMS-2031883 and \#814639  and the TILOS institute (NSF CCF-2112665).
This work used NVIDIA V100 GPUs NVLINK and HDR IB (Expanse GPU) at SDSC Dell Cluster through allocation TG-CIS220009 and also, Delta system at the National Center for Supercomputing Applications through allocation bbjr-delta-gpu from the Advanced Cyberinfrastructure Coordination Ecosystem: Services \& Support (ACCESS) program, which is supported by National Science Foundation grants \#2138259, \#2138286, \#2138307, \#2137603, and \#2138296.

 %For the future direction, it would be interesting to connect AGOP alignment with other feature learning measurements, e.g., functional properties observed in neural networks, e.g.,~\cite{jacot2023bottleneck}.  
%An important direction of future investigation is understanding the catapult dynamics with other loss functions, such as cross-entropy. For example, see our discussion on cross-entropy loss in  Appendix~\ref{app:logistic}.

% \section*{Broader Impact}
% Our work sheds light on certain  phenomena which have been empirically in neural networks. We hope it will be useful for developing new methods and analyses. The authors are unable to foresee the potential societal consequences of this particular work  within the broad landscape of future AI. 

%This could potentially
%lead to more efficient training algorithms and make machine learning applications more reliable and transparent. There are many potential societal consequences of our work, none of which we feel must be specifically highlighted here.

\printbibliography

%%%%%%%%%%%%%%%%%%%%%%%%%%%%%%%%%%%%%%%%%%%%%%%%%%%%%%%%%%%%%%%%%%%%%%%%%%%%%%%
%%%%%%%%%%%%%%%%%%%%%%%%%%%%%%%%%%%%%%%%%%%%%%%%%%%%%%%%%%%%%%%%%%%%%%%%%%%%%%%
% APPENDIX
%%%%%%%%%%%%%%%%%%%%%%%%%%%%%%%%%%%%%%%%%%%%%%%%%%%%%%%%%%%%%%%%%%%%%%%%%%%%%%%
%%%%%%%%%%%%%%%%%%%%%%%%%%%%%%%%%%%%%%%%%%%%%%%%%%%%%%%%%%%%%%%%%%%%%%%%%%%%%%%
\newpage
\appendix
\onecolumn
\input{appendix}

\end{document}

%% file: math_command.tex
\renewcommand\L{{\mathcal{L}}}

\newcommand\vx{{\boldsymbol{x}}}
\newcommand\vy{{\boldsymbol{y}}}
\newcommand\vz{{\boldsymbol{z}}}

\newcommand{\ceil}[1]{\left\lceil #1 \right\rceil}

\renewcommand\P{{\mathcal{P}}}
\newcommand{\inner}[1]{\left<#1\right>}
\newcommand{\norm}[1]{\left\|#1\right\|}
\newcommand{\round}[1]{\left(#1\right)}

\def\rvf{{\mathbf{f}}}

\def\rvu{{\mathbf{i}}}

\def\rvu{{\mathbf{u}}}
\def\rvv{{\mathbf{v}}}
\def\rvw{{\mathbf{w}}}
\def\rvx{{\mathbf{x}}}
\def\vy{{\mathbf{y}}}

\renewcommand\L{{\mathcal{L}}}

\newcommand\etc{{\eta_{\mathrm{crit}}}}

\newcommand\etm{{\eta_{\mathrm{max}}}}

\newcommand\batch{{\mathrm{batch}}}
\newcommand\AGOP{AGOP~}
\usepackage{physics}
\usepackage{xfrac}

\newtheorem{lemma}{Lemma}%[section]
\newtheorem{claim}{Claim}%[section]
\newtheorem{definition}{Definition}%[section]
\newtheorem{remark}{Remark}%[section]

%% file: appendix.tex
\section*{Appendix}

\section{The critical learning rate can be well approximated using NTK for wide neural networks}\label{sec:h_K}

In this section, we show that the critical learning rate $\etc :=\frac{2}{\lambda_{\max}(H_\L)}$ can be well approximated using NTK, i.e., $\tilde{\eta}_{\mathrm{crit}} := \frac{n}{\lambda_{\max}(K)(\rvw)}$. Note  that $\norm{K} = \lambda_{\max}(K)(\rvw)$. 

\subsection{Approximation of the critical learning rate using NTK during training with a small constant learning rate}\label{subsec:h_K_init}
For MSE  $\L(\rvw;X) = \frac{1}{n}\sum_{i=1}^n (f(\rvw;\rvx_i)-y_i)^2$, we can compute its $H_\L$ by the chain rule:
\begin{align*}
    H_\L(\rvw) = \frac{2}{n}\underbrace{\sum_{i=1}^n\round{\frac{\partial f(\rvw;\rvx_i)}{\partial \rvw}}^T \frac{\partial f(\rvw;\rvx_i)}{\partial \rvw}}_{\mathcal{A(\rvw)}}  + \underbrace{\frac{2}{n}\sum_{i=1}^n  (f(\rvw;\rvx_i)-y_i)\frac{\partial^2 f(\rvw;\rvx_i)}{\partial \rvw^2}}_{\mathcal{B}(\rvw)}.
\end{align*}

Assume $\|\vx_i\| = O(1)$ and $|y_i| = O(1)$ for all $i\in [n]$. For $\mathcal{B}(\rvw_0)$, by random initialization of weights $\rvw_0$, with high probability, we have $|f(\rvw_0;\rvx_i)-y_i| = O(\log m)$, and $\norm{\frac{\partial^2 f(\rvw_0;\rvx_i)}{\partial \rvw^2}}_2 = \tilde{O}(1/\sqrt{m})$~\citep{liu2020linearity,zhu2022transition} where $m$ denotes the width of the network. Therefore, by the union bound, with high probability, we have $\mathcal{B}(\rvw_0) = \tilde{O}(1/\sqrt{m})$.

Note that $\lambda_{\max}(\mathcal{A}(\rvw)) = \frac{2}{n}\lambda_{\max}(K(\rvw))$ for any $\rvw$. Combining all the bounds together, we have $\left|\lambda_{\max}(H_\L)(\rvw_0) -\frac{2}{n}\lambda_{\max}(K)(\rvw_0)\right| = \tilde{O}(1/\sqrt{m})$. Then we have
\begin{align*}
    |\etc - \tilde{\eta}_{\mathrm{crit}}|= \left|\frac{2}{\lambda_{\max}(H_\L)(\rvw_0)} -\frac{n}{\lambda_{\max}(K)(\rvw_0)}\right| = \tilde{O}(1/\sqrt{m})
\end{align*}
 as long as $\lambda_{\max}(K)(\rvw_0) = \Omega(1)$, which is true with high probability over random initialization for wide networks~\citep{nguyen2018loss,banerjee2023neural}.

For wide neural networks trained with a small constant learning rate, $\norm{\frac{\partial^2 f(\rvw_0;\rvx_i)}{\partial \rvw^2}}_2 = \tilde{O}(1/\sqrt{m})$ holds during the whole training process of GD/SGD, hence this approximation holds~\cite{liu2020linearity}.

\subsection{Approximation of the critical learning rate using NTK during training with a large learning rate}\label{subsec:h_K_whole}

In this section, we provide further evidence for SGD that $\tilde{\eta}_{\mathrm{crit}}$  approximates $\etc$ during training even with a large learning rate. Recall that $\tilde{\eta}_{\mathrm{crit}}= {b}/{\lambda_{\max}(K(\rvw;X_\batch))}$ where $b$ is the batch size.  We consider the same network architectures as the shallow network in Fig.~\ref{fig:sign_cata} and deep networks in Fig.~\ref{fig:sgd_deep}.

\begin{figure}[htb!]
\vspace{-10pt}
     \centering
     % \begin{subfigure}[b]{0.4\textwidth}
     %     \centering
     %     \includegraphics[width=\textwidth]{figure/lr_match_prec.pdf}\caption{Sign match between $\eta-\etc$ and $\P\Diff_1$}
     % \end{subfigure}
     \begin{subfigure}[b]{0.32\textwidth}
         \centering
         \includegraphics[width=\textwidth]{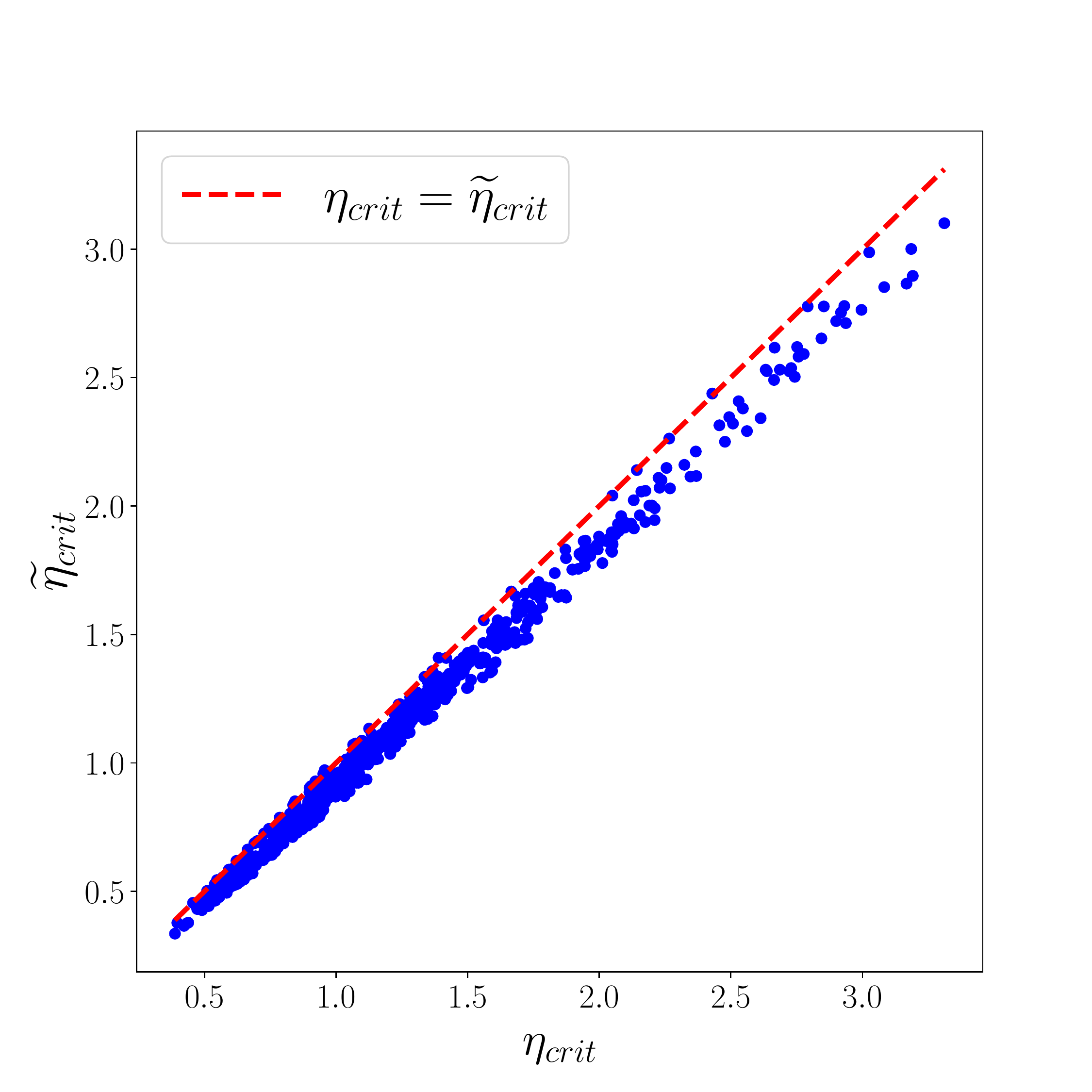}\caption{ Shallow network}
     \end{subfigure}
          \begin{subfigure}[b]{0.32\textwidth}
         \centering
         \includegraphics[width=\textwidth]{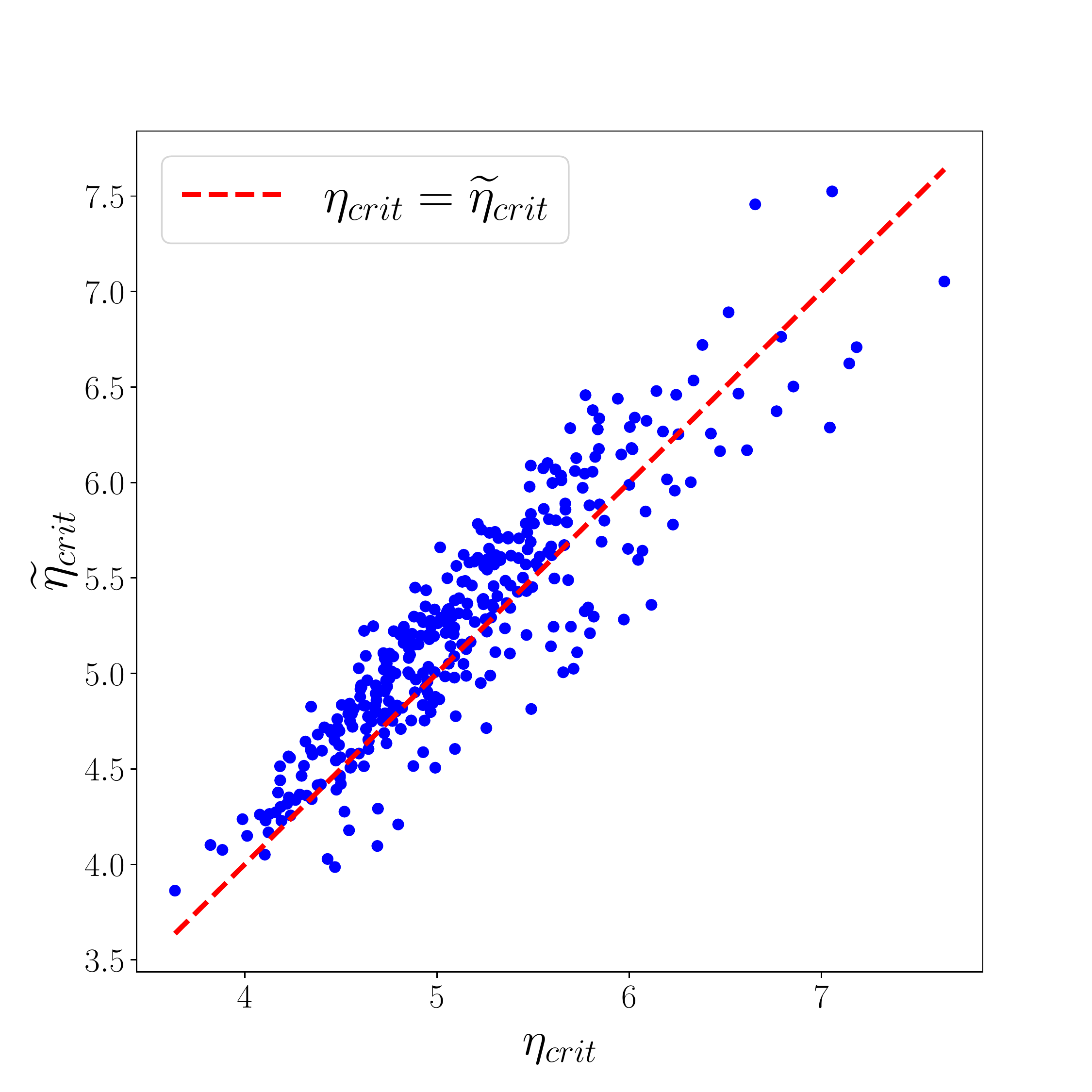}\caption{5-layer FCN}
     \end{subfigure}
          \begin{subfigure}[b]{0.32\textwidth}
         \centering
         \includegraphics[width=\textwidth]{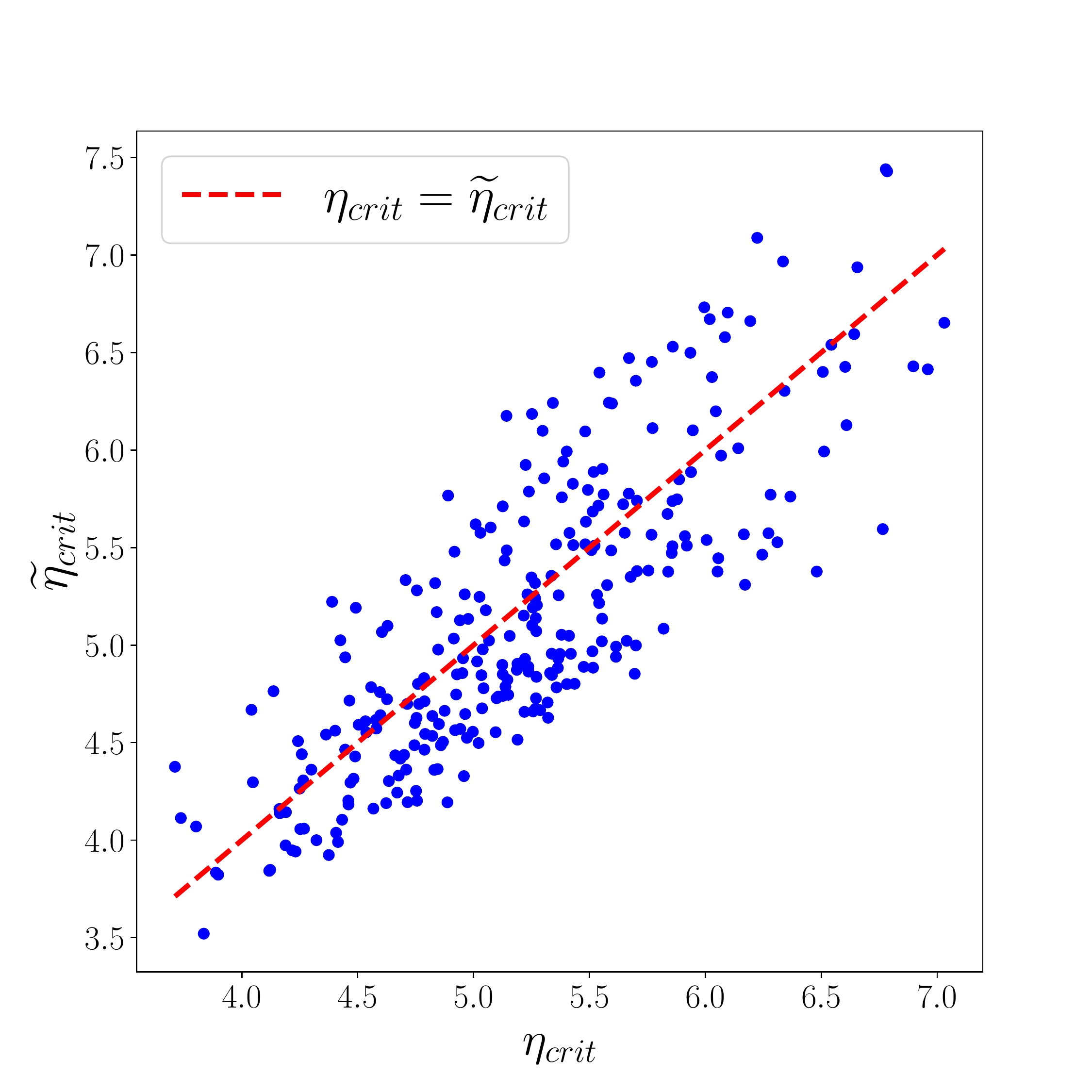}\caption{5-layer CNN}
     \end{subfigure}
     \caption{{\bf Validation of $\etc \approx \tilde{\eta}_{\mathrm{crit}}$ during SGD with catapults.}  Plot of points $(\etc,\tilde{\eta}_{\mathrm{crit}})$ at each iteration of SGD for the shallow network, 5-layer FCN and CNN. The models are trained on $128$ data points from CIFAR-10 by SGD with batch size $32$. The settings are the same with Table~\ref{tab:match_rate}. }\label{fig:sign_match_tilde_crit}
\end{figure}

We can see Fig.~\ref{fig:sign_match_tilde_crit} shows that $\etc$ is close to  $\tilde{\eta}_{\mathrm{crit}}$ during training with SGD.
% Similar to Fig.~\ref{fig:sign_cata}(a), to verify the claim,  we compare the sign of $\P\Diff_1$ and $\eta-\tilde{\eta}_{\mathrm{crit}}$ for each batch. 
% In Fig.~\ref{fig:sign_match_tilde_crit}, we show the number of events where $\sgn(\Delta\L_{\pi_1}(X_\batch))=\sgn(\eta-\tilde{\eta}_{\mathrm{crit}})$ divided by the number of iterations until convergence.
% We allow a small perturbation, $\epsilon$, in $\eta$ to account for error in estimating $\eta$ due to using finite width models.

% In Fig.~\ref{fig:sign_match_tilde_crit}(a), we can see that the sign of $\Delta\L_{\pi_1}(\rvf^t(X_\batch))$ and $\eta - \tilde{\eta}_{\mathrm{crit}}(\rvf^t(X_{\batch}))$ is well matched throughout the training process. 

% Furthermore, we show that $\etc$ is close to $\tilde{\eta}_{\mathrm{crit}}$ during the whole training process. 

\newpage 
\section{Additional experiments for the catapult in GD}\label{sec:gd_add}
\subsection{Catapults occur in the top eigenspace of NTK}
In this section, we provide additional empirical evidence to verify Claim~\ref{claim: Claim 1}. In particular, we consider three neural network architectures: a 5-layer Fully Connected Neural Network (FCN), a 5-layer Convolutional Neural Network (CNN), and  Wide ResNets 10-10; and three datasets CIFAR-10, SVHN, and a synthetic dataset. For the synthetic dataset, we consider the rank-2 regression task with training size $128$.

\begin{figure}[H]\vspace{-10pt}
     \centering
     \begin{subfigure}[b]{0.35\textwidth}
         \centering
    \includegraphics[width=\textwidth]{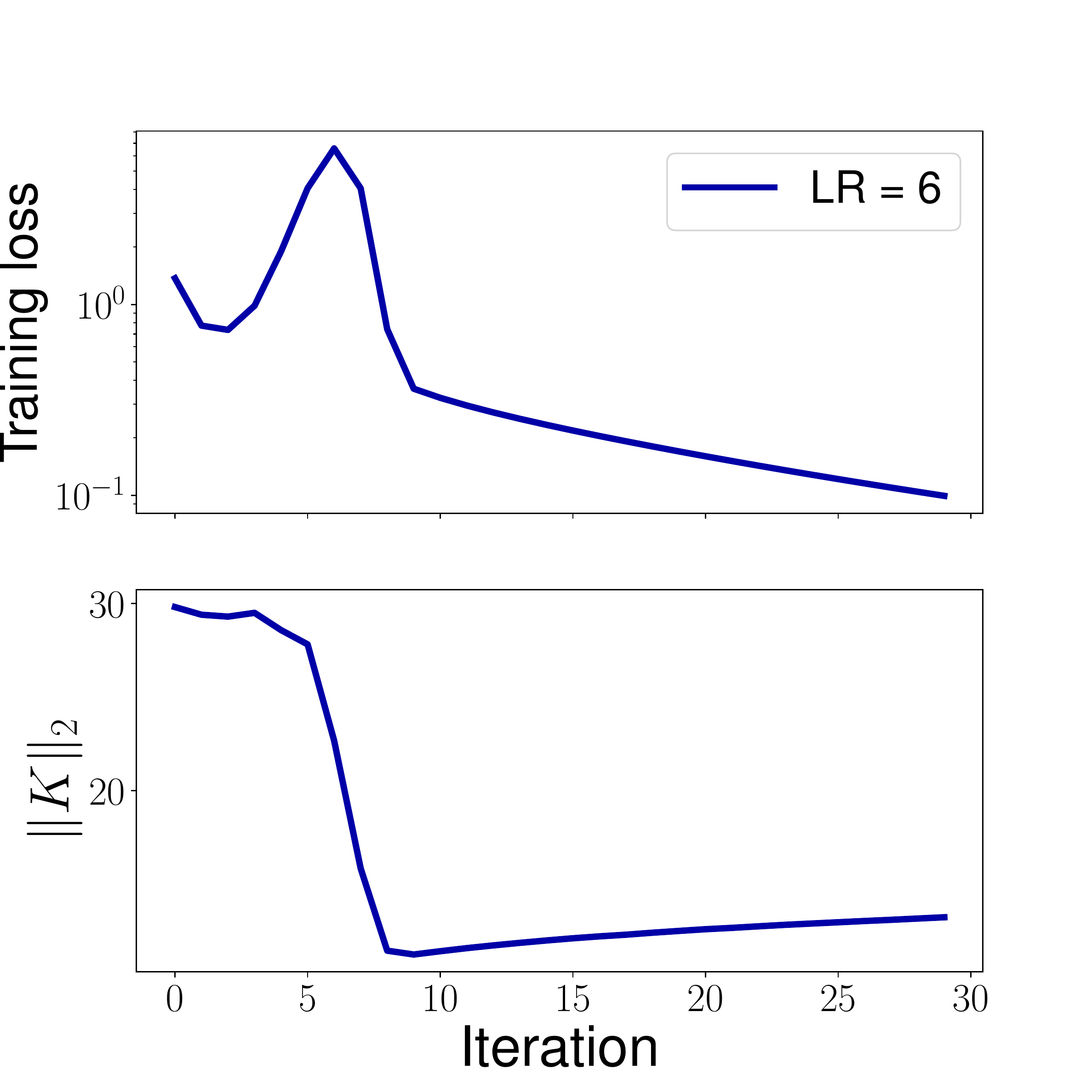}
         \caption{5-layer FCN}
     \end{subfigure}
         %  \begin{subfigure}[b]{0.4\textwidth}
         % \centering
         % \includegraphics[width=\textwidth]{figure/loss_fcn_proj.pdf}
         % \caption{Loss decomposition}
         % \end{subfigure}
    \begin{subfigure}[b]{0.35\textwidth}
         \centering
         \includegraphics[width=\textwidth]{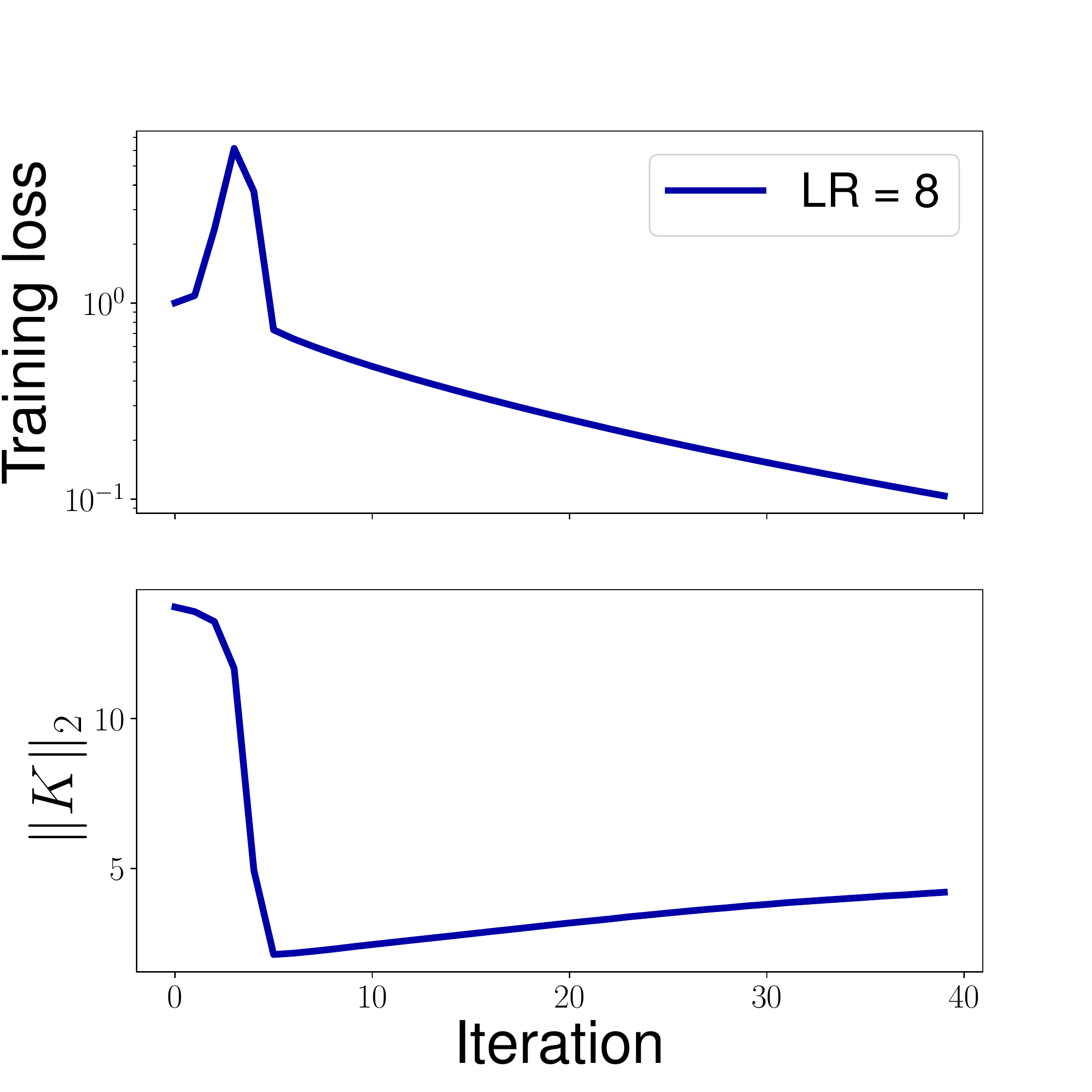}
        \caption{5-layer CNN}
     \end{subfigure}
%           \begin{subfigure}[b]{0.4\textwidth}
%          \centering
% \includegraphics[width=\textwidth]{figure/loss_cnn_proj.pdf}
%           \caption{Loss decomposition}
%      \end{subfigure}
     \caption{{\bf The training loss and the spectral norm of the tangent kernel during catapult for 5-layer FCN (a) and CNN (b) on CIFAR-10 dataset}. Both networks are trained under the same experimental setting with Fig.~\ref{fig:cata_fcn}.    }\label{fig:cata_fcn_cifar}\vspace{-20pt}
\end{figure}

\begin{figure}[H]
     \centering
     \begin{subfigure}[b]{0.4\textwidth}
         \centering
         \includegraphics[width=\textwidth]{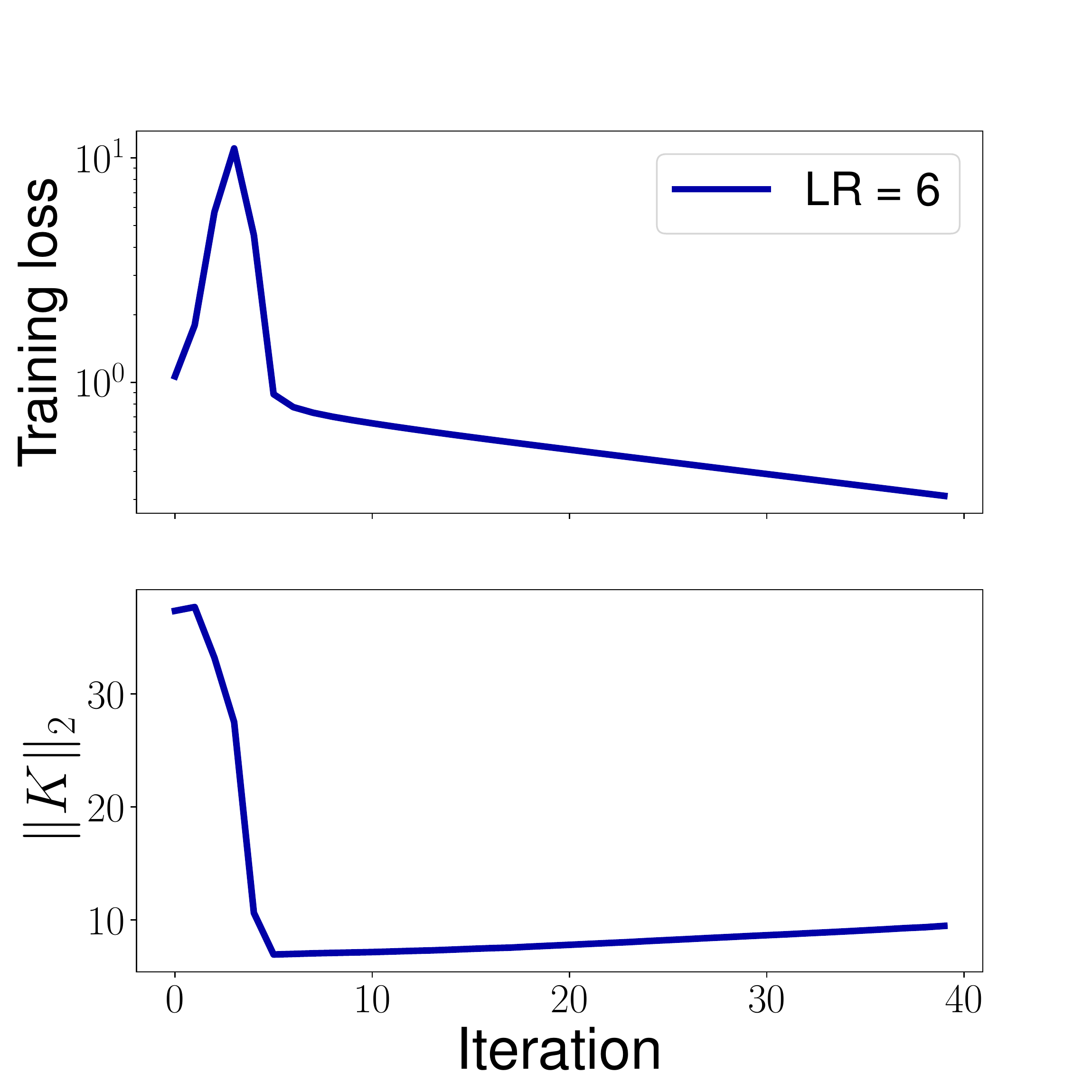}
         \caption{Training loss(FCN)}
     \end{subfigure}
          \begin{subfigure}[b]{0.4\textwidth}
         \centering
         \includegraphics[width=\textwidth]{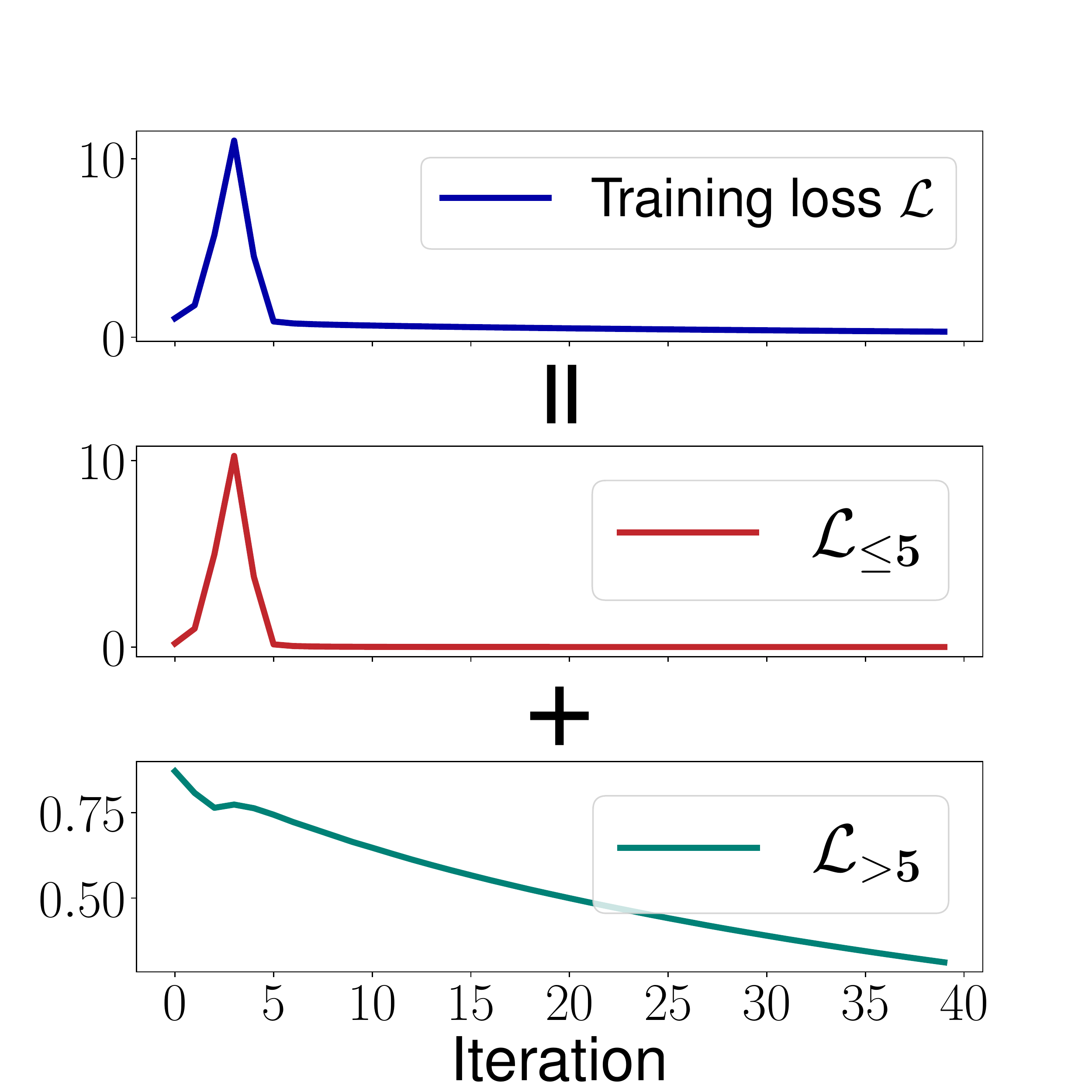}
         \caption{Loss decomposition}
         \end{subfigure}
    \begin{subfigure}[b]{0.4\textwidth}
         \centering
         \includegraphics[width=\textwidth]{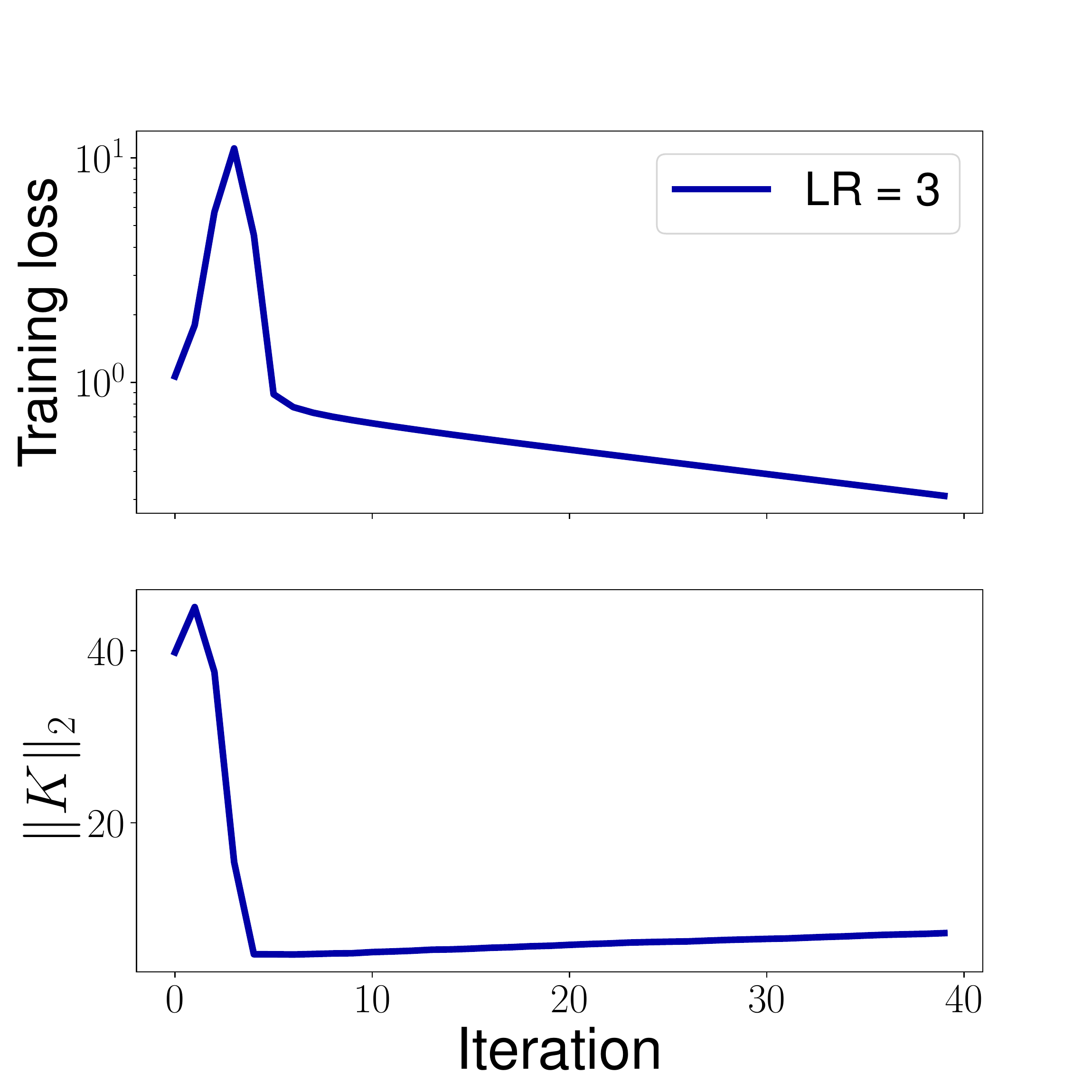}
        \caption{Training loss(CNN)}
     \end{subfigure}
          \begin{subfigure}[b]{0.4\textwidth}
         \centering
         \includegraphics[width=\textwidth]{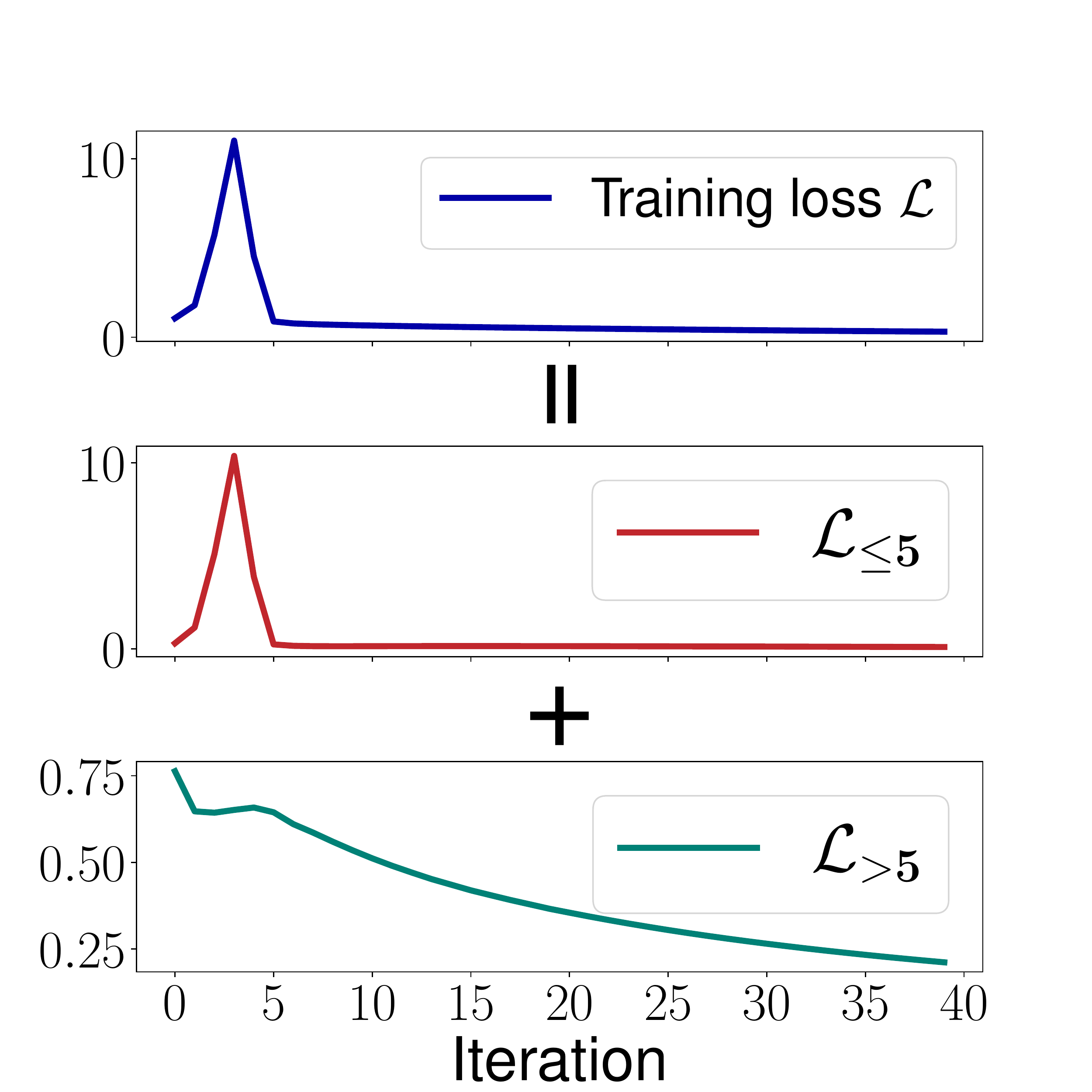}
          \caption{Loss decomposition}
     \end{subfigure}
     \caption{{\bf  Catapult dynamics for 5-layer FCN (a-b) and CNN (c-d) on SVHN dataset}. Panel (a) and (c) are the training loss and the spectral norm of the tangent kernel with learning rate $6.0$ and $3.0$ respectively,  and Panel (b) and (d) are the training loss decomposed into the top eigendirections of the tangent kernel, $\L_{\leq 5}$ and the remaining eigendirections, $\L_{> 5}$.  All the networks are trained on a subset of SVHN with $128$ data points. In this experiment, the critical learning rates for FCN and CNN are $3.4$ and $1.6$  respectively.\label{fig:cata_fcn_svhn} }\vspace{-10pt}
\end{figure}

\begin{figure}[H]
     \centering
     \begin{subfigure}[b]{0.4\textwidth}
         \centering
         \includegraphics[width=\textwidth]{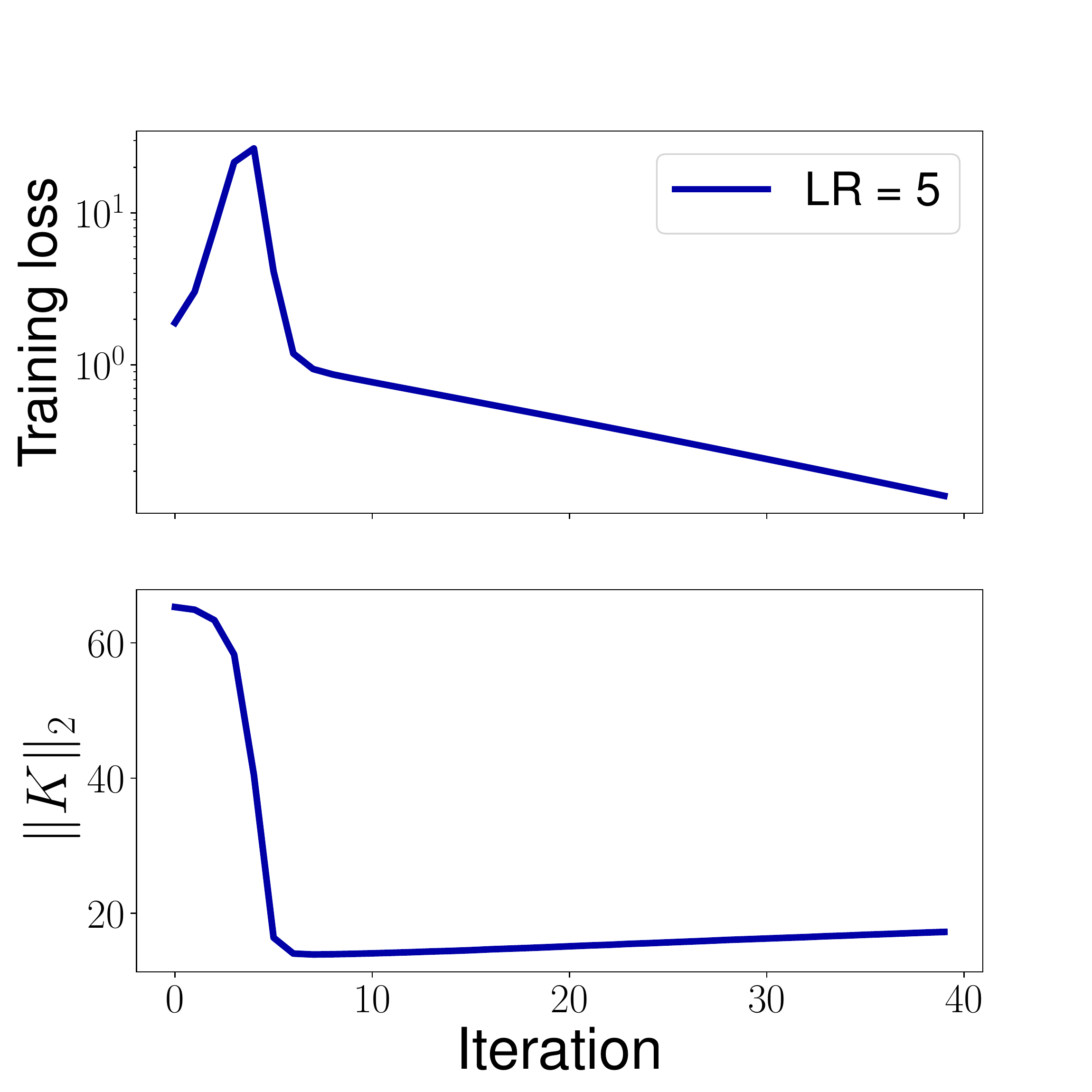}
         \caption{Training loss(FCN)}
     \end{subfigure}
          \begin{subfigure}[b]{0.4\textwidth}
         \centering
         \includegraphics[width=\textwidth]{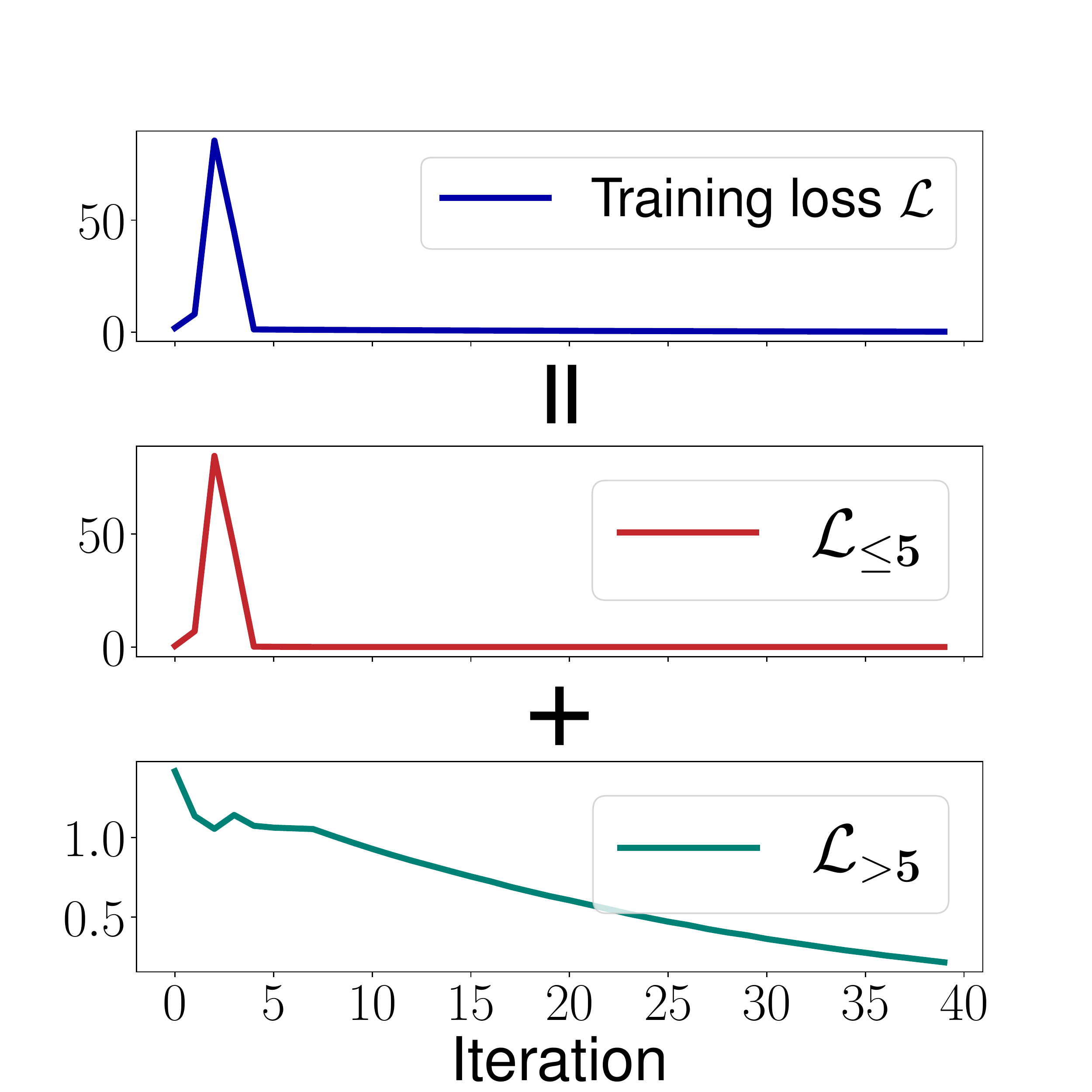}
         \caption{Loss decomposition}
         \end{subfigure}
    \begin{subfigure}[b]{0.4\textwidth}
         \centering
         \includegraphics[width=\textwidth]{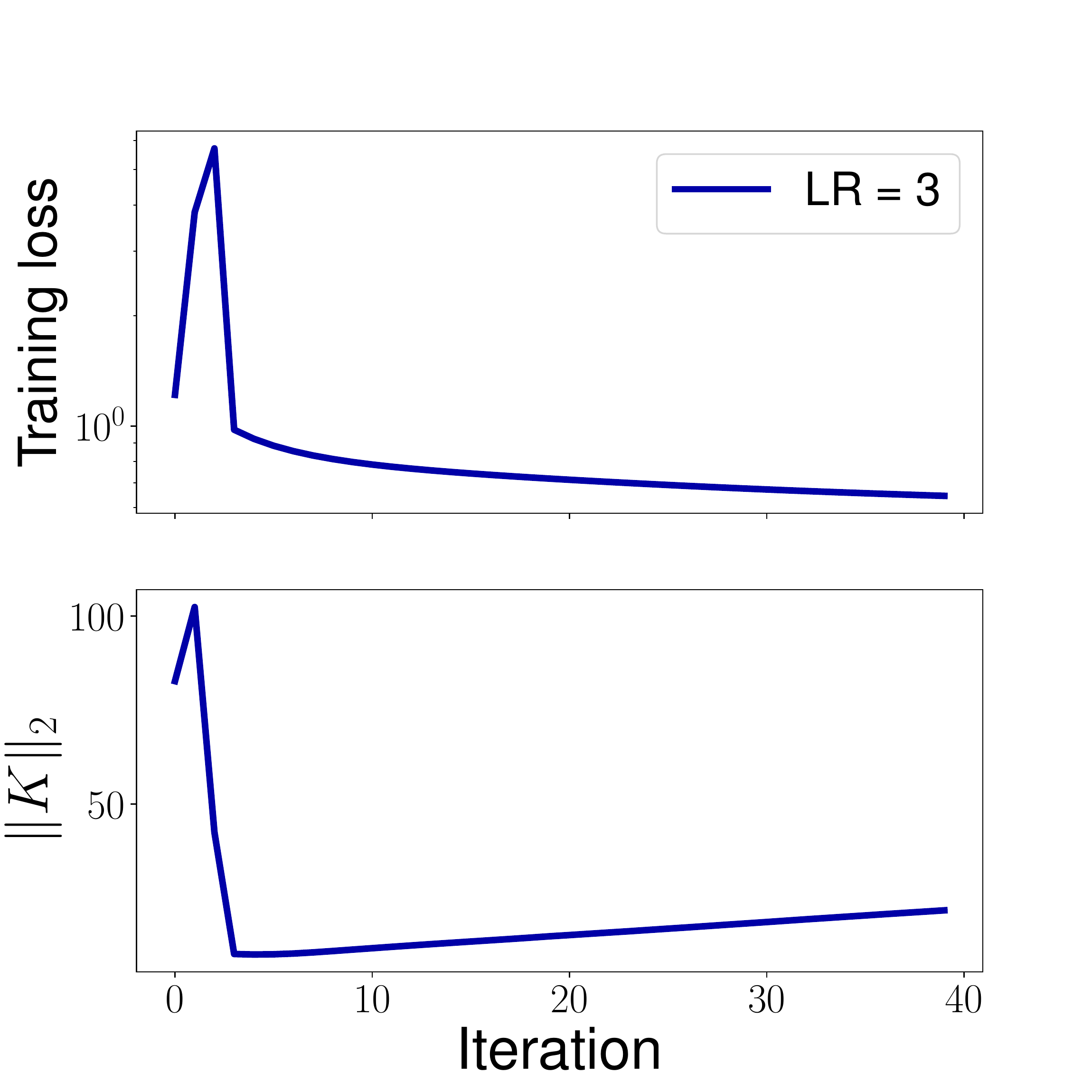}
        \caption{Training loss(WRN)}
     \end{subfigure}
          \begin{subfigure}[b]{0.4\textwidth}
         \centering
         \includegraphics[width=\textwidth]{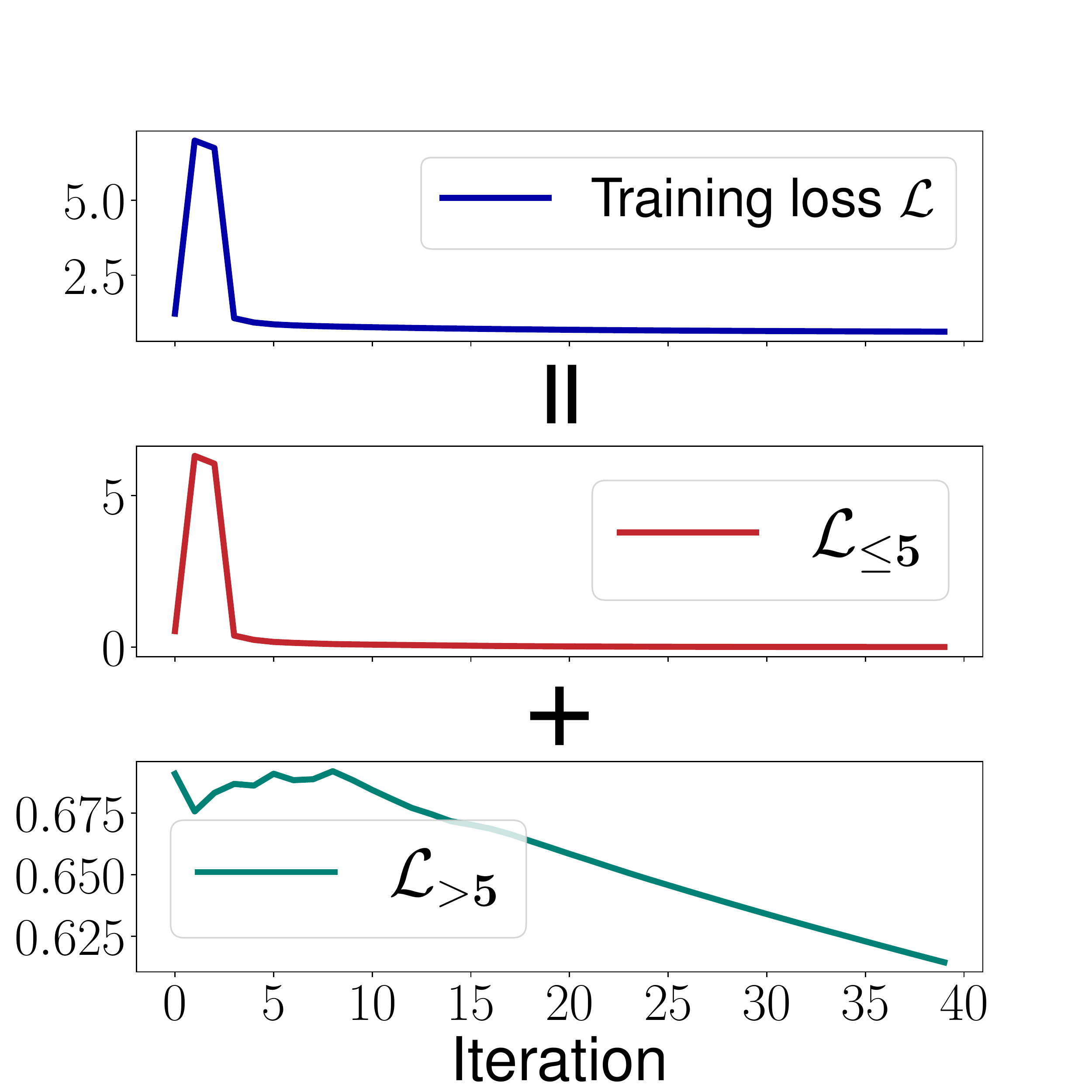}
          \caption{Loss decomposition}
     \end{subfigure}
     \caption{{\bf  Catapult dynamics for FCN (a-b) on a synthetic dataset and Wide ResNets 10-10 (c-d) on CIFAR-10 dataset}. Panel (a) and (c) are the training loss and the spectral norm of the tangent kernel with learning rates $5.0$ and $3.0$ respectively,  and Panel (b) and (d) are the training loss decomposed into the top eigendirections of the tangent kernel, $\L_{\leq 5}$ and the remaining eigendirections, $\L_{> 5}$.   For the synthetic dataset, we use the rank-2 regression task considered in Section~\ref{sec:feature_learning}. The size of the training set is $128$.  In this experiment, the critical learning rates for FCN and WRN are $1.9$ and $1.5$ respectively. }\label{fig:cata_wrn}\vspace{-10pt}
\end{figure}

From the experimental results, we can see that for a large learning rate that causes catapult dynamics, the loss spike occurs in the top eigenspace of the tangent kernel. 
% Furthermore, the magnitude of the loss spike decreases in $\L_{>s}$ as $s$ increases.   
See Fig.~\ref{fig:cata_fcn_cifar} for 5-layer FCN and CNN on CIFAR-10 dataset and \ref{fig:cata_fcn_svhn} on SVHN dataset,  and \ref{fig:cata_wrn} for Wide-ResNets on CIFAR-10 dataset.

\begin{figure}[H]
     \centering
     \begin{subfigure}[b]{0.4\textwidth}
         \centering
         \includegraphics[width=\textwidth]{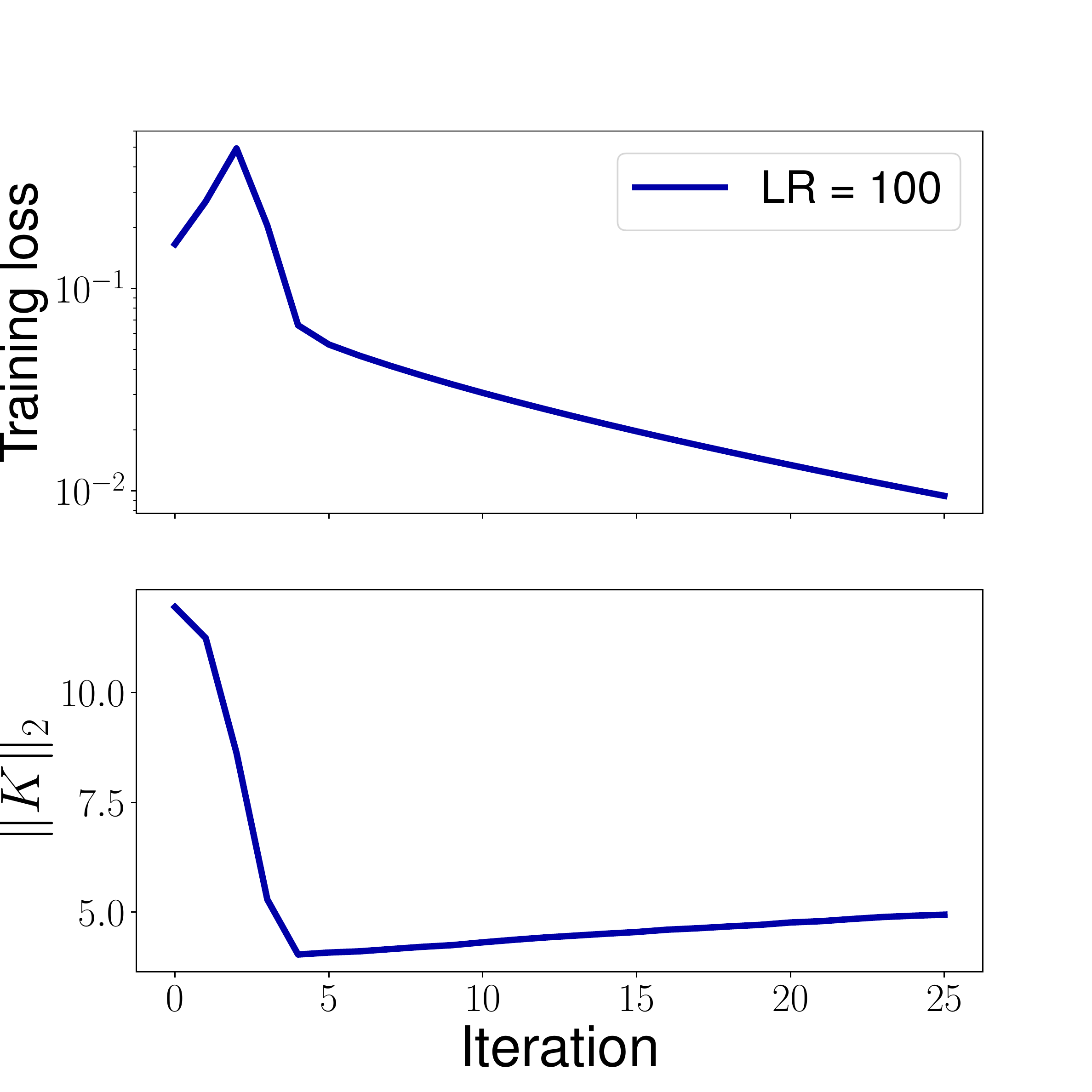}
         \caption{Training loss(FCN)}
     \end{subfigure}
          \begin{subfigure}[b]{0.4\textwidth}
         \centering
         \includegraphics[width=\textwidth]{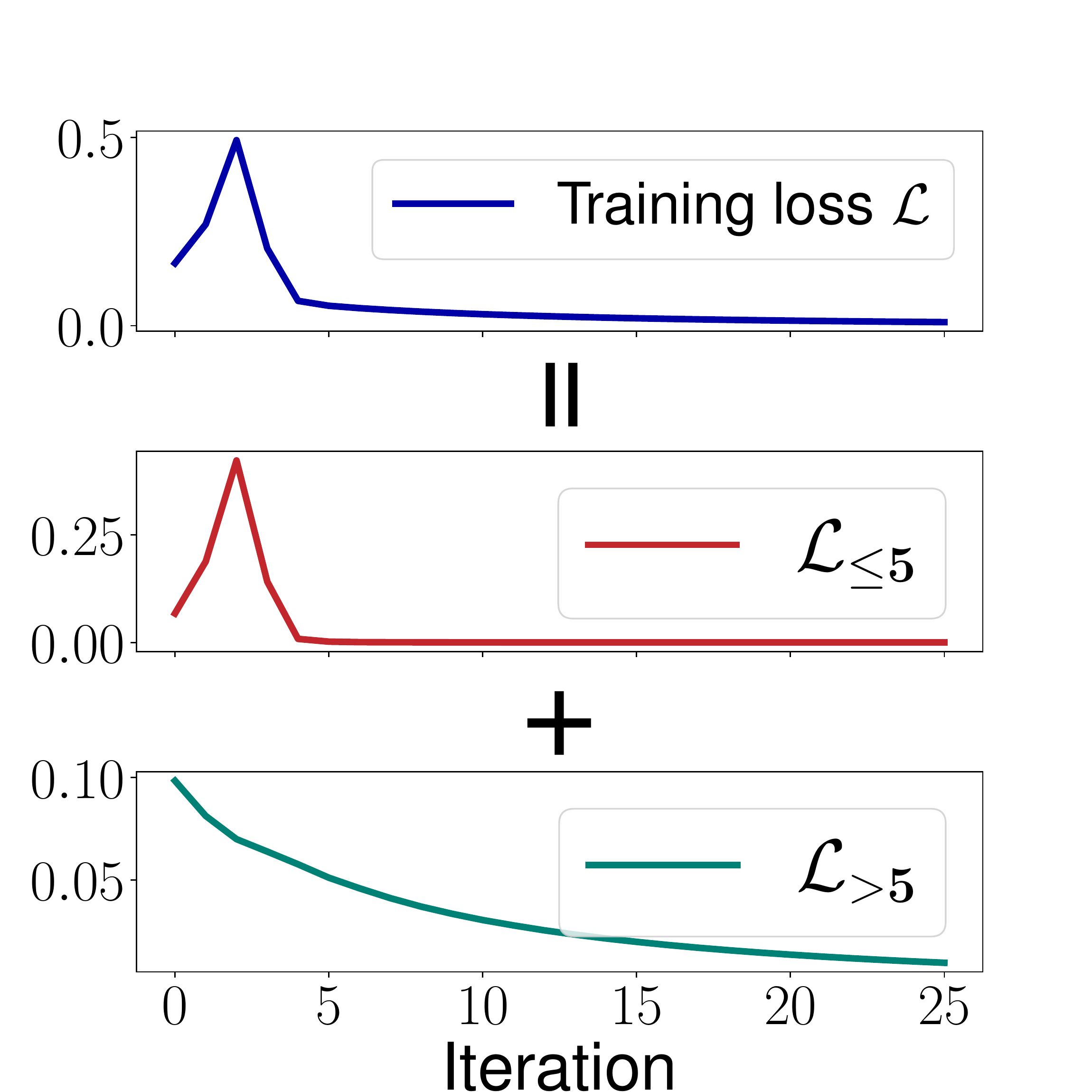}
         \caption{Loss decomposition}
         \end{subfigure}
    \begin{subfigure}[b]{0.4\textwidth}
         \centering
         \includegraphics[width=\textwidth]{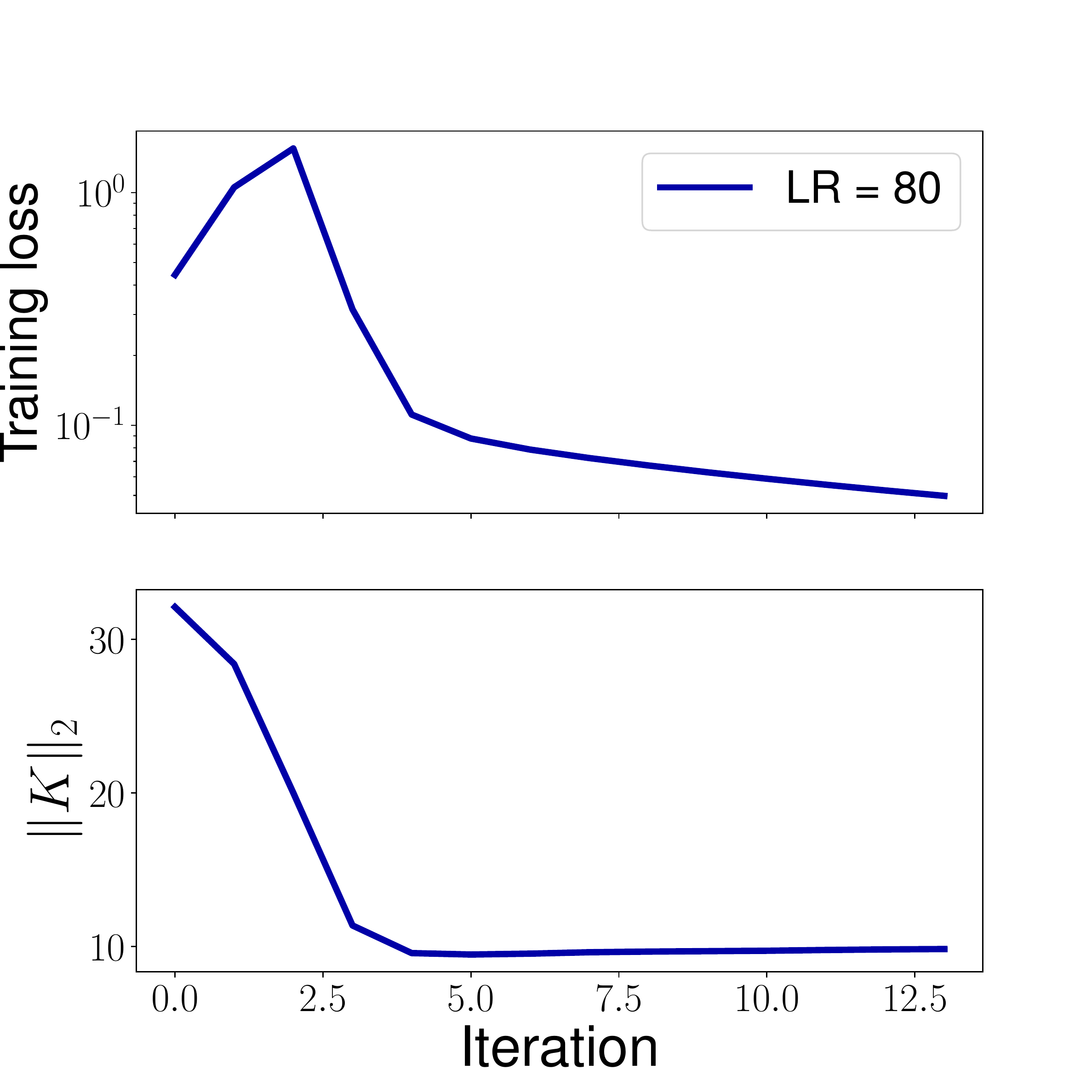}
        \caption{Training loss(CNN)}
     \end{subfigure}
          \begin{subfigure}[b]{0.4\textwidth}
         \centering
         \includegraphics[width=\textwidth]{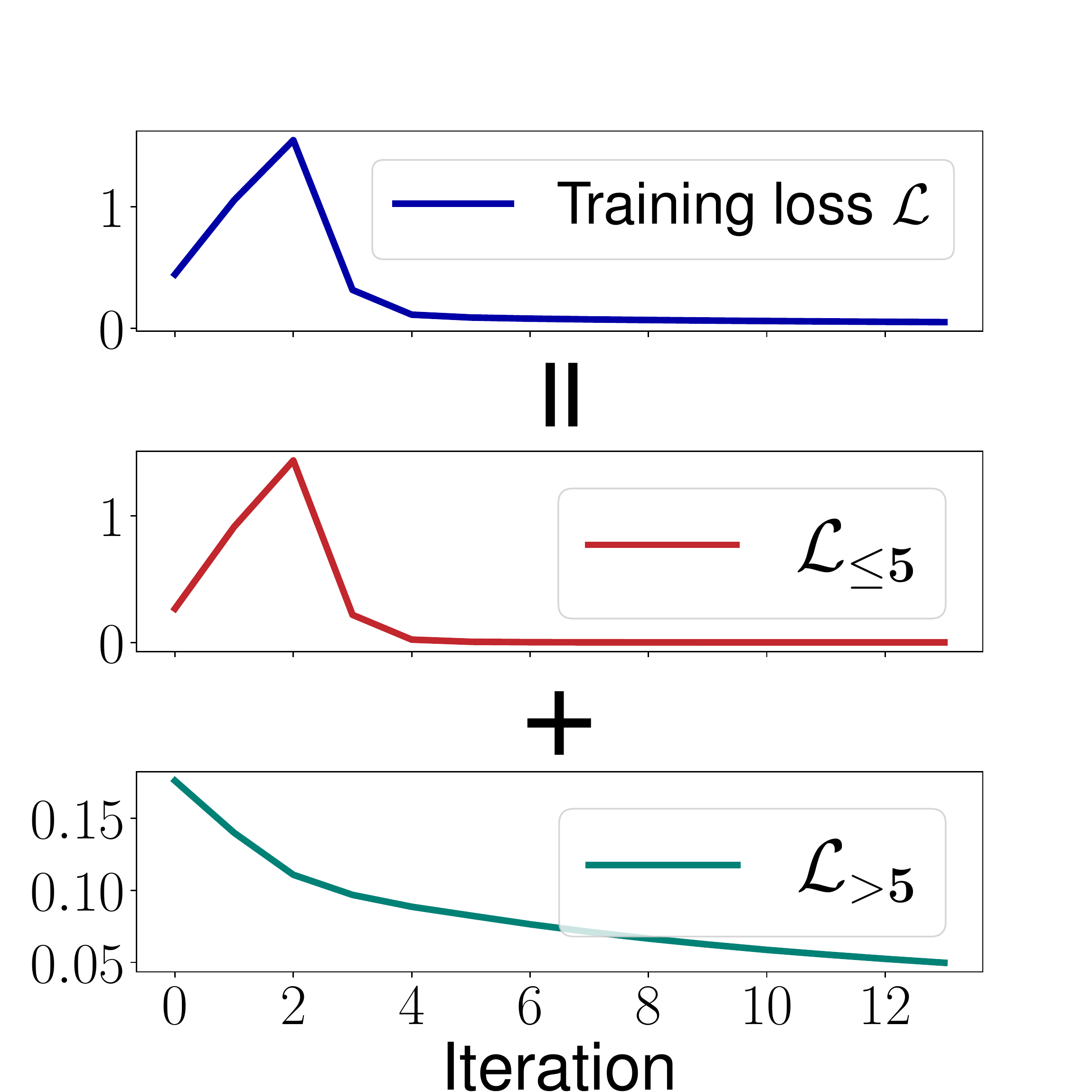}
          \caption{Loss decomposition}
     \end{subfigure}
     \caption{{\bf  Catapult dynamics for 5-layer FCN (a-b) and CNN (c-d) on multiclass classification tasks.} Panel (a) and (c) are the training loss and the spectral norm of the tangent kernel with learning rate $100$ and $80$ respectively,   and Panel (b) and (d) are the training loss decomposed into the top eigendirections of the tangent kernel, $\L_{\leq 5}$ and the remaining eigendirections, $\L_{> 5}$.  All the networks are trained on a subset of CIFAR-10 with 10 classes. Here the dimension of the eigenspace $s=1,3,5$ refers to $10,30,50$ respectively due to the output dimension $10$. The critical learning rate for FCN and CNN are $34$ and $16$ respectively. }\label{fig:cata_fcn_full}
\end{figure}

We further show  Claim~\ref{claim: Claim 1} holds for multidimensional outputs in Fig.~\ref{fig:cata_fcn_full}. In particular, for $k$-class classification tasks, we project the flattened vector of predictions of size $kn$ to the top eigenspaces of the empirical NTK, which is of size $kn \times kn$.
Correspondingly, we empirically observe that catapults occur in the top $ks$ eigenspace with a small $s$. 

\subsection{Multiple catapults in GD occur in the top eigenspace of NTK}
\begin{figure}[H]
     \centering
     \begin{subfigure}[b]{0.4\textwidth}
         \centering
         \includegraphics[width=\textwidth]{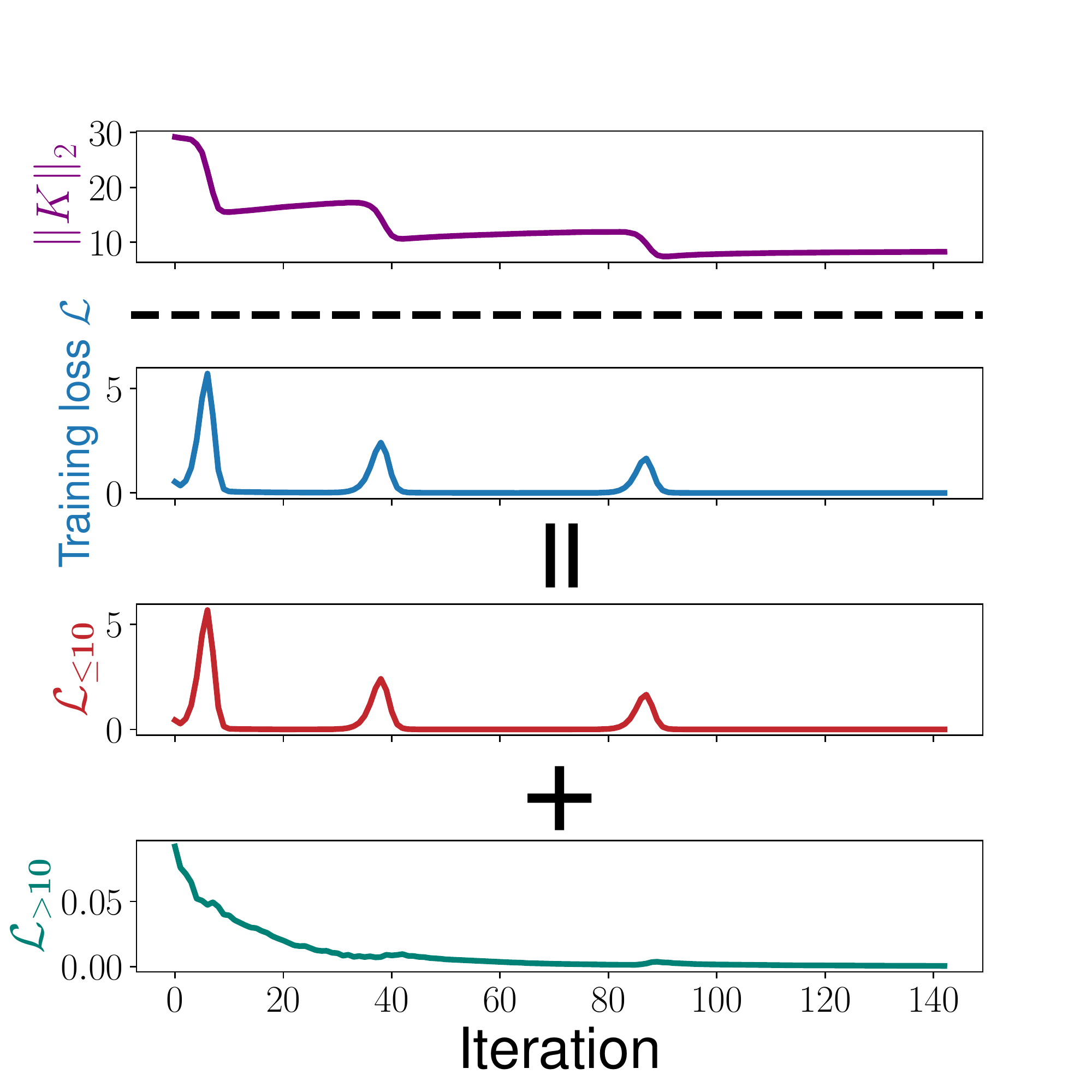}\caption{5-layer FCN}
     \end{subfigure}
          \begin{subfigure}[b]{0.4\textwidth}
         \centering
         \includegraphics[width=\textwidth]{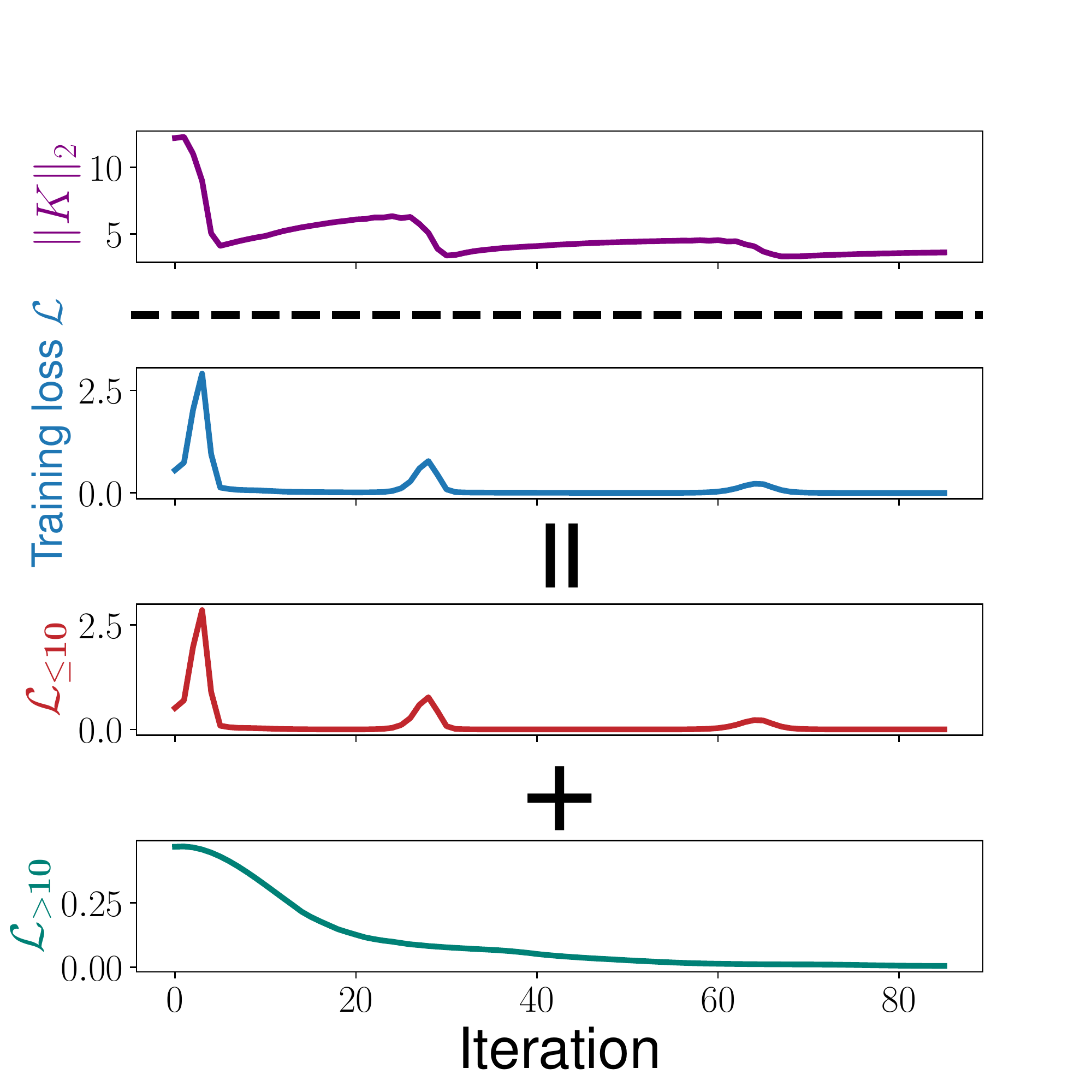}\caption{5-layer CNN}
     \end{subfigure}
     \caption{{\bf Multiple catapults in GD with increased learning rates.} With the same setting of Fig.~\ref{fig:multi_cata}, the training loss is decomposed into the top eigendirections of the tangent kernel, $\L_{\leq 10}$ and the remaining eigendirections, $\L_{> 10}$.}    \label{fig:multi_cata_loss_decomp}
\end{figure}

For the multiple catapults shown in Fig.~\ref{fig:multi_cata}, similar to a single catapult, we show that the catapults occur in the top eigenspace of NTK. See Fig.~\ref{fig:multi_cata_loss_decomp}.

\subsection{Multiple catapults allow a larger learning rate at convergence}\label{app:multi_cata}

Corresponding to Fig.~\ref{fig:multi_cata} in Section~\ref{subsec:multi_cata}, we show that if the neural networks are trained with the learning rate at the convergence, i.e., after multiple catapults, the GD will diverge.

\begin{figure}[H]
     \centering
     % \begin{subfigure}[b]{0.4\textwidth}
     %     \centering
     %     \includegraphics[width=\textwidth]{figure/lr_match_prec.pdf}\caption{Sign match between $\eta-\etc$ and $\P\Diff_1$}
     % \end{subfigure}
     \begin{subfigure}[b]{0.4\textwidth}
         \centering
         \includegraphics[width=\textwidth]{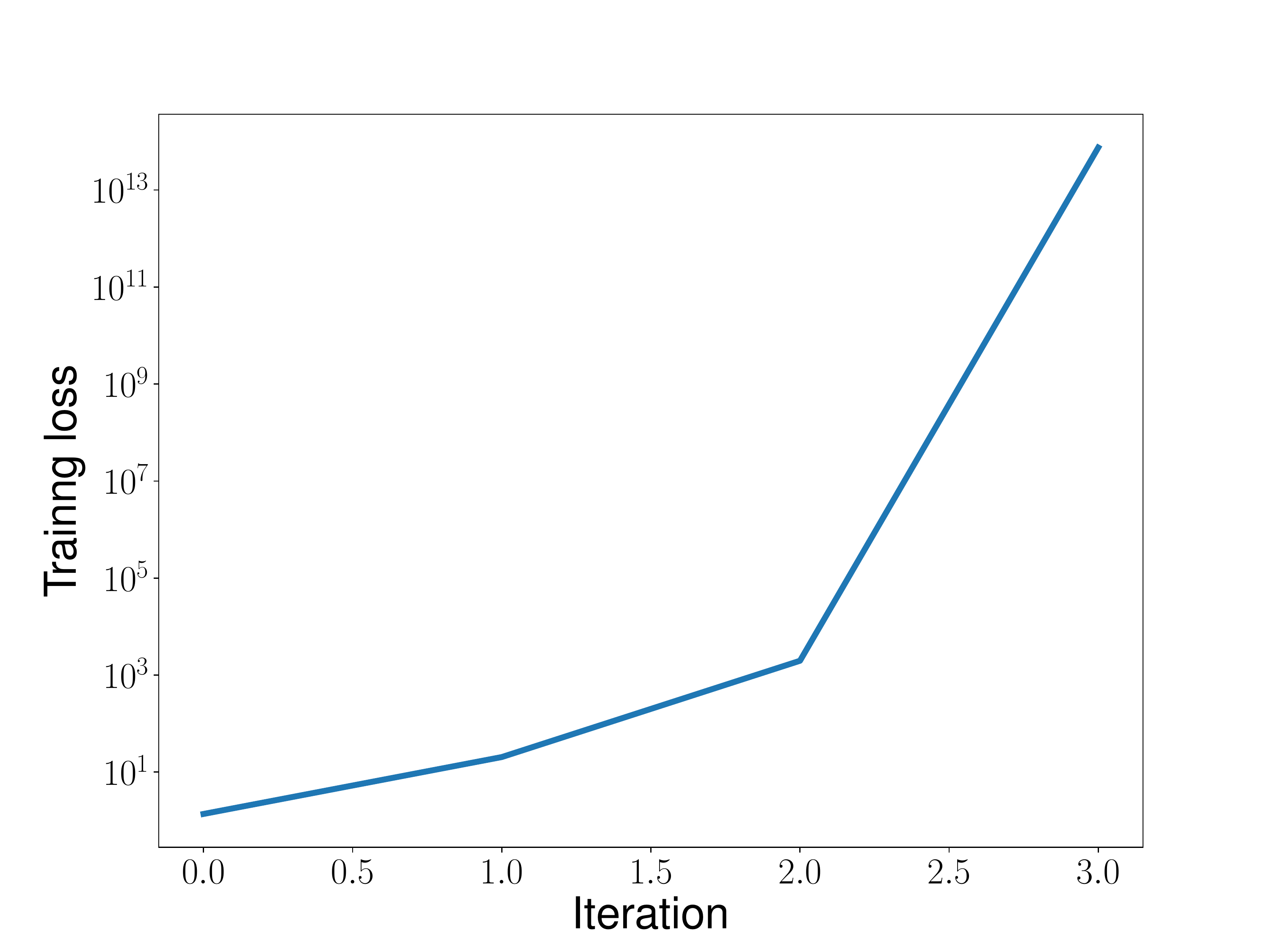}\caption{5-layer FCN}
     \end{subfigure}
          \begin{subfigure}[b]{0.4\textwidth}
         \centering
         \includegraphics[width=\textwidth]{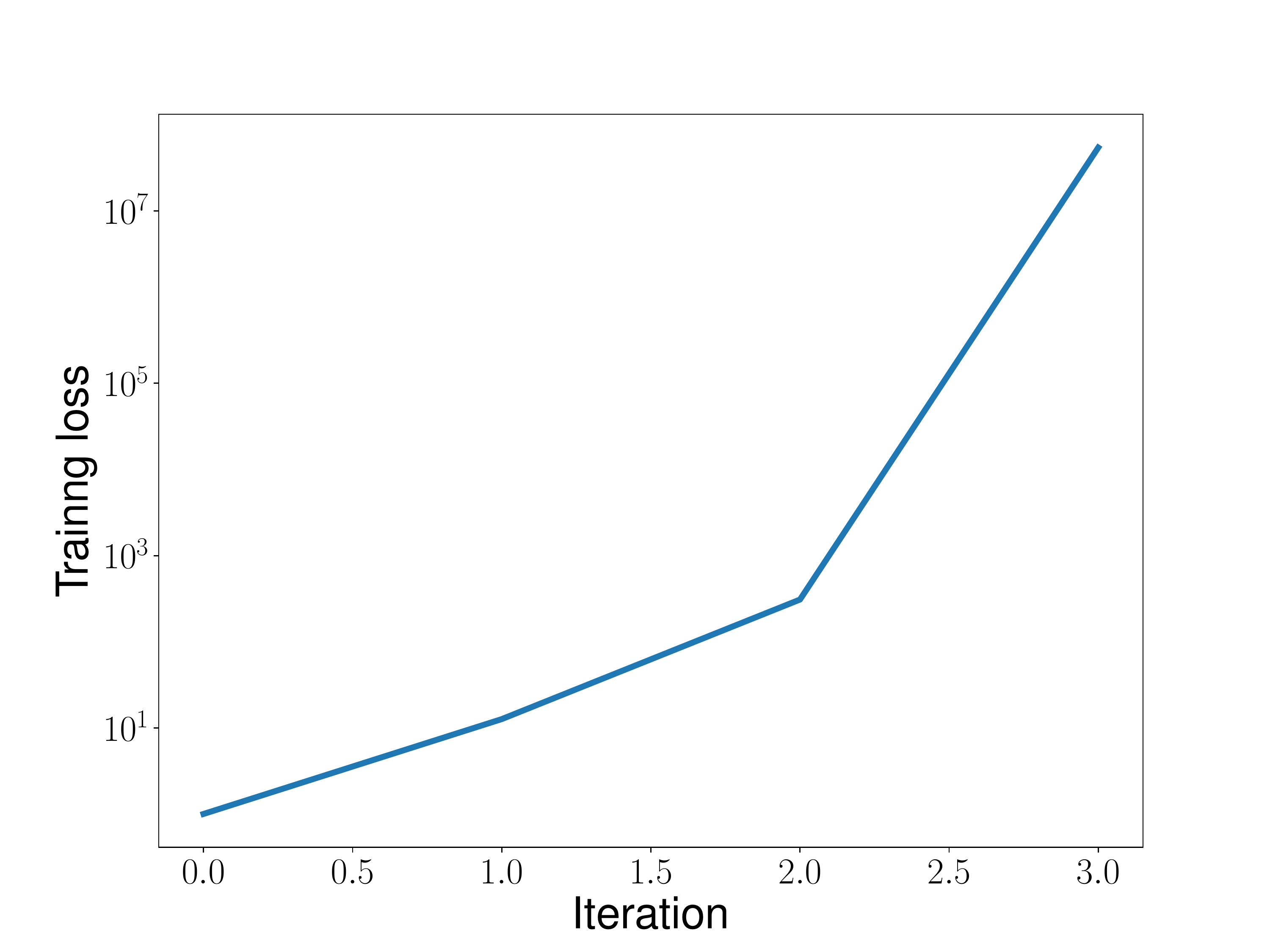}\caption{5-layer CNN}
     \end{subfigure}
     \caption{{\bf GD diverges when trained with the learning rate after multiple catapults.}  Corresponding to Fig.~\ref{fig:multi_cata}, we train the model using GD with learning rate at convergence, $60$ and $40$ respectively for the 5-layer FCN and CNN. }\label{fig:multi_cata_diverge}
\end{figure}

\section{Additional experiments for catapults in SGD}\label{sec:sgd_add}

\subsection{Full training process visualization corresponding to Fig.~\ref{fig:sgd_deep}}

\begin{figure}[H]
     \centering
       \begin{subfigure}[b]{0.4\textwidth}
         \centering
         \includegraphics[width=\textwidth]{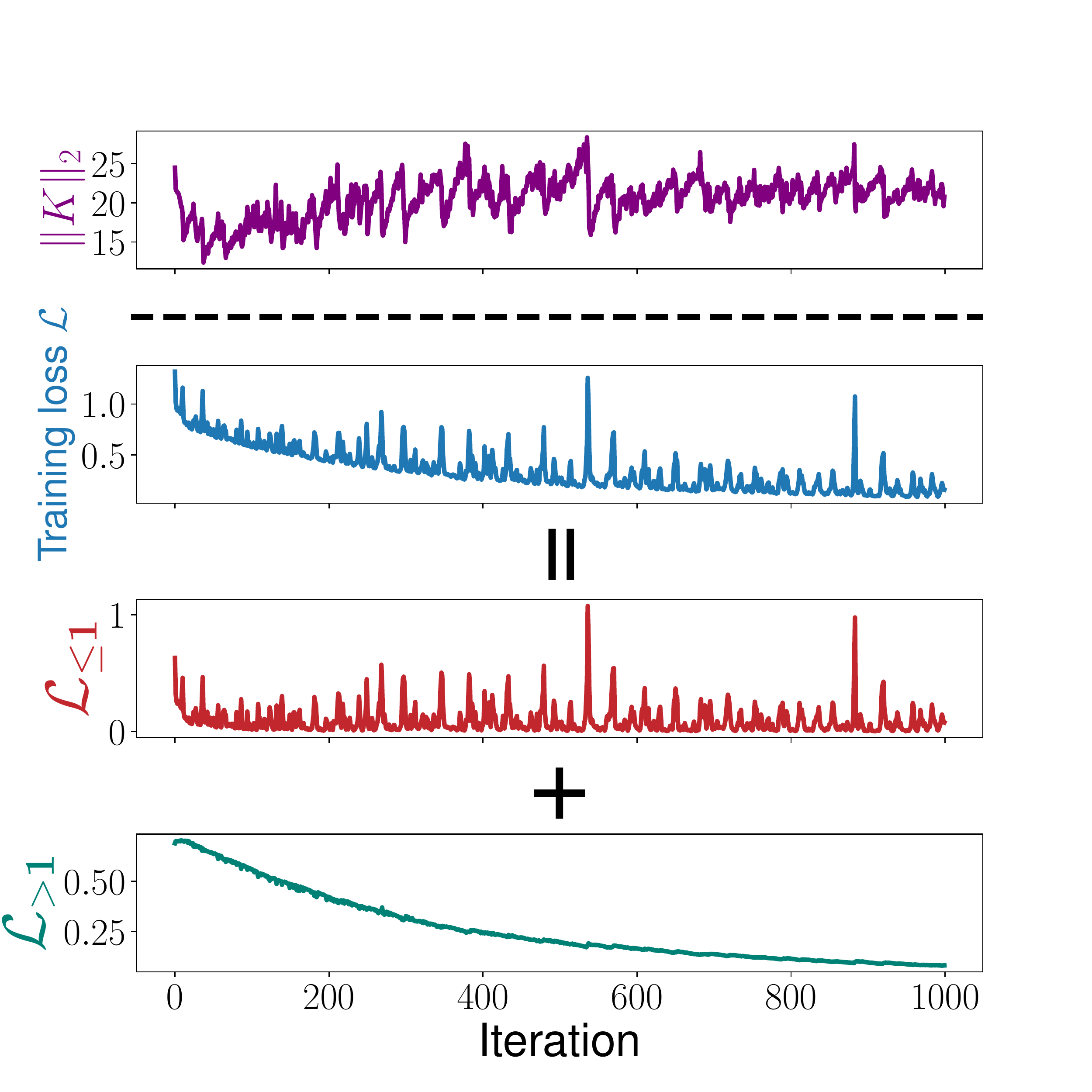}
         \caption{Wide ResNets 10-10}
     \end{subfigure}
    \begin{subfigure}[b]{0.4\textwidth}
         \centering
         \includegraphics[width=\textwidth]{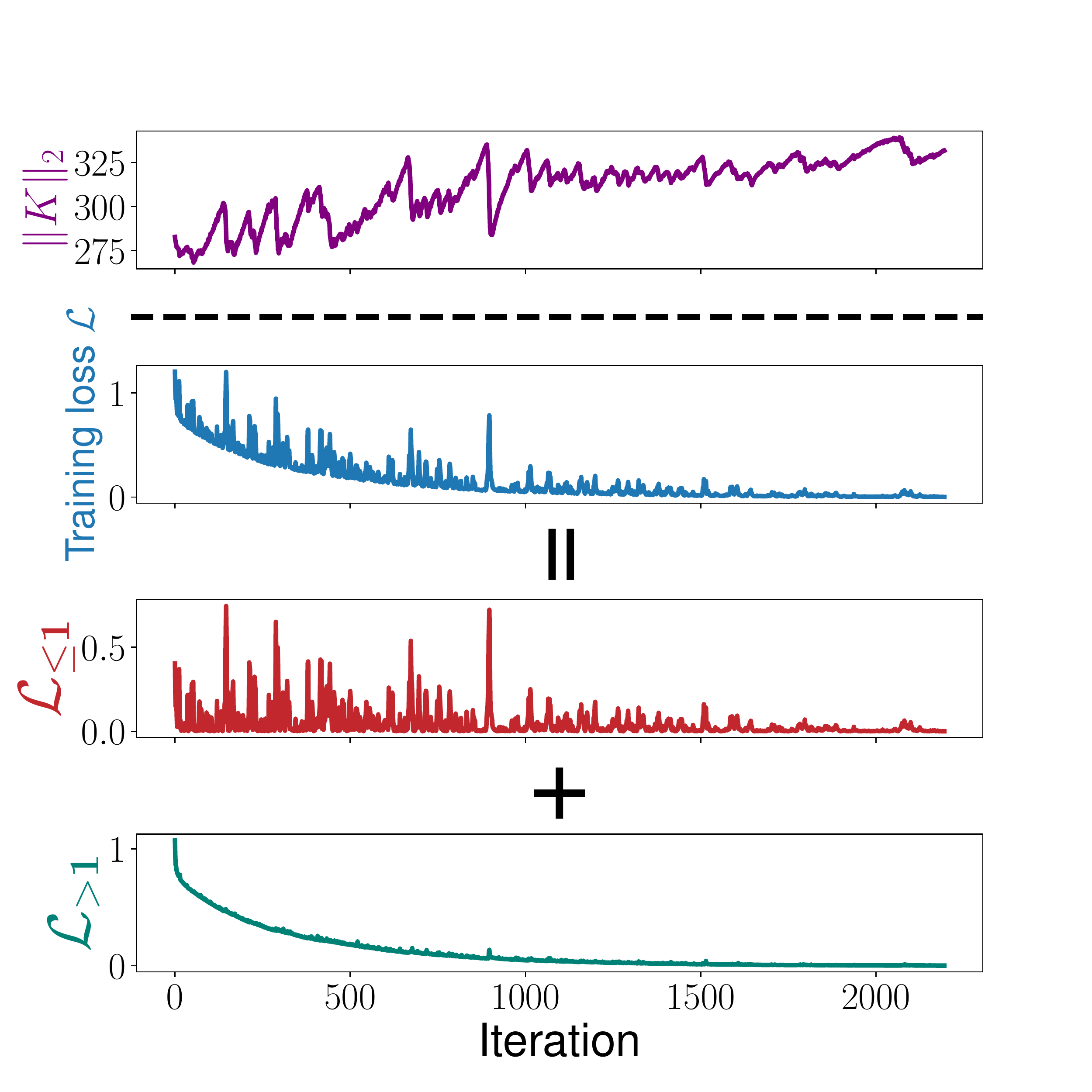}
         \caption{ViT-4}
     \end{subfigure}
     \caption{{\bf Cataput dynamics in SGD for modern deep architectures.} The complete versions corresponding to Fig.~\ref{fig:sgd_deep}(c,d). The training loss is decomposed based on the eigendirections of the NTK: $\L_{\leq 1}$ and $\L_{>1}$.}\label{fig:sgd_deep_complete}
\end{figure}

We present the complete training loss and the spectrum norm of the NTK corresponding to Fig.~\ref{fig:sgd_deep}(c,d) in Fig.~\ref{fig:sgd_deep_complete}.

\subsection{Catapults in SGD with Pytorch default parameterization}

\begin{figure}[H]
     \centering
     \begin{subfigure}[b]{0.4\textwidth}
         \centering
         \includegraphics[width=\textwidth]{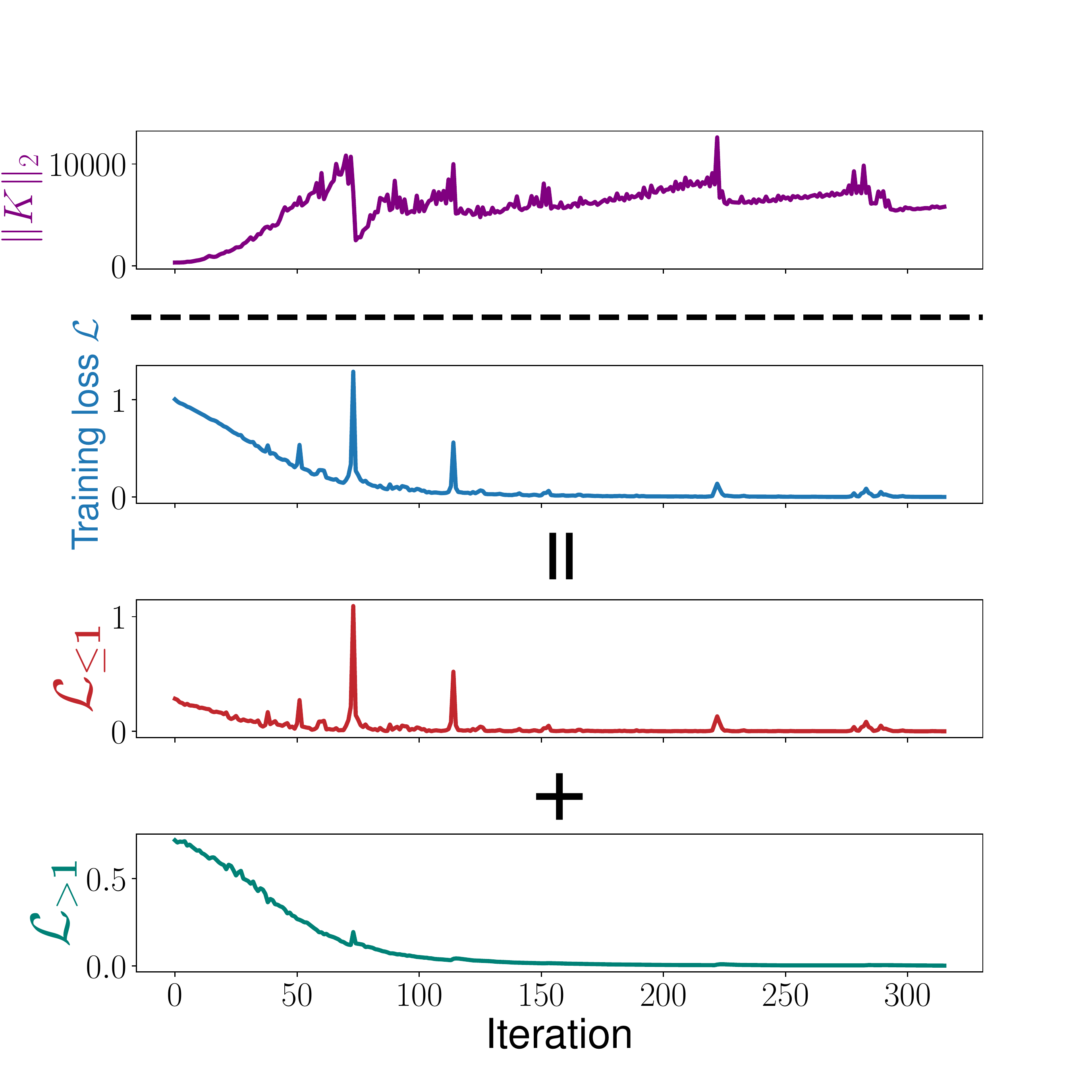}
         \caption{FCN}
     \end{subfigure}
    \begin{subfigure}[b]{0.4\textwidth}
         \centering
         \includegraphics[width=\textwidth]{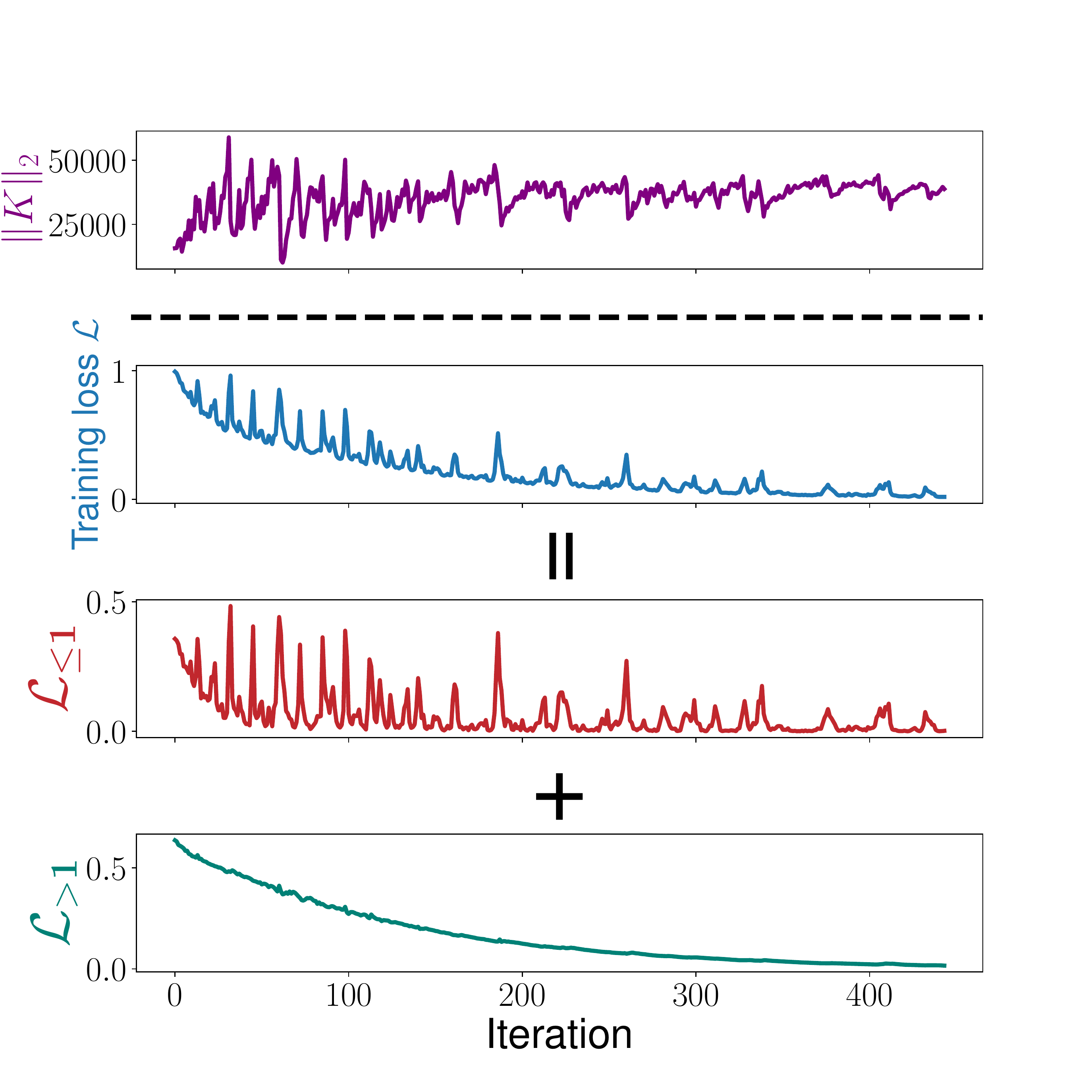}
         \caption{CNN}
     \end{subfigure}
       \begin{subfigure}[b]{0.4\textwidth}
         \centering
         \includegraphics[width=\textwidth]{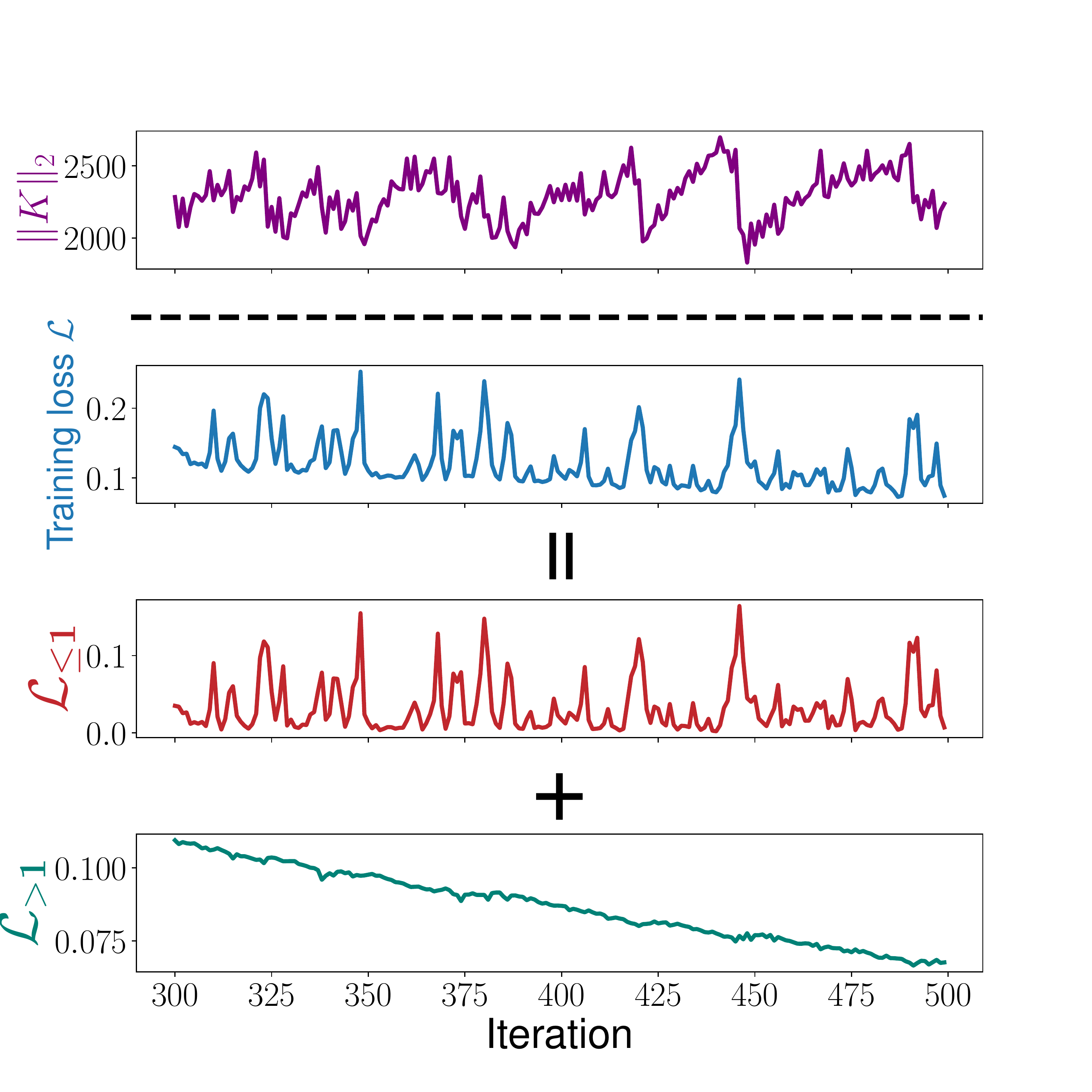}
         \caption{Wide ResNets 10-10}
     \end{subfigure}
    \begin{subfigure}[b]{0.4\textwidth}
         \centering
         \includegraphics[width=\textwidth]{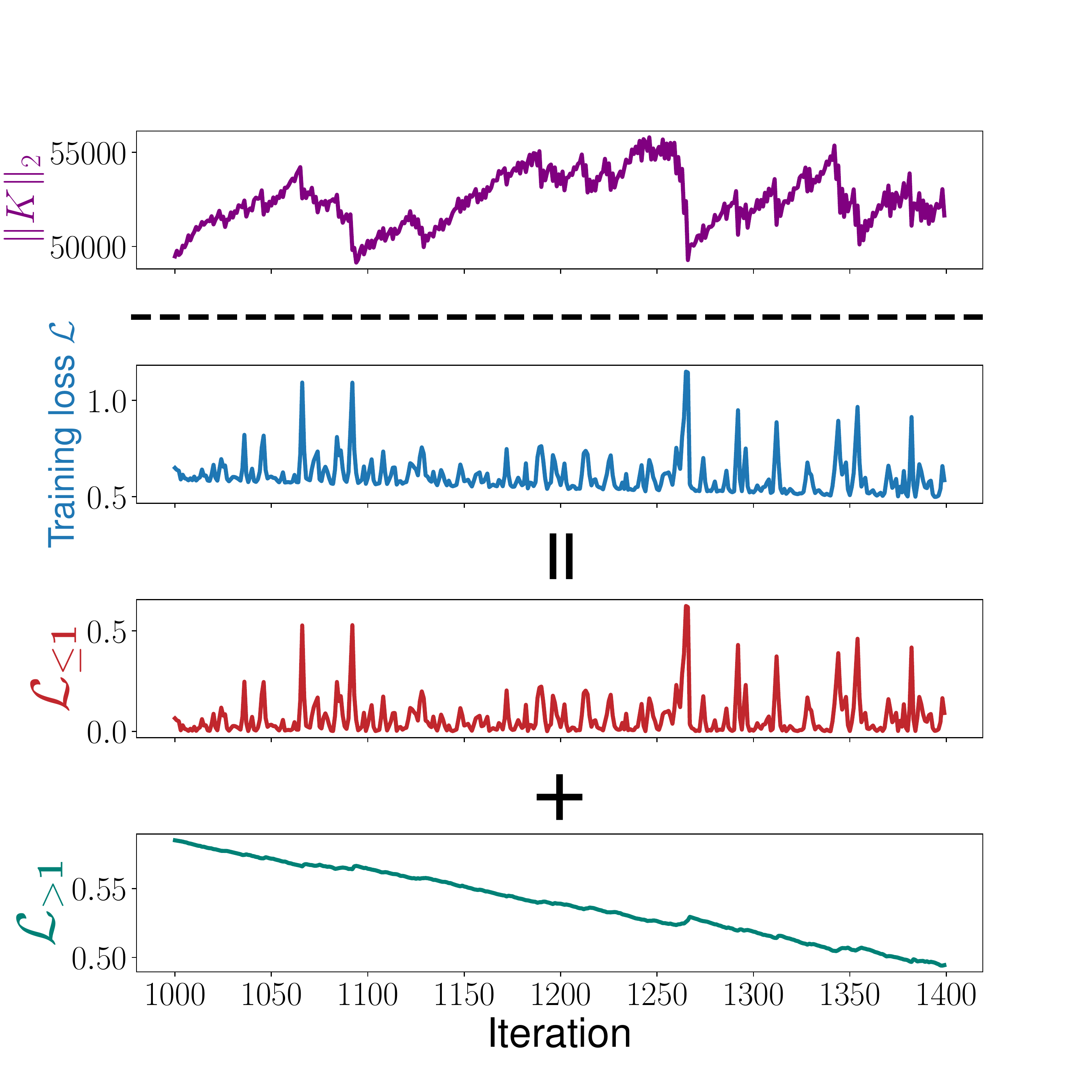}
         \caption{ViT-4}
     \end{subfigure}
     \caption{{\bf Cataput dynamics in SGD for modern deep architectures with Pytorch default parameterization.} The tasks are the same with Fig.~\ref{fig:sgd_deep} except that we use Pytorch default parameterization. The training loss is decomposed based on the eigendirections of the NTK: $\L_{\leq 1}$ and $\L_{>1}$. }\label{fig:sgd_deep_std}
\end{figure}

In Fig.~\ref{fig:sgd_deep}, we used NTK parameterization (see the definition in Appendix~\ref{sec:exp_details}) for the neural networks.  We further validate our empirical observations on (1) the occurrence of the loss spikes of SGD in the top eigenspace of the tangent kernel and (2)  the decrease in the spectral norm of the tangent kernel during loss spikes in the setting with Pytorch default parameterization,  under which the wide networks are still close to their linear approximations~\citep{liu2020linearity,yang2021tensor} in Fig.~\ref{fig:sgd_deep_std}.

\subsection{Catapults in SGD with additional datasets}
\begin{figure}[H]
     \centering
     \begin{subfigure}[b]{0.4\textwidth}
         \centering
         \includegraphics[width=\textwidth]{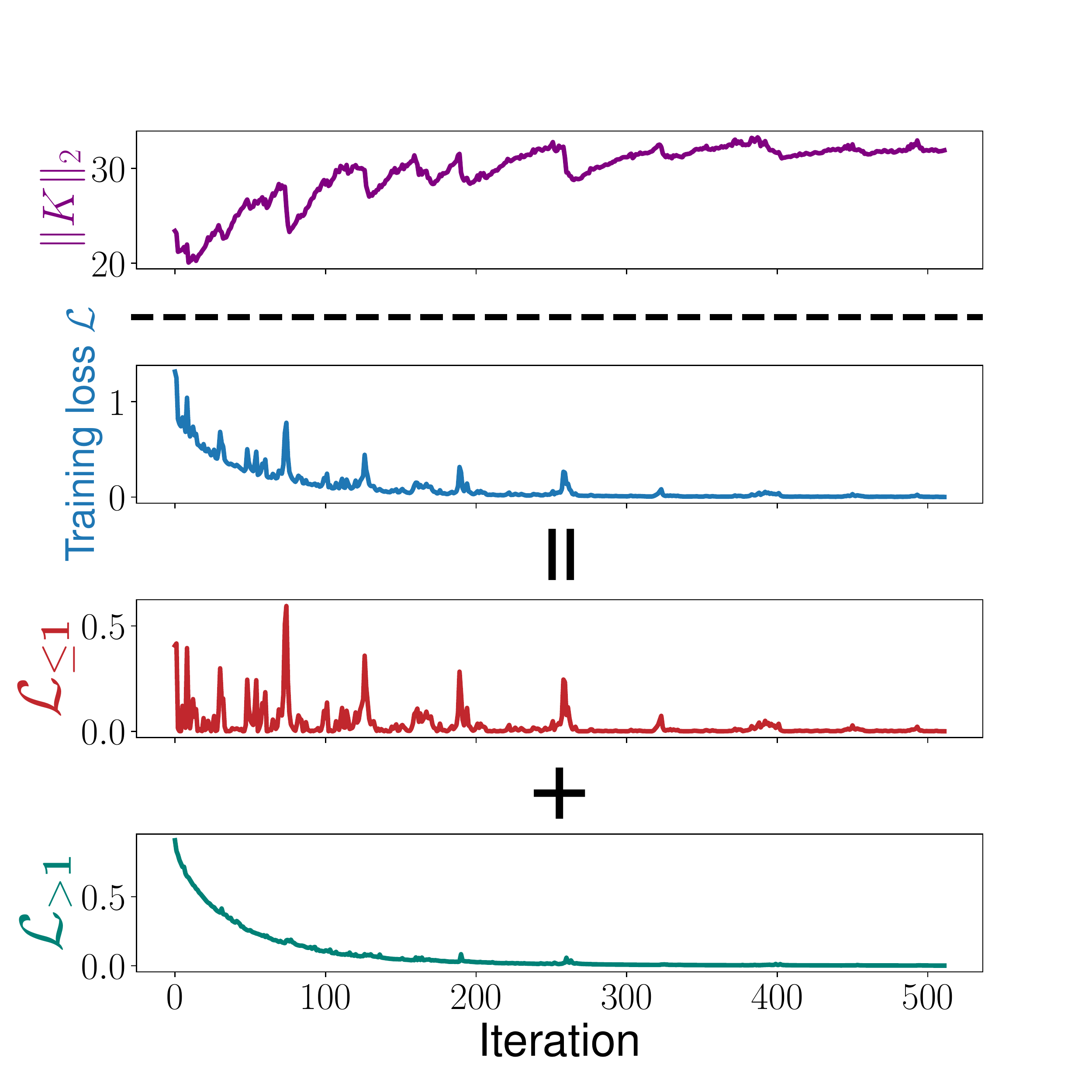}
         \caption{5-layer FCN}
     \end{subfigure}
    \begin{subfigure}[b]{0.4\textwidth}
         \centering
         \includegraphics[width=\textwidth]{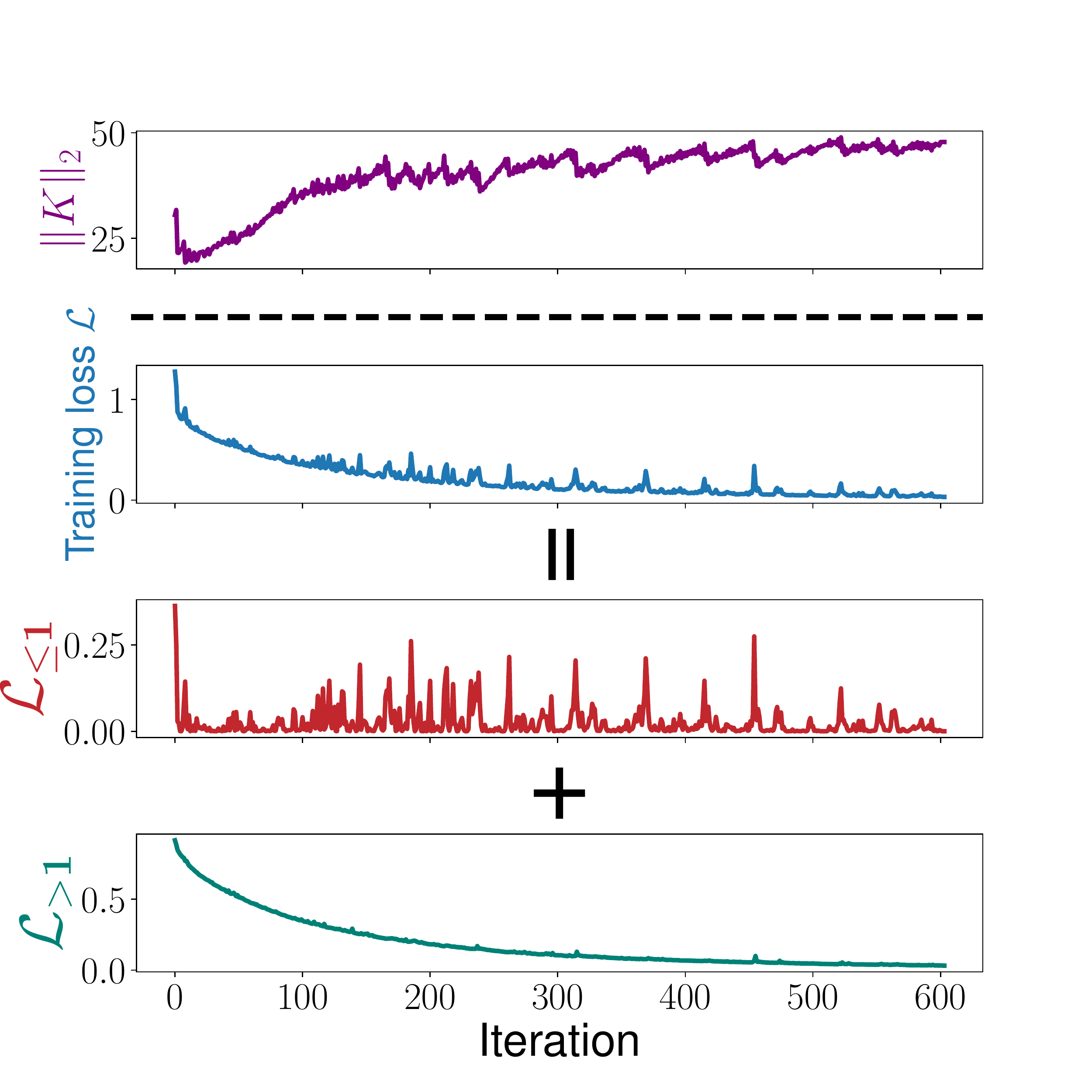}
         \caption{5-layer CNN}
     \end{subfigure}
       \begin{subfigure}[b]{0.4\textwidth}
         \centering
         \includegraphics[width=\textwidth]{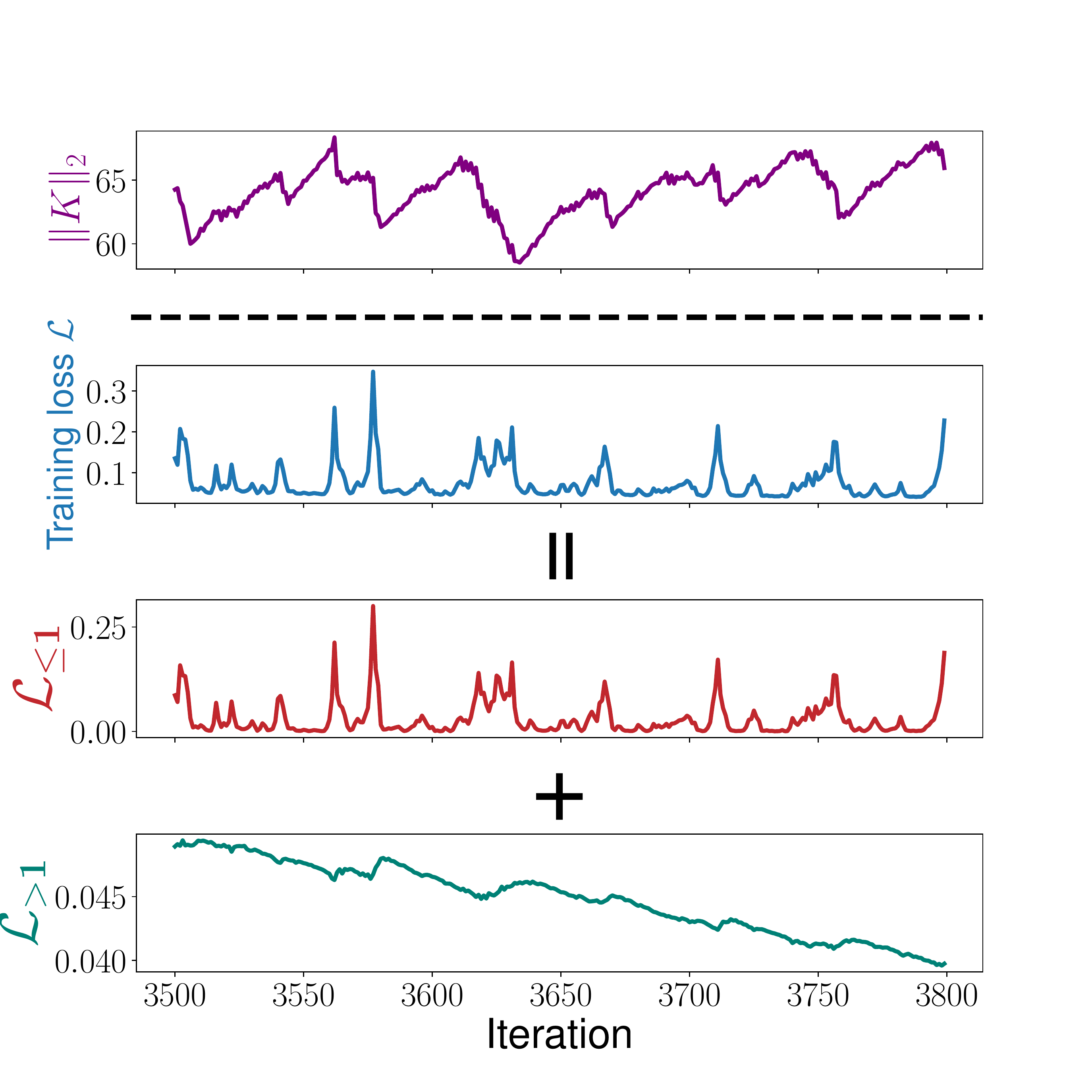}
         \caption{Wide ResNets 10-10 (zoomed-in)}
     \end{subfigure}
    \begin{subfigure}[b]{0.4\textwidth}
         \centering
         \includegraphics[width=\textwidth]{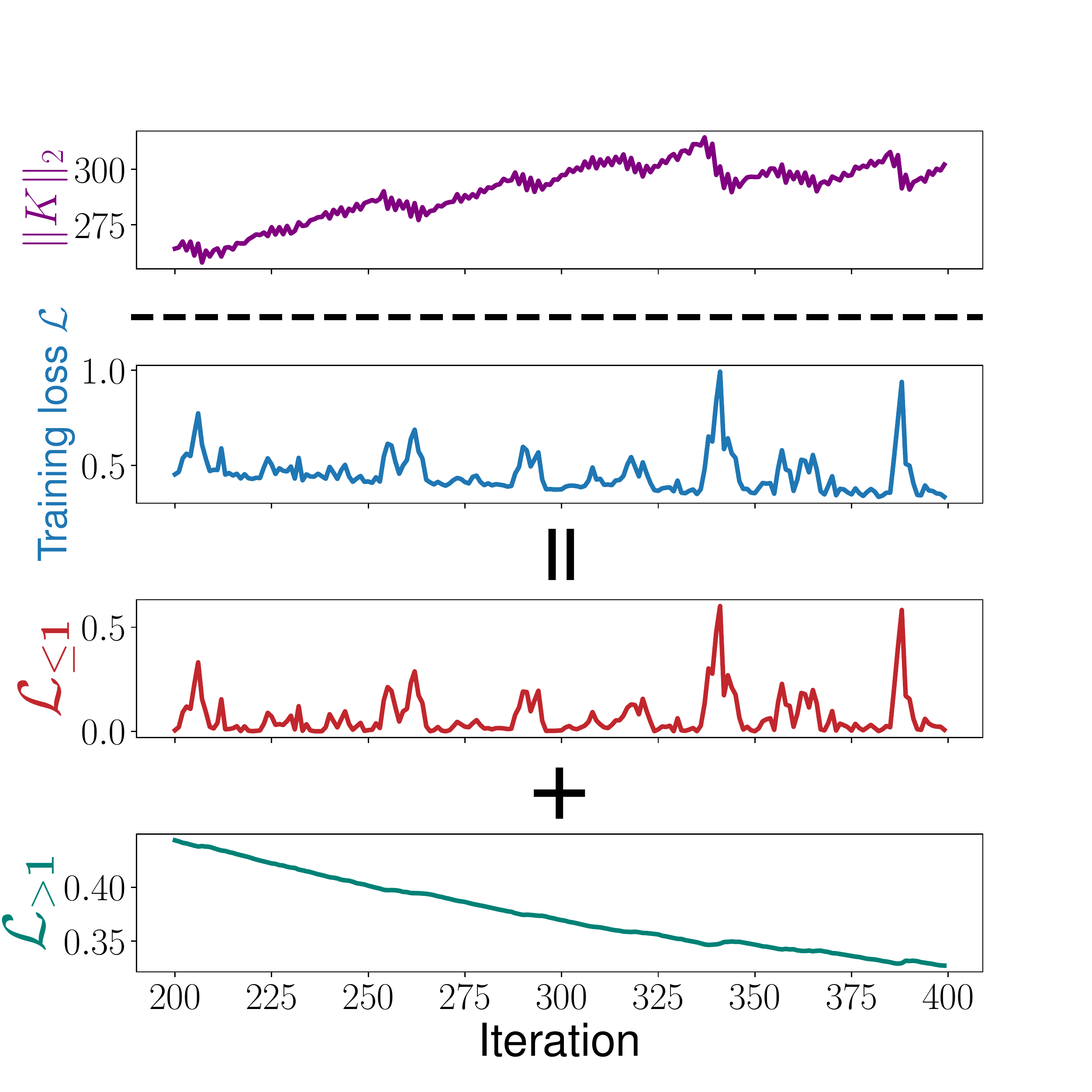}
         \caption{ViT-4 (zoomed-in)}
     \end{subfigure}
     \caption{{\bf Cataput dynamics in SGD for modern deep architectures on 2-class SVHN.} The tasks are the same with Fig.~\ref{fig:sgd_deep} except that we train the neural networks on a subset of SVHN dataset. {The training loss is decomposed into the top eigenspace of the tangent kernel $\L_{\leq 1}$ and its complement $\L_{>1}$. Here $\L = \L_{\leq 1} + \L_{>1} $. } \label{fig:sgd_deep_svhn}}
\end{figure}

\begin{figure}[H]
     \centering\hspace*{-2em}
     \begin{subfigure}[b]{0.35\textwidth}
         \centering
         \includegraphics[width=\textwidth]{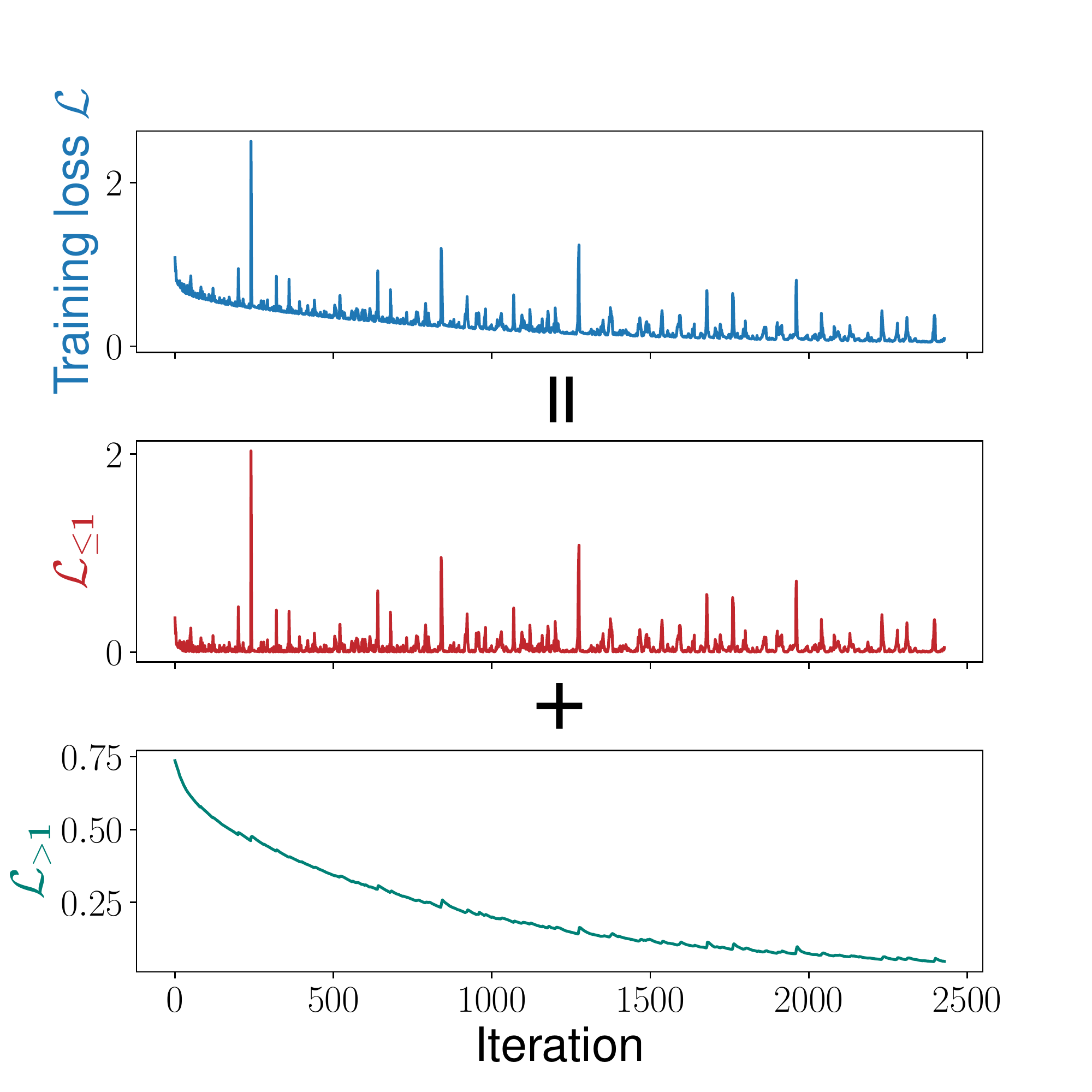}\caption{5-layer FCN}
     \end{subfigure}\hspace*{-0.9em}
     \begin{subfigure}[b]{0.35\textwidth}
         \centering
         \includegraphics[width=\textwidth]{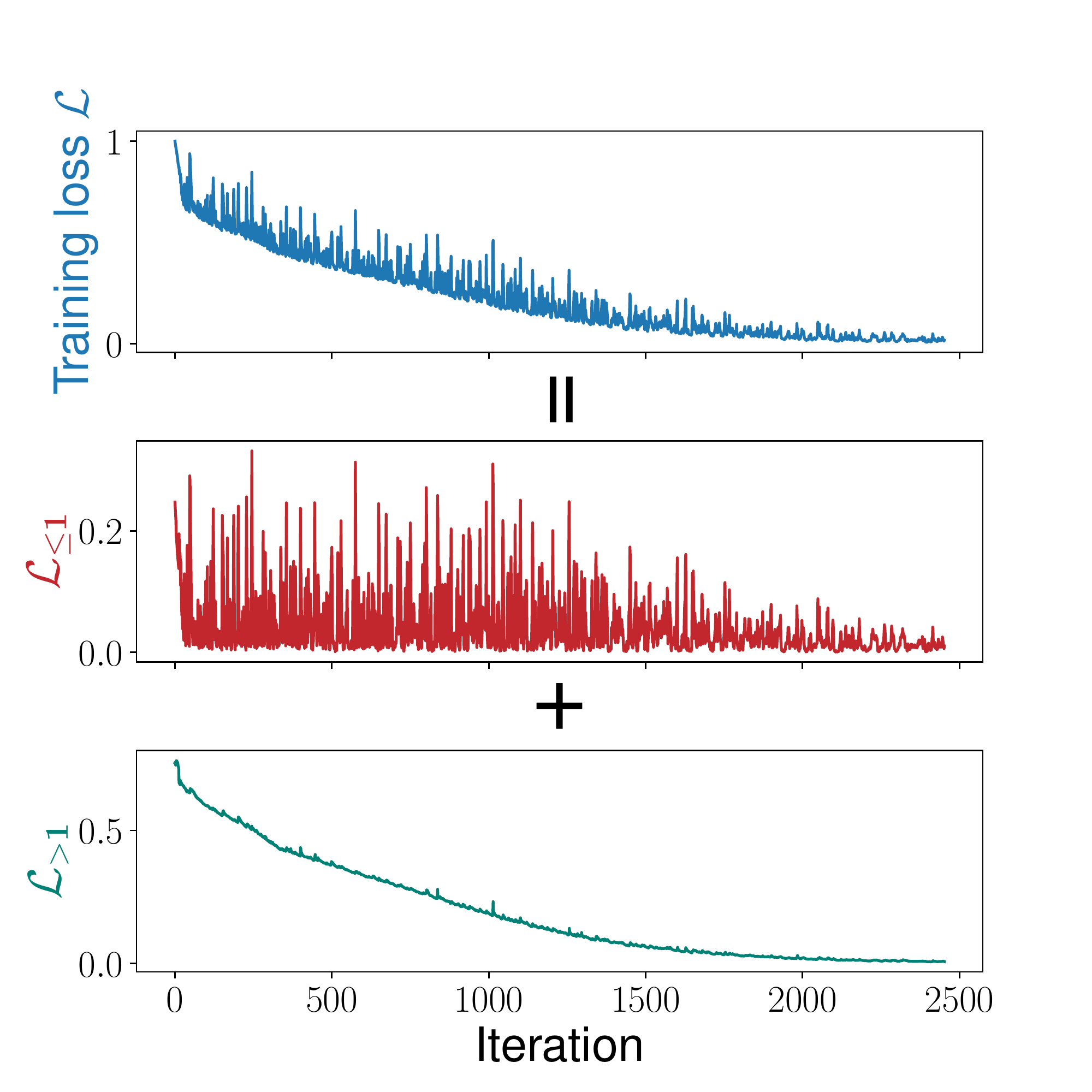}\caption{5-layer CNN}
     \end{subfigure}\hspace*{-0.9em}
     \begin{subfigure}[b]{0.35\textwidth}
         \centering
         \includegraphics[width=\textwidth]{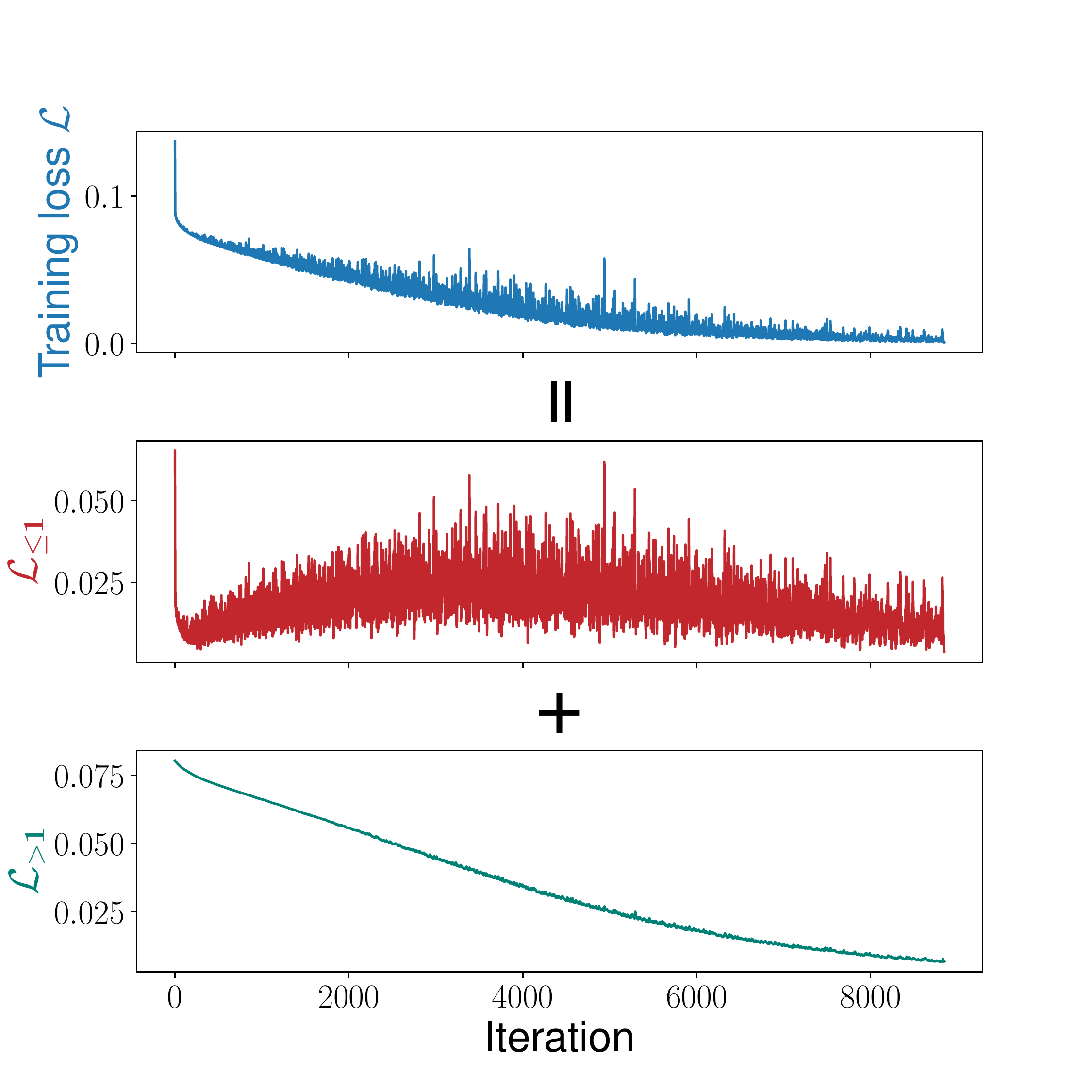}\caption{Myrtle network~\cite{myrtle}}
     \end{subfigure}\hspace*{-2em}
     \caption{{\bf Catapult dynamics in SGD for large datasets (Panel (a) and (b)) and multi-class classification problems (Panel(c)).} Panel(a,b): The networks are trained on $5,000$ data points from CIFAR-2. Panel(c): The network is trained on $128$ points from CIFAR-10. {The training loss is decomposed into the top eigenspace of the tangent kernel $\L_{\leq 1}$ and its complement $\L_{>1}$. Here $\L = \L_{\leq 1} + \L_{>1} $. } }\label{fig:large_sgd}
\end{figure}

We show that the findings in Fig.~\ref{fig:sgd_deep} hold for a subset of SVHN dataset (see Fig.~\ref{fig:sgd_deep_svhn}) and for a larger dataset ($5,000$ data points from CIFAR-2) and for multi-class classification problems (see Fig.~\ref{fig:large_sgd}).

\subsection{Catapults occur in training with cyclical learning rates}\label{subsec:cyclic}

% Training neural networks with cyclical learning rates where the learning rate cyclically varies between reasonable boundary values was shown to improve the generalization performance of neural networks with less tuning~\cite{izmailov2018averaging,smith2017cyclical}. We empirically show that cyclical learning rates in SGD work similarly to the increased learning rate in GD, e.g., Fig~\ref{fig:multi_cata} which can induce multiple catapults. Specifically, we observe that there is a loss spike in the training loss when the learning rate is increased in the cyclic learning rate schedule. We show that the loss spikes are catapults, by similarly showing the loss occurs in the top eigenspace of the tangent kernel and there is a decrease in the spectral norm of the tangent kernel according to each loss spike. See the results in Fig.~\ref{fig:cyc_lr}. 
% In the following Section~\ref{sec:feature_learning}, we show that catapults will lead to better generalization through feature learning, which may account for better generalization performance with cyclical learning rates.

In this section, we show that catapults occur in SGD with a cyclical learning rate schedule. Specifically, we show that loss spikes occur in the top eigenspace of the tangent kernel and there is a decrease in the spectral norm of the tangent kernel according to each loss spike. 

\begin{figure}[H]
     \centering
     % \begin{subfigure}[b]{0.4\textwidth}
     %     \centering
     %     \includegraphics[width=\textwidth]{figure/multi_toy_loss.pdf}
     %     \caption{Shallow network}
     % \end{subfigure}
     \begin{subfigure}[b]{0.45\textwidth}
         \centering
         \includegraphics[width=\textwidth]{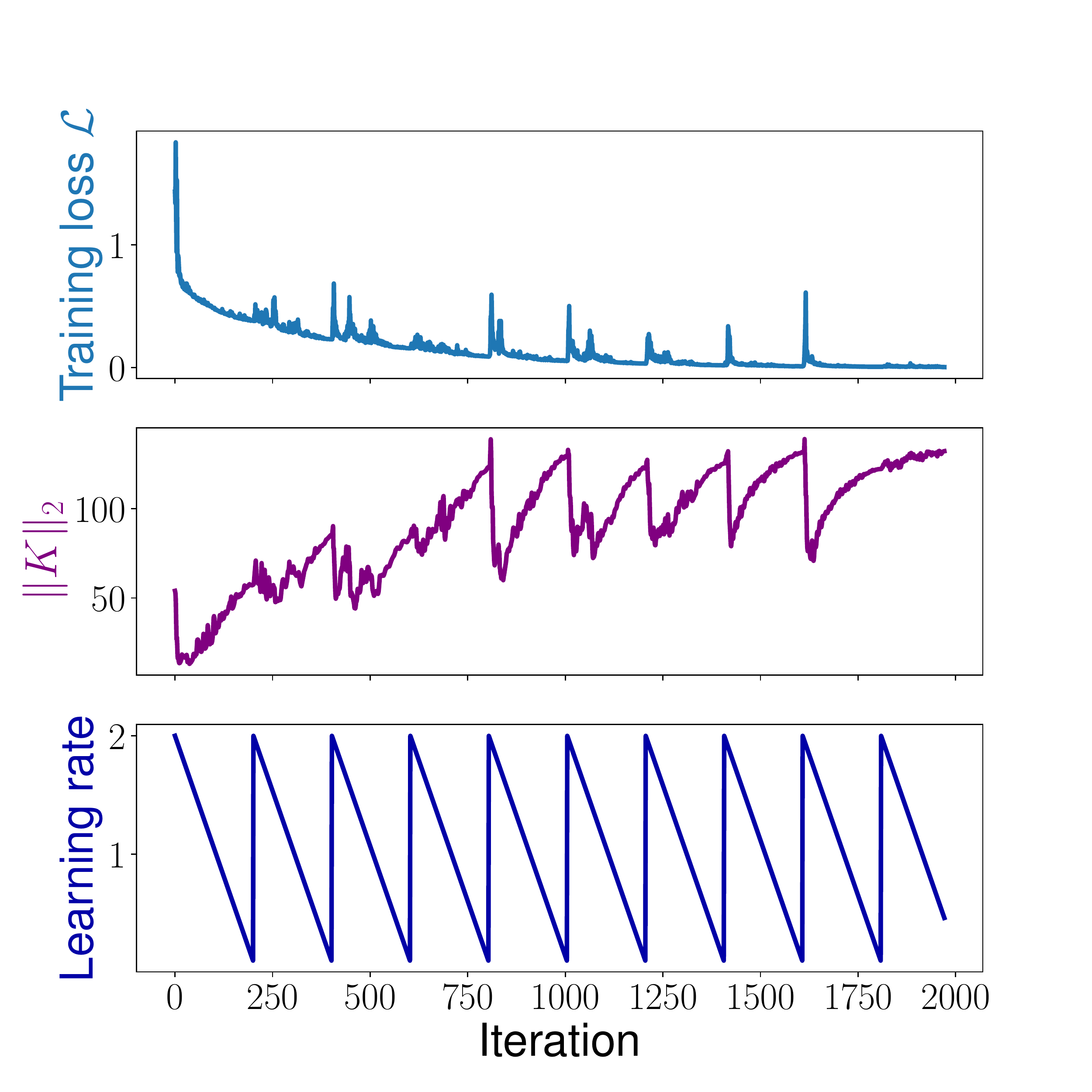}
         \caption{Training loss, $\norm{K}_2$ and learning rate}
     \end{subfigure}
     \begin{subfigure}[b]{0.45\textwidth}
         \centering
         \includegraphics[width=\textwidth]{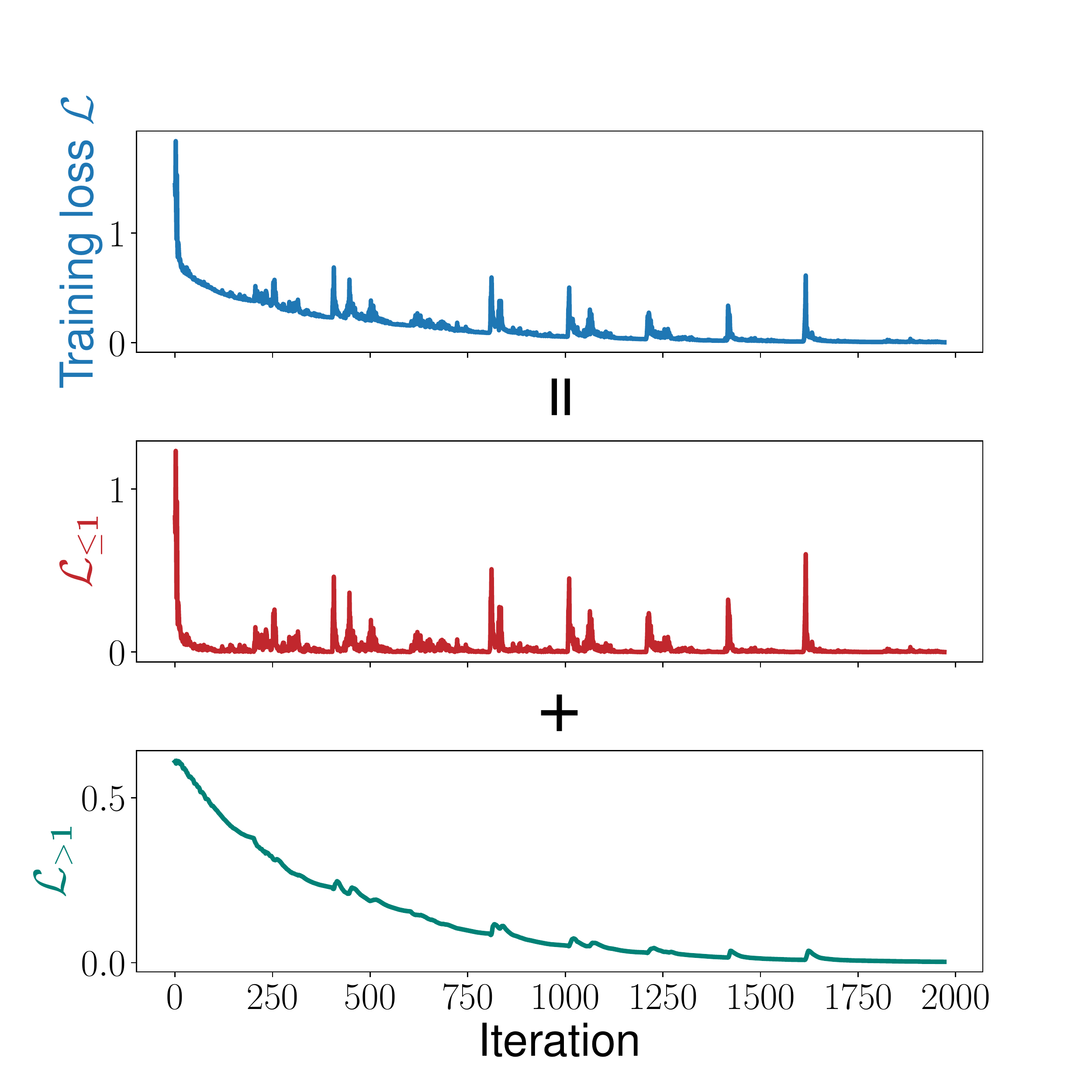}
         \caption{Loss decomposition}
     \end{subfigure}
     \caption{{\bf Catapults in SGD with cyclical learning rates}. \ Panel (a): The plot of the training loss and the spectral norm of the tangent kernel corresponding to the whole training set with a cyclic learning rate schedule.  Panel (b): The training loss is decomposed into the top and non-top eigenspace of the tangent kernel, i.e.,$\L_{\leq 1}$ and $\L_{>1}$. Here $\L = \L_{\leq 1} + \L_{>1}$.   We train Wide ResNets 10-10 on a subset of CIFAR-10. The setting is the same with Fig.~\ref{fig:sgd_deep}c except for a cyclical learning rate schedule.
     \label{fig:cyc_lr}}
\end{figure}

\section{Additional experiments for feature learning in GD}\label{sec:add_agop_gd}

\subsection{Validation loss/error for multiple catapults corresponding to Fig.~\ref{fig:multi_cata_gd_low_rank}}

\begin{figure}[H]
     \centering
     \begin{subfigure}[b]{0.4\textwidth}
         \centering
\includegraphics[width=\textwidth]{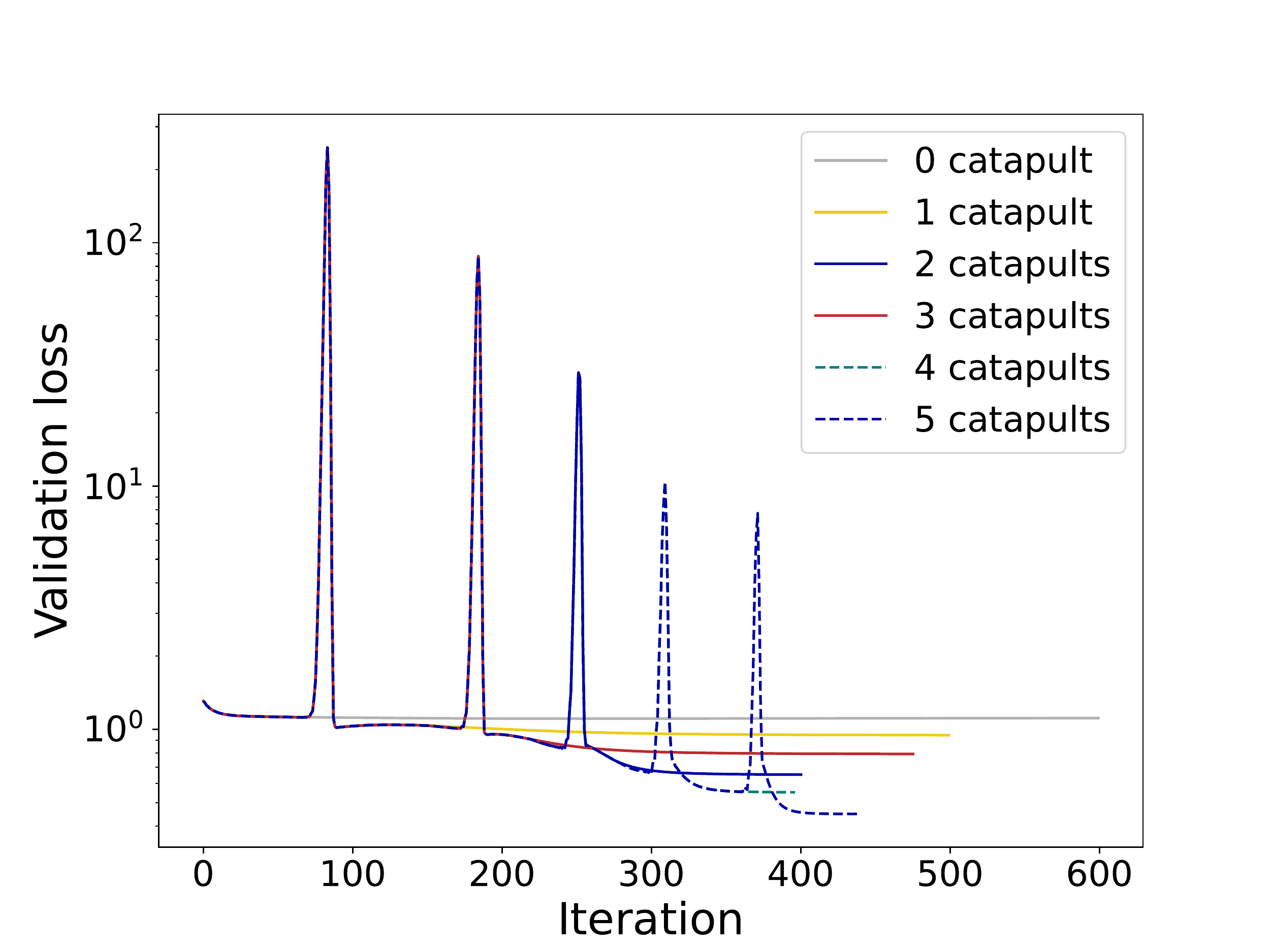}
         \caption{Rank-2 regression}
     \end{subfigure}
     \begin{subfigure}[b]{0.4\textwidth}
         \centering
\includegraphics[width=\textwidth]{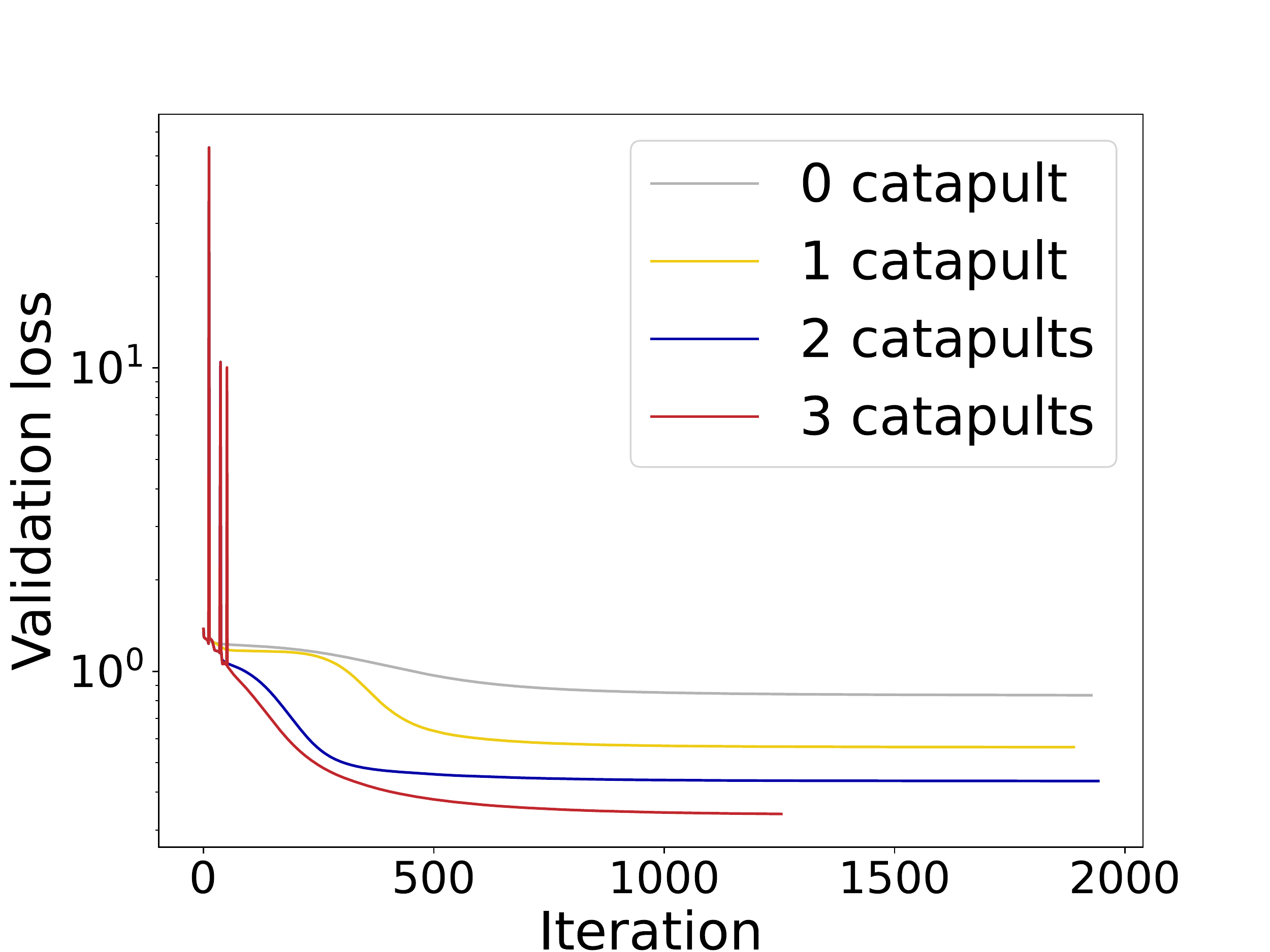}
         \caption{Rank-3 regression}
     \end{subfigure}
     % \begin{subfigure}[b]{0.24\textwidth}
     %     \centering
     %     \includegraphics[width=\textwidth]{figure/fl_gd_test_4_digit.pdf}
     %     \caption{Rank-4 regression}
     % \end{subfigure}\hspace*{-0.9em}
     \begin{subfigure}[b]{0.4\textwidth}
         \centering
    \includegraphics[width=\textwidth]{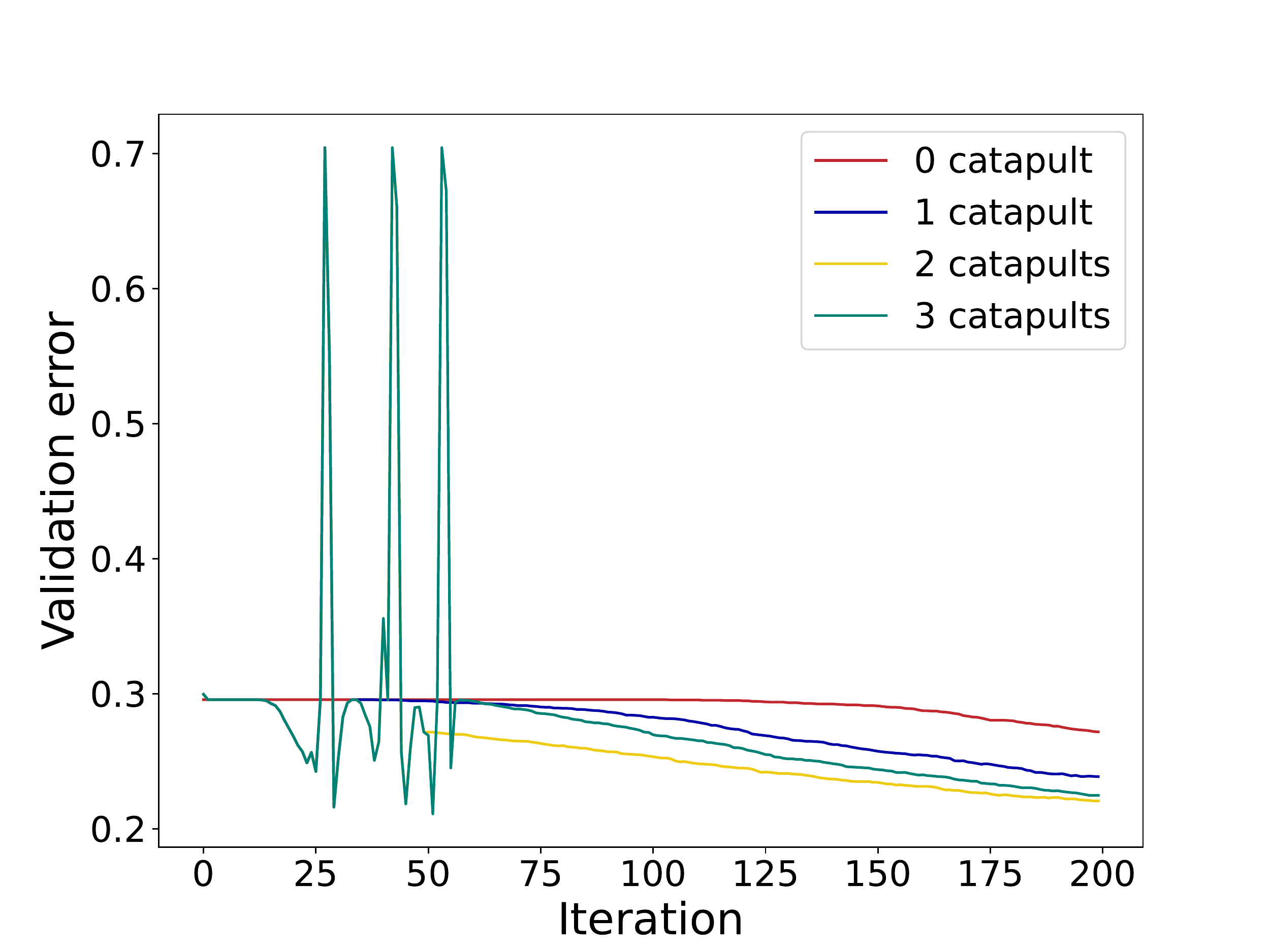}
         \caption{SVHN-2}
     \end{subfigure}
     \begin{subfigure}[b]{0.4\textwidth}
         \centering
\includegraphics[width=\textwidth]{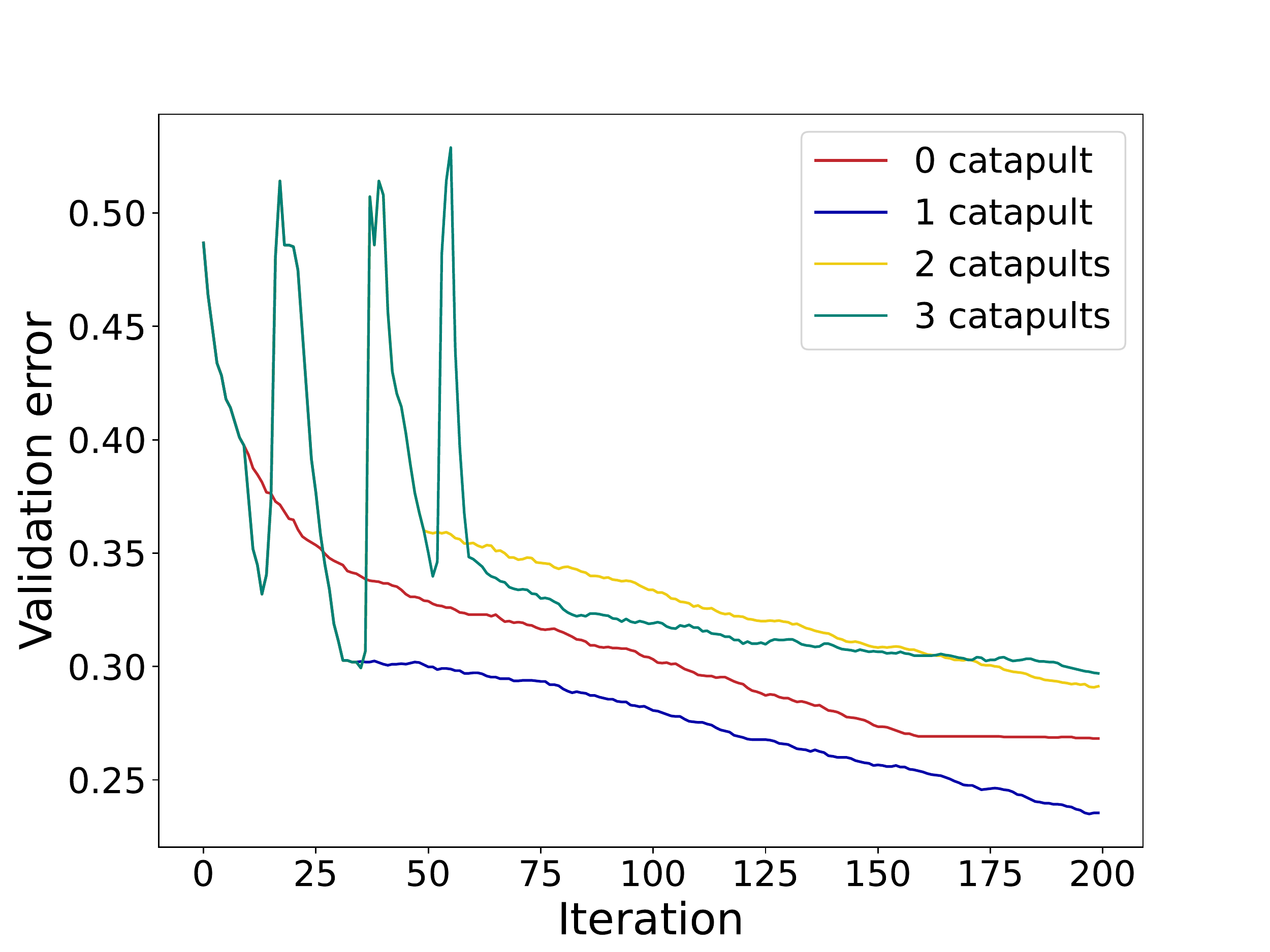}
         \caption{CelebA}
     \end{subfigure}
\caption{{\bf Validation loss/error of multiple catapults in GD corresponding to} Fig.~\ref{fig:multi_cata_gd_low_rank}. Panel(c)\&(d) only present first $200$ iterations. \label{fig:fl_gd_loss_spikes}}
\end{figure}

We present the validation loss/error in Fig.~\ref{fig:fl_gd_loss_spikes} for the tasks corresponding to Fig.~\ref{fig:multi_cata_gd_low_rank}. The learning rate is increased during training to generate multiple catapults.

\subsection{Feature learning with near zero initialization}

We compare the performance of networks exhibiting multiple catapults with those initialized using near zero initialization scheme, i.e., each weight is sampled i.i.d. from $\mathcal{N}(0,\sigma^2)$ with $\sigma = 0.1$.  This is in contrast to the NTK parameterization where we use $\sigma = 1$. It was argued in~\cite{yang2021tensor} that feature learning occurs with near zero initialization.   We can see that small initialization achieves the smallest test loss/error as well as the best \AGOP alignment, which indicates that learning AGOP correlates strongly with the test performance.

\begin{figure}[H]
     \centering
     \begin{subfigure}[b]{0.4\textwidth}
         \centering
    \includegraphics[width=\textwidth]{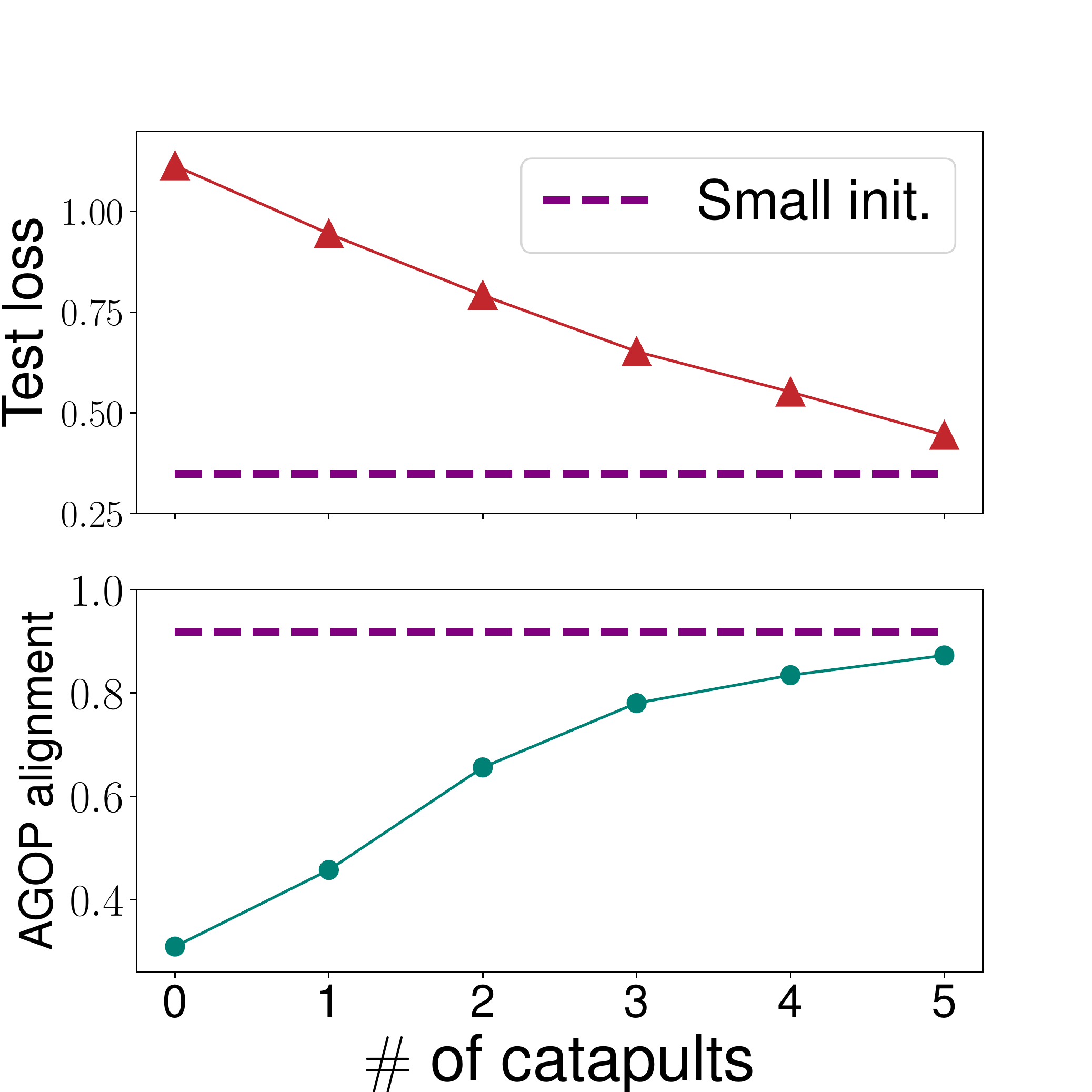}
         \caption{Rank-2 regression}
     \end{subfigure}
     %      \begin{subfigure}[b]{0.24\textwidth}
     %     \centering
     %     \includegraphics[width=\textwidth]{figure/fl_gd_test_agop_3_digit.pdf}
     %     \caption{Rank-4 regression}
     % \end{subfigure}
     \begin{subfigure}[b]{0.4\textwidth}
         \centering
         \includegraphics[width=\textwidth]{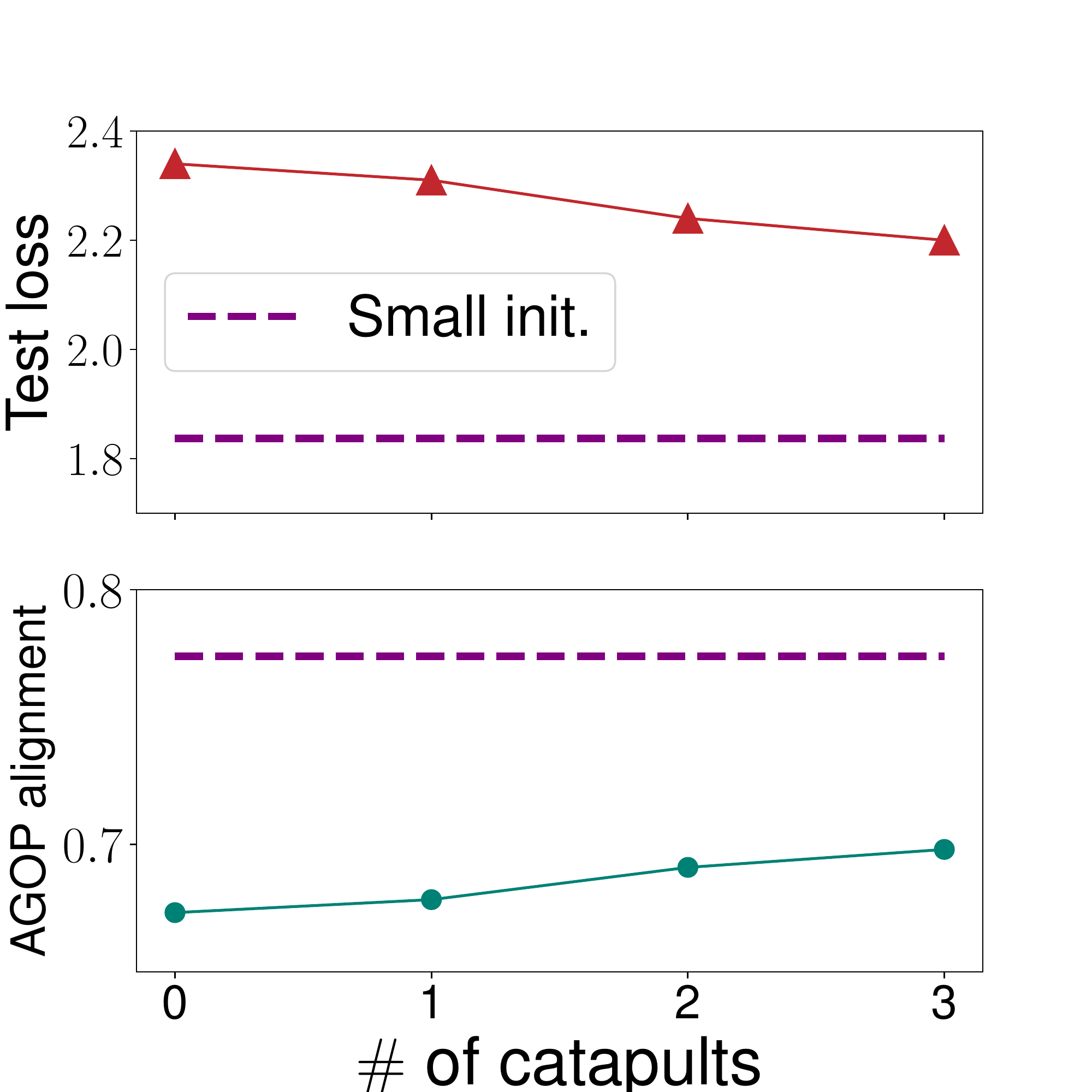}
         \caption{Rank-4 regression}
     \end{subfigure}
     \begin{subfigure}[b]{0.4\textwidth}
         \centering
         \includegraphics[width=\textwidth]{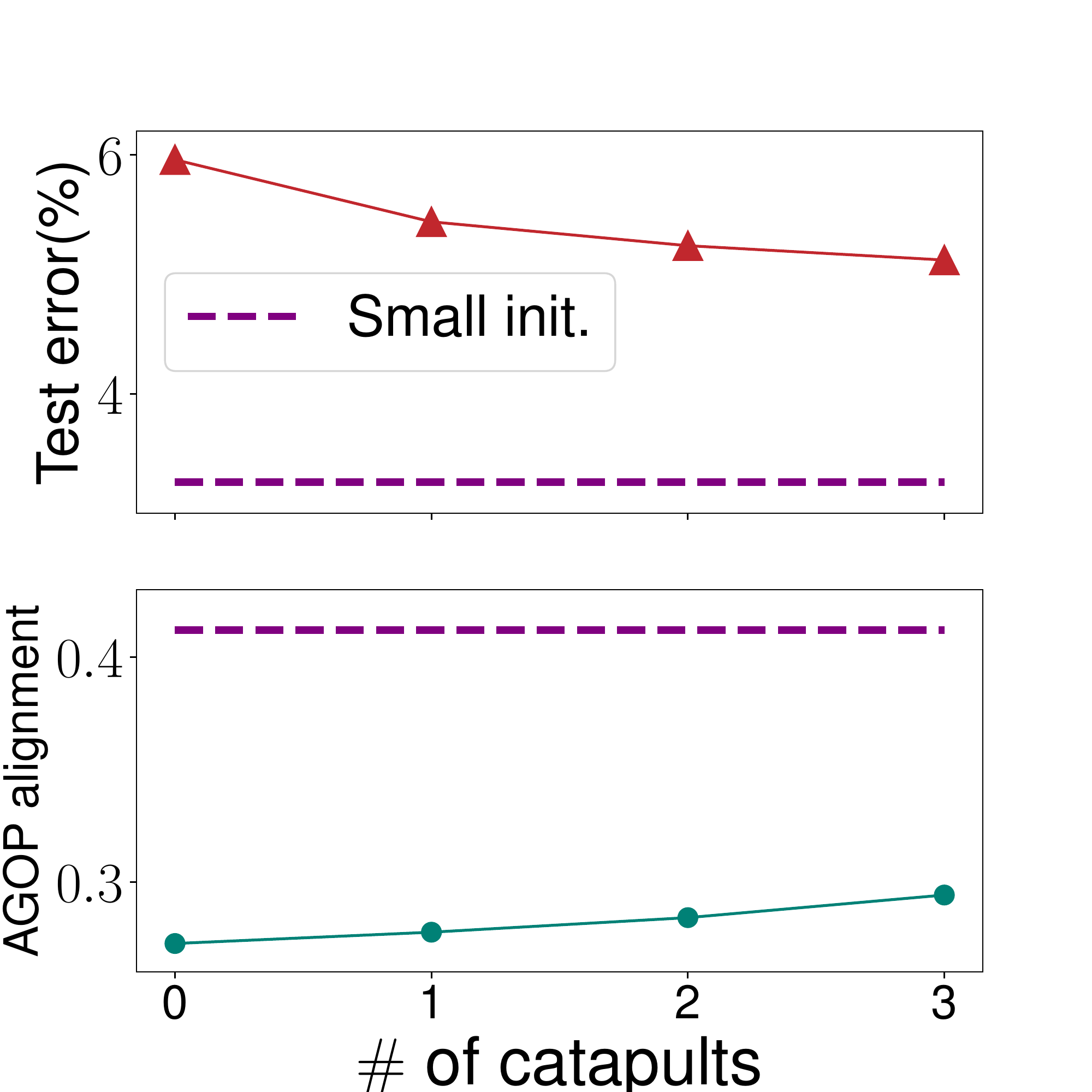}
         \caption{SVHN-2}
     \end{subfigure}
     \begin{subfigure}[b]{0.4\textwidth}
         \centering
 \includegraphics[width=\textwidth]{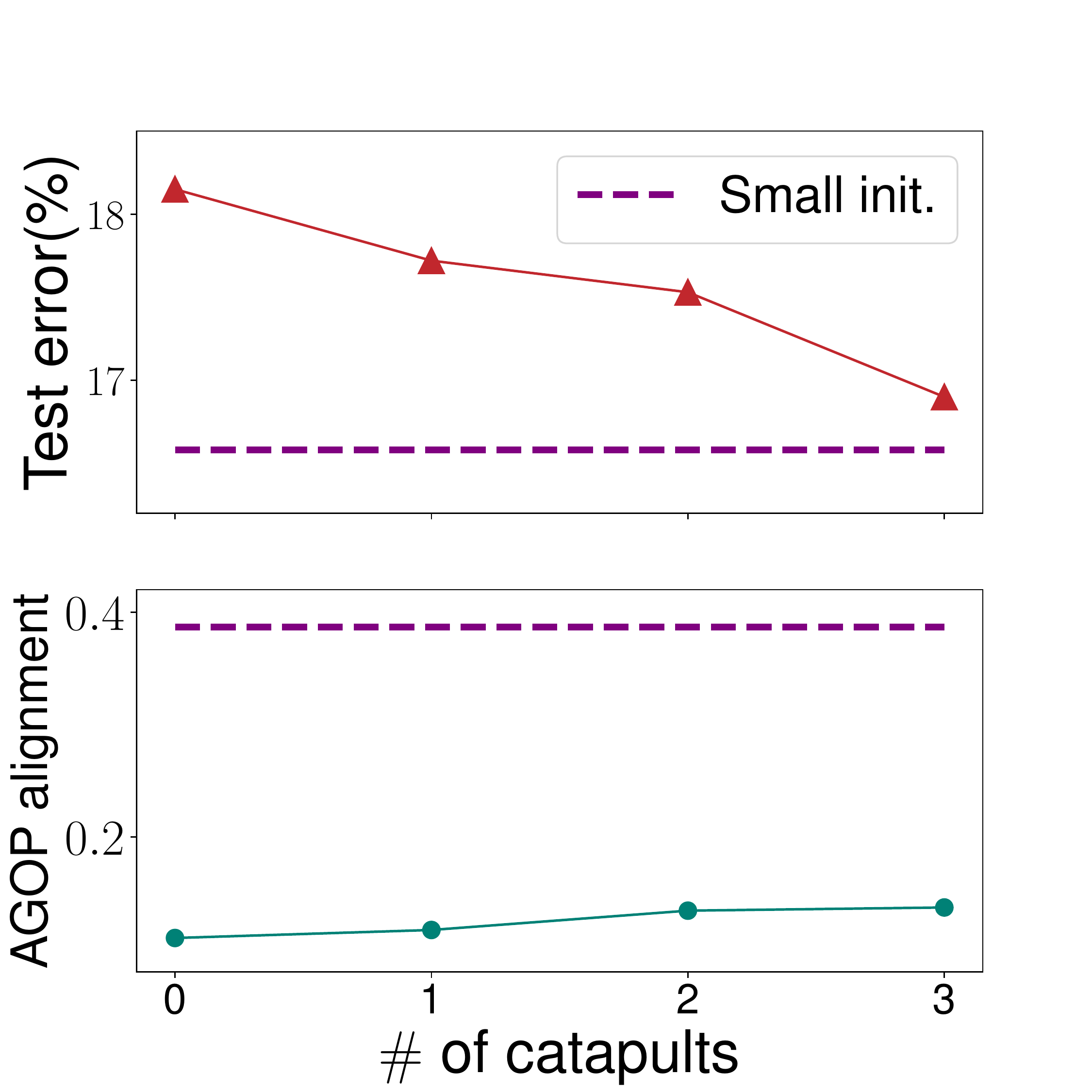}
         \caption{CelebA}
     \end{subfigure} 
\caption{{\bf Multiple catapults in GD compared to the small initialization scheme. }
We train a 2-layer FCN in Panel(a), a 4-layer FCN in Panel(b,d) and a 5-layer CNN in Panel(c). For small initialization, each weight parameter is i.i.d. from $\mathcal{N}(0,\sigma^2)$ with $\sigma = 0.1$.  The experimental setup is the same as Fig.~\ref{fig:multi_cata_gd_low_rank}\label{fig:multi_cata_gd_low_rank_with_small}.}
\end{figure}

For the Rank-2 regression task, we visualize the AGOP in the following Fig.~\ref{fig:multi_cata_shallow_vis}, where we can see that the features are learned better, i.e., closer to the True AGOP,  with a greater number of catapults.

\begin{figure}[H]
     \centering\hspace*{-5em}
     \begin{subfigure}[b]{1.2\textwidth}
         \centering         \includegraphics[width=\textwidth]{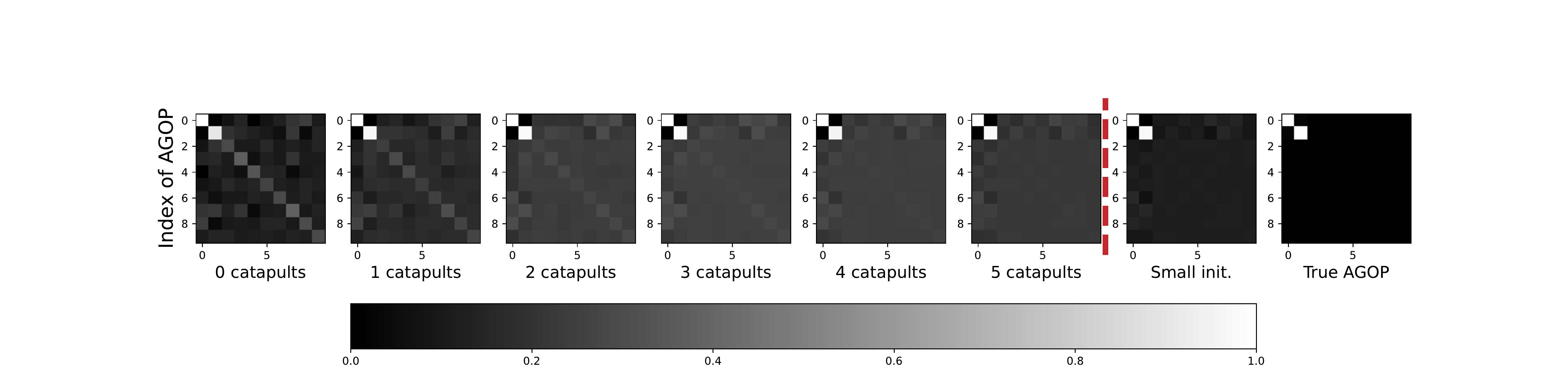}
     \end{subfigure}
\caption{{\bf Visualization of \AGOP{} for rank-2 regression task.} All pixels are normalized to the range $[0,1]$ and the top 10 rows and columns of the \AGOP{} are plotted.}\label{fig:multi_cata_shallow_vis}
\end{figure}

\subsection{Feature learning in GD for additional datasets}
\begin{figure}[H]
     \centering\hspace*{-2em}
     \begin{subfigure}[b]{0.33\textwidth}
         \centering
         \includegraphics[width=\textwidth]{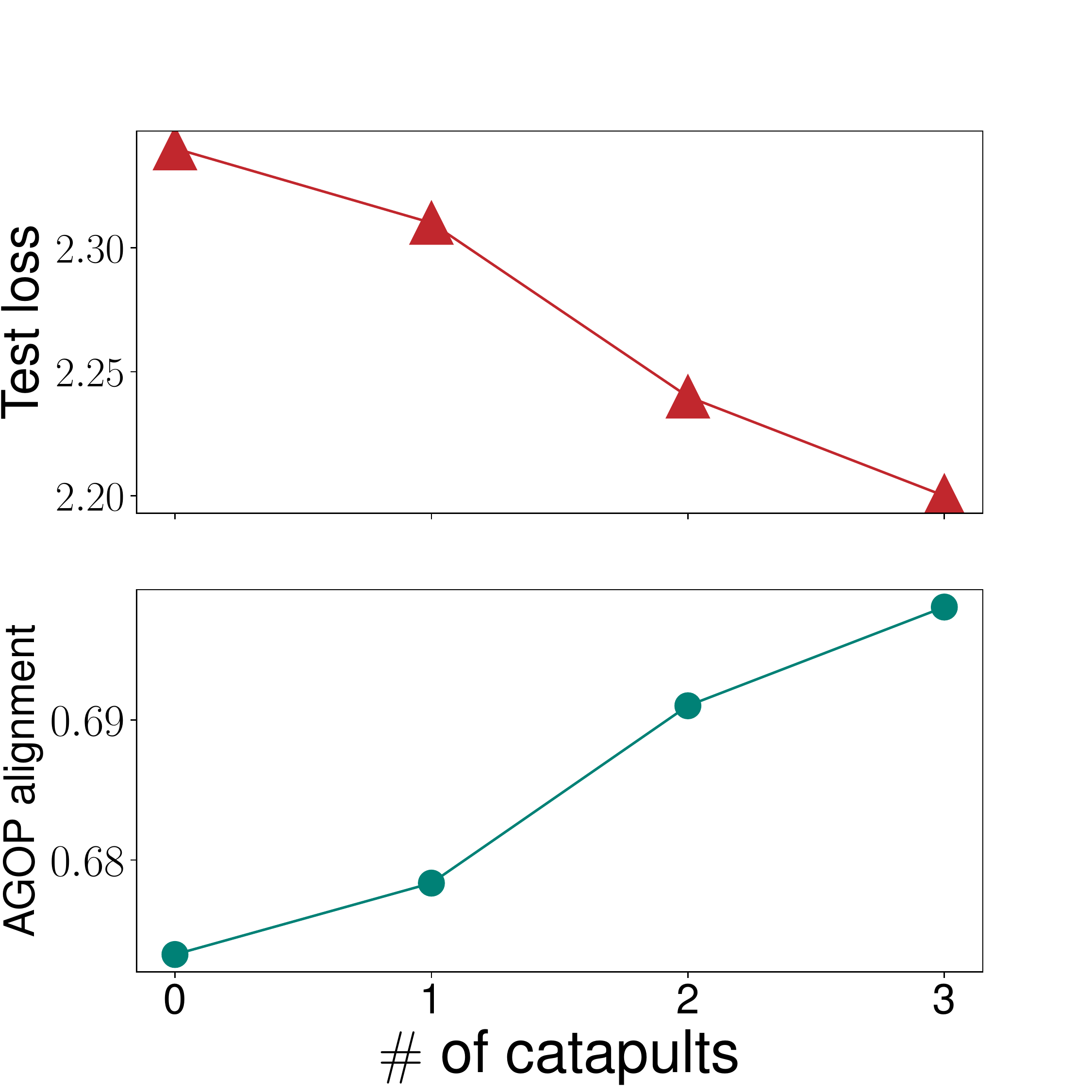}
         \caption{Rank-4 regression}
     \end{subfigure}\hspace*{-0.9em}
     \begin{subfigure}[b]{0.33\textwidth}
         \centering
         \includegraphics[width=\textwidth]{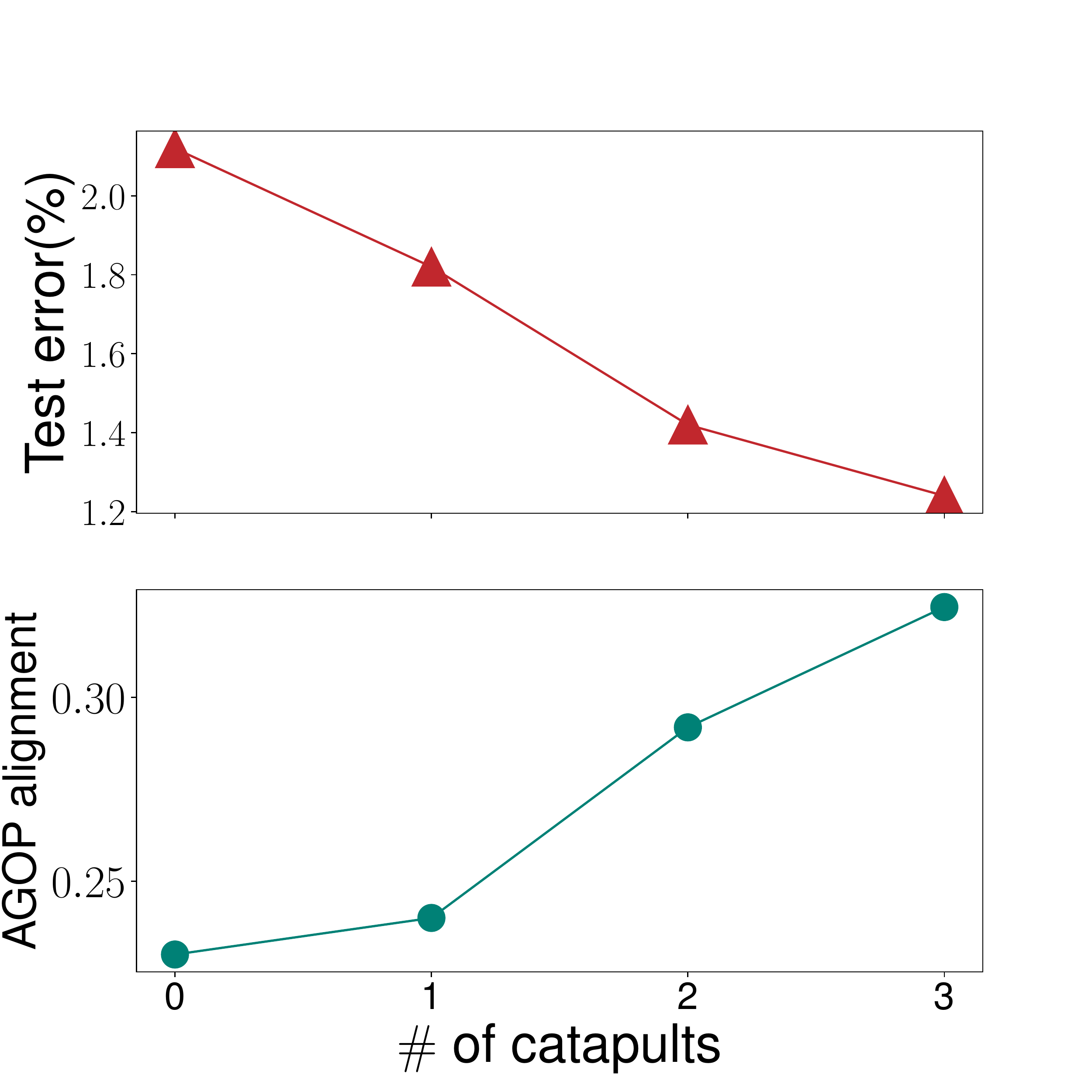}
         \caption{USPS}
     \end{subfigure}\hspace*{-0.9em}
     \begin{subfigure}[b]{0.33\textwidth}
         \centering
         \includegraphics[width=\textwidth]{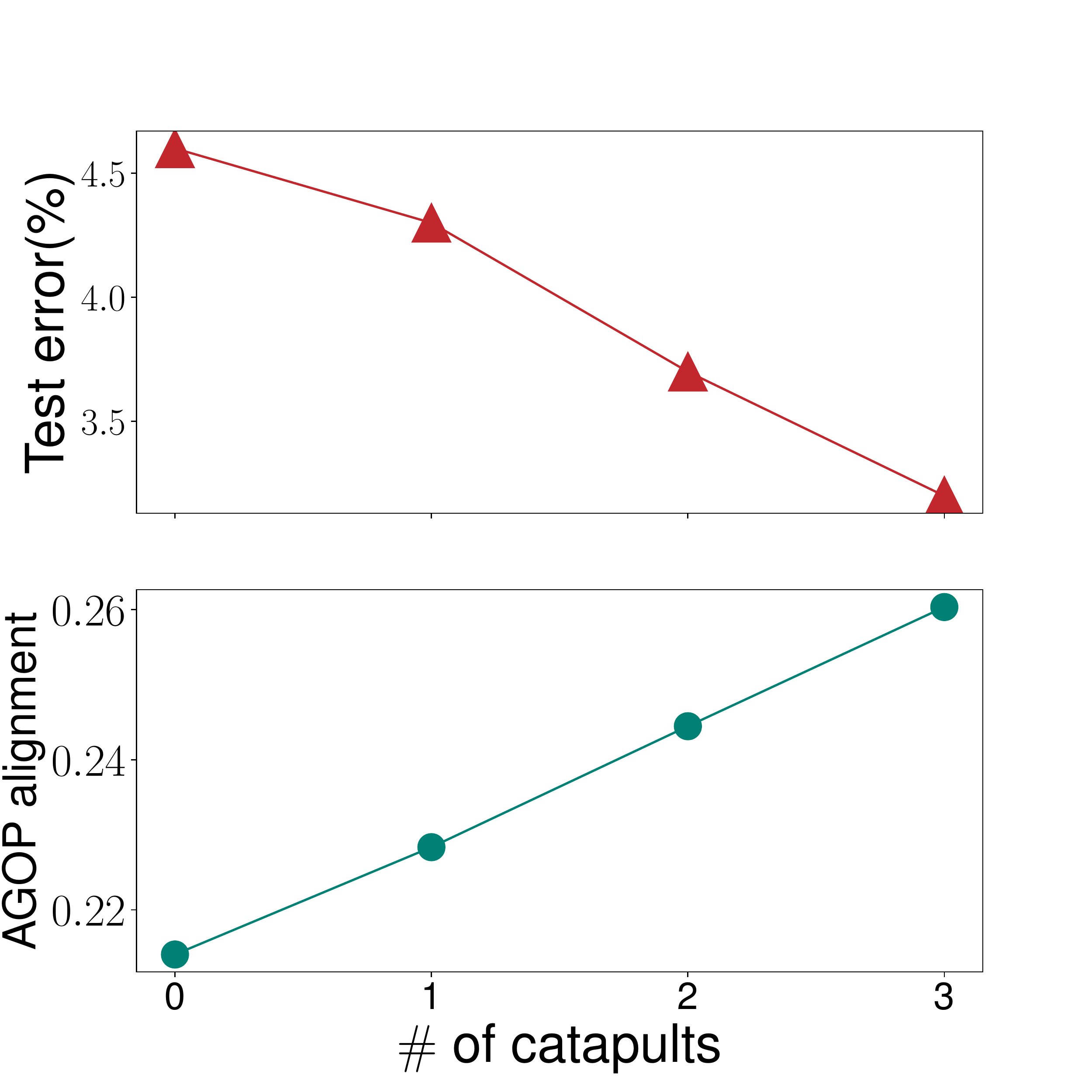}
         \caption{Fashion MNIST}
     \end{subfigure}\hspace*{-2em} 
\caption{{\bf Correlation between AGOP alignment and test performance in GD with multiple catapults on additional datasets.  } 
We train a 4-layer FCN using GD for all tasks.  The learning rate is increased multiple times during training to generate multiple catapults.  Experimental details can be found in Appendix~\ref{exp:feature_learning}.\label{fig:multi_cata_gd_low_rank_add}}
\end{figure}

In this section, we show the findings observed in Fig.~\ref{fig:multi_cata_gd_low_rank} hold for Rank-4 regression, USPS dataset and Fashion MNIST dataset. See Fig.~\ref{fig:multi_cata_gd_low_rank_add}.

\subsection{No feature learning for full rank task}

\begin{figure}[H]
     \centering\hspace*{-0.9em}
               \begin{subfigure}[b]{0.30\textwidth}
         \centering
         \includegraphics[width=\textwidth]{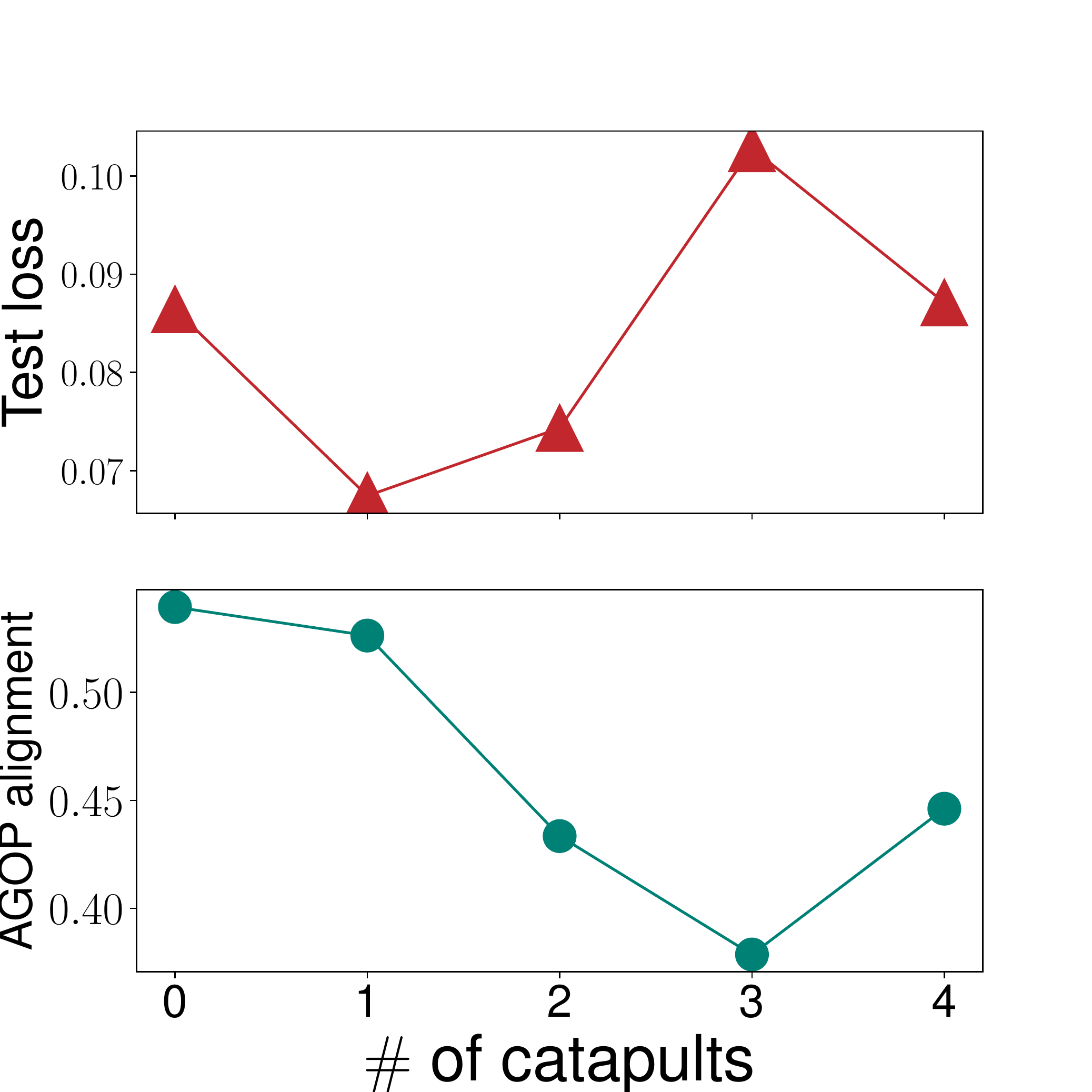}
         \caption{Validation loss vs. iteration}
     \end{subfigure}
     \begin{subfigure}[b]{0.4\textwidth}
         \centering
         \includegraphics[width=\textwidth]{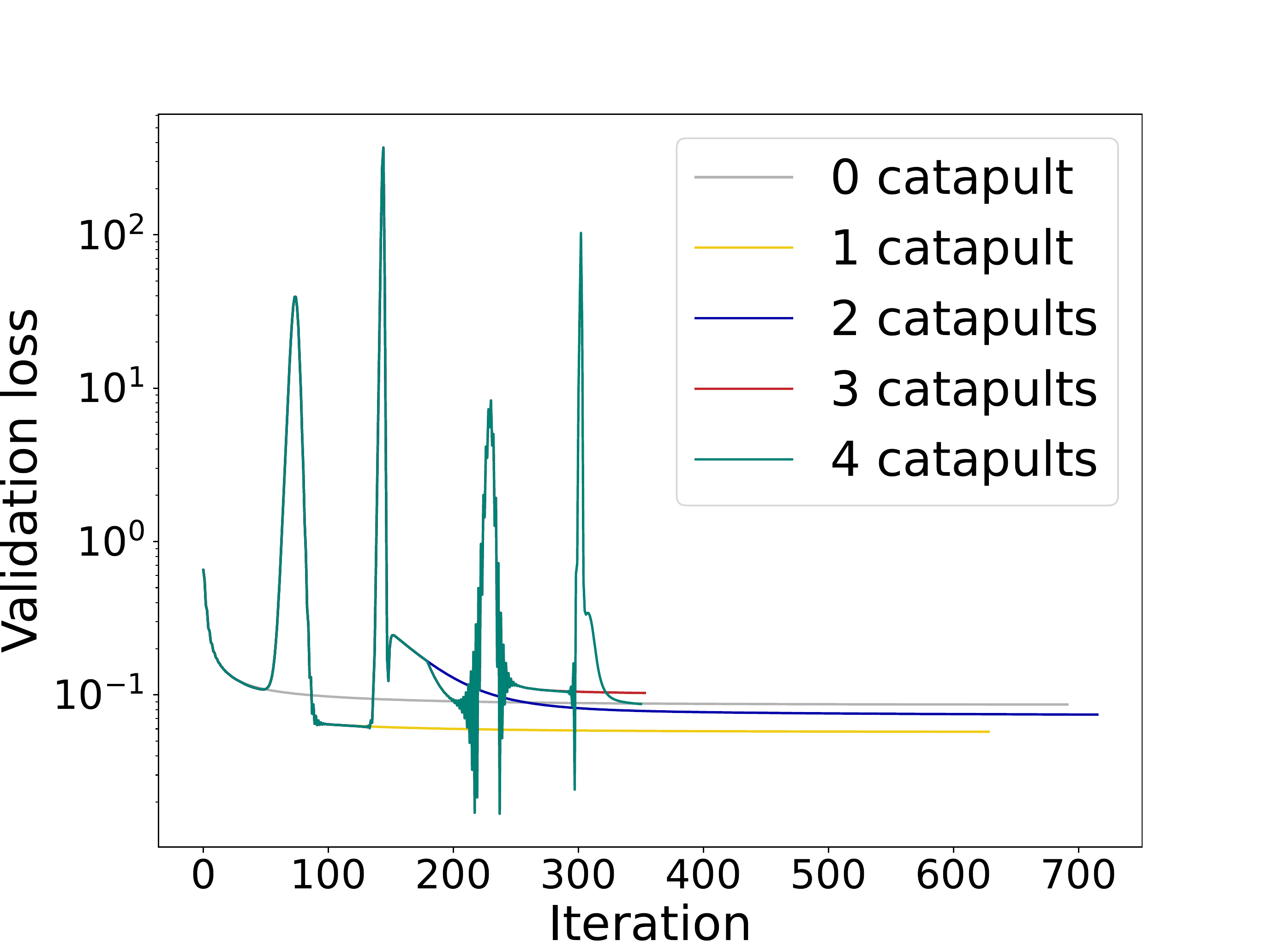}
         \caption{Test loss and AGOP alignment}
     \end{subfigure}
\caption{{\bf Multiple catapults in GD for a full rank task. }
We train a 2-layer FCN on a synthetic dataset with a full rank target function $f^*(\vx) = \frac{1}{\sqrt{d}}\norm{\vx}$ using GD.The learning rate is increased multiple times during training to generate multiple catapults. The experimental details can be found in Appendix~\ref{exp:add_agop_gd}.  \label{fig:multi_cata_gd_full_rank}}
\end{figure}

In Fig.~\ref{fig:multi_cata_gd_full_rank}, we show that for a full-rank task where the target function is $f^*(\vx) = \frac{1}{\sqrt{d}}\norm{\vx}$, catapults do not improve the test performance or the AGOP alignment.

% \FloatBarrier
% \begin{figure}[htb!]
%      \centering
%      \captionsetup[subfigure]{justification=centering}
%      \begin{subfigure}[b]{0.38\textwidth}
%          \centering
%          \includegraphics[width=\textwidth]{figure/fl_gd_test_agop_3_digit.pdf}
%          \caption{Test loss/ AGOP alignment vs. number of catapults}
%         \end{subfigure}
%         \begin{subfigure}[b]{0.5\textwidth}
%          \centering
%          \includegraphics[width=\textwidth]{figure/fl_gd_test_3_digit.pdf}
%          \caption{Validation loss vs. iteration}
%      \end{subfigure}
% \caption{{\bf Multiple catapults in GD improve generalization through AGOP alignment.} Similar to the results in Fig.~\ref{fig:multi_cata_gd_low_rank}, a greater number of catapults leads to better generalization by promoting AGOP alignment. The true predictor is  rank-3 $f^*(\vx) = x_1x_2(\sum_{i=1}^{10}x_i)$. The experimental details can be found in  Appendix~\ref{sec:exp_details}.\label{fig:fl_gd_3_digit}}
% \end{figure}
% \FloatBarrier

\section{Additional experiments for feature learning in SGD}\label{sec:add_agop_sgd}

\subsection{Feature learning of catapults in SGD with Pytorch parameterization}
In this section, we further verify our observation on the feature learning of SGD with Pytorch default parameterization on the same tasks with Fig.~\ref{fig:fl_sgd_agop} in Section~\ref{sec:feature_learning}.

\begin{figure}[H]
     \centering\hspace*{-1em}
     \begin{subfigure}[b]{0.26\textwidth}
         \centering
         \includegraphics[width=\textwidth]{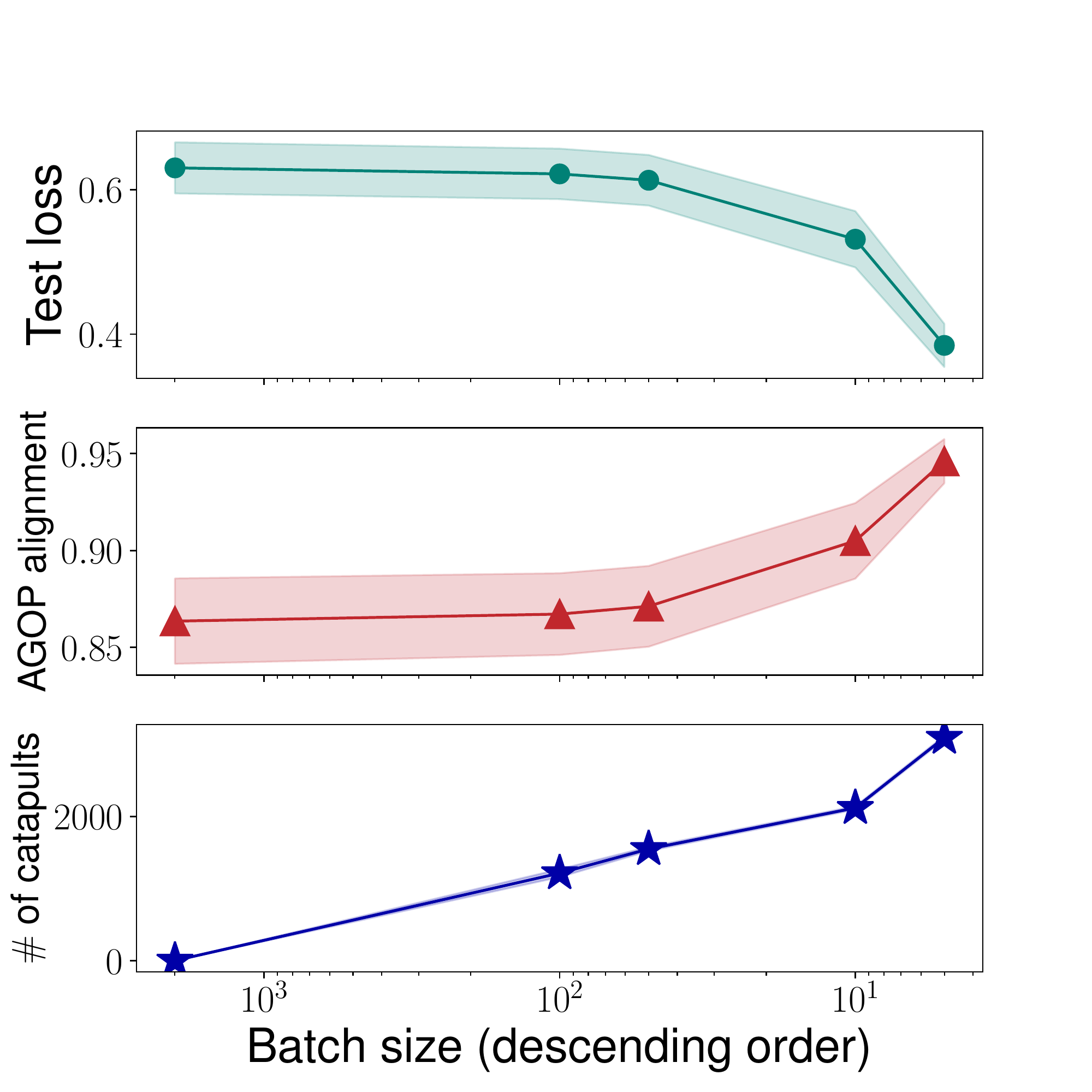}
             \caption{Rank-2 regression}
     \end{subfigure}\hspace*{-0.9em}
     \begin{subfigure}[b]{0.26\textwidth}
         \centering
         \includegraphics[width=\textwidth]{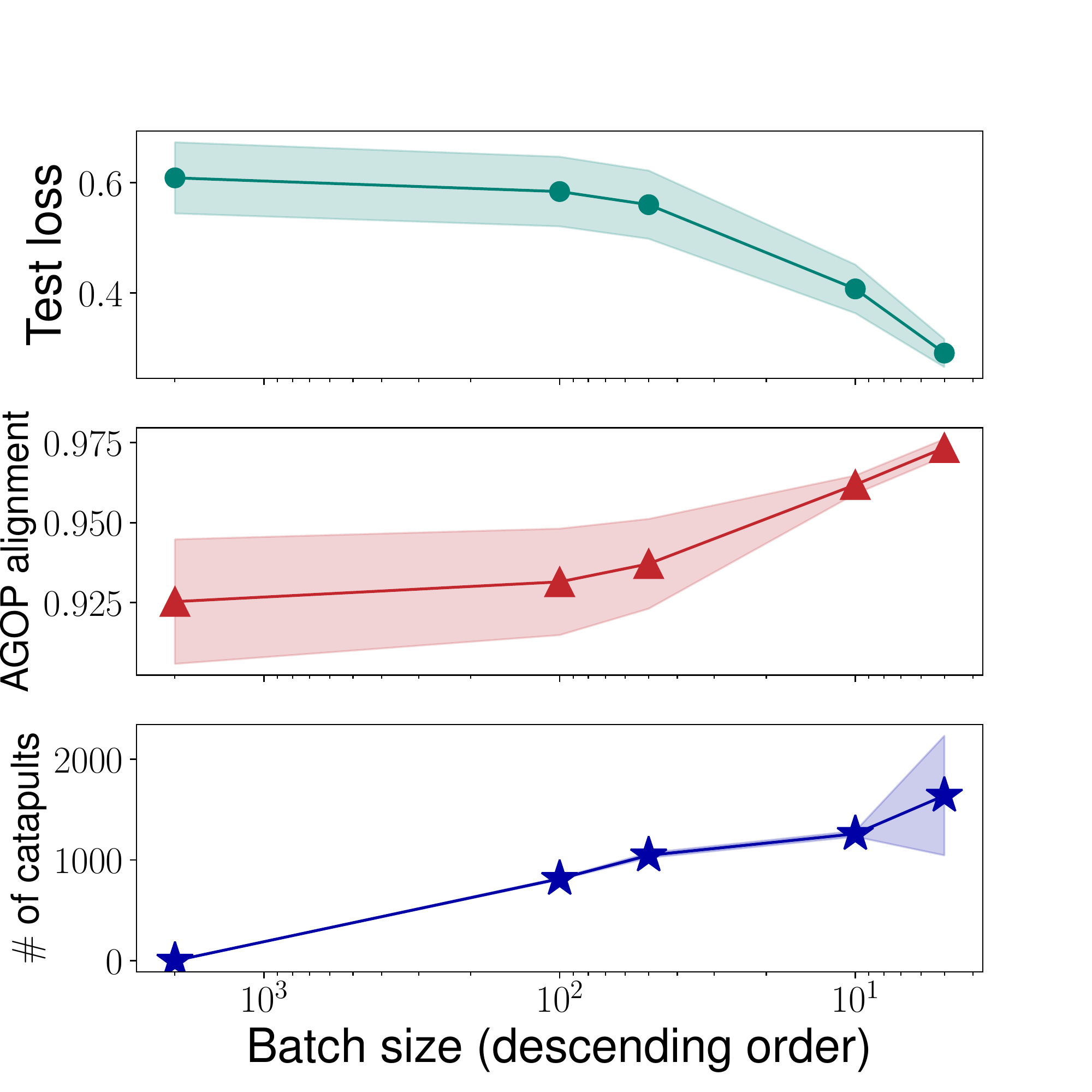}
         \caption{Rank-3 regression}
     \end{subfigure}\hspace*{-0.9em}
     % \begin{subfigure}[b]{0.24\textwidth}
     %     \centering
     %     \includegraphics[width=\textwidth]{figure/fl_sgd_test_loss_cata_agop_4_digit_std.pdf}
     %     \caption{Rank-4 regression}
     % \end{subfigure}\hspace*{-0.9em}
     \begin{subfigure}[b]{0.26\textwidth}
         \centering
         \includegraphics[width=\textwidth]{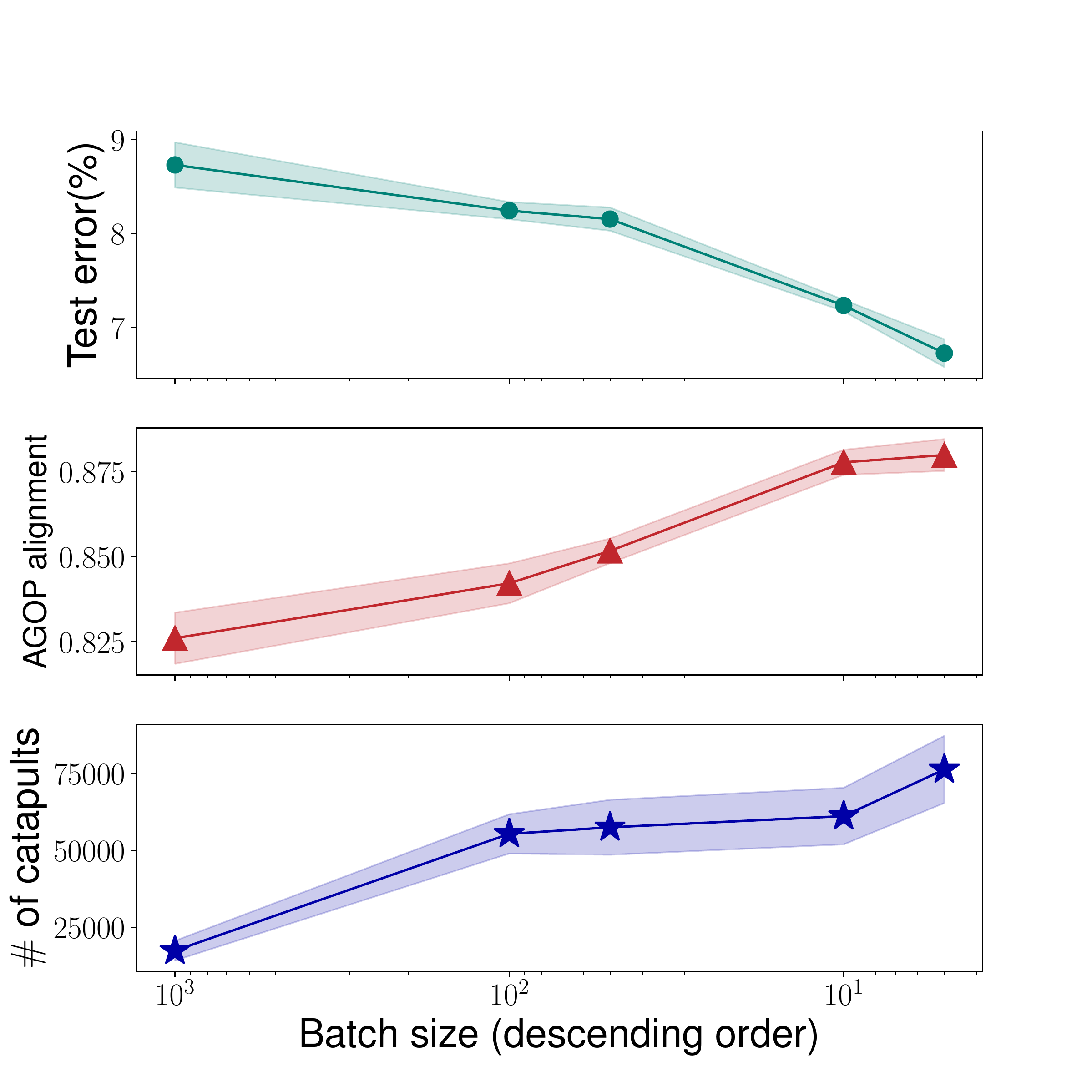}
         \caption{SVHN}
     \end{subfigure}\hspace*{-0.9em}
        \begin{subfigure}[b]{0.26\textwidth}
         \centering
         \includegraphics[width=\textwidth]{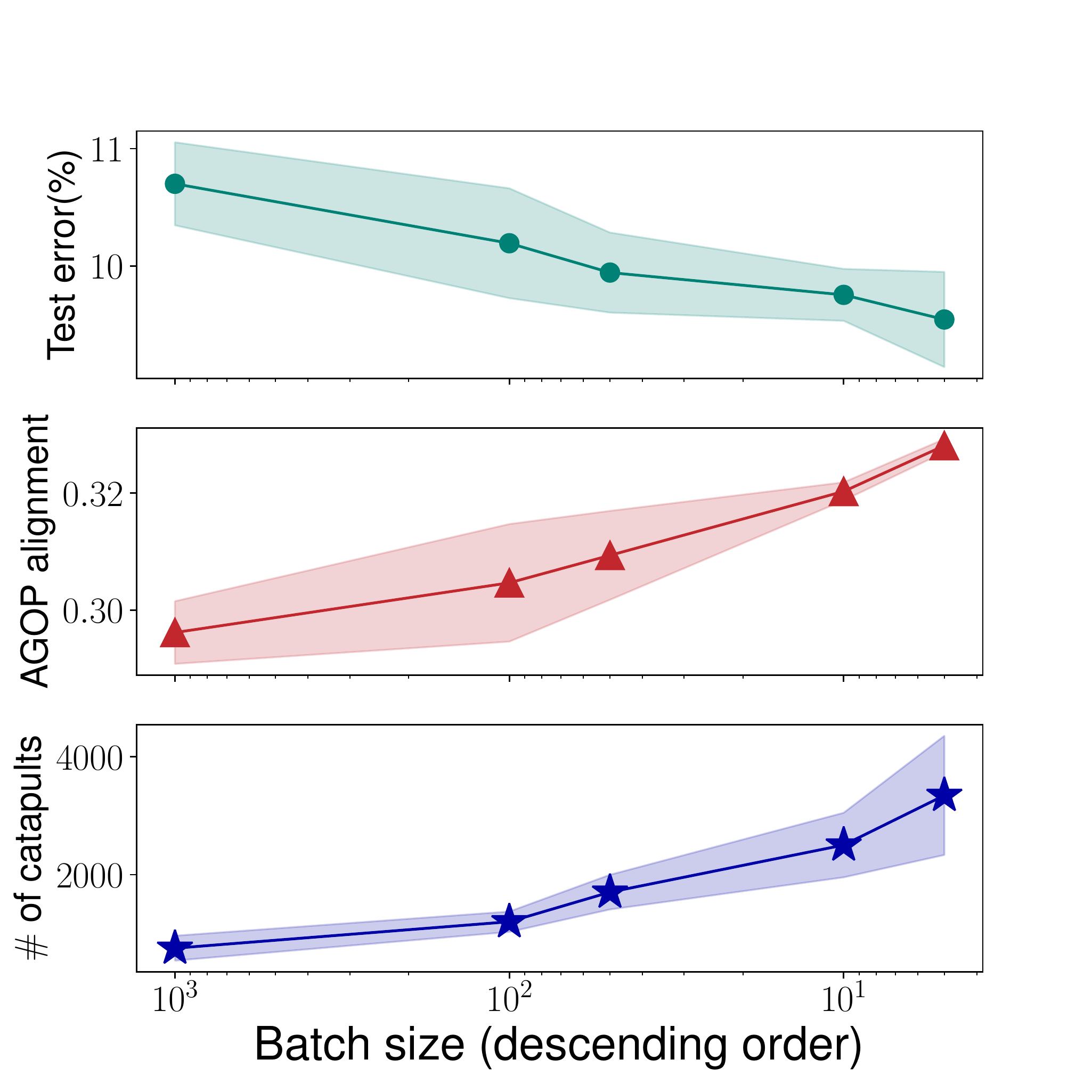}
         \caption{CelebA}
     \end{subfigure}\hspace*{-1em}
\caption{{\bf Correlation between AGOP alignment and test performance in SGD with Pytorch default parameterization.} The tasks are the same with Fig.~\ref{fig:fl_sgd_agop} except that we use  Pytorch default parameterization. \label{fig:fl_sgd_agop_std}}
\end{figure}

\subsection{Validation loss/error of SGD corresponding to Fig.~\ref{fig:fl_sgd_agop} and \ref{fig:fl_sgd_agop_std}}

\begin{figure}[H]
     \centering
     \begin{subfigure}[b]{0.4\textwidth}
\centering\includegraphics[width=\textwidth]{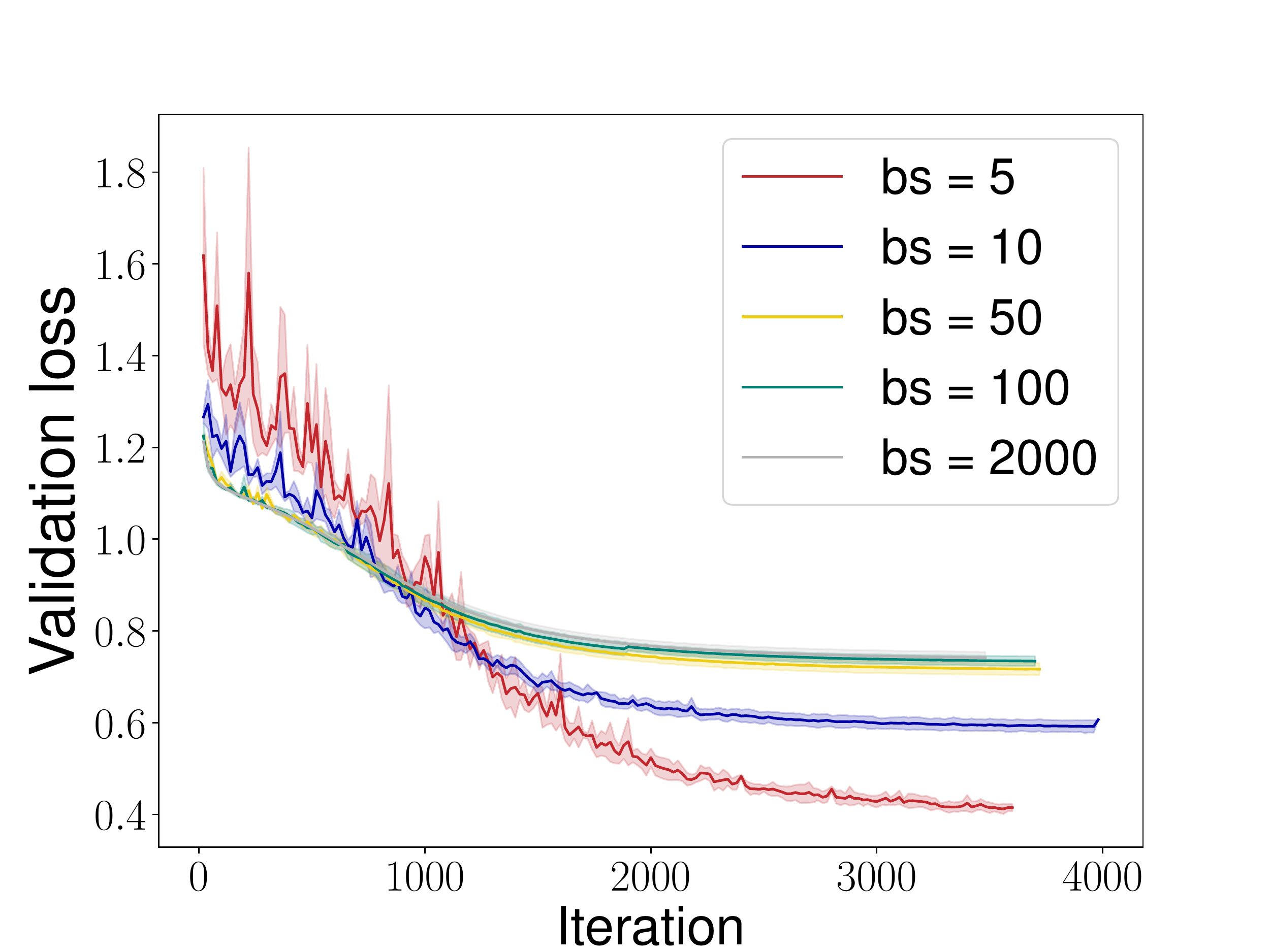}
             \caption{Rank-2 regression}
     \end{subfigure}\hspace*{-0.9em}
      \begin{subfigure}[b]{0.4\textwidth}
         \centering
         \includegraphics[width=\textwidth]{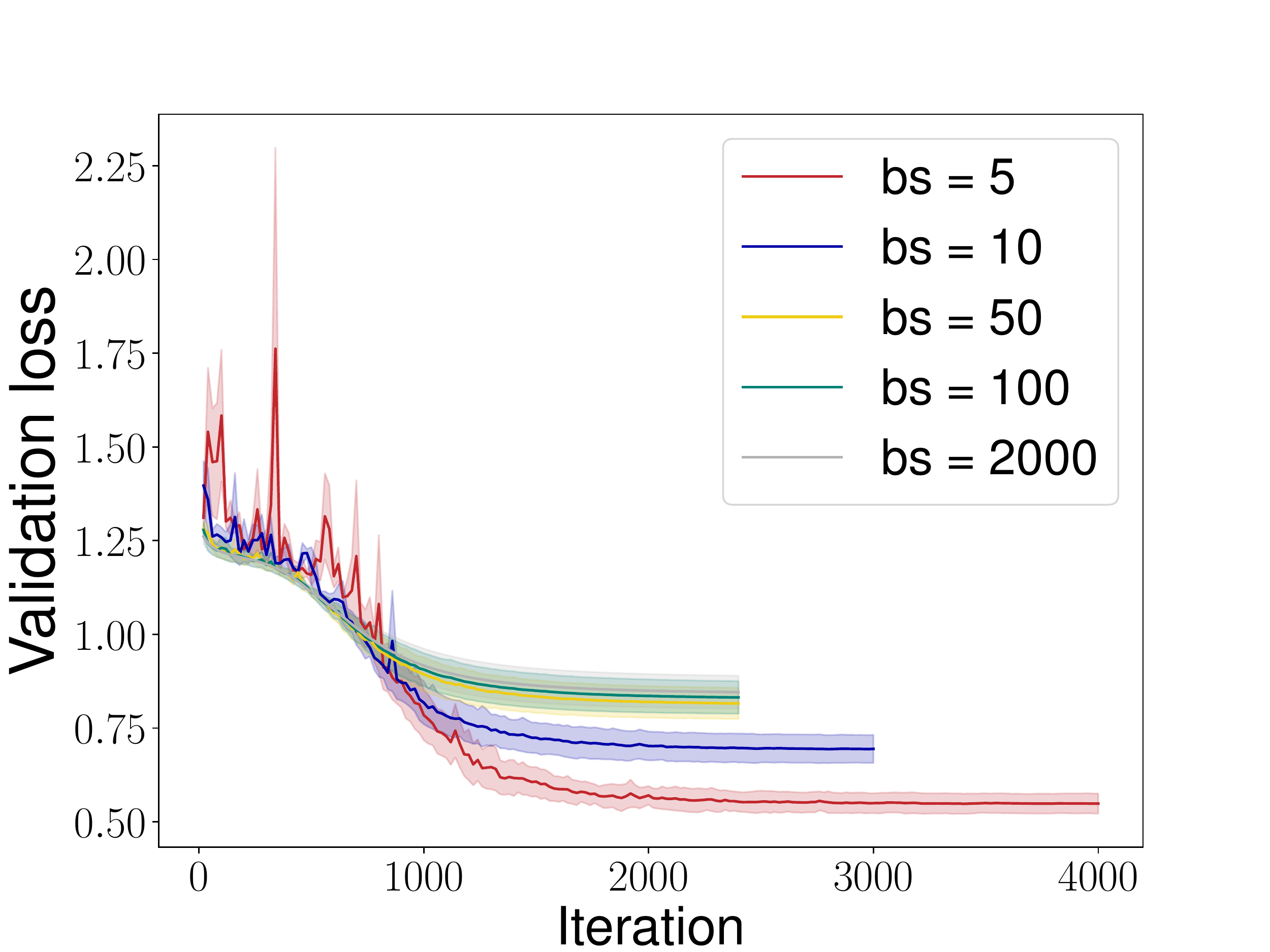}
             \caption{Rank-3 regression}
     \end{subfigure}
     % \begin{subfigure}[b]{0.33\textwidth}
     %     \centering
     %     \includegraphics[width=\textwidth]{figure/fl_sgd/fl_sgd_test_loss_ite_4_digit.pdf}
     %     \caption{Rank-4 regression}
     % \end{subfigure}\hspace*{-0.9em}
     \begin{subfigure}[b]{0.4\textwidth}
         \centering
         \includegraphics[width=\textwidth]{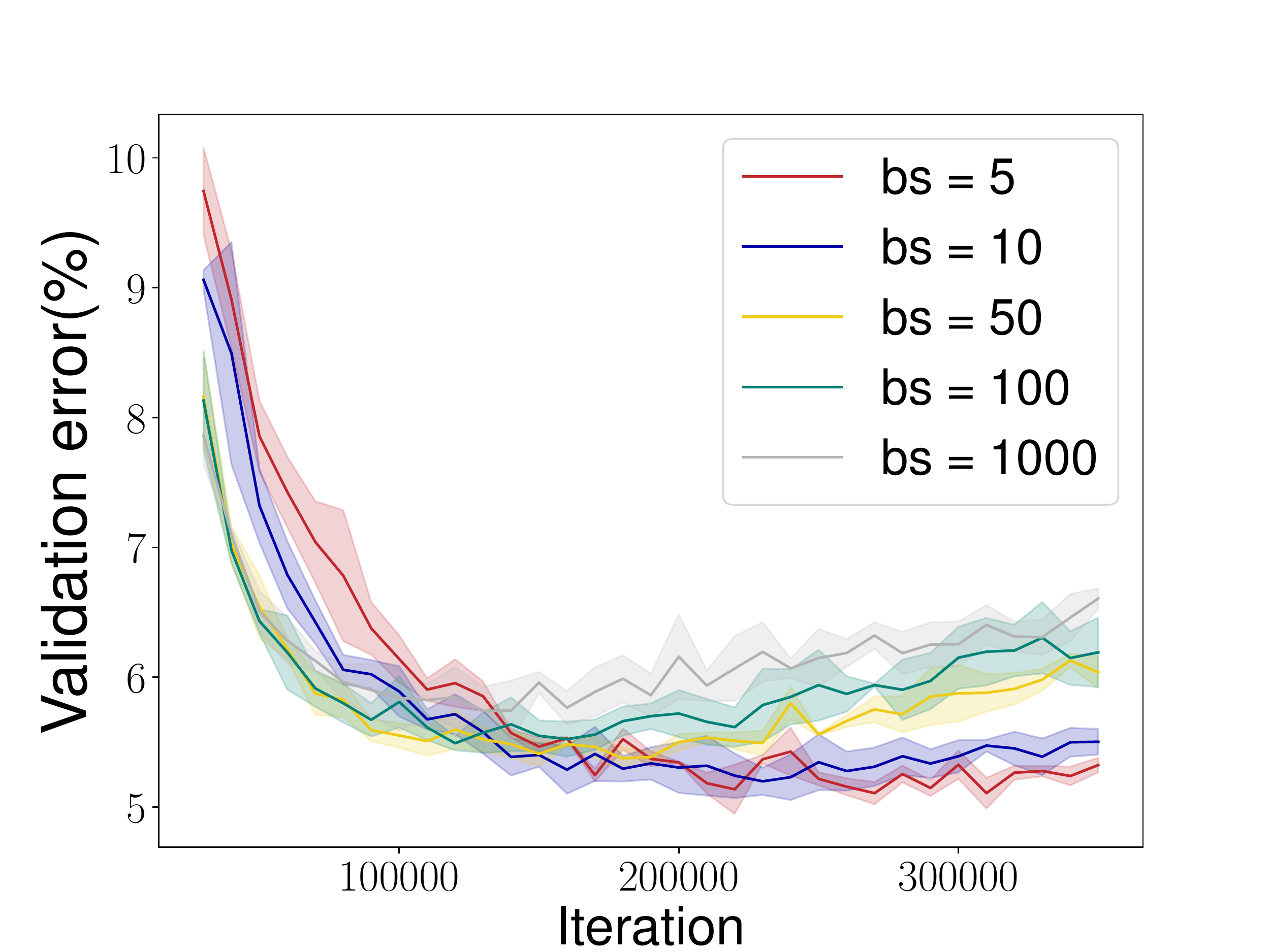}
         \caption{SVHN}
     \end{subfigure}
        \begin{subfigure}[b]{0.4\textwidth}
         \centering
         \includegraphics[width=\textwidth]{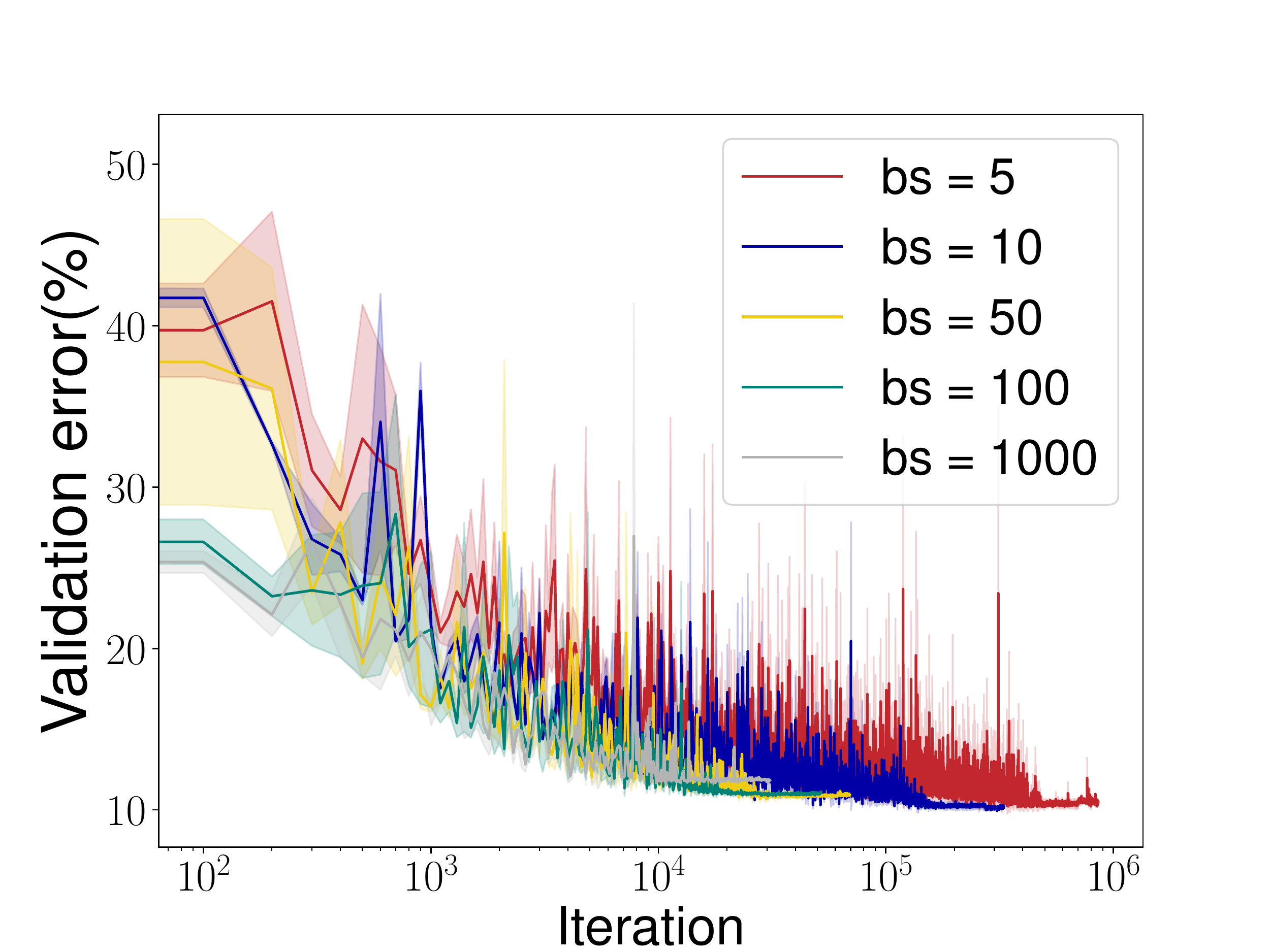}
         \caption{CelebA}
     \end{subfigure}
\caption{{\bf Validation loss/error corresponding to }Fig.~\ref{fig:fl_sgd_agop}. Panel(c) presents the validation error from iteration 4000.\label{fig:val_fl_sgd}} 
\end{figure}

\begin{figure}[H]
     \centering
     \begin{subfigure}[b]{0.4\textwidth}
         \centering
         \includegraphics[width=\textwidth]{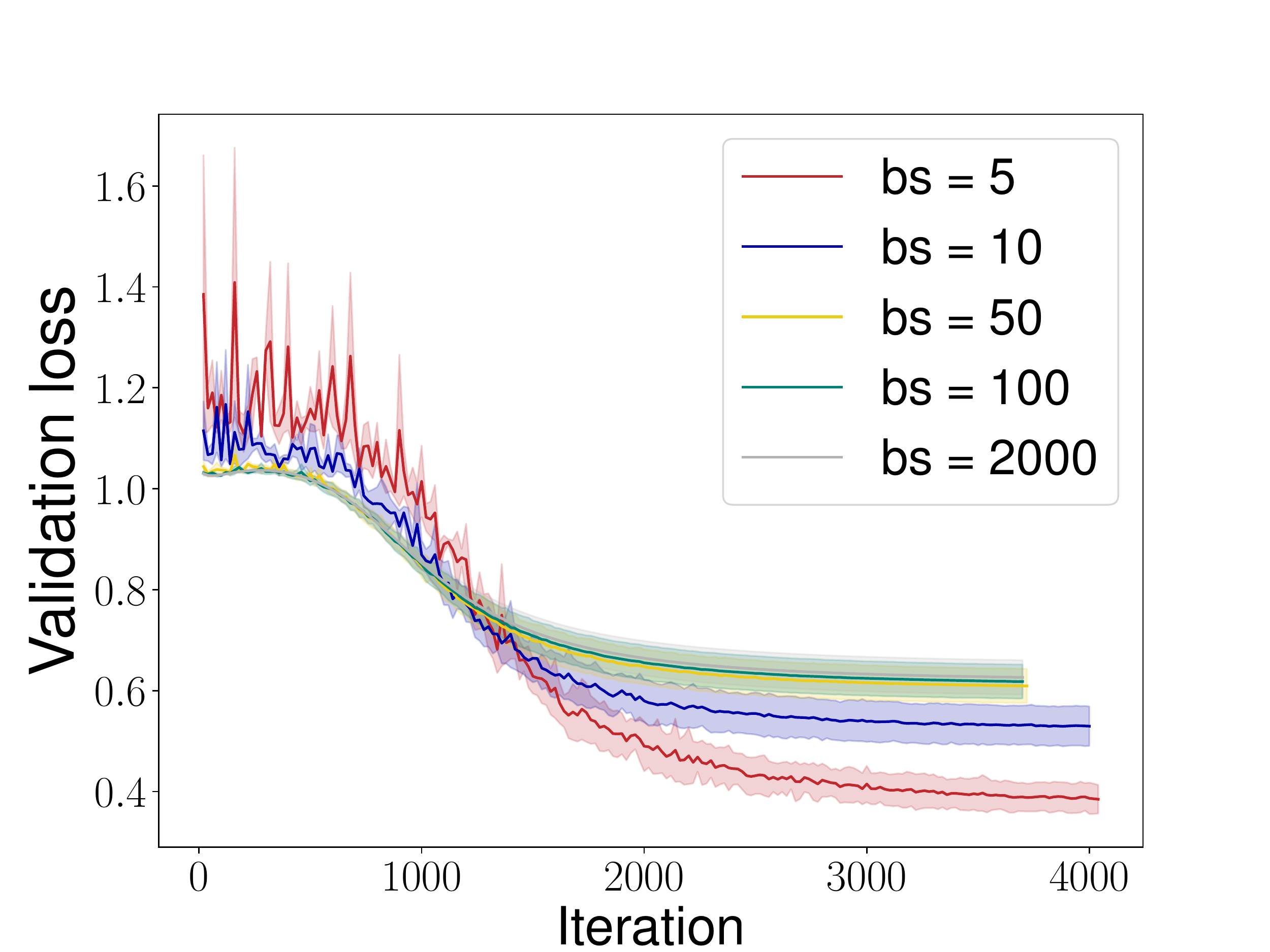}
         \caption{Rank-2}
     \end{subfigure}
          \begin{subfigure}[b]{0.4\textwidth}
         \centering
         \includegraphics[width=\textwidth]{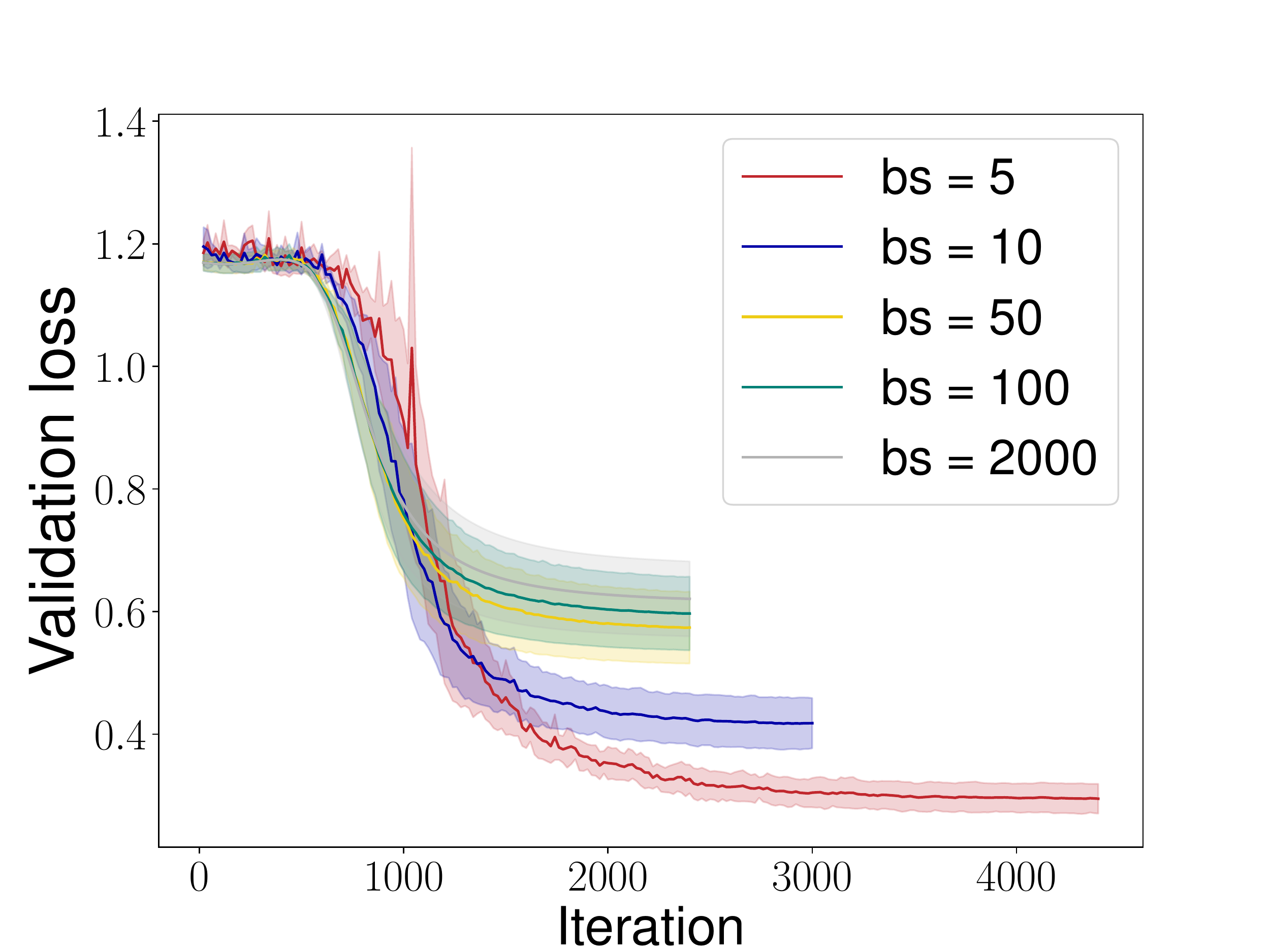}
         \caption{Rank-3}
     \end{subfigure}
     % \begin{subfigure}[b]{0.33\textwidth}
     %     \centering
     %     \includegraphics[width=\textwidth]{figure/fl_sgd_std/fl_sgd_test_loss_ite_4_digit_std.pdf}
     %     \caption{Rank-4}
     % \end{subfigure}\hspace*{-0.9em}
          \begin{subfigure}[b]{0.4\textwidth}
         \centering
         \includegraphics[width=\textwidth]{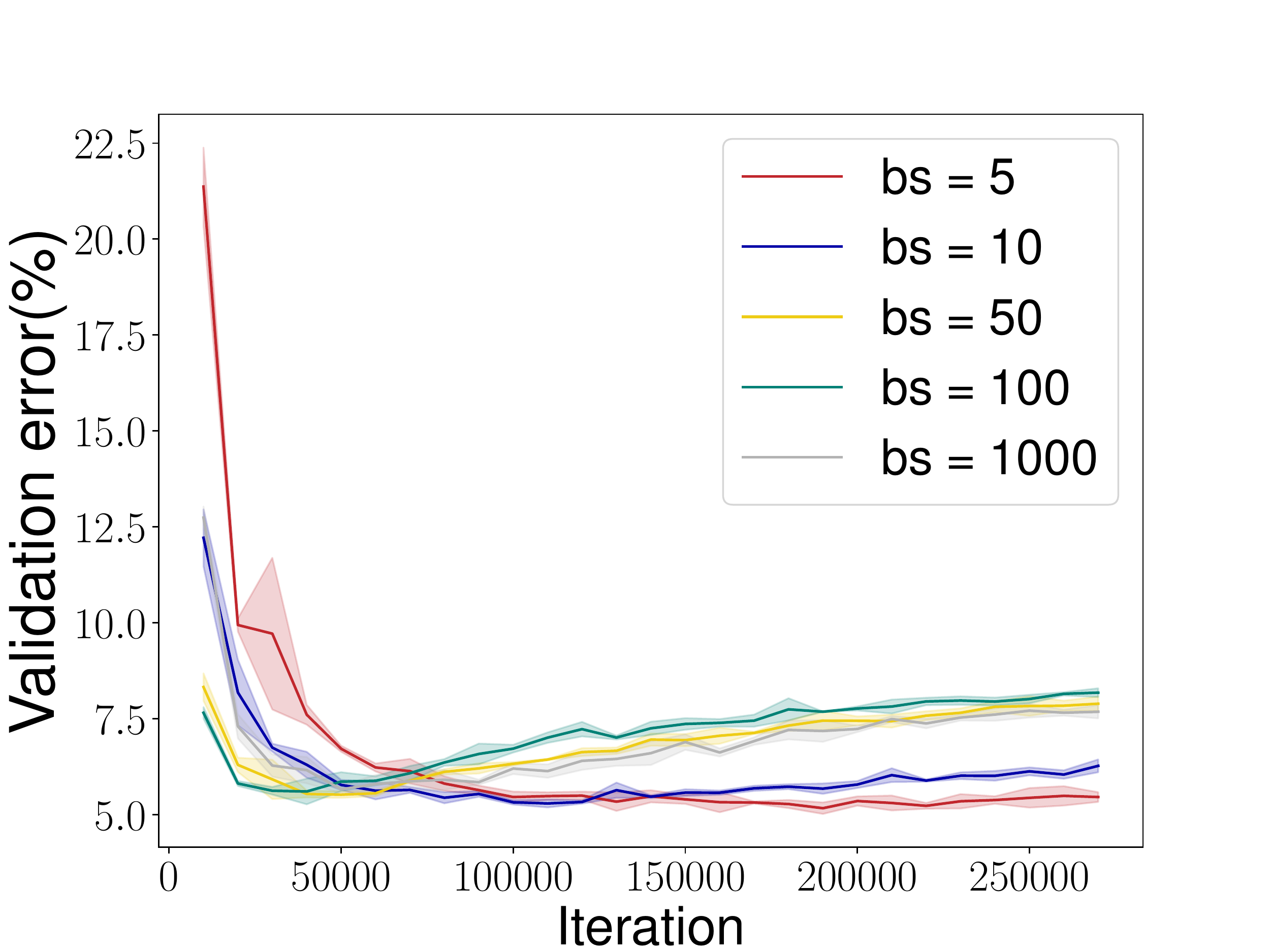}
         \caption{SVHN}
     \end{subfigure}
        \begin{subfigure}[b]{0.4\textwidth}
         \centering
         \includegraphics[width=\textwidth]{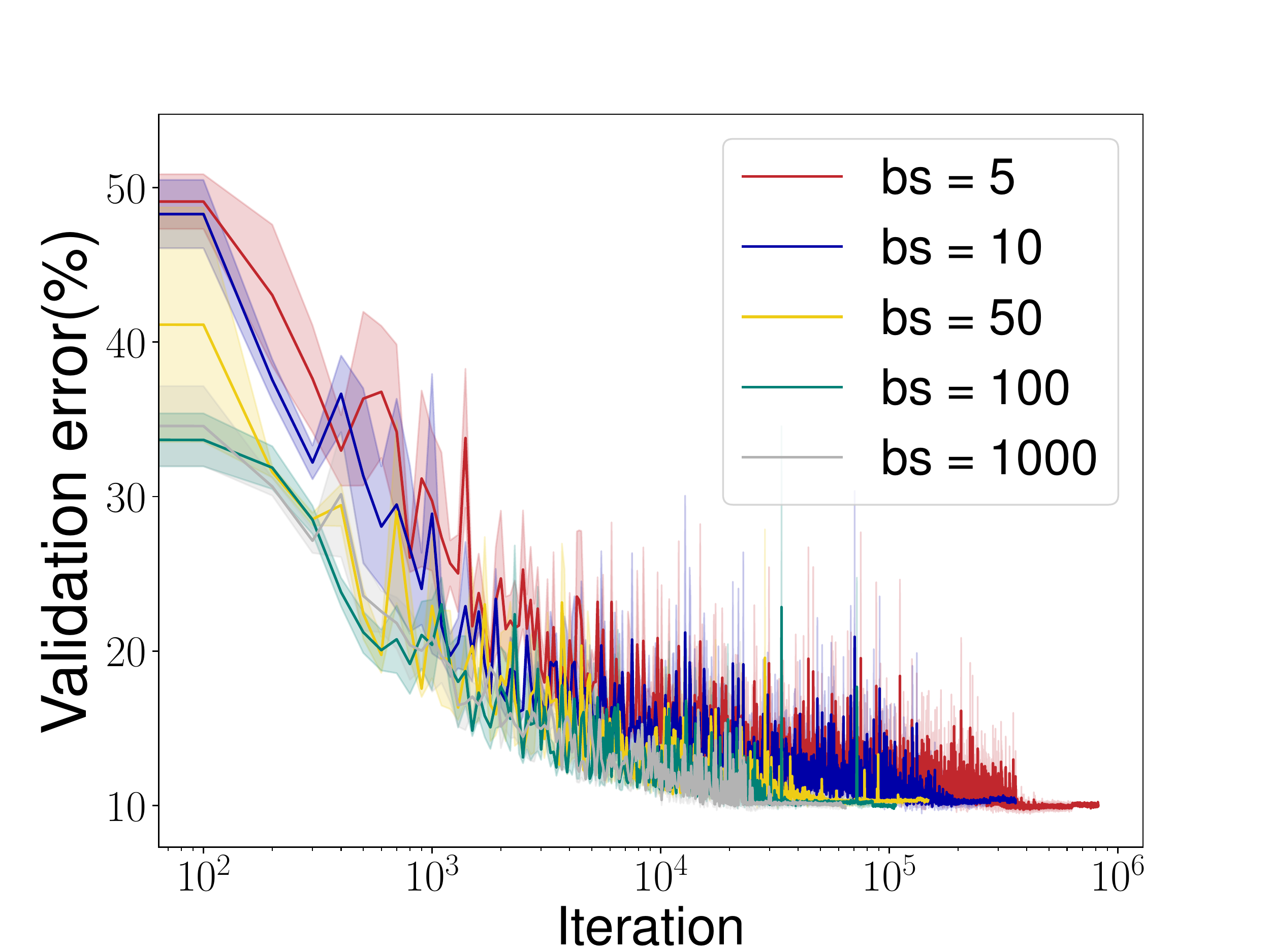}
         \caption{CelebA}
     \end{subfigure}
\caption{{\bf Validation loss/error with Pytorch default parameterization corresponding to }Fig.~\ref{fig:fl_sgd_agop_std}. Panel(c) presents the validation error from iteration 2000.\label{fig:val_fl_sgd_std}} 
\end{figure}

\subsection{Feature learning in SGD for additional datasets}\label{sec:add_data}
\begin{figure}[H]
     \centering\hspace*{-1em}
     \begin{subfigure}[b]{0.33\textwidth}
         \centering
         \includegraphics[width=\textwidth]{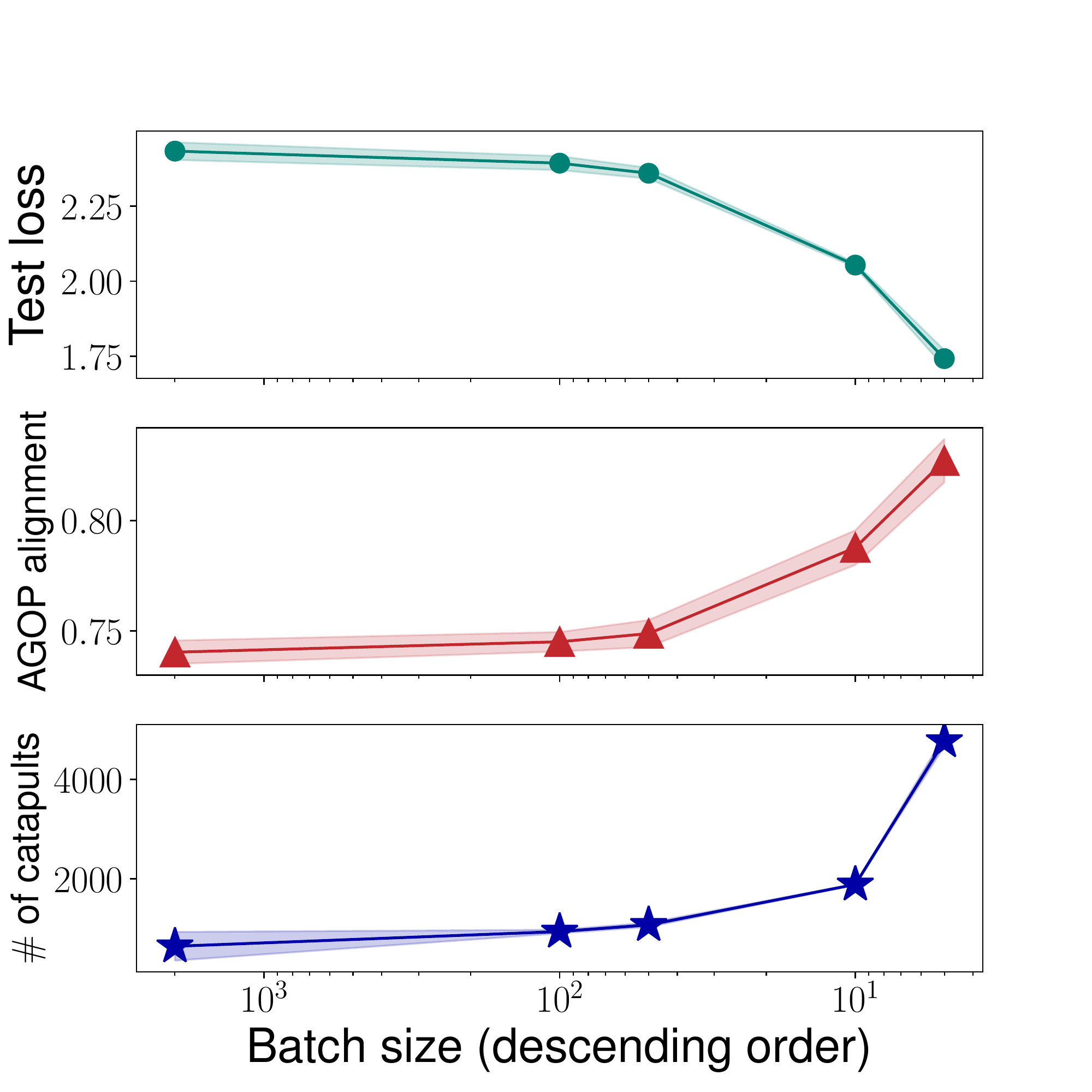}
         \caption{Rank-4 regression}
     \end{subfigure}\hspace*{-0.9em}
     \begin{subfigure}[b]{0.33\textwidth}
         \centering
         \includegraphics[width=\textwidth]{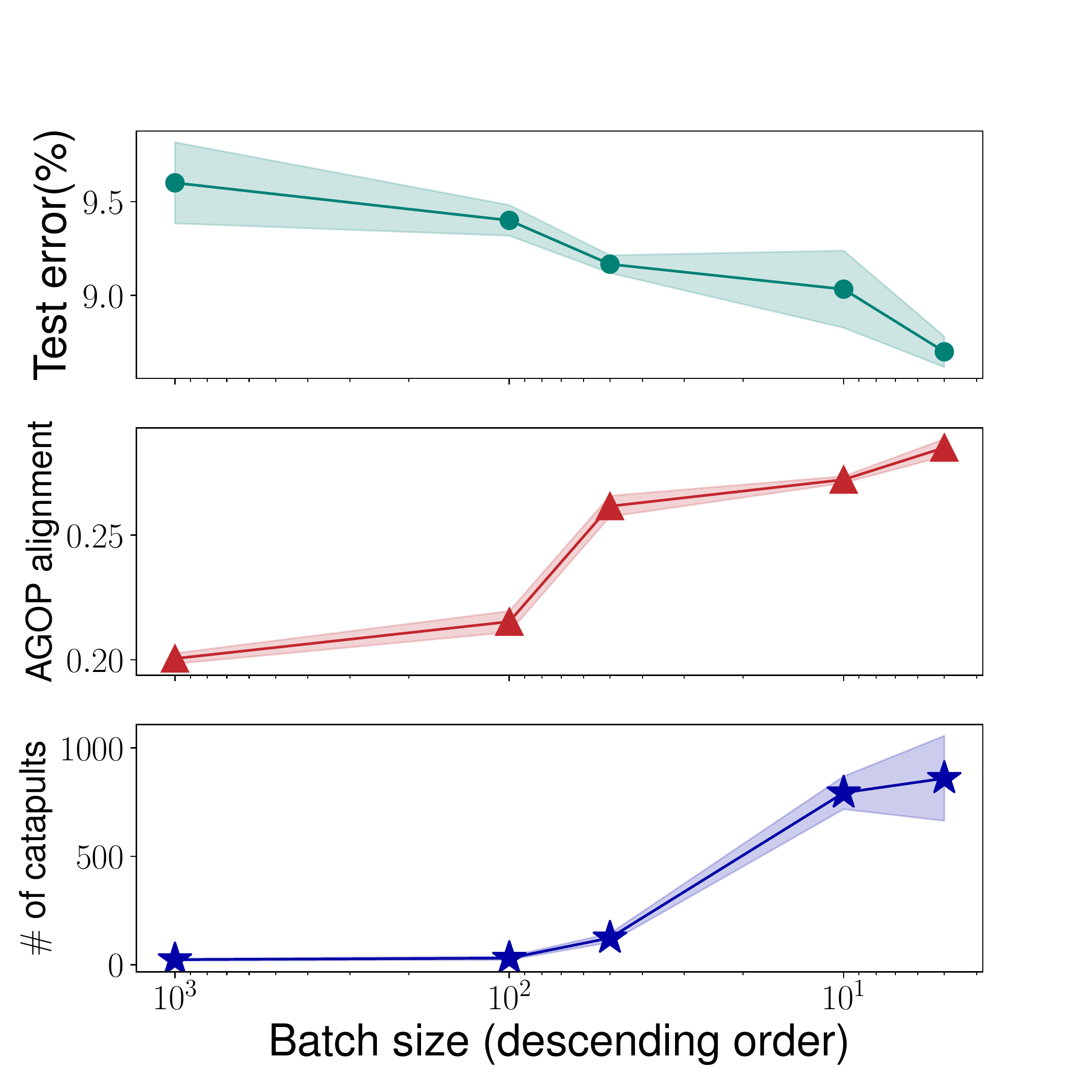}
         \caption{USPS}
     \end{subfigure}\hspace*{-0.9em}
        \begin{subfigure}[b]{0.33\textwidth}
         \centering
         \includegraphics[width=\textwidth]{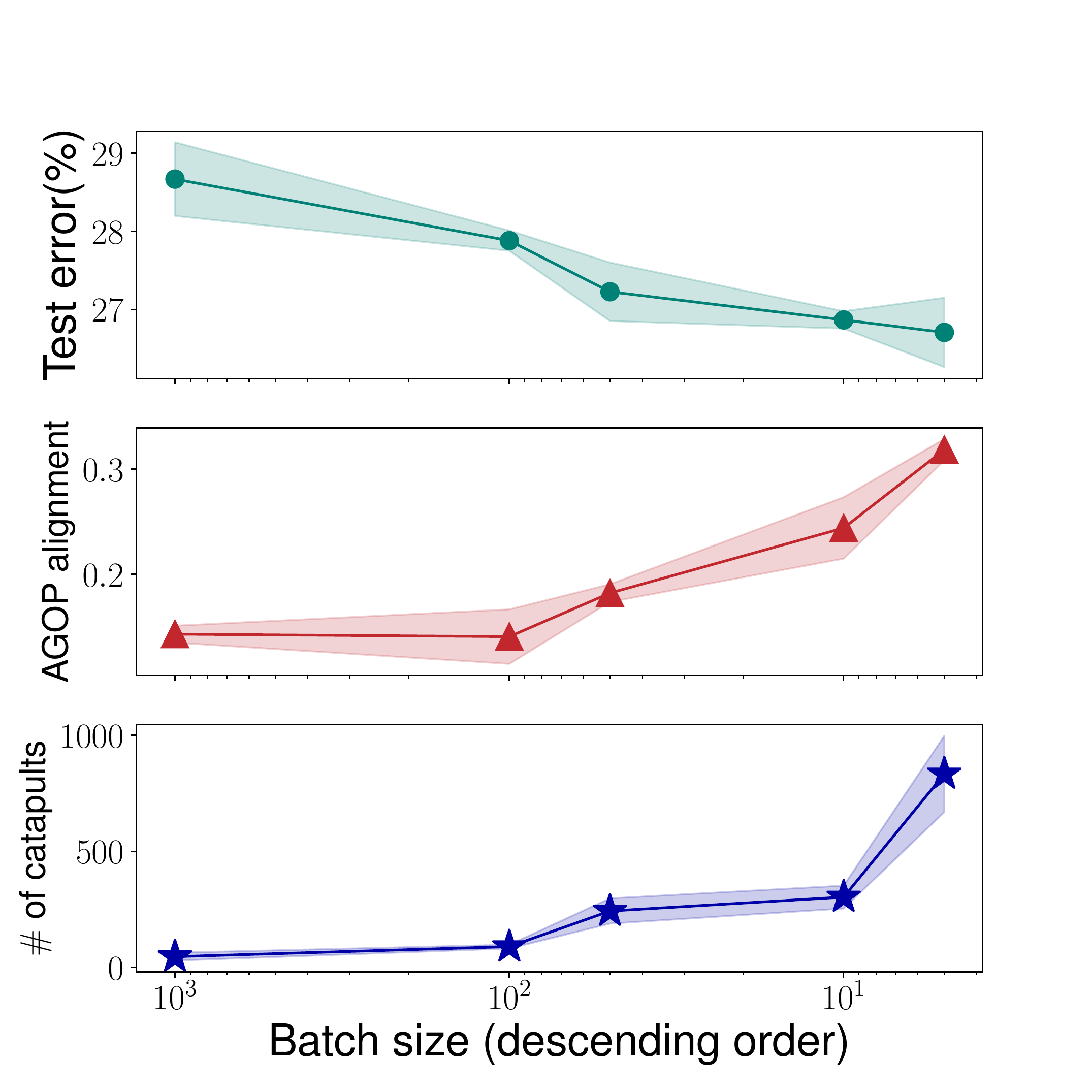}
         \caption{Fashion MNIST}
     \end{subfigure}\hspace*{-1em}
\caption{{\bf Correlation between AGOP alignment and test performance in SGD. } We train a 4-layer fully connected neural network using SGD. \label{fig:fl_sgd_agop_add}}
\end{figure}

In this section, we show the findings observed in Fig.~\ref{fig:fl_sgd_agop} hold for Rank-4 regression, USPS dataset and Fashion MNIST dataset. See Fig.~\ref{fig:fl_sgd_agop_add}.

\subsection{Verification of catapults in SGD}
\begin{figure}[H]
     \centering
        \begin{subfigure}[b]{0.45\textwidth}
         \centering
         \includegraphics[width=\textwidth]{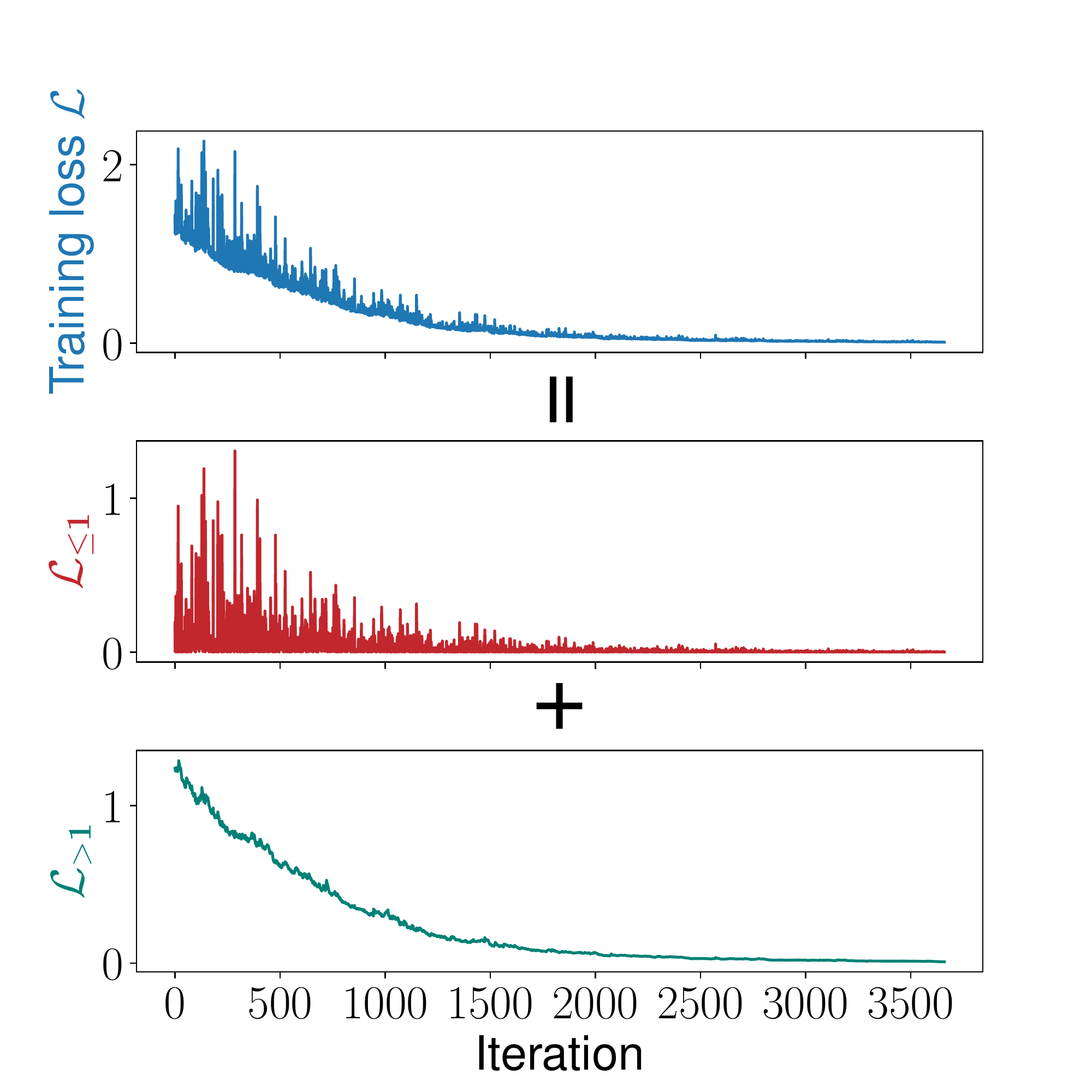}
         \caption{Loss decomposition (Rank-2)}
     \end{subfigure}
     \begin{subfigure}[b]{0.45\textwidth}
         \centering
         \includegraphics[width=\textwidth]{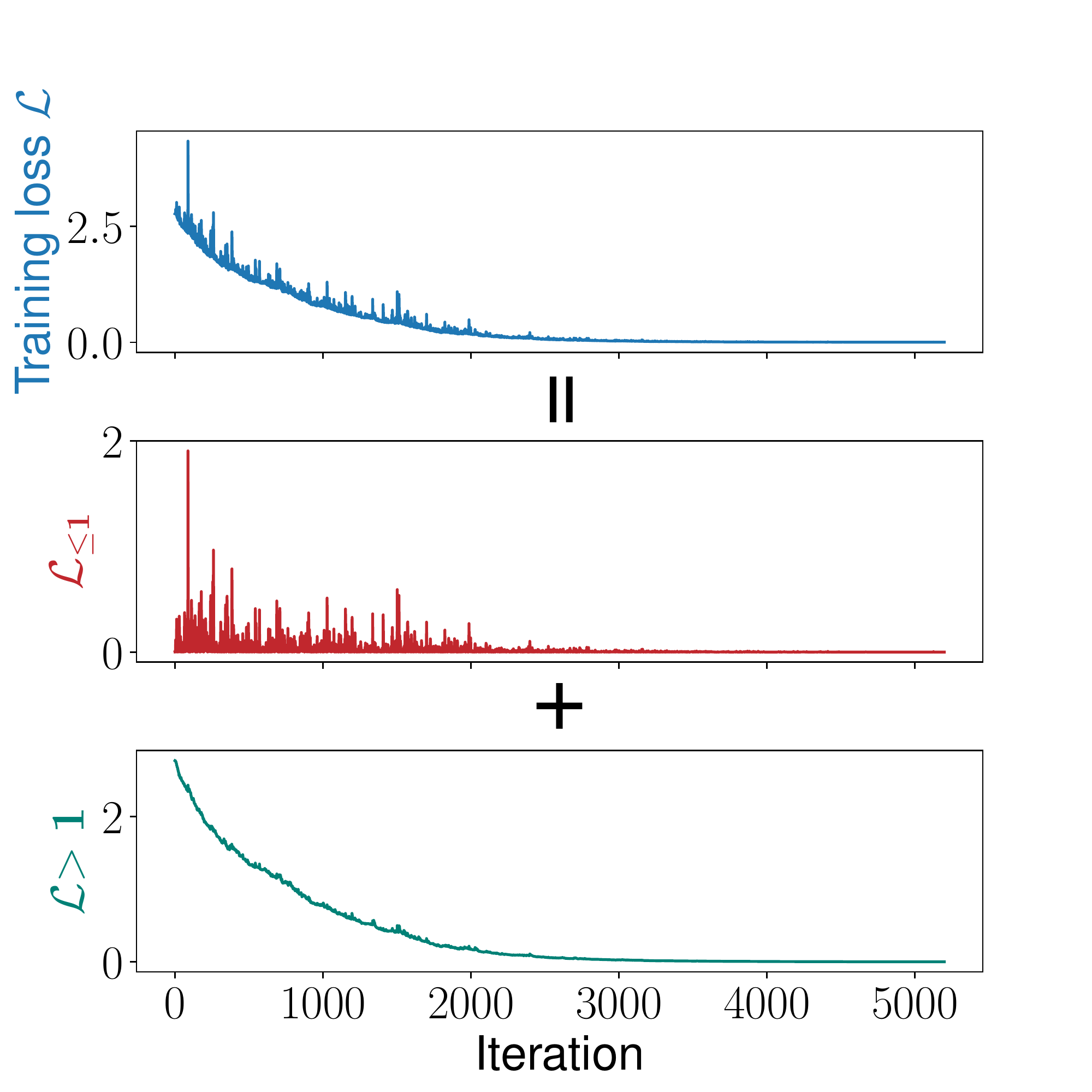}
         \caption{Loss decomposition (Rank-4)}
        \end{subfigure}
\caption{{\bf Verification of catapult dynamics: loss decomposition of Rank-2 and Rank-4 regression tasks corresponding to Fig.~\ref{fig:fl_sgd_agop} with batch size $5$.} The training loss is decomposed into the top eigenspace of the tangent kernel $\L_{\leq 1}$ and its complement $\L_{>1}$. Here $\L = \L_{\leq 1} + \L_{>1}$.\label{fig:fl_sgd_2_digit_decom}}
\end{figure}

In Fig.~\ref{fig:fl_sgd_2_digit_decom}, we verify that the spikes in the training loss of SGD with small batch sizes are caused by catapult dynamics. Specifically, we show that the spikes occur in the top eigendirection of the NTK.

\subsection{No feature learning with a small learning rate for SGD}
\begin{figure}[H]
     \centering
        \begin{subfigure}[b]{0.45\textwidth}
         \centering
         \includegraphics[width=\textwidth]{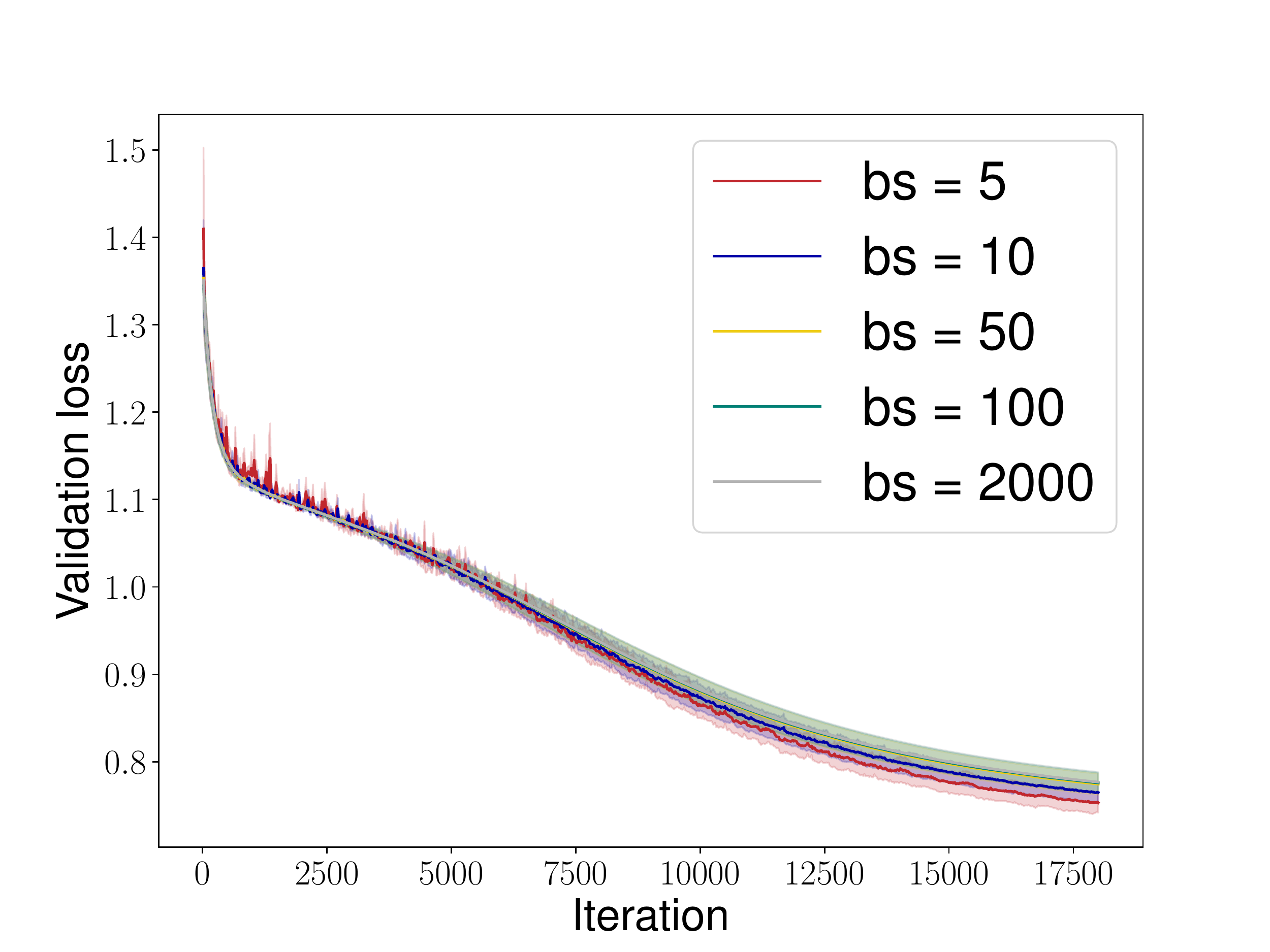}
         \caption{Rank-2}
     \end{subfigure}
     \begin{subfigure}[b]{0.45\textwidth}
         \centering
         \includegraphics[width=\textwidth]{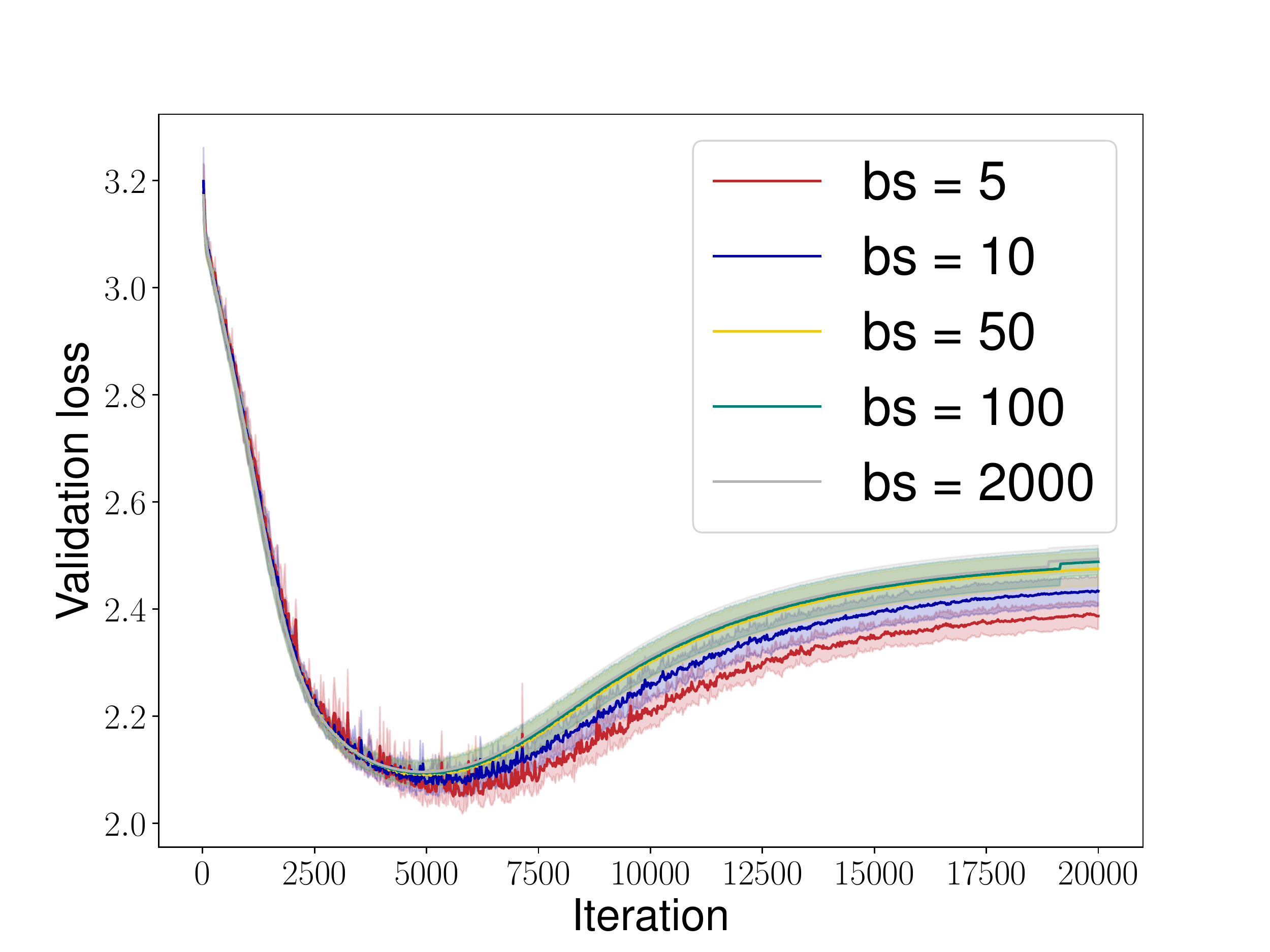}
         \caption{Rank-4}
        \end{subfigure}
\caption{{\bf The networks are trained with a smaller learning rate corresponding to Fig.~\ref{fig:val_fl_sgd}a \&b.}  We consider the same setting as in Fig.~\ref{fig:val_fl_sgd} except for selecting a smaller learning rate $\etc/40$ compared to  $ \etc/2$ in Fig.~\ref{fig:val_fl_sgd}.  Here $\etc$ is the critical learning rate for the whole dataset.\label{fig:fl_sgd_small_lr}}
\end{figure}

In Fig.~\ref{fig:fl_sgd_agop}, we have shown that a smaller batch size leads to more catapults, hence resulting in better test performance.  In this section, we show that the test performance with different batch sizes is similar when training with a small learning rate, where no catapults occur. This further verifies that a greater number of catapults accounts for better test performance for small batch sizes. See Fig.~\ref{fig:fl_sgd_small_lr}.

% Recall that the algorithmic learning rate needs to be large enough to induce catapults in SGD, as discussed in Section~\ref{sec:cata_sgd}. Specifically, the algorithmic learning rate needs to be larger than the critical learning rates for some of the batches.

% \begin{figure}[h]
%      \centering
%      \begin{subfigure}[b]{0.48\textwidth}
%          \centering
%      \includegraphics[width=\textwidth]{figure/fl_sgd_test_loss_svhn.png}\caption{}
%      \end{subfigure}
%      \begin{subfigure}[b]{0.48\textwidth}
%          \centering
%          \includegraphics[width=\textwidth]{figure/fl_sgd_agop_svhn.png}\caption{}
%      \end{subfigure}
% \end{figure}

\section{Experimental details}\label{sec:exp_details}
For all the networks considered in this paper, we use ReLU activation functions. We parameterize the networks by NTK parameterization~\cite{jacot2018neural}.  Note that NTK parametrization is widely used for understanding neural networks~\cite{lee2020finite,du2019gradient,lewkowycz2020large}. We also verify our results with Pytorch~\citep{paszke2019pytorch} default parameterization for the experiments shown in Fig.~\ref{fig:sgd_deep_std} and \ref{fig:fl_sgd_agop_std}.

\paragraph{NTK parameterization.} Given a neural network with NTK parameterization, all the trainable weight parameters are i.i.d. from $\mathcal{N}(0,1)$. For a fully connected layer, it takes the form $f^{\ell+1} =\mathsf{ReLU}\round{\frac{1}{\sqrt{m_{\ell}}} W^\ell f^{\ell} + b^\ell}$ where $W^\ell \in\mathbb{R}^{m_{\ell+1}\times m_\ell}, f^{\ell}\in\mathbb{R}^{m_\ell}, b^\ell \in\mathbb{R}^{m_{\ell+1}}$. For a convolutional layer, it takes the form $f^{\ell+1}_{i,j,k} = \mathsf{ReLU}\round{\frac{1}{\sqrt{m_\ell s^2}}\sum_{p=0}^{\ceil{\frac{s+1}{2}}} \sum_{q=0}^{\ceil{\frac{s+1}{2}}} \sum_{o = 1}^{m_\ell} W^\ell_{p,q,o,k} f^\ell_{i-\ceil{\frac{s-1}{2}},j-\ceil{\frac{s-1}{2}},o}+ b_k^\ell}$, where $W^\ell \in \mathbb{R}^{s\times s\times m_\ell \times m_{\ell+1}}, f^{\ell} \in \mathbb{R}^{h\times w\times m_\ell }, b^\ell \in \mathbb{R}^{m_{\ell+1}}$. Note that $s$ is the filter size and we assume the stride to be $1$ in this case. For $f^{\ell}$ with negative indices, we let it be $0$, i.e., zero padding. For the output layer, we use a linear layer without activation functions. 
% For CNNs, the output before the last layer is flattened into a vector then fed into a fully connected layer.

\paragraph{Dataset.}For the synthetic datasets, we generate data $\{(\vx_i,y_i)\}_{i=1}^n$ by i.i.d. $\vx_i\sim \mathcal{N}(0,I_{100})$ and $y_i = f^*(\vx)+\epsilon$ with $\epsilon~\sim\mathcal{N}(0,0.1^2)$. For two real-world datasets, we consider a subset of CelebA dataset with glasses as the label,  the Street View House Numbers (SVHN) dataset, USPS dataset and Fashion MNIST dataset. Due to computational limitations with GD, for some tasks, we select two classes (number $0$ and $2$) of SVHN dataset, USPS dataset and Fashion MNIST dataset.

\paragraph{EGOP (Epexcted Gradient Outer Product).}Note that for these low-rank polynomial regression tasks,  we know the analytical form of target functions hence we can calculate the EGOP by $G^* = \mathbb{E}_{\vx}  \frac{\partial f^*}{\partial \vx}\frac{\partial f^*}{\partial\vx}^T$. For real-world datasets, we estimate the EGOP by using the AGOP of one of the state-of-the-art models $\hat{f}$ that achieve high test accuracy: $\hat{G} = \frac{1}{n} \sum_{i=1}^n\frac{\partial \hat{f}}{\partial \vx_i}\frac{\partial \hat{f}}{\partial\vx_i}^T$. 

In the following, we provide the detailed experimental setup for each experiment. Note that in the classification tasks, i.e. CelebA and SVHN datasets, the test error refers to the classification error on the test split.

\subsection{Experiments in Section~\ref{sec:cata_gd}}\label{exp:cata_gd}
\paragraph{Fig.~\ref{fig:cata_fcn}:} We use a 2-class subset of CIFAR-10 dataset~\cite{krizhevsky2009learning} (class 7 and class 9) and randomly select 128 data points out of it. For the network architectures, we use a 5-layer FCN with width $1024$ and 5-layer CNN with $512$ channels per layer. For CNN, we flatten the image into a one-dimensional vector before the last fully connected layer. 

\subsection{Experiments in Section~\ref{subsec:multi_cata}}\label{exp:muli_cata}
\paragraph{Fig.~\ref{fig:multi_cata}:} We use the same training tasks as in Fig.~\ref{fig:cata_fcn}. For FCN, we start with a learning rate $6$ and we increase the learning rate to $[10,15]$ at iteration $[15,60]$.  For CNN, we start with a learning rate $8$ and we  increase the learning rate to $[15,20]$ at iteration $[10,40]$.

\subsection{Experiments in Section~\ref{sec:cata_sgd}}\label{exp:cata_sgd}
\paragraph{Fig.~\ref{fig:loss_lr_match_sgd}:} We consider a synthetic dataset ${(x_i,y_i)}$ where $x_i$ are sampled i.i.d. on unit sphere and $y_i = 1$ with training size and dimension both equal to $100$. We train a wide two-layer ReLU network with second-layer weights fixed, using SGD with batch size one. The critical learning rate $\etc(x_i)$ for (a minibatch of size one) $x_i$ is proportional to $1/\norm{x_i}_2^2 = const$. We select one data point $(x_*, y_*)$ from the training set and multiply both $x_*$ and $y_*$ by $2$. This makes the critical learning rate corresponding to the sample $x_*$ four times smaller. We choose the (constant) learning rate $\eta$ between the critical learning rate of (the mini-batch) $x_*$ and the critical learning rates of the rest of the data points. Thus SGD with learning rate $\eta$ induces catapult on $x_*$ but not on any other data points. 
% During training, we observe that there is indeed a spike in the total training loss when the gradient is computed on (minibatch) $x_*$, otherwise, as expected, the loss decreases monotonically. 

\paragraph{Fig.~\ref{fig:sign_cata}:} For the shallow network, we use a 2-layer FCN with width $1024$.  We train the model on $128$ data points from CIFAR 2 using SGD with batch size $32$. We use a constant learning rate $0.8$.  We stop training when the training loss is less than $10^{-3}$.

\paragraph{Table~\ref{tab:match_rate}:} The 5-layer FCN and CNN are the same as in Fig.~\ref{fig:cata_fcn}.  We train the model on $128$ data points from CIFAR 2 using SGD with batch size $32$. We use a constant learning rate $6$ and $8$ for 5-layer FCN and CNN respectively. We stop training when the training loss is less than $10^{-3}$.

\paragraph{Fig.~\ref{fig:sgd_deep}:} The 5-layer FCN and CNN are the same as in Fig.~\ref{fig:cata_fcn}. And we use the standard Wide ResNets 10-10 and ViT-4 architectures.  The learning rates for 5-layer FCN, 5-layer CNN,  are $6,8,3,0.2$ respectively. We train the model with a constant learning rate, and we stop training when the training loss is less than $10^{-3}$. All the models are trained on $128$ data points from CIFAR-2 using SGD with batch size $32$.

% \paragraph{Fig.~\ref{fig:cyc_lr}:}We use the standard Wide ResNets architectures, the same with Fig.~\ref{fig:sgd_deep}c. The models is trained on $128$ data points from CIFAR-2 using SGD with batch size $32$.

\subsection{Experiments in Section~\ref{sec:feature_learning}}\label{exp:feature_learning}
\paragraph{Fig.~\ref{fig:multi_cata_gd_low_rank}:} 
For rank-2 task, we train a 2-layer FCN with width $1024$. The size of the training set, testing set and validation set are $2000, 5000$ and $5000$ respectively.

For rank-3 task, CelebA tasks, we train a 4-layer FCN with width $256$. The size of the training set, testing set and validation set are $1000, 5000$ and $5000$ respectively.

For SVHN-2 tasks, we train a 5-layer CNN with width $256$. We select class 0 and class 2 out of the full SVHN datasets as SVHN-2. The size of the training set, testing set and validation set are $1000, 5000$ and $5000$ respectively.

 We  increase the learning rate during training. For Rank-2 task, we increase the learning rate to $[8,16,30,50,75,80]$ at iteration $[50,150,220,280,350,400]$. For Rank-3 task, we increase the learning rate to $[40,100,150]$ at iteration $[20,60,80]$.  For SVHN-2 task, we increase the learning rate to $[30,60,90]$ at iteration $[10,35,50]$. For CelebA task, we increase the learning rate to $[40,70,100]$ at iteration $[10,35,50]$. We decay the learning rate if necessary after the catapult to avoid extra catapults until the end of training.

\paragraph{Fig.~\ref{fig:fl_sgd_agop}:} For both Rank-2 and Rank-3 tasks, we let the size of training set, testing set and validation set be $2000, 5000$ and $5000$. For the SVHN task, we train the full SVHN using the 5-layer Myrtle network. For the CelebA task, we train the full 2-class CelebA dataset with glasses feature using 4-layer FCN with width $256$.
 To obtain the true \AGOP{}, we use one of the SOTA models (WideResNet 16-2) which achieves $97.2\%$ test accuracy on SVHN and 5-layer Myrtle network which achieves $95.7\%$ test accuracy on CelebA.  

We use the same learning rate across batch sizes for each task. The learning rate is chosen as $\frac{1}{2}\etc$ corresponding to the whole training set. For SVHN and CelebA tasks, we estimate $\etc$ using a subset with size $5000$ of the whole training set. We train the model with batch size $[5,10,50,100,2000]$.  For all tasks,  we stop training when the training loss is less than $10^{-3}$. We report the average of 3 independent runs.

\paragraph{Fig.~\ref{fig:optim_agop}:} 
We use the same network architectures and training/validation/testing sets as in Fig.~\ref{fig:multi_cata_gd_low_rank}.

For all the tasks, except for GD, all the optimizers use a mini-batch size $100$.

We stop training when the training loss is less than $10^{-3}$. We report the average of 3 independent runs. 

For the rank-2 task and rank-4 task, we know the target function hence we can analytically compute the exact true \AGOP{}. For SVHN-2 task and CelebA task, to estimate the true \AGOP{}, we use one of the SOTA models, Myrtle-5 which achieves $98.4\%$ test accuracy on two-class SVHN dataset and $95.7\%$  test accuracy on CelebA dataset.

The following table is the learning rate we choose for the experiments:

\begin{table}[H]
\begin{center}
\begin{tabular}{cccccccc}
\multicolumn{1}{c}{\bf Task}&\multicolumn{1}{c}{\bf SGD}  &\multicolumn{1}{c}{\bf GD} &\multicolumn{1}{c}{\bf SGD+M} &\multicolumn{1}{c}{\bf Adadelta}  &\multicolumn{1}{c}{\bf Adagrad } &\multicolumn{1}{c}{\bf RMSprop } &\multicolumn{1}{c}{\bf Adam }
\\  \midrule \midrule
Rank-2 &2.0&2.0&2.0&2.0&0.1&$10^{-2}$&$10^{-2}$ \\  \midrule
Rank-3  &2.0&2.0&2.0&2.0&$10^{-2}$&$10^{-2}$&$10^{-3}$ \\\midrule
Rank-4  &1.0&1.0&1.0&1.0&$5\times10^{-3}$&$10^{-3}$&$10^{-3}$ \\\midrule
SVHN-2  &5.0&5.0&5.0&5.0&$5\times10^{-3}$&$10^{-4}$&$10^{-3}$ \\\midrule
CelebA & 10.0&10.0&10.0&10.0&$5\times10^{-3}$ &$10^{-3}$ &$10^{-3}$\\
\end{tabular}
\end{center}
\caption{{\bf Choice of learning rates for Fig.~\ref{fig:optim_agop}.} }\label{tab:test_acc}
\end{table}

The experiment is to demonstrate the correlation between AGOP alignment and test performance. For this reason, we did not fine-tune the learning rate to achieve the best test performance.

% \subsection{Experiments in Appendix~\ref{sec:h_K}}
% \paragraph{Fig.~\ref{fig:sign_match_tilde_crit}:}

\subsection{Experiments in Appendix~\ref{sec:sgd_add}}\label{exp:sgd_add}
\paragraph{Fig.~\ref{fig:cata_fcn_full}:} We use the same network architectures as in Fig.~\ref{fig:cata_fcn} and we train $128$ data point from CIFAR-10. 

\paragraph{Fig.~\ref{fig:sgd_deep_svhn}:} We use the same setting as Fig.~\ref{fig:sgd_deep}, except we train the networks on $128$ data points from SVHN-2(number 0 and 2).  

\paragraph{Fig.~\ref{fig:large_sgd}:} For panel(a) and panel(b), we train the same 5-layer FCN and CNN as in Fig.~\ref{fig:cata_fcn} and on 5,000 data points from CIFAR-2. For panel(c), we train a 5-layer Myrtle network on 128 points from CIFAR-10.

% \paragraph{Fig.~\ref{fig:fl_gd_3_digit}:}We train a 3-layer FCN with width $1024$. The size of the training set, testing set and validation set are $2000, 5000$ and $5000$ respectively. We increase the learning rate to $[40,100,150]$ at iteration $[20,60,80]$ to generate multiple catapults.

\subsection{Experiments in Appendix~\ref{sec:add_agop_gd}}\label{exp:add_agop_gd}
\paragraph{Fig.~\ref{fig:multi_cata_gd_low_rank_add}:} For rank-4 task, USPS dataset and Fashion MNIST dataset, we train a 4-layer FCN with width $256$. The size of the training set, testing set and validation set are $1000, 5000$ and $5000$ respectively.

For rank-4 task,   we increase the learning rate to $[15,40,60]$ at iteration $[50,75,110]$. For USPS dataset, we increase the learning rate to $[15,30,40]$ at iteration $[10,30,45]$. For Fashion MNIST dataset, we increase the learning rate to $[10,40,55]$ at iteration $[6,20,30]$.

\paragraph{Fig.~\ref{fig:multi_cata_gd_full_rank}:} We train a 2-layer FCN with width $1024$. We consider a synthetic dataset, where $f^*(\vx) = \frac{1}{\sqrt{d}}\norm{\vx}$. The size of the training set and validation set is $128, 2000$ respectively. During training, we start with lr=$6$ and increase the learning rate to $[7,12,40,80]$ at iteration $[30,120,180,280] $.

\subsection{Experiments in Appendix~\ref{sec:add_agop_sgd}}\label{exp:add_agop_sgd}
\paragraph{Fig.~\ref{fig:fl_sgd_agop_std}:} We use the same setup with Fig.~\ref{fig:fl_sgd_agop} except that all the networks are parameterized with Pytorch default parameterization. The learning rates are $0.01$, $0.01$, $0.05$ and $1.0$ for each task.
\paragraph{Fig.~\ref{fig:fl_sgd_agop_add}:} For rank-4 task, USPS dataset and Fashion MNIST dataset, we train a 4-layer FCN with width $256$. The size of the training set, testing set and validation set are $2000, 5000$ and $5000$ respectively. We add $10\%$ label noise for the USPS dataset and Fashion MNIST dataset. 
To obtain the true \AGOP, we use one of the SOTA models (5-layer CNN) which achieves $99.2\%$ test accuracy on USPS and 5-layer Myrtle network which achieves $91.8\%$ test accuracy on Fashion MNIST.

%% file: example_paper.bib
@inproceedings{jacot2018neural,
  title={Neural tangent kernel: Convergence and generalization in neural networks},
  author={Jacot, Arthur and Gabriel, Franck and Hongler, Cl{\'e}ment},
  booktitle={Advances in neural information processing systems},
  pages={8571--8580},
  year={2018}
}

@inproceedings{
nguyen2018loss,
title={On the loss landscape of a class of deep neural networks with no bad local valleys},
author={Quynh Nguyen and Mahesh Chandra Mukkamala and Matthias Hein},
booktitle={International Conference on Learning Representations},
year={2019},
url={https://openreview.net/forum?id=HJgXsjA5tQ},
}

@inproceedings{du2019gradient,
  title={Gradient descent finds global minima of deep neural networks},
  author={Du, Simon and Lee, Jason and Li, Haochuan and Wang, Liwei and Zhai, Xiyu},
  booktitle={International conference on machine learning},
  pages={1675--1685},
  year={2019},
  organization={PMLR}
}

@inproceedings{lee2019wide,
  title={Wide neural networks of any depth evolve as linear models under gradient descent},
  author={Lee, Jaehoon and Xiao, Lechao and Schoenholz, Samuel and Bahri, Yasaman and Novak, Roman and Sohl-Dickstein, Jascha and Pennington, Jeffrey},
  booktitle={Advances in neural information processing systems},
  pages={8570--8581},
  year={2019}
}

@inproceedings{nesterov1983method,
  title={A method for unconstrained convex minimization problem with the rate of convergence O (1/k\^{} 2)},
  author={Nesterov, Yurii},
  booktitle={Doklady AN USSR},
  volume={269},
  pages={543--547},
  year={1983}
}

@inproceedings{zou2019improved,
  title={An improved analysis of training over-parameterized deep neural networks},
  author={Zou, Difan and Gu, Quanquan},
  booktitle={Advances in Neural Information Processing Systems},
  pages={2053--2062},
  year={2019}
}

@article{liu2020linearity,
  title={On the linearity of large non-linear models: when and why the tangent kernel is constant},
  author={Liu, Chaoyue and Zhu, Libin and Belkin, Misha},
  journal={Advances in Neural Information Processing Systems},
  volume={33},
  pages={15954--15964},
  year={2020}
}

@InProceedings{KingBa15,
  author    = {Kingma, Diederik and Ba, Jimmy},
  booktitle = {International Conference on Learning Representations (ICLR)},
  title     = {Adam: A Method for Stochastic Optimization},
  year      = {2015},
  address   = {San Diego, CA, USA},
  optmonth  = {12},
}

@article{lewkowycz2020large,
  title={The large learning rate phase of deep learning: the catapult mechanism},
  author={Lewkowycz, Aitor and Bahri, Yasaman and Dyer, Ethan and Sohl-Dickstein, Jascha and Gur-Ari, Guy},
  journal={arXiv preprint arXiv:2003.02218},
  year={2020}
}

@inproceedings{
cohen2021gradient,
title={Gradient Descent on Neural Networks Typically Occurs at the Edge of Stability},
author={Jeremy Cohen and Simran Kaur and Yuanzhi Li and J Zico Kolter and Ameet Talwalkar},
booktitle={International Conference on Learning Representations},
year={2021},
url={https://openreview.net/forum?id=jh-rTtvkGeM}
}

@article{liu2020loss,
  title={Loss landscapes and optimization in over-parameterized non-linear systems and neural networks},
  author={Liu, Chaoyue and Zhu, Libin and Belkin, Mikhail},
  journal={Applied and Computational Harmonic Analysis},
  year={2022},
  publisher={Elsevier}
}

@article{lee2020finite,
  title={Finite versus infinite neural networks: an empirical study},
  author={Lee, Jaehoon and Schoenholz, Samuel and Pennington, Jeffrey and Adlam, Ben and Xiao, Lechao and Novak, Roman and Sohl-Dickstein, Jascha},
  journal={Advances in Neural Information Processing Systems},
  volume={33},
  pages={15156--15172},
  year={2020}
}

@inproceedings{yang2021tensor,
  title={Tensor programs iv: Feature learning in infinite-width neural networks},
  author={Yang, Greg and Hu, Edward J},
  booktitle={International Conference on Machine Learning},
  pages={11727--11737},
  year={2021},
  organization={PMLR}
}

@article{fort2020deep,
  title={Deep learning versus kernel learning: an empirical study of loss landscape geometry and the time evolution of the neural tangent kernel},
  author={Fort, Stanislav and Dziugaite, Gintare Karolina and Paul, Mansheej and Kharaghani, Sepideh and Roy, Daniel M and Ganguli, Surya},
  journal={Advances in Neural Information Processing Systems},
  volume={33},
  pages={5850--5861},
  year={2020}
}

@article{ortiz2021can,
  title={What can linearized neural networks actually say about generalization?},
  author={Ortiz-Jim{\'e}nez, Guillermo and Moosavi-Dezfooli, Seyed-Mohsen and Frossard, Pascal},
  journal={Advances in Neural Information Processing Systems},
  volume={34},
  year={2021}
}

@article{netzer2011reading,
  title={Reading digits in natural images with unsupervised feature learning},
  author={Netzer, Yuval and Wang, Tao and Coates, Adam and Bissacco, Alessandro and Wu, Bo and Ng, Andrew Y},
  year={2011}
}

@article{krizhevsky2009learning,
  title={Learning multiple layers of features from tiny images},
  author={Krizhevsky, Alex and Hinton, Geoffrey and others},
  year={2009},
  publisher={Citeseer}
}

@inproceedings{
zhu2024quadratic,
title={Quadratic models for understanding catapult dynamics of neural networks},
author={Libin Zhu and Chaoyue Liu and Adityanarayanan Radhakrishnan and Mikhail Belkin},
booktitle={The Twelfth International Conference on Learning Representations},
year={2024},
url={https://openreview.net/forum?id=PvJnX3dwsD}
}

@article{lecun2015deep,
  title={Deep learning},
  author={LeCun, Yann and Bengio, Yoshua and Hinton, Geoffrey},
  journal={nature},
  volume={521},
  number={7553},
  pages={436--444},
  year={2015},
  publisher={Nature Publishing Group UK London}
}

@article{xing2018walk,
  title={A walk with sgd},
  author={Xing, Chen and Arpit, Devansh and Tsirigotis, Christos and Bengio, Yoshua},
  journal={arXiv preprint arXiv:1802.08770},
  year={2018}
}

@article{ruder2016overview,
  title={An overview of gradient descent optimization algorithms},
  author={Ruder, Sebastian},
  journal={arXiv preprint arXiv:1609.04747},
  year={2016}
}

@article{keskar2017improving,
  title={Improving generalization performance by switching from adam to sgd},
  author={Keskar, Nitish Shirish and Socher, Richard},
  journal={arXiv preprint arXiv:1712.07628},
  year={2017}
}

@article{paszke2019pytorch,
  title={Pytorch: An imperative style, high-performance deep learning library},
  author={Paszke, Adam and Gross, Sam and Massa, Francisco and Lerer, Adam and Bradbury, James and Chanan, Gregory and Killeen, Trevor and Lin, Zeming and Gimelshein, Natalia and Antiga, Luca and others},
  journal={Advances in neural information processing systems},
  volume={32},
  year={2019}
}

@article{abbe2021staircase,
  title={The staircase property: How hierarchical structure can guide deep learning},
  author={Abbe, Emmanuel and Boix-Adsera, Enric and Brennan, Matthew S and Bresler, Guy and Nagaraj, Dheeraj},
  journal={Advances in Neural Information Processing Systems},
  volume={34},
  pages={26989--27002},
  year={2021}
}

@article{kandel2020effect,
  title={The effect of batch size on the generalizability of the convolutional neural networks on a histopathology dataset},
  author={Kandel, Ibrahem and Castelli, Mauro},
  journal={ICT express},
  volume={6},
  number={4},
  pages={312--315},
  year={2020},
  publisher={Elsevier}
}

@article{masters2018revisiting,
  title={Revisiting small batch training for deep neural networks},
  author={Masters, Dominic and Luschi, Carlo},
  journal={arXiv preprint arXiv:1804.07612},
  year={2018}
}

@inproceedings{
keskar2017on,
title={On Large-Batch Training for Deep Learning: Generalization Gap and Sharp Minima},
author={Nitish Shirish Keskar and Dheevatsa Mudigere and Jorge Nocedal and Mikhail Smelyanskiy and Ping Tak Peter Tang},
booktitle={International Conference on Learning Representations},
year={2017},
url={https://openreview.net/forum?id=H1oyRlYgg}
}

@incollection{lecun2002efficient,
  title={Efficient backprop},
  author={LeCun, Yann and Bottou, L{\'e}on and Orr, Genevieve B and M{\"u}ller, Klaus-Robert},
  booktitle={Neural networks: Tricks of the trade},
  pages={9--50},
  year={2002},
  publisher={Springer}
}

@inproceedings{
smith2020origin,
title={On the Origin of Implicit Regularization in Stochastic Gradient Descent},
author={Samuel L Smith and Benoit Dherin and David Barrett and Soham De},
booktitle={International Conference on Learning Representations},
year={2021},
url={https://openreview.net/forum?id=rq_Qr0c1Hyo}
}

@article{hochreiter1994simplifying,
  title={Simplifying neural nets by discovering flat minima},
  author={Hochreiter, Sepp and Schmidhuber, J{\"u}rgen},
  journal={Advances in neural information processing systems},
  volume={7},
  year={1994}
}

@article{hochreiter1997flat,
  title={Flat minima},
  author={Hochreiter, Sepp and Schmidhuber, J{\"u}rgen},
  journal={Neural computation},
  volume={9},
  number={1},
  pages={1--42},
  year={1997},
  publisher={MIT Press One Rogers Street, Cambridge, MA 02142-1209, USA journals-info~…}
}

@inproceedings{dinh2017sharp,
  title={Sharp minima can generalize for deep nets},
  author={Dinh, Laurent and Pascanu, Razvan and Bengio, Samy and Bengio, Yoshua},
  booktitle={International Conference on Machine Learning},
  pages={1019--1028},
  year={2017},
  organization={PMLR}
}

@article{neyshabur2017exploring,
  title={Exploring generalization in deep learning},
  author={Neyshabur, Behnam and Bhojanapalli, Srinadh and McAllester, David and Srebro, Nati},
  journal={Advances in neural information processing systems},
  volume={30},
  year={2017}
}

@article{wu2017towards,
  title={Towards understanding generalization of deep learning: Perspective of loss landscapes},
  author={Wu, Lei and Zhu, Zhanxing and others},
  journal={arXiv preprint arXiv:1706.10239},
  year={2017}
}

@inproceedings{kleinberg2018alternative,
  title={An alternative view: When does SGD escape local minima?},
  author={Kleinberg, Bobby and Li, Yuanzhi and Yuan, Yang},
  booktitle={International conference on machine learning},
  pages={2698--2707},
  year={2018},
  organization={PMLR}
}

@inproceedings{
xie2020diffusion,
title={A Diffusion Theory For Deep Learning Dynamics: Stochastic Gradient Descent Exponentially Favors Flat Minima},
author={Zeke Xie and Issei Sato and Masashi Sugiyama},
booktitle={International Conference on Learning Representations},
year={2021},
url={https://openreview.net/forum?id=wXgk_iCiYGo}
}

@inproceedings{
Jiang*2020Fantastic,
title={Fantastic Generalization Measures and Where to Find Them},
author={Yiding Jiang* and Behnam Neyshabur* and Hossein Mobahi and Dilip Krishnan and Samy Bengio},
booktitle={International Conference on Learning Representations},
year={2020},
url={https://openreview.net/forum?id=SJgIPJBFvH}
}

@inproceedings{
foret2021sharpnessaware,
title={Sharpness-aware Minimization for Efficiently Improving Generalization},
author={Pierre Foret and Ariel Kleiner and Hossein Mobahi and Behnam Neyshabur},
booktitle={International Conference on Learning Representations},
year={2021},
url={https://openreview.net/forum?id=6Tm1mposlrM}
}

@inproceedings{arora2022understanding,
  title={Understanding gradient descent on the edge of stability in deep learning},
  author={Arora, Sanjeev and Li, Zhiyuan and Panigrahi, Abhishek},
  booktitle={International Conference on Machine Learning},
  pages={948--1024},
  year={2022},
  organization={PMLR}
}

@inproceedings{ahn2022understanding,
  title={Understanding the unstable convergence of gradient descent},
  author={Ahn, Kwangjun and Zhang, Jingzhao and Sra, Suvrit},
  booktitle={International Conference on Machine Learning},
  pages={247--257},
  year={2022},
  organization={PMLR}
}

@inproceedings{
damian2023selfstabilization,
title={Self-Stabilization: The Implicit Bias of Gradient Descent at the Edge of Stability},
author={Alex Damian and Eshaan Nichani and Jason D. Lee},
booktitle={The Eleventh International Conference on Learning Representations },
year={2023},
url={https://openreview.net/forum?id=nhKHA59gXz}
}

@inproceedings{
Frankle2020The,
title={The Early Phase of Neural Network Training},
author={Jonathan Frankle and David J. Schwab and Ari S. Morcos},
booktitle={International Conference on Learning Representations},
year={2020},
url={https://openreview.net/forum?id=Hkl1iRNFwS}
}

@inproceedings{smith2019super,
  title={Super-convergence: Very fast training of neural networks using large learning rates},
  author={Smith, Leslie N and Topin, Nicholay},
  booktitle={Artificial intelligence and machine learning for multi-domain operations applications},
  volume={11006},
  pages={369--386},
  year={2019},
  organization={SPIE}
}

@article{gilmer2021loss,
  title={A loss curvature perspective on training instability in deep learning},
  author={Gilmer, Justin and Ghorbani, Behrooz and Garg, Ankush and Kudugunta, Sneha and Neyshabur, Behnam and Cardoze, David and Dahl, George and Nado, Zachary and Firat, Orhan},
  journal={arXiv preprint arXiv:2110.04369},
  year={2021}
}

@article{zhu2022transition,
  title={Transition to linearity of general neural networks with directed acyclic graph architecture},
  author={Zhu, Libin and Liu, Chaoyue and Belkin, Misha},
  journal={Advances in Neural Information Processing Systems},
  volume={35},
  pages={5363--5375},
  year={2022}
}

@article{kalra2023phase,
  title={Phase diagram of early training dynamics in deep neural networks: effect of the learning rate, depth, and width},
  author={Kalra, Dayal Singh and Barkeshli, Maissam},
  journal={Advances in Neural Information Processing Systems},
  volume={37},
  year={2023}
}

@inproceedings{zagoruyko2016wide,
  title={Wide Residual Networks},
  author={Zagoruyko, Sergey and Komodakis, Nikos},
  booktitle={British Machine Vision Conference 2016},
  year={2016},
  organization={British Machine Vision Association}
}

@inproceedings{
dosovitskiy2021an,
title={An Image is Worth 16x16 Words: Transformers for Image Recognition at Scale},
author={Alexey Dosovitskiy and Lucas Beyer and Alexander Kolesnikov and Dirk Weissenborn and Xiaohua Zhai and Thomas Unterthiner and Mostafa Dehghani and Matthias Minderer and Georg Heigold and Sylvain Gelly and Jakob Uszkoreit and Neil Houlsby},
booktitle={International Conference on Learning Representations},
year={2021},
url={https://openreview.net/forum?id=YicbFdNTTy}
}

@misc{myrtle,
  author = {Myrtle.ai},
  title = {Myrtle Network},
  year = {2018},
  howpublished = {\url{https://myrtle.ai/}}
}

@misc{rmsprop,
 author = {Geoffrey Hinton},
 title = {},
 url = {http://www.cs.toronto.edu/~tijmen/csc321/slides/lecture_slides_lec6.pdf},
 year = {2014}
}

@article{radhakrishnan2024mechanism,
  title={Mechanism for feature learning in neural networks and backpropagation-free machine learning models},
  author={Radhakrishnan, Adityanarayanan and Beaglehole, Daniel and Pandit, Parthe and Belkin, Mikhail},
  journal={Science},
  volume={383},
  number={6690},
  pages={1461--1467},
  year={2024},
  publisher={American Association for the Advancement of Science}
}

@inproceedings{trivedi2014consistent,
  title={A consistent estimator of the expected gradient outerproduct},
  author={Trivedi, Shubhendu and Wang, Jialei and Kpotufe, Samory and Shakhnarovich, Gregory},
  booktitle={Proceedings of the Thirtieth Conference on Uncertainty in Artificial Intelligence},
  pages={819--828},
  year={2014}
}

@inproceedings{liu2015faceattributes,
  title = {Deep Learning Face Attributes in the Wild},
  author = {Liu, Ziwei and Luo, Ping and Wang, Xiaogang and Tang, Xiaoou},
  booktitle = {Proceedings of International Conference on Computer Vision (ICCV)},
  month = {December},
  year = {2015} 
}

@article{qian1999momentum,
  title={On the momentum term in gradient descent learning algorithms},
  author={Qian, Ning},
  journal={Neural networks},
  volume={12},
  number={1},
  pages={145--151},
  year={1999},
  publisher={Elsevier}
}

@article{zeiler2012adadelta,
  title={Adadelta: an adaptive learning rate method},
  author={Zeiler, Matthew D},
  journal={arXiv preprint arXiv:1212.5701},
  year={2012}
}

@article{duchi2011adaptive,
  title={Adaptive subgradient methods for online learning and stochastic optimization.},
  author={Duchi, John and Hazan, Elad and Singer, Yoram},
  journal={Journal of machine learning research},
  volume={12},
  number={7},
  year={2011}
}

@inproceedings{
jastrzębski2018on,
title={On the Relation Between the Sharpest Directions of {DNN} Loss and the {SGD} Step Length},
author={Stanisław Jastrzębski and Zachary Kenton and Nicolas Ballas and Asja Fischer and Yoshua Bengio and Amost Storkey},
booktitle={International Conference on Learning Representations},
year={2019},
url={https://openreview.net/forum?id=SkgEaj05t7},
}

@inproceedings{
Jastrzebski2020The,
title={The Break-Even Point on Optimization Trajectories of Deep Neural Networks},
author={Stanislaw Jastrzebski and Maciej Szymczak and Stanislav Fort and Devansh Arpit and Jacek Tabor and Kyunghyun Cho* and Krzysztof Geras*},
booktitle={International Conference on Learning Representations},
year={2020},
url={https://openreview.net/forum?id=r1g87C4KwB}
}

@article{goyal2017accurate,
  title={Accurate, large minibatch sgd: Training imagenet in 1 hour},
  author={Goyal, Priya and Doll{\'a}r, Piotr and Girshick, Ross and Noordhuis, Pieter and Wesolowski, Lukasz and Kyrola, Aapo and Tulloch, Andrew and Jia, Yangqing and He, Kaiming},
  journal={arXiv preprint arXiv:1706.02677},
  year={2017}
}

@article{jastrzkebski2017three,
  title={Three factors influencing minima in sgd},
  author={Jastrz{\k{e}}bski, Stanis{\l}aw and Kenton, Zachary and Arpit, Devansh and Ballas, Nicolas and Fischer, Asja and Bengio, Yoshua and Storkey, Amos},
  journal={arXiv preprint arXiv:1711.04623},
  year={2017}
}

@article{robbins1951stochastic,
  title={A stochastic approximation method},
  author={Robbins, Herbert and Monro, Sutton},
  journal={The annals of mathematical statistics},
  pages={400--407},
  year={1951},
  publisher={JSTOR}
}

@inproceedings{he2016deep,
  title={Deep residual learning for image recognition},
  author={He, Kaiming and Zhang, Xiangyu and Ren, Shaoqing and Sun, Jian},
  booktitle={Proceedings of the IEEE conference on computer vision and pattern recognition},
  pages={770--778},
  year={2016}
}

@inproceedings{huang2017densely,
  title={Densely connected convolutional networks},
  author={Huang, Gao and Liu, Zhuang and Van Der Maaten, Laurens and Weinberger, Kilian Q},
  booktitle={Proceedings of the IEEE conference on computer vision and pattern recognition},
  pages={4700--4708},
  year={2017}
}

@article{hardle1989investigating,
  title={Investigating smooth multiple regression by the method of average derivatives},
  author={H{\"a}rdle, Wolfgang and Stoker, Thomas M},
  journal={Journal of the American statistical Association},
  volume={84},
  number={408},
  pages={986--995},
  year={1989},
  publisher={Taylor \& Francis}
}

@article{hristache2001structure,
  title={Structure adaptive approach for dimension reduction},
  author={Hristache, Marian and Juditsky, Anatoli and Polzehl, Jorg and Spokoiny, Vladimir},
  journal={Annals of Statistics},
  pages={1537--1566},
  year={2001},
  publisher={JSTOR}
}

@article{xia2002adaptive,
  title={An adaptive estimation of dimension reduction space},
  author={Xia, Yingcun and Tong, Howell and Li, Wai Keung and Zhu, Li-Xing},
  journal={Journal of the Royal Statistical Society: Series B (Statistical Methodology)},
  volume={64},
  number={3},
  pages={363--410},
  year={2002},
  publisher={Wiley Online Library}
}

@article{ding2024flat,
  title={Flat minima generalize for low-rank matrix recovery},
  author={Ding, Lijun and Drusvyatskiy, Dmitriy and Fazel, Maryam and Harchaoui, Zaid},
  journal={Information and Inference: A Journal of the IMA},
  volume={13},
  number={2},
  pages={iaae009},
  year={2024},
  publisher={Oxford University Press}
}

@inproceedings{
liu2022transition,
title={Transition to Linearity of Wide Neural Networks is an Emerging Property of Assembling Weak Models},
author={Chaoyue Liu and Libin Zhu and Misha Belkin},
booktitle={International Conference on Learning Representations},
year={2022},
url={https://openreview.net/forum?id=CyKHoKyvgnp}
}

@inproceedings{agarwala2023sam,
  title={SAM operates far from home: eigenvalue regularization as a dynamical phenomenon},
  author={Agarwala, Atish and Dauphin, Yann},
  booktitle={International Conference on Machine Learning},
  pages={152--168},
  year={2023},
  organization={PMLR}
}

@inproceedings{agarwala2023second,
  title={Second-order regression models exhibit progressive sharpening to the edge of stability},
  author={Agarwala, Atish and Pedregosa, Fabian and Pennington, Jeffrey},
  booktitle={International Conference on Machine Learning},
  pages={169--195},
  year={2023},
  organization={PMLR}
}

@inproceedings{banerjee2023neural,
  title={Neural tangent kernel at initialization: linear width suffices},
  author={Banerjee, Arindam and Cisneros-Velarde, Pedro and Zhu, Libin and Belkin, Mikhail},
  booktitle={Uncertainty in Artificial Intelligence},
  pages={110--118},
  year={2023},
  organization={PMLR}
}

@article{meltzer2023catapult,
  title={Catapult Dynamics and Phase Transitions in Quadratic Nets},
  author={Meltzer, David and Liu, Junyu},
  journal={arXiv preprint arXiv:2301.07737},
  year={2023}
}

@inproceedings{smith2017cyclical,
  title={Cyclical learning rates for training neural networks},
  author={Smith, Leslie N},
  booktitle={2017 IEEE winter conference on applications of computer vision (WACV)},
  pages={464--472},
  year={2017},
  organization={IEEE}
}

@inproceedings{izmailov2018averaging,
  title={Averaging weights leads to wider optima and better generalization},
  author={Izmailov, P and Wilson, AG and Podoprikhin, D and Vetrov, D and Garipov, T},
  booktitle={34th Conference on Uncertainty in Artificial Intelligence 2018, UAI 2018},
  pages={876--885},
  year={2018}
}

@inproceedings{
geiping2022stochastic,
title={Stochastic Training is Not Necessary for Generalization},
author={Jonas Geiping and Micah Goldblum and Phil Pope and Michael Moeller and Tom Goldstein},
booktitle={International Conference on Learning Representations},
year={2022},
url={https://openreview.net/forum?id=ZBESeIUB5k}
}

@article{zhang2023loss,
  title={Loss Spike in Training Neural Networks},
  author={Zhang, Zhongwang and Xu, Zhi-Qin John},
  journal={arXiv preprint arXiv:2305.12133},
  year={2023}
}

@article{wang2022analyzing,
  title={Analyzing sharpness along gd trajectory: Progressive sharpening and edge of stability},
  author={Wang, Zixuan and Li, Zhouzi and Li, Jian},
  journal={Advances in Neural Information Processing Systems},
  volume={35},
  pages={9983--9994},
  year={2022}
}

@inproceedings{papyan2019measurements,
  title={Measurements of Three-Level Hierarchical Structure in the Outliers in the Spectrum of Deepnet Hessians},
  author={Papyan, Vardan},
  booktitle={International Conference on Machine Learning},
  pages={5012--5021},
  year={2019},
  organization={PMLR}
}

@article{loo2022evolution,
  title={Evolution of neural tangent kernels under benign and adversarial training},
  author={Loo, Noel and Hasani, Ramin and Amini, Alexander and Rus, Daniela},
  journal={Advances in Neural Information Processing Systems},
  volume={35},
  pages={11642--11657},
  year={2022}
}

@inproceedings{
atanasov2022neural,
title={Neural Networks as Kernel Learners: The Silent Alignment Effect},
author={Alexander Atanasov and Blake Bordelon and Cengiz Pehlevan},
booktitle={International Conference on Learning Representations},
year={2022},
url={https://openreview.net/forum?id=1NvflqAdoom}
}

@article{yuan2023efficient,
  title={Efficient Estimation of the Central Mean Subspace via Smoothed Gradient Outer Products},
  author={Yuan, Gan and Xu, Mingyue and Kpotufe, Samory and Hsu, Daniel},
  journal={arXiv preprint arXiv:2312.15469},
  year={2023}
}

@article{beaglehole2023mechanism,
  title={Mechanism of feature learning in convolutional neural networks},
  author={Beaglehole, Daniel and Radhakrishnan, Adityanarayanan and Pandit, Parthe and Belkin, Mikhail},
  journal={arXiv preprint arXiv:2309.00570},
  year={2023}
}

@article{radhakrishnan2024linear,
  title={Linear Recursive Feature Machines provably recover low-rank matrices},
  author={Radhakrishnan, Adityanarayanan and Belkin, Mikhail and Drusvyatskiy, Dmitriy},
  journal={arXiv preprint arXiv:2401.04553},
  year={2024}
}

@article{xiao2017fashion,
  title={Fashion-mnist: a novel image dataset for benchmarking machine learning algorithms},
  author={Xiao, Han and Rasul, Kashif and Vollgraf, Roland},
  journal={arXiv preprint arXiv:1708.07747},
  year={2017}
}

@ARTICLE{uspsdataset,
  author={J. J. {Hull}},
  journal={IEEE Transactions on Pattern Analysis and Machine Intelligence}, 
  title={A database for handwritten text recognition research}, 
  year={1994},
  volume={16},
  number={5},
  pages={550-554},
  doi={10.1109/34.291440}
}

@inproceedings{
wang2021large,
title={Large Learning Rate Tames Homogeneity: Convergence and Balancing Effect},
author={Yuqing Wang and Minshuo Chen and Tuo Zhao and Molei Tao},
booktitle={International Conference on Learning Representations},
year={2022},
url={https://openreview.net/forum?id=3tbDrs77LJ5}
}
